\documentclass{article}
\usepackage{arxiv,times}

\usepackage{amsmath,amssymb,amsthm,mathtools,mathrsfs}
\usepackage{graphicx}
\usepackage{booktabs}
\usepackage{multirow}
\usepackage{xcolor}
\usepackage{hyperref}
\hypersetup{hidelinks}
\usepackage{url}
\usepackage[capitalize,nameinlink]{cleveref}
\usepackage{thm-restate}

\usepackage{todonotes}
\newcommand{\Acc}{\mathcal{A}}

\newtheorem{theorem}{Theorem}[section]
\newtheorem{proposition}[theorem]{Proposition}

\newtheorem{lemma}[theorem]{Lemma}
\newtheorem{corollary}[theorem]{Corollary}

\theoremstyle{definition}

\newenvironment{linkedtheorem}[2]{%
  \refstepcounter{theorem}%
  \par\addvspace{\topsep}%
  \noindent
  \textbf{Theorem~\thetheorem}\,
  {\normalfont[\hyperref[#2]{\ref*{#2}}]}\,
  \textbf{(#1).}\enspace
  \itshape
}{%
  \par\addvspace{\topsep}%
}

\title{The Accuracy Well: Nonmonotone Accuracy Dynamics at Vanishing Initialization}
\author{%
    Yaroslav~Gusev\\
    Applied AI Institute\\
    Moscow, Russia \\
    \texttt{i.gusev@applied-ai.ru} \\
    \And
    Eugene~Golikov\\
    Applied AI Institute\\
    Moscow, Russia \\
    \texttt{e.golikov@applied-ai.ru} \\
    \And
    Dmitry Yarotsky\\
    Applied AI Institute \& \\ Steklov Mathematical Institute of Russian Academy of Sciences\\
    Moscow, Russia \\
    \texttt{yarotsky@gmail.com} \\
}

\begin{document}
\maketitle

\begin{abstract}
We study classification dynamics under small initialization and identify a pronounced non-monotone behavior of accuracy that we call the \emph{accuracy well}: accuracy rises rapidly early in training, subsequently drops, and only much later recovers. We observe this phenomenon in deep linear networks, fully connected ReLU networks on MNIST, deep convolutional networks on CIFAR-10, and controlled synthetic tasks, across different losses and optimizers and in both training and test accuracy. We propose a three-stage explanation in terms of target-driven organization, a degeneracy bottleneck, and subsequent recruitment and recovery. We develop a rigorous theory of these dynamics for shallow and deep linear networks, showing in particular that high accuracy can emerge while the predictor and the change in loss remain asymptotically vanishing, before continued optimization causes accuracy to deteriorate. For a simple synthetic classification task, we prove an extreme instance in which accuracy asymptotically rises to \(100\%\), collapses to \(0\), and later recovers permanently to \(100\%\). Our results show that classification accuracy can reveal aspects of learning
dynamics that are not captured by the training loss alone.
\end{abstract}

\section{Introduction}
\label{sec:intro}

Initialization scale can qualitatively change the dynamics of gradient-based
learning.  In suitable parameterizations and scalings, training can enter a lazy,
kernel-like regime in which the nonlinear trajectory remains close to its
linearization around initialization
\citep{jacot2018ntk,chizat2019lazy}, whereas small or vanishing
initialization can expose substantially richer evolution of the learned
predictor \citep{woodworth2020kernel}. This regime has been studied from
several complementary perspectives.  In linear and homogeneous models it
can produce sequential learning of spectral components, incremental
low-rank structure, and saddle-to-saddle dynamics
\citep{saxe2014exact,gidel2019implicit, gissin2020implicit,jacot2021saddle,zhang2026saddle};
in nonlinear networks, substantial directional or feature organization can
already occur before appreciable loss decay
\citep{atanasov2022silent,boursier2022gradient,min2024early}.  Small initialization is therefore
not merely a choice of scale: it can separate qualitatively different stages
of learning that are compressed together at larger initialization.

\begin{figure}[t]
    \centering
    \includegraphics[width=0.6\linewidth]
    {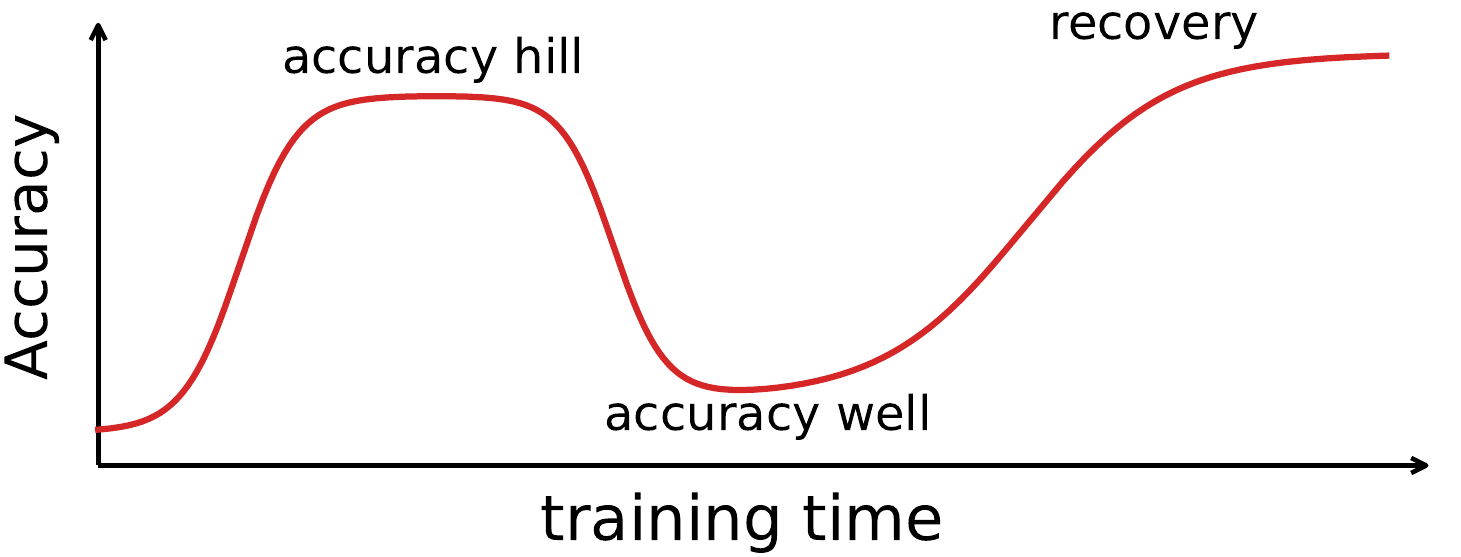}

    \caption{
    Schematic accuracy well: non-monotone accuracy dynamics at small initialization.
    }
    \label{main:accuracy_well}
\end{figure}

In this paper we describe and analyze a small-initialization phenomenon that
we call the \emph{accuracy well}.  We consider classification problems in
which a model is trained from small initialization and follow its
classification accuracy (i.e., the fraction of examples classified
correctly) throughout training.  At sufficiently small initialization, we
repeatedly observe a pronounced non-monotone accuracy trajectory: it first rises
rapidly, then significantly drops, and only much later recovers.  We call
the early rise an \emph{accuracy hill}; together with the subsequent drop
and recovery, it forms an \emph{accuracy well}; see
Figure~\ref{main:accuracy_well}. The phenomenon appears in deep linear networks, fully connected ReLU
networks on MNIST, and deep convolutional networks on CIFAR-10, is visible in both training and test accuracy, persists across different losses and optimization methods, and disappears in several informative control settings.

Related non-monotone predictive phenomena include transient
prediction-level effects arising from unequal spectral learning rates in
deep linear networks \citep{saxe2019semantic}, example forgetting and
subset-specific accuracy dynamics
\citep{toneva2019forgetting,schneider2025learning}, as well as epoch-wise
double descent, where the pronounced non-monotonicity is primarily observed
in test or generalization performance
\citep{nakkiran2020deep,heckel2021early,
olmin2024towards}. Existing theoretical accounts of these neighboring phenomena do not directly characterize the pronounced non-monotone evolution of the training
accuracy studied here. We instead follow how the induced classifier and its class ordering evolve from small initialization, and develop rigorous results for the resulting sequence of early improvement, intermediate deterioration, and later recovery. To our knowledge, such dynamics have not previously been analyzed directly under full-batch training. Detailed connections to prior work are discussed in Appendix~\ref{sec:related}.

\paragraph{Our contributions.}
\begin{itemize}

     \item
    We \emph{systematically analyze the phenomenology of the accuracy well} across linear and nonlinear models, including classification on MNIST, CIFAR-10, and controlled synthetic tasks;
    see Section~\ref{sec:phenomenology}.

    \item
    We \emph{propose a general heuristic explanation of the accuracy well} in terms of a target-driven organization, a degeneracy bottleneck, and subsequent recruitment; see Section~\ref{sec:heuristic}.

    \item
    We \emph{develop a rigorous theory of the accuracy well} in shallow and
    deep linear networks.  We obtain general results for classification
    dynamics under small initialization; see Section~\ref{sec:linear}.

\end{itemize}

\section{Phenomenology}
\label{sec:phenomenology}

\begin{figure}[t]
    \centering

    \includegraphics[width=0.32\linewidth]
    {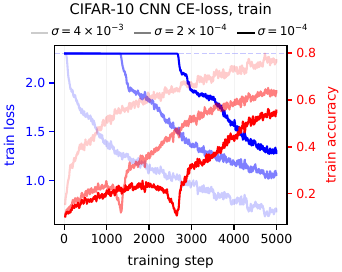}
    \hfill
    \includegraphics[width=0.32\linewidth]
    {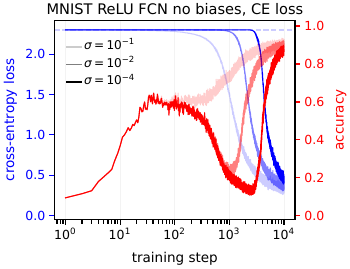}
    \hfill
    \includegraphics[width=0.32\linewidth]
    {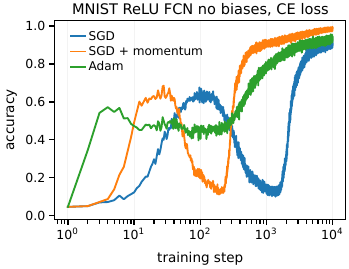}

    \vspace{0.35em}

    \includegraphics[width=0.32\linewidth]
    {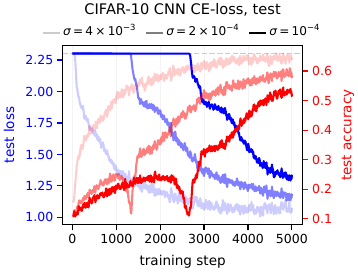}
    \hfill
    \includegraphics[width=0.32\linewidth]
    {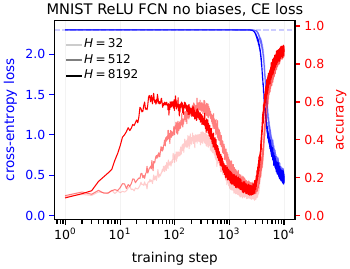}
    \hfill
    \includegraphics[width=0.32\linewidth]
    {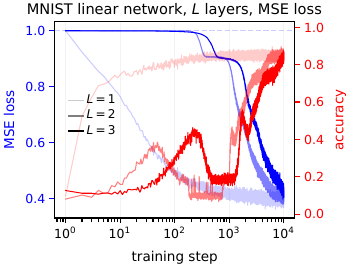}

    \caption{
    \textbf{Accuracy-well phenomenology on CIFAR-10 and MNIST.}
    \emph{Left}: train and test dynamics of a two-hidden-layer CIFAR-10 CNN
    under an initialization-scale sweep.
    \emph{Middle}: MNIST ReLU networks without biases under cross-entropy,
    varying initialization scale (top) and width (bottom).
    \emph{Right}: optimizer comparison for the MNIST ReLU network (top) and
    a comparison of linear networks $f=\mathbf W_L\cdots \mathbf W_1 $ on MNIST with
    \(L=1,2,3\) (bottom).
    }
    \label{fig:main-cifar-mnist}
\end{figure}

\begin{figure}[t]
    \centering
    \includegraphics[width=0.32\linewidth]
    {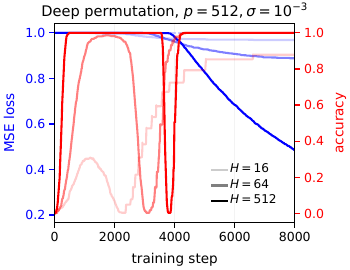}
    \hfill
    \includegraphics[width=0.32\linewidth]
    {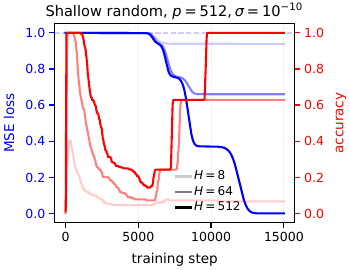}
    \hfill
    \includegraphics[width=0.32\linewidth]
    {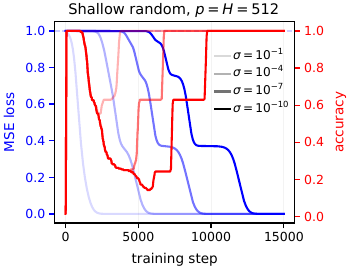}

    \caption{
    \textbf{Representative unary-operation experiments.}
    \emph{Left}: a deep linear network ($L=3$ layers) trained on a permutation
    operation develops a pronounced accuracy well as width increases.
    \emph{Middle}: a shallow network ($L=2$ layers) trained on a random unary
    operation develops a well in which the early classifier can substantially
    outperform the late-time predictor.
    \emph{Right}: decreasing the initialization scale \(\sigma\) separates
    the rise, collapse, and recovery over increasingly distinct time scales.
    }
    \label{fig:main-unary}
\end{figure}

We first analyze the accuracy well empirically across a range of models
and training settings.  Figures~\ref{fig:main-cifar-mnist}
and~\ref{fig:main-unary} collect representative examples and controlled
sweeps that reveal several recurring features of the phenomenon.  Our aim is
not to provide an exhaustive taxonomy of when an accuracy well occurs, but to identify the main empirical regularities that motivate our theoretical study.

\paragraph{Experimental settings and breadth of the phenomenon.}
We consider both standard image-classification tasks and simple synthetic
problems, using architectures ranging from fully connected linear and ReLU
networks to a convolutional network.  For natural data we use MNIST and
CIFAR-10.  As controlled synthetic tasks, we consider unary operations
\(f:[p]\to[p]\), where \([p]=\{1,\ldots,p\}\), represented using one-hot
inputs and outputs.  We study random maps, where the values \(f(i)\) are
chosen independently and uniformly, and permutations, where \(f\) is
bijective.  Across these different settings we repeatedly observe pronounced accuracy
wells, while several controls show that the phenomenon is not universal.
On MNIST, the accuracy-well pattern is also reproducible across five
independent random seeds. On CIFAR-10, we also observe an accuracy well in a five-hidden-layer convolutional network.
Experimental details and additional figures are given in
Appendix~\ref{app:experiments}.

\paragraph{Relation to loss decay.}
A recurring pattern is that the early accuracy hill forms while the loss does not visibly change. The accuracy then starts to fall, and the
low point of the well typically occurs around the time when the loss begins
to decrease noticeably. This behavior is visible across several sweeps in
Figures~\ref{fig:main-cifar-mnist} and~\ref{fig:main-unary}. Thus large
changes in classification can happen before substantial loss optimization,
while the deepest part of the well tends to align with the onset of loss
decay.

\paragraph{Initialization scale.}
The most consistent control parameter is the initialization scale. At
relatively large \(\sigma\), accuracy typically evolves more monotonically,
whereas decreasing \(\sigma\) progressively separates the early hill, the
collapse, and the eventual recovery. This is especially clear in the
top-middle panel of Figure~\ref{fig:main-cifar-mnist} and the right panel of
Figure~\ref{fig:main-unary}, where the same qualitative phases become
increasingly well separated in training time as \(\sigma\) decreases.

Small initialization is nevertheless not sufficient. The shallow permutation model remains free of a well at small initialization, while the directly optimized
linear MNIST model (bottom-right panel of
Figure~\ref{fig:main-cifar-mnist}) also exhibits monotone accuracy dynamics. Thus small initialization makes the well sharply visible in many
settings, but does not by itself determine whether one occurs.

\paragraph{Model capacity.}
Increasing width generally strengthens the early classifier and deepens the
resulting well. This is visible for MNIST in the bottom-middle panel of
Figure~\ref{fig:main-cifar-mnist} and in the width sweeps in the left and
middle panels of Figure~\ref{fig:main-unary}. We do not claim a universal
width-scaling law, since the quantitative effect is visibly task- and
architecture-dependent.

\paragraph{Loss, train--test behavior, and optimization.}
The well appears across different losses and optimizers.  The unary
experiments use mean-squared error, while CIFAR-10 and most MNIST experiments
use cross-entropy.  On CIFAR-10, train and test accuracy show closely aligned
hills, collapses, and recoveries, so the phenomenon is not specific to either
training or test accuracy.  On MNIST, the well is present with SGD, SGD with
momentum, and Adam, although the drop is smaller for Adam under the
hyperparameters used here.  A systematic optimizer comparison is beyond our
scope.

\paragraph{Depth and target structure.}
Depth can qualitatively change accuracy dynamics even when the
end-to-end function class is unchanged.  The bottom-right panel of
Figure~\ref{fig:main-cifar-mnist} compares linear networks
\(f(\mathbf{x})=\mathbf{W}_L\cdots \mathbf{W}_1\mathbf{x}\) trained on MNIST with \(L=1,2,3\).  All represent
linear maps \(\mathbb R^{784}\to\mathbb R^{10}\) and start from small initialization, but the deeper models
(\(L>1\)) develop pronounced accuracy wells while direct optimization
(\(L=1\)) does not.

Target structure also matters.  For a two-layer linear network, a
permutation target retains its early high accuracy, whereas a random unary
target develops a pronounced well
(Figure~\ref{fig:main-unary}).
At greater depth, however, even the permutation target can develop a well.
Thus neither small initialization nor target structure alone determines the
phenomenon; the depth matters as well.  Additional architectural
controls, including bias ablations, are reported in
Appendix~\ref{app:experiments}.

\paragraph{Early versus late classifiers.}
The middle panel of Figure~\ref{fig:main-unary} shows a particularly
striking effect: narrow random-unary networks can reach very high accuracy
early in training but settle at substantially lower accuracy after much
longer optimization of the MSE.  Thus continued loss minimization does not
necessarily improve classification accuracy: an early predictor can be a
better classifier than a much later predictor with substantially smaller
loss.

These observations raise the central question: what mechanism
produces the accuracy well?

\section{A heuristic picture of the accuracy well}
\label{sec:heuristic}

We propose a three-stage picture for the accuracy well:
\emph{target-driven organization}, a \emph{degeneracy bottleneck}, and
\emph{recruitment and recovery}. The first two stages occur on a microscopic
scale: the classifier can reorganize substantially while the predictor
remains small and the loss is nearly unchanged. The final recovery instead
occurs on the macroscopic training scale, together with substantial growth of
the predictor and loss decay.

\paragraph{Target-driven organization.}
The first stage can be understood from a local linearization of the initial
dynamics, closely related to the picture underlying lazy learning
\citep{jacot2018ntk,chizat2019lazy}. Here, however, we use this only as a short-time description suitable for
illustrating the first stage. Subsequent learning may involve dynamics beyond this initial first-order behavior.

Consider a \(C\)-class model
\(f_{\boldsymbol{\theta}}:\mathbb R^p\to\mathbb R^C\), parametrized by
\(\boldsymbol{\theta}\in\mathbb R^d\), and stack its outputs and the
corresponding ground-truth targets over the training set into
\(\mathbf F_{\boldsymbol{\theta}},\mathbf Y\in\mathbb R^{NC}\).
For squared loss,
\begin{equation}
\dot{\mathbf F}_{\boldsymbol{\theta}}
=
\mathbf K_{\boldsymbol{\theta}}
\left(
\mathbf Y-\mathbf F_{\boldsymbol{\theta}}
\right),
\qquad
\mathbf K_{\boldsymbol{\theta}}
=
\mathbf J_{\boldsymbol{\theta}}
\mathbf J_{\boldsymbol{\theta}}^{\top},
\qquad
\mathbf J_{\boldsymbol{\theta}}
=
\frac{\partial\mathbf F_{\boldsymbol{\theta}}}
{\partial\boldsymbol{\theta}},
\label{eq:heuristic-predictor-flow}
\end{equation}
where \(\mathbf K_{\boldsymbol{\theta}}\) is the empirical tangent kernel.
If vanishing initialization makes
\(\mathbf F_{\boldsymbol{\theta}(0)}\) small, then over a sufficiently short
initial interval the tangent kernel is approximately frozen to first order,
and
\begin{equation}
\mathbf F_{\boldsymbol{\theta}(\tau)}
\approx
\underbrace{\mathbf F_{\boldsymbol{\theta}(0)}}_{\text{initialization noise}}
+
\underbrace{\tau \mathbf K_{\boldsymbol{\theta}(0)}\mathbf Y}_{\text{target-driven signal}}.
\label{eq:heuristic-early-target-motion}
\end{equation}
Equation~\eqref{eq:heuristic-early-target-motion} suggests a simple
competition during the first stage. The predictor initially contains a
random contribution inherited from initialization, while training begins to
build a coherent component aligned with the targets. At very early times the
former may still dominate, but the target-driven component grows with
\(\tau\). Increasing the width favors the target-driven component in this competition:
initialization-dependent fluctuations are reduced by self-averaging, whereas
the target-driven contribution adds coherently. Thus
there can be an early window in which \(\tau\) is already large enough for
the target-driven signal to control the score ordering, but still small
enough for the local linearized description above to remain accurate. In
this window the predictor itself may remain microscopic---indeed, under
vanishing initialization the tangent kernel and hence the target-driven
signal may also vanish in absolute magnitude---yet the classifier can already
achieve high accuracy while the loss has barely changed. Our linear theory in
Section~\ref{sec:linear} makes this target-driven organization rigorous, with
the precise near-origin dynamics depending on the model and depth. The same
qualitative mechanism may also extend beyond squared loss.

\begin{figure}[t]
    \centering
    \includegraphics[width=0.67\linewidth]{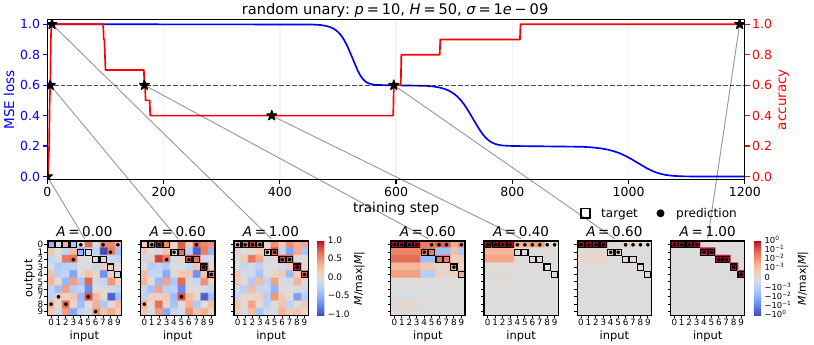}
    \hfill
    \includegraphics[width=0.32\linewidth]{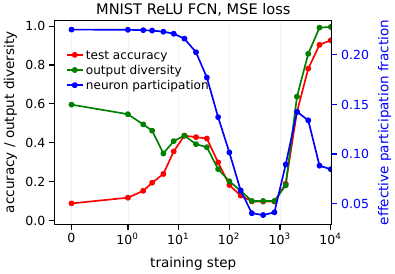}
    \caption{
\textbf{Diagnostics of the proposed accuracy-well mechanism.}
\emph{Left}: evolution of a two-layer linear network
\(\mathbf M=\mathbf{W}_2 \mathbf{W}_1\) on a random unary operation.  Snapshots of the normalized
score matrix show how the model reorganizes along the accuracy-well
trajectory. Notably, snapshots with the same
    accuracy (\(A=0.6\)) have qualitatively different score-matrix
    structure.
\emph{Right}: test accuracy, effective hidden-neuron participation
    fraction, and output diversity in a one-hidden-layer ReLU network on
    MNIST. The participation fraction measures the effective fraction of hidden
    neurons contributing to class-dependent scores, while output diversity
    measures the effective fraction of output classes used by the predictions; see equations \eqref{eq:mnist-participation-fraction},~\eqref{eq:mnist-output-diversity} in Appendix~\ref{app:mnist-experiments}.
}
    \label{fig:main-mechanism}
\end{figure}

\paragraph{Degeneracy bottleneck.}
We associate the subsequent loss of accuracy with the predictor becoming
temporarily \emph{degenerate} while still remaining microscopic. We call
this stage a \emph{degeneracy bottleneck}. Its precise form is
model-dependent. In two-layer linear networks, the predictor can be viewed
as a sum of rank-one singular components whose microscopic growth rates may
separate strongly. The dynamics can therefore become dominated by only a
small subset of these components, possibly containing several equally
dominant modes. If this dominant subset does not preserve the class ordering
produced by their earlier combination, accuracy falls. This mechanism is
visible in Figure~\ref{fig:main-mechanism} (left) and is made rigorous in
Section~\ref{sec:linear}.

For deeper linear networks, our rigorous result uses the stronger
nondegeneracy assumption that the leading singular direction of the target
is unique. Under generic Gaussian initialization, the first saddle escape is then asymptotically rank one, with the probability guarantees described in
Section~\ref{sec:linear}, by the result of \citet{jacot2021saddle}. For
nonlinear networks we do not claim an analogous spectral characterization
and instead use degeneracy as an empirical diagnostic. In the
one-hidden-layer ReLU experiment, both the effective hidden-neuron
participation fraction and the output diversity decrease sharply near the
bottom of the accuracy well; see Figure~\ref{fig:main-mechanism} (right) and
Appendix~\ref{app:mnist-experiments}. Thus, across these settings, the
bottleneck corresponds to a transient loss of predictive diversity, although
the microscopic mechanism producing it need not be universal.

\paragraph{Recruitment and recovery.}
The final stage is qualitatively different from the first two. Recovery
takes place on the macroscopic training scale, when the predictor grows
substantially and the loss begins to decrease. At this stage, predictive
structure that was suppressed or absent at the bottleneck can become
dynamically relevant. We call this re-entry or emergence of complementary
predictive structure \emph{recruitment}. In linear models this corresponds
to the progressive incorporation of additional singular modes, closely
related to classical mode-learning and incremental-learning dynamics
\citep{saxe2014exact}. In nonlinear networks, the analogous phenomenon is naturally viewed in
terms of feature learning: task-relevant features or directions can emerge
progressively over training \citep{dandi2024twolayer}. We use
\emph{recruitment} to emphasize specifically this recovery-side increase in predictive diversity.

Both panels of Figure~\ref{fig:main-mechanism} are consistent with this
picture. In the linear network example, the score matrix regains a diverse, target-like structure as accuracy recovers. In the ReLU network, effective neuron participation and the output diversity rise again during the recovery stage. Thus the accuracy well separates two microscopic stages, target-driven organization
and a degeneracy bottleneck, from a later macroscopic stage of recruitment
and recovery.

\section[
Accuracy Dynamics in Deep Linear Networks under Vanishing Initialization
]{
Accuracy Dynamics in Deep Linear Networks under Vanishing Initialization
\texorpdfstring{[\hyperref[app:linear]{Appendix~\ref*{app:linear}}]}{}
}
\label{sec:linear}

We now develop a rigorous realization of this picture in deep linear
networks.
Consider a \(C\)-class training set
\(\{(\mathbf{x}_i,\mathbf{y}_i)\}_{i=1}^N\), with
\(\mathbf{x}_i\in\mathbb R^p, ~\mathbf{y}_i=\mathbf{e}_{c_i}\in\mathbb R^C,\)
\(c_i\in\{1,\ldots,C\}\),
and a linear classifier
\[
f(\mathbf{x})
=
\mathbf{W}_L\cdots\mathbf{W}_1\mathbf{x}
=
\mathbf{M}\mathbf{x}.
\]
For \(L\ge2\), all hidden layers have width \(H\):
\(\mathbf{W}_1\in\mathbb R^{H\times p}\),
\(\mathbf{W}_L\in\mathbb R^{C\times H}\), and
\(\mathbf{W}_\ell\in\mathbb R^{H\times H}, 1<\ell<L\).
For \(L=1\), the end-to-end predictor
\(\mathbf{M}\in\mathbb R^{C\times p}\) is optimized directly.
The weights are initialized independently with
\(\mathcal N(0,\sigma^2)\) entries and trained by gradient flow on
\(\mathcal L(\mathbf{M})
=
(2N)^{-1}\sum_i
\|\mathbf{M}\mathbf{x}_i-\mathbf{y}_i\|_2^2\).
Although cross-entropy is more common for classification, we use mean squared loss for
analytical tractability.

Our main object is the training accuracy
\begin{equation}
\Acc(\mathbf{M})
=
\frac1N\sum_{i=1}^N
\mathbf{1}\!\left\{
(\mathbf{M}\mathbf{x}_i)_{c_i}
>
\max_{c\neq c_i}(\mathbf{M}\mathbf{x}_i)_c
\right\}.
\label{eq:acc-main}
\end{equation}
We focus on small initialization limit, i.e. we first take \(\sigma\downarrow0\), followed by the appropriate
large-system limit; precise orders of limits are stated in the corresponding
theorems.
Throughout this section we assume isotropic input covariance \(\frac1N\sum_{i=1}^N
\mathbf{x}_i\mathbf{x}_i^{\top}
= \gamma
\mathbf{I}_p,\) and
define the effective target
\begin{equation}
\mathbf{T}
=
\frac{1}{N\gamma}\sum_{i=1}^N
\mathbf{y}_i\mathbf{x}_i^{\top},
\label{eq:main-linear-target}
\end{equation}
so the loss can be expressed as \(\mathcal L(\mathbf{M})
=
\frac{\gamma}{2}\|\mathbf{M}-\mathbf{T}\|_{\mathrm F}^2+\mathrm{const}.\)
We will work with rescaled time $\tau=\gamma t$ throughout the section. We consider SVD and the truncated SVD, respectively:
\begin{equation}
\mathbf{T}
=
\sum_{j=1}^r
s_j\mathbf{p}_j\mathbf{q}_j^{\top},
\qquad
s_1\ge\cdots\ge s_r>0,
\qquad
\mathbf{T}_k
:=
\sum_{j=1}^k
s_j\mathbf{p}_j\mathbf{q}_j^{\top}.
\label{eq:main-linear-svd}
\end{equation}
Let \(k_\star\) be the multiplicity of the largest singular value,
\(
s_1=\cdots=s_{k_\star}>s_{k_\star+1},~
s_{r+1}:=0,
\)
so \(\mathbf{T}_{k_\star}\) is the complete top singular component and is
independent of the choice of singular-vector basis within the degenerate top
subspace. We consider two accuracy levels:
\begin{equation}
A_{\mathbf T}:=\Acc(\mathbf{T}),
\qquad
A_{k_\star}:=\Acc(\mathbf{T}_{k_\star}).
\label{eq:main-linear-accuracy-levels}
\end{equation}
Note that both cases \(A_{\mathbf T}>A_{k_\star}\) and \(A_{\mathbf T}<A_{k_\star}\) are possible
(see \cref{app:ordering}), although the former is observed consistently
across the standard image datasets in
Table~\ref{tab:linear-L2-natural-ordering} in Appendix~\ref{app:ordering}.
For \(L=2\), we show that the microscopic dynamics can pass from the
classifier of \(\mathbf{T}\) to that of
\(\mathbf{T}_{k_\star}\). When \(k_\star=1\), this is the rank-one component
\(\mathbf{T}_1=s_1\mathbf{p}_1\mathbf{q}_1^{\top}\).
Our \(L\ge3\) saddle-escape result requires this simple-leading-mode case,
\(k_\star=1\).
Table~\ref{tab:linear-summary} in Appendix~\ref{app:linear}
summarizes the results and their precise regimes.

\paragraph{Direct linear model: \(L=1\)
[\hyperref[app:linear-L1]{\ref*{app:linear-L1}}].}
We begin with direct optimization of the end-to-end predictor
\(\mathbf M\).  Its gradient-flow trajectory is explicit:
\begin{equation}
\mathbf M(\tau)
=
e^{-\tau}\mathbf M(0)
+
(1-e^{-\tau})\mathbf T.
\label{eq:main-linear-direct-flow}
\end{equation}
Thus \(\mathbf M(\tau)\) moves along the line segment from its
initialization to the least-squares solution \(\mathbf T\).  Along this
trajectory, agreement with the classifier induced by \(\mathbf T\) is
monotone on every training example: once
\(\mathbf M(\tau)\) and \(\mathbf T\) assign the same top-scoring class to
an example, they continue to agree on that example at all later times (Lemma~\ref{lem:linear-L1-monotone-agreement}).
This yields a direct bound on any possible nonmonotonicity of the
training accuracy.

\begin{linkedtheorem}
{Accuracy well bound for \(L=1\)}
{thm:linear-L1-accuracy-bound}
\label{thm:main-linear-L1-accuracy-bound}
Suppose that \(\mathbf T\) has a unique top-scoring class on every
training example and
\(
\Acc(\mathbf T)=1-\varepsilon.
\)
Then, for every \(0\le\tau_1\le\tau_2\),
\begin{equation}
\Acc(\mathbf M(\tau_2))
\ge
\Acc(\mathbf M(\tau_1))
-
\varepsilon.
\label{eq:main-linear-L1-drawdown}
\end{equation}
In particular, if \(\Acc(\mathbf T)=1\), then
\(\Acc(\mathbf M(\tau))\) is monotone nondecreasing.
\end{linkedtheorem}

Furthermore, under vanishing initialization, any nontrivial classifier dynamics for
\(L=1\) are confined to the microscopic regime.  For
\(\mathbf M(0)=\sigma\mathbf Z\), all classifier changes occur while
\(\|\mathbf M(\tau)\|_{\mathrm F}=O(\sigma)\), and by the time the
classifier stabilizes only an \(O(\sigma)\) fraction of the total loss
decrease from initialization to the least-squares solution has occurred
(Theorem~\ref{thm:linear-L1-boundary-layer}).  After this initial interval,
the classifier agrees permanently with that of \(\mathbf T\).
Thus the direct linear model provides a sharply constrained baseline.  As
we show next, factorized linear networks with \(L\ge2\) can exhibit
qualitatively richer microscopic classifier dynamics, including a
target-driven rise followed by a degeneracy bottleneck before the later
macroscopic recovery.

\paragraph{Two-layer networks: \(L=2\)
[\hyperref[app:linear-L2]{\ref*{app:linear-L2}}].}
We consider
\(\mathbf M=\mathbf W_2\mathbf W_1\). Gradient flow satisfies
\begin{equation}
\dot{\mathbf W}_1
=
\mathbf W_2^{\top}\mathbf T
-
\mathbf W_2^{\top}\mathbf W_2\mathbf W_1,
\qquad
\dot{\mathbf W}_2
=
\mathbf T\mathbf W_1^{\top}
-
\mathbf W_2\mathbf W_1\mathbf W_1^{\top}.
\label{eq:main-linear-L2-flow}
\end{equation}
The cubic terms vanish as \(\sigma\downarrow0\) on fixed time intervals, leaving a linear limiting dynamics. Integrating the linear limiting dynamics, consider the normalized mean predictor
\begin{equation}
\overline{\mathbf M}(\tau)
:=
\lim_{\sigma\downarrow0}
H^{-1} \sigma^{-2}\,
\mathbb E\!\left[\mathbf M(\tau)\right]
=
\sum_{j=1}^r
\sinh(2s_j\tau)\,
\mathbf p_j\mathbf q_j^{\top},
\label{eq:main-linear-L2-microscopic-predictor}
\end{equation}
At early times,
\(\overline{\mathbf M}(\tau)=2\tau\mathbf T+O(\tau^3)\), while finite width
introduces initialization-dependent fluctuations. Schematically, at the level of individual classification scores,
\begin{equation}
\mathbf M(\tau)
\approx
2H\sigma^2\tau
\left[
\mathbf T
+
O_{\mathbb P}\!\left(H^{-1/2}\tau^{-1}\right)
+
O(\tau^2)
\right].
\label{eq:main-linear-L2-early-target}
\end{equation}
Thus, for sufficiently large width, there is an early regime in which
\(\tau\) is large enough for the target-driven component to dominate the
random fluctuations, yet small enough for the higher-order dynamical correction to remain negligible. We call a time in this regime
\(\tau_{\mathrm{org}}\), the \emph{organization time}; the predictor then
induces the classifier of \(\mathbf T\), with accuracy \(A_{\mathbf T}\),
while remaining microscopic.

At later microscopic times, the factors \(\sinh(2s_j\tau)\) increasingly
favor the largest singular modes. Hence, at a suitable
\emph{bottleneck time} \(\tau_{\mathrm{bot}}>\tau_{\mathrm{org}}\),
\begin{equation}
\mathbf M(\tau_{\mathrm{bot}})
\approx
H\sigma^2s_1^{-1}\sinh(2s_1\tau_{\mathrm{bot}})
\,\mathbf T_{k_\star},
\label{eq:main-linear-L2-bottleneck}
\end{equation}
with the slower modes and finite-width fluctuations negligible relative to
the dominant block. The predictor therefore induces the classifier of
\(\mathbf T_{k_\star}\), with accuracy \(A_{k_\star}\). Thus target-driven
organization is followed by a rank-\(k_\star\) degeneracy bottleneck, while
the predictor norm and loss change remain vanishing in the
small-initialization limit. At later times the full gradient flow converges
to \(\mathbf T\), restoring the target classifier.

To make it rigorous, we consider a large-system sequence indexed by \(H\to\infty\), where
\(p\), \(C\), and \(N\) may also depend on \(H\) and grow sufficiently
mildly. We assume bounded input norms and
\(\|\mathbf T\|_{\mathrm{op}}\), sufficient width
\(H\ge\operatorname{rank}(\mathbf T)\), a nonvanishing spectral gap between
the top singular block and the remaining modes, and decision margins of
\(\mathbf T\) and \(\mathbf T_{k_\star}\) bounded away from zero.
These uniform margin assumptions can be restrictive in growing-dimensional
settings, particularly when the rank \(k_\star\) of the top singular block
remains fixed; the more quantitative margin-dependent conditions underlying
the result are stated in Appendix~\ref{app:linear-L2}.

\begin{linkedtheorem}
{Three-stage accuracy dynamics for \(L=2\)}
{thm:linear-L2-three-stage-accuracy-dynamics}
\label{thm:main-linear-L2-accuracy-dynamics}
There exist deterministic times
\(\tau_{\mathrm{org}}<\tau_{\mathrm{bot}}\), which may depend on
\(N,p,C,H\) and the dataset \(\{(\mathbf{x}_i,\mathbf{y}_i)\}_{i=1}^N\) but not on \(\sigma\) or on the random
initialization, such that, in the sequential limit
\(\sigma\downarrow0\) followed by \(H\to\infty\),
\begin{equation}
\mathbb P\!\left(
\Acc(\mathbf{M}(\tau_{\mathrm{org}}))=A_{\mathbf T},\;
\Acc(\mathbf{M}(\tau_{\mathrm{bot}}))=A_{k_\star}
\right)
\longrightarrow1.
\label{eq:main-linear-L2-two-stages}
\end{equation}

Moreover, for each fixed member of the sequence, almost surely as
\(\sigma\downarrow0\), the entire trajectory up to the bottleneck time
remains microscopic:
\begin{equation}
\sup_{0\le \tau\le\tau_{\mathrm{bot}}}
\|\mathbf M(\tau)\|_{\mathrm{op}}
\longrightarrow0,
\qquad
\sup_{0\le \tau\le\tau_{\mathrm{bot}}}
\left|
\mathcal L(\tau)-\mathcal L(0)
\right|
\longrightarrow0.
\label{eq:main-linear-L2-microscopic-stages}
\end{equation}
\end{linkedtheorem}

Whenever \(A_{\mathbf T}>A_{k_\star}\), these stages establish an accuracy well:
an early target-level classifier, a degeneracy bottleneck, and
eventual permanent recovery.

\paragraph{Unary operations.}
We now specialize the two-layer model to a unary operation
\(f:[p]\to[p]\), represented by one-hot inputs and outputs:
\(N=C=p\), \(\mathbf x_i=\mathbf e_i\), and
\(\mathbf y_i=\mathbf e_{f(i)}\). Let
\(n_c:=|\{i:f(i)=c\}|\) denote the occupancy of output \(c\). Then
\begin{equation}
\mathbf T = \sum_{i=1}^p
\mathbf e_{f(i)}\mathbf e_i^{\top}, \qquad
\mathbf T\mathbf e_i=\mathbf e_{f(i)},
\qquad
\mathbf T\mathbf T^{\top}
=
\operatorname{diag}(n_1,\ldots,n_p).
\label{eq:main-unary-target-structure}
\end{equation}
Thus \(\mathbf T\) has perfect accuracy, and its nonzero singular values are
exactly \(\sqrt{n_c}\). The occupancy structure therefore determines the
microscopic growth rates exactly.

Consider first a random unary operation,
\(f(1),\ldots,f(p)\stackrel{\mathrm{iid}}{\sim}\operatorname{Unif}[p]\).
The occupancies are asymptotically Poisson. This setting does not satisfy the nonvanishing decision-gap
assumption for the top singular block in
Theorem~\ref{thm:main-linear-L2-accuracy-dynamics}: on every submaximal
occupancy fiber, \(\mathbf T_{k_\star}\) vanishes and hence has zero decision
margin. Consequently, the collapse is not visible in the mean predictor
\eqref{eq:main-linear-L2-microscopic-predictor} alone: for every finite
\(\tau>0\), its nonzero entry in each column remains on the correct output, so
its accuracy is still one. The bottleneck is instead a finite-width effect.
Modes associated with larger occupancies grow faster, and on examples
associated with slower modes the corresponding mean margin becomes
asymptotically negligible relative to the dominant modes, allowing amplified
finite-width fluctuations to determine the predicted class. As the problem
size grows, this mechanism drives the bottleneck accuracy to zero; the
detailed occupancy argument is given in Appendix~\ref{app:linear-L2}.

\begin{linkedtheorem}
{Random unary operation}
{thm:linear-L2-random-unary}
\label{thm:main-random-unary}
For every \(\alpha>0\), there exist deterministic times
\(\tau_{\mathrm{org}}<\tau_{\mathrm{bot}}\), which may depend on \(p\) and \(H\), but not on \(\sigma\), the random operation, or the random
initialization, such that, in the sequential limit
\(\sigma\downarrow0\) followed by
\(p,H\to\infty\) with \(H/p\to\alpha\),
\begin{equation}
\Acc\!\left(\mathbf M(\tau_{\mathrm{org}})\right)\longrightarrow1,
\qquad
\Acc\!\left(\mathbf M(\tau_{\mathrm{bot}})\right)\longrightarrow0
\label{eq:main-random-unary-rise-drop}
\end{equation}
in probability.
Moreover, for each fixed \(p,H\), almost surely as
\(\sigma\downarrow0\), the entire trajectory up to the bottleneck remains
microscopic:
\begin{equation}
\sup_{0\le\tau\le\tau_{\mathrm{bot}}}
\|\mathbf M(\tau)\|_{\mathrm{op}}
\longrightarrow0,
\qquad
\sup_{0\le\tau\le\tau_{\mathrm{bot}}}
\left|\mathcal L(\tau)-\mathcal L(0)\right|
\longrightarrow0.
\label{eq:main-random-unary-small-predictor}
\end{equation}

If, in addition, \(\alpha>1-e^{-1}\), then with probability tending to one
over the random operation, for every fixed \(\sigma>0\) there almost surely
exists a finite, possibly \(\sigma\)- and realization-dependent recovery time
\(\tau_{\mathrm{rec}}>\tau_{\mathrm{bot}}\), such that
\begin{equation}
\Acc(\mathbf M(\tau))=1
\qquad
\text{for all }\tau\ge\tau_{\mathrm{rec}}.
\label{eq:main-random-unary-recovery}
\end{equation}
\end{linkedtheorem}

The random-unary well is driven by heterogeneity of the output occupancies, which produces different singular values and hence different microscopic growth rates. It is therefore natural to ask what happens when this source of spectral separation is removed. A permutation provides the opposite extreme: \(n_c=1\) for every \(c\), so all singular values of \(\mathbf T\) are equal to one. There is consequently no spectral hierarchy capable of producing the bottleneck above. In fact, the behavior is stronger: the accuracy is nondecreasing, with all correct-versus-incorrect margins increasing until perfect accuracy is reached and then retained permanently.

\begin{linkedtheorem}
{Permutation operation: monotone accuracy}
{thm:linear-L2-permutation}
\label{thm:main-permutation-unary}
For every \(\alpha>0\), there exists a deterministic time
\(\tau_{\mathrm{org}}\), which may depend on \(p\) and \(H\), but not
on \(\sigma\), the permutation, or the random initialization, such that, in the sequential limit \(\sigma\downarrow0\) followed by
\(p,H\to\infty\) with \(H/p\to\alpha\), with probability tending to one,
\begin{equation}
\Acc(\mathbf M(\tau))
\quad\text{is nondecreasing for all }\tau\ge0,
\qquad
\Acc(\mathbf M(\tau))=1
\quad
\text{for all }\tau\ge\tau_{\mathrm{org}}.
\label{eq:main-permutation-monotone}
\end{equation}
Moreover, for each fixed \(p,H\), almost surely as \(\sigma\downarrow0\),
\begin{equation}
\sup_{0\le\tau\le\tau_{\mathrm{org}}}
\|\mathbf M(\tau)\|_{\mathrm{op}}
\longrightarrow0,
\qquad
\sup_{0\le\tau\le\tau_{\mathrm{org}}}
\left|
\mathcal L(\tau)-\mathcal L(0)
\right|
\longrightarrow0.
\label{eq:main-permutation-early-flat-loss}
\end{equation}
\end{linkedtheorem}

Thus heterogeneous occupancies produce a nearly complete accuracy well,
whereas the flat spectrum of a permutation eliminates the bottleneck
altogether. Remarkably, permanent perfect classification for a permutation
holds even for \(\alpha<1\), where
\(\operatorname{rank}(\mathbf M)\le H<p=\operatorname{rank}(\mathbf T)\)
and exact fitting is impossible.

\paragraph{Deeper networks: \(L\ge3\)
[\hyperref[app:linear-deep]{\ref*{app:linear-deep}}].}
For generic Gaussian initialization, we combine the same target-driven
organization mechanism as for \(L=2\) with the rank-one saddle-escape result
of \citet{jacot2021saddle}. At an early stage, the predictor develops a
coherent component aligned with the full target \(\mathbf T\), while
finite-width fluctuations self-average, so the classifier of \(\mathbf T\)
emerges while the predictor norm and the change in loss remain asymptotically
vanishing. At a later stage, when \(s_1>s_2\), the trajectory escapes the
initial rank-zero saddle through the leading rank-one singular component,
producing the classifier of the leading rank-one component
\(\mathbf T_1\).

To make this rigorous, we consider fixed \(L\ge3\) and a large-system
sequence indexed by \(H\to\infty\). We assume bounded input norms and
\(\|\mathbf T\|_{\mathrm{op}}\), a nonvanishing decision margin for
\(\mathbf T\), unique top-scoring classes for \(\mathbf T_1\) on each
finite problem, an isolated leading singular value \(s_1>s_2\), and
sufficiently mild growth of the problem dimensions:
\(p/H=O(1)\), \(C/H=O(1)\), and
\(\log^{L-1}(2NCH)=o(\sqrt H)\).

\begin{linkedtheorem}
{Target-driven organization and predictive bottleneck for \(L\ge3\)}
{thm:linear-deep-accuracy-well}
\label{thm:main-linear-deep-accuracy-well}
Under the assumptions above, there exist times
\(
\tau_{\mathrm{org}}<\tau_{\mathrm{bot}},
\)
where \(\tau_{\mathrm{bot}}\) may depend on the initialization, such that the following holds. In the sequential limit \(\sigma\downarrow0\) followed by
\(H\to\infty\), with probability tending to one for \(L=3\), and with
asymptotic probability at least \(1/2\) for \(L>3\), we have
\begin{equation}
\Acc\!\left(\mathbf M(\tau_{\mathrm{org}})\right)=A_{\mathbf T},
\qquad
\Acc\!\left(\mathbf M(\tau_{\mathrm{bot}})\right)=A_1.
\label{eq:main-deep-two-stages}
\end{equation}
Moreover, in the same sequential limit, uniformly for
\(0\le\tau\le\tau_{\mathrm{bot}}\), the predictor norm vanishes and the
loss converges to its value at the zero predictor.
\end{linkedtheorem}

\paragraph{Spectral-mode recruitment and recovery under aligned initialization
[\hyperref[app:recovery-L3]{\ref*{app:recovery-L3}}].}
For generic Gaussian initialization, our theory rigorously establishes the
first two stages---target-driven organization followed by a rank-one
bottleneck---but recovery of the same trajectory remains open, since
available convergence results do not exclude lower-rank limiting critical
points \citep{bah2022learning,nguegnang2024convergence}. To isolate the
recovery mechanism, we therefore consider the spectrally aligned and balanced
dynamics of \citet{saxe2014exact}, where
\begin{equation}
\mathbf M(\tau)
=
\sum_{k=1}^r m_k(\tau)\mathbf p_k\mathbf q_k^{\top},
\qquad
\dot m_k
=
L(s_k-m_k)m_k^{2-2/L}.
\label{eq:main-deep-aligned-modes}
\end{equation}
The leading mode can dominate transiently, producing the same low-rank
bottleneck, but its growth slows as \(m_1\to s_1\), while the remaining
modes are progressively recruited. Eventually
\(\mathbf M(\tau)\to\mathbf T\), and the target classifier is recovered
permanently. Corollary~\ref{cor:linear-deep-aligned-recovery} makes this
precise: whenever the aligned trajectory enters the leading-mode bottleneck,
its accuracy equals \(A_1\) there and subsequently returns permanently to
\(A_{\mathbf T}\). Hence, if \(A_{\mathbf T}>A_1\), the aligned dynamics
rigorously exhibit the bottleneck-and-recovery part of an accuracy well.
Together with the generic-initialization result above, this establishes the
three proposed stages in complementary rigorous settings, while a complete
three-stage theorem for one generic Gaussian-initialized \(L\ge3\) trajectory
remains open.

\section{Discussion}
\label{sec:discussion}

\paragraph{Accuracy need not mirror optimization progress.}
Training loss decreases monotonically under gradient flow, yet classification
accuracy can follow a qualitatively different trajectory. We prove that a
high-accuracy classifier can emerge while both the predictor and loss change
remain microscopic, and can subsequently be destroyed by continued
optimization before recovering. In the random-unary model this separation is
extreme: accuracy rises asymptotically to \(100\%\), collapses to \(0\), and
later recovers permanently to \(100\%\). Our deep-linear analysis therefore
reveals predictive structure that appears before the familiar rank-one saddle
escape and before appreciable progress in the optimized objective.

\paragraph{Limitations and future directions.}
Our rigorous analysis is restricted to linear networks, whereas the
experiments show pronounced accuracy wells also in nonlinear models on
MNIST and CIFAR-10. Even within linear networks, the pronounced well observed
for deep permutation targets is not covered by our \(L\ge3\) theory, which
assumes an isolated leading singular value. Understanding these flat-spectrum
deep dynamics, identifying the mechanisms responsible for the well in
nonlinear networks, and determining whether the
organization--bottleneck--recruitment picture extends to broader
architectures are natural directions for future work.

\subsection*{AI use statement}

Generative AI tools were used extensively throughout this work to assist with
research brainstorming, the development and refinement of theoretical ideas
and hypotheses, mathematical derivations and proof development, the design
and analysis of numerical experiments, interpretation of results, and
literature search. They were also used to assist with experimental code,
figure generation, and the drafting, revision, organization, and presentation
of the manuscript.

All AI-assisted mathematical arguments were treated as provisional and were
independently checked by the authors, including their assumptions,
derivations, theorem statements, and proofs. AI-assisted code and numerical
results were likewise inspected and validated, and references identified with
the help of AI tools were checked against the original sources before
inclusion. The authors reviewed all AI-assisted material and take full
responsibility for the mathematical claims, experimental results, citations,
and conclusions presented in this work.

\subsubsection*{Acknowledgments}
The research was supported by the Russian Science Foundation grant No.~25-11-00355,\\\url{https://rscf.ru/project/25-11-00355/}.

\bibliography{iclr2027_conference}
\bibliographystyle{iclr2027_conference}

\newpage

\tableofcontents

\newpage

\appendix

\section{Related Work}
\label{sec:related}

\subsection{Small initialization and stagewise dynamics in linear networks}

Deep linear networks provide a classical setting in which gradient-based
learning separates into distinct dynamical stages. In the spectrally aligned
setting, \citet{saxe2014exact} derive exact gradient-flow solutions in which
different singular modes are learned on different time scales, producing
plateaus and rapid transitions. Building on these dynamics,
\citet{saxe2019semantic} show that unequal mode-learning time scales can also
produce non-monotone prediction-level behavior, including transient
``illusory correlations'' in which individual item--property predictions
first strengthen and later reverse. This provides a particularly close
precedent for the predictive effects studied here. More general exact
analyses of two-layer linear networks extend this picture beyond specially aligned
initializations and characterize the accompanying evolution of
representations and alignment \citep{domine2023exact}. These works establish spectral organization as a central description of
deep-linear learning.  We study a complementary question that is not captured by the spectral trajectories alone: how those dynamics organize the induced multiclass classifier.  In particular, we show that an accurate class ordering can emerge at microscopic predictor scale, be destroyed by subsequent concentration onto dominant spectral directions, and later recover as additional directions are recruited.

Small or infinitesimal initialization makes this stagewise structure
particularly sharp.  \citet{gidel2019implicit} show that
vanishing-initialization two-layer linear network dynamics can successively approximate
reduced-rank regression solutions of increasing rank, while
\citet{gissin2020implicit} formalize incremental learning in deep linear
networks and show that depth substantially relaxes the initialization-scale
requirements needed for successive learning of increasingly complex
solutions.  In matrix factorization, \citet{li2021greedy} connect
infinitesimal-initialization gradient flow to a greedy low-rank learning
procedure.  Related results show sequential recovery of principal components
under small random initialization \citep{jiang2023algorithmic}, and
\citet{chou2024gradient} rigorously identify intermediate time intervals in
deep matrix factorization where the iterate has effective rank close to that
of low-rank projections of the target.  Thus, in linear and matrix-factorization settings, preferential growth of spectral
components and passage through effectively low-rank predictors are
well-established features of small-initialization dynamics.

Our results reveal a classifier-level stage that precedes this familiar
spectral concentration.  For deep linear networks, we show that sufficiently
early microscopic dynamics can already produce the same classifier as the
full least-squares target, i.e., the global minimizer of the end-to-end MSE
objective, while both the predictor norm and the change in loss remain
asymptotically vanishing.  We refer to this early stage as
\emph{target-driven organization}.  Subsequent dynamics can then concentrate onto a smaller set of dominant spectral directions and destroy this already formed class ordering.  We refer to this transient loss of predictive
diversity as a \emph{degeneracy bottleneck}.  In this sense, our results complement the incremental-learning literature by resolving the dynamics at the level of aggregate multiclass accuracy and by showing that spectral
concentration can temporarily degrade a classifier that had already attained the accuracy of the full least-squares solution.

The saddle-to-saddle viewpoint provides another closely related description
of small-initialization dynamics.  For deep linear networks,
\citet{jacot2021saddle} characterize the first escape from the rank-zero
saddle and conjecture a subsequent sequence of increasing-rank saddles.
More recent work develops this picture across a broader class of neural
architectures \citep{zhang2026saddle}.  Our deep-linear result fits naturally
into this framework but identifies an additional, earlier predictive stage.
Before the leading rank-one mode governs the saddle escape, target-driven
organization can already establish the classifier of the full target at
microscopic scale. In our setting, the subsequent rank-one escape can therefore act as a degeneracy bottleneck, replacing an already accurate classifier by the
classifier of the leading singular component.  Thus, from the viewpoint of classification, escape from the origin need not mark the first emergence of useful predictive structure.

The final recovery stage raises a separate convergence question.  In the two-layer case, available convergence theory implies that for almost every
initialization gradient flow reaches a global minimizer
\citep{bah2022learning}.  We use this endpoint result to turn the microscopic target-driven-organization--bottleneck transition into a complete
rise--collapse--recovery trajectory: after spectral concentration degrades
the classifier, the remaining target directions are eventually recovered and the full target classifier is restored permanently.  At greater depth, gradient-flow convergence results identify a limiting predictor that is
globally optimal within a manifold of some rank \(k\), but do not determine
that limiting rank in general \citep{bah2022learning}; an analogous
limitation holds for gradient descent \citep{nguegnang2024convergence}.
Accordingly, our generic \(L\ge3\) theory establishes target-driven
organization and the subsequent degeneracy bottleneck, while the later recruitment-and-recovery mechanism is proved separately under the spectrally aligned dynamics of \citet{saxe2014exact}.

\subsection{Vanishing initialization and early feature dynamics beyond linear models}

Small initialization also plays a fundamental role in nonlinear networks.
The lazy-training analysis of \citet{chizat2019lazy} and the kernel--rich
regimes of \citet{woodworth2020kernel} show that initialization scale and parameterization can determine whether training remains close to a fixed
linearization or instead develops substantial feature learning.  Our use of
vanishing initialization has a different emphasis.  A small predictor norm
does not by itself imply lazy dynamics; rather, we exploit the early
small-output regime to resolve changes in class ordering that can occur while
the predictor and the loss remain nearly unchanged.

A growing literature shows that substantial task-dependent organization can
already occur in this early regime.  \citet{maennel2018quantizes} describe
the tendency of ReLU neurons at small initialization to concentrate along a
small set of data-dependent directions.  \citet{atanasov2022silent}
identify \emph{silent alignment}, in which the directional structure of the
neural tangent kernel becomes increasingly adapted to the task before
appreciable loss decay, and demonstrate related behavior in nonlinear
networks.  Rigorous analyses of shallow ReLU networks likewise establish
early directional alignment and staged dynamics under small initialization
and structured data assumptions
\citep{boursier2022gradient,min2024early}.  These results show that
substantial task-dependent organization can emerge before ordinary loss
convergence.

This behavior is closely related to what we call \emph{target-driven
organization}. Our emphasis, however, is on the classifier induced by this
early structure. In our linear theory, target-driven organization can
already establish the classifier of the full target while both the predictor
and the loss change remain asymptotically vanishing. Subsequent dynamics can
then temporarily destroy this class ordering rather than merely refine it.
This distinction is also conceptually related to
\citet{boursier2025early}, who show that early alignment can strongly restrict the represented directions and, in suitable settings, steer an overparameterized
two-layer ReLU network toward a suboptimal stationary point rather than a global minimizer. Relatedly, \citet{pezeshki2021gradient} identify
\emph{gradient starvation}, in which optimization preferentially captures a
subset of predictive features while other useful features remain
underlearned. This feature imbalance is close in spirit to our degeneracy
bottleneck, although the mechanisms and dynamical claims are distinct.
A different notion of bottleneck appears in \citet{jacot2023bottleneck},
who show that large-depth nonlinear networks under norm regularization can
favor low-dimensional hidden representations.  Their bottleneck is thus a structural property of intermediate feature
representations, rather than the transient concentration of the end-to-end predictor studied here.

Recent work further shows that stagewise dynamics beyond the initial
alignment phase are not specific to linear factorization.
\citet{kumar2024directional} show that, near small initialization, parameter
directions align toward task-dependent limiting directions, and establish an
analogous directional-alignment phenomenon near certain structured saddles
for two-homogeneous neural networks,
while \citet{kumar2025beyond} follow homogeneous gradient flow beyond its
escape from the origin and characterize, under suitable assumptions, the
subsequent saddle reached by the trajectory.  For deep ReLU networks,
\citet{bantzis2026saddle} characterize optimal escape directions from the
origin and show that they exhibit a pronounced low-rank bias in the deeper
layers, motivating a corresponding saddle-to-saddle picture.  \citet{zhang2026saddle} provide a
broader account of progressively more complex saddle-to-saddle dynamics
across several neural architectures.

Together, these works establish that directional selection, low-rank
structural bias, and saddle-to-saddle progression extend well beyond deep
linear networks.  Our organization--bottleneck--recruitment picture concerns
a complementary classifier-level question: whether early target-driven
organization produces a useful class ordering, whether subsequent
concentration of the dynamics onto a smaller effective set of predictive
directions can destroy that ordering, and whether later recruitment of
additional directions can restore it.  Our linear theory proves concrete
instances of this mechanism.  For the nonlinear networks studied
experimentally here, we regard this broader interpretation as a hypothesis
rather than a consequence of existing feature-alignment or saddle-to-saddle
theory.

\subsection{Accuracy dynamics and neighboring non-monotone phenomena}
\label{sec:related_accuracy}

Non-monotone classification behavior during training has several important
precedents, although the observables and mechanisms considered in prior work
differ from the aggregate training-accuracy dynamics studied here.

At the level of individual examples, \citet{toneva2019forgetting} show that
training examples can undergo \emph{forgetting events}, changing from correct
to incorrect classification during ordinary single-task training.  This
establishes that correctness need not evolve monotonically even when the
training set and objective are fixed.  Our object is different: rather than
tracking individual examples, we study a coherent non-monotone trajectory of
the aggregate classifier, in which a large fraction of the training set can
become correctly classified, subsequently lose that classification, and later
recover.

A closer analogue is provided by \citet{schneider2025learning}. They show
empirically that geometrically selected atypical subsets can lose accuracy
during training and later recover, and analyze the associated competition
between typical and atypical examples through their losses and learning
dynamics. Their theoretical results do not establish a non-monotone
trajectory of the aggregate top-one training accuracy itself.  Our mechanism
is also different: in our linear theory, target-driven organization can first
establish the classifier of the full target at microscopic predictor scale,
after which unequal growth of predictive directions can concentrate the
predictor onto a smaller effective subspace and produce a \emph{degeneracy
bottleneck}.  Thus the deterioration occurs through changes in the spectral
composition of the predictor rather than through competition between
different subsets of examples.

Another neighboring phenomenon is epoch-wise double descent, in which test or
generalization error decreases, subsequently rises, and then decreases again
as training proceeds \citep{nakkiran2020deep}.  Analytical accounts relate
this behavior to different parts of the model being learned on different time scales
\citep{heckel2021early}.  Particularly close mathematically is
\citet{olmin2024towards}, who analyze epoch-wise double descent in two-layer
linear neural networks and derive necessary conditions for non-monotone
generalization MSE from decoupled spectral-mode dynamics.  These mechanisms
are related to the unequal spectral growth that appears in our linear analysis, but the observables are different: epoch-wise double descent
concerns generalization loss or error, whereas we follow top-one class ordering directly.

A complementary line of work studies \emph{monotonic linear interpolation}
(MLI), namely the behavior along the line segment connecting an initialization
to a trained solution.  \citet{lucas2021monotonic} provide broad empirical
evidence for monotonic loss interpolation and derive sufficient conditions
under mean-squared error, while \citet{wang2023plateau} study both loss and
accuracy along interpolation paths and emphasize the role of biases and
extended plateaus.  Our \(L=1\) analysis intersects this literature in a
particularly rigid setting: under isotropic inputs, the actual gradient-flow
trajectory in predictor space is itself exactly the line segment from the
initialization to the least-squares predictor \(\mathbf T\), rather than a
post-hoc interpolation between two parameter vectors.  This exact structure
allows us to resolve the classifier along the optimization trajectory itself.
Agreement with the terminal least-squares classifier evolves monotonically,
which sharply constrains any possible accuracy well; when the terminal
classifier is perfect, non-monotone accuracy is ruled out altogether.  Under
vanishing initialization, all remaining classifier changes are confined to
the early microscopic regime in which the predictor and the loss have barely
changed.

Related work on stagewise prediction provides another useful contrast.
\citet{kalimeris2019sgd} find that the simple, approximately linear structure
that explains SGD's early predictions is largely retained as training
subsequently acquires more complex structure.  Our results establish the
complementary possibility: an early classifier can already be highly
accurate, yet continued optimization can temporarily destroy its class
ordering before later recovery.

Grokking provides a particularly relevant neighboring example of widely
separated predictive time scales.  In grokking, training performance
typically improves early while generalization emerges only much later
\citep{power2022grokking}, and theoretical work has shown that such delayed
transitions can be induced by distinct implicit biases operating at different
stages of training \citep{lyu2024dichotomy}.  The accuracy well differs in an
important way: it contains an intermediate deterioration of the training
classifier itself, and its early rise and collapse can both occur before
substantial progress in the optimized loss.  Thus the separation of time
scales concerns the evolution of class ordering within the training dynamics,
rather than only delayed generalization.

Thus our contribution is not the general observation that predictive
quantities can evolve non-monotonically; such behavior already appears, for
example, in the illusory-correlation dynamics of
\citet{saxe2019semantic}, as well as in the neighboring phenomena discussed
above. We identify a regime in which the \emph{aggregate training classifier}
undergoes a pronounced rise--collapse--recovery along the optimization
trajectory on a fixed training set,
with the first two stages occurring at microscopic predictor scale and
essentially constant loss. Our linear theory further identifies a concrete
mechanism for these stages: target-driven organization can first establish
the class ordering of the full target, after which unequal spectral growth can
produce a degeneracy bottleneck that destroys this ordering, followed
eventually by recruitment and recovery. The direct \(L=1\) dynamics provide
a complementary control: agreement with the terminal least-squares classifier
is monotone, any accuracy drawdown is bounded by its terminal error, and all
classifier changes under vanishing initialization are confined to an
\(O(\sigma)\) boundary layer. Factorization therefore does not merely change
the rate of approach to the same linear function class; it can enable
qualitatively richer aggregate classifier trajectories that are impossible,
or sharply restricted, under direct linear optimization.

\section[
Accuracy Dynamics in Deep Linear Networks under Vanishing Initialization
]{
Accuracy Dynamics in Deep Linear Networks under Vanishing Initialization
\texorpdfstring{[\hyperref[sec:linear]{Main Section~\ref*{sec:linear}}]}{}
}
\label{app:linear}

\begin{table}[t]
\centering
\small
\setlength{\tabcolsep}{3.5pt}
\renewcommand{\arraystretch}{0.95}
\caption{
Summary of the deep-linear accuracy theory.
``Microscopic'' means that the predictor norm and loss change vanish uniformly
up to the stated time. Recovery statements are for fixed \(\sigma>0\).
}
\label{tab:linear-summary}
\begin{tabular}{@{}p{0.065\linewidth}p{0.17\linewidth}p{0.72\linewidth}@{}}
\toprule
\textbf{Depth} & \textbf{Setting} & \textbf{Result} \\
\midrule

\(L=1\)
&
General
&
\(\exists\,\tau_\ast=O(\sigma)\) such that
\(\Acc(\mathbf M(\tau))=A_{\mathbf T}\) for all \(\tau\ge\tau_\ast\).
Any accuracy drawdown is at most \(1-A_{\mathbf T}\).
All classifier dynamics occur in a microscopic \(O(\sigma)\) boundary layer.
\\

\midrule

\multirow{3}{*}{\(L=2\)}
&
General
&
As \(\sigma\downarrow0\), then \(H\to\infty\),
\(\exists\,\tau_{\rm org}<\tau_{\rm bot}\) such that, with probability tending
to one,
\(\Acc(\mathbf M(\tau_{\rm org}))=A_{\mathbf T}\) and
\(\Acc(\mathbf M(\tau_{\rm bot}))=A_{k_\star}\).
The trajectory through \(\tau_{\rm bot}\) is microscopic; for fixed
\(\sigma>0\), it eventually recovers permanently to \(A_{\mathbf T}\).
\\

\cmidrule(lr){2-3}

&
Random unary,
\(H/p\to\alpha\)
&
As \(\sigma\downarrow0\), then \(p,H\to\infty\),
\(\exists\,\tau_{\rm org}<\tau_{\rm bot}\) such that
\(\Acc(\mathbf M(\tau_{\rm org}))\to1\) and
\(\Acc(\mathbf M(\tau_{\rm bot}))\to0\) in probability.
The trajectory is microscopic through \(\tau_{\rm bot}\).
If \(\alpha>1-e^{-1}\), fixed-\(\sigma\) recovery to \(1\) is permanent
with probability tending to one.
\\

\cmidrule(lr){2-3}

&
Permutation,
\(H/p\to\alpha\)
&
As \(\sigma\downarrow0\), then \(p,H\to\infty\), accuracy is
nondecreasing with probability tending to one and reaches \(1\) during the
microscopic regime, remaining \(1\) thereafter.
\\

\midrule

\multirow{2}{*}{\(L\ge3\)}
&
Gaussian,
\(s_1>s_2\)
&
As \(\sigma\downarrow0\), then \(H\to\infty\),
\(\exists\,\tau_{\rm org}<\tau_{\rm bot}\) such that
\(\Acc(\mathbf M(\tau_{\rm org}))=A_{\mathbf T}\) and
\(\Acc(\mathbf M(\tau_{\rm bot}))=A_1\).
The trajectory through \(\tau_{\rm bot}\) is microscopic.
The probability tends to \(1\) for \(L=3\), and is asymptotically at least
\(1/2\) for \(L>3\). Generic recovery remains open.
\\

\cmidrule(lr){2-3}

&
Spectrally aligned
&
If a leading-mode bottleneck occurs, its accuracy is \(A_1\).
Subsequent mode recruitment gives
\(\mathbf M(\tau)\to\mathbf T\), hence permanent recovery to
\(A_{\mathbf T}\).
Thus \(A_{\mathbf T}>A_1\) yields a bottleneck followed by recovery.
\\

\bottomrule
\end{tabular}
\end{table}

This appendix develops the linear-network theory underlying the results in
Section~\ref{sec:linear}. We track both the mean-squared training loss and the
classification accuracy along the learning trajectory, with particular emphasis
on the latter. This joint perspective is important because substantial changes
in classification accuracy can occur while the loss changes only negligibly.
In particular, we ask how the parameterization depth of a linear predictor can
change its accuracy dynamics, even though the realized function class remains
linear.

We consider a supervised classification dataset
\[
\{(\mathbf x_i, \mathbf y_i)\}_{i=1}^N,
\qquad
\mathbf x_i\in\mathbb R^p,
\qquad
\mathbf y_i= \mathbf e_{c_i}\in\mathbb R^C,
\qquad
c_i\in[C],
\]
where \(\mathbf e_{c_i}\) is the one-hot vector corresponding to the correct class
\(c_i\). A depth-\(L\) linear network has the form
\begin{equation}
f(\mathbf x)
=
\mathbf W_L \mathbf W_{L-1}\cdots \mathbf W_1 \mathbf x
=
\mathbf M \mathbf x,
\label{eq:linear-general-predictor}
\end{equation}
where
\begin{equation}
\mathbf M
=
\mathbf W_L \mathbf W_{L-1}\cdots \mathbf W_1
\in\mathbb R^{C\times p}
\label{eq:linear-general-end-to-end}
\end{equation}
is the end-to-end predictor. The weights $\mathbf W_{\ell}$ are trained by gradient flow on the
mean-squared error
\begin{equation}
\mathcal L
=
\frac{1}{2N}
\sum_{i=1}^N
\|\mathbf M \mathbf x_i-\mathbf y_i\|_2^2.
\label{eq:linear-general-loss}
\end{equation}
Although cross-entropy loss is more common for classification problems, we use MSE for analytical tractability.
Classification depends only on the ordering of the \(C\) output scores.
We consider the strict training accuracy
\begin{equation}
\Acc(\mathbf M)
=
\frac{1}{N}
\sum_{i=1}^N
\mathbf 1
\left\{
(\mathbf M \mathbf x_i)_{c_i}
>
\max_{c\neq c_i}(\mathbf M \mathbf x_i)_c
\right\},
\label{eq:linear-general-accuracy}
\end{equation}
so ties are counted as errors. Thus the loss and the accuracy probe very
different aspects of the same trajectory: the loss depends on the magnitudes
of the prediction errors, whereas the accuracy depends only on whether the
correct score exceeds every competing score. At small initialization, these
two observables can evolve on parametrically different scales. A predictor
whose norm is still vanishingly small can already have nontrivial or even
perfect accuracy, and subsequent changes in the relative strengths of its
predictive components can alter the class ordering while producing little
visible change in the loss.

Unless stated otherwise, the trainable weights are initialized independently
with centered Gaussian entries of variance \(\sigma^2\), and we are interested
in the small-initialization regime \(\sigma\downarrow0\). The end-to-end
predictor \(\mathbf x\mapsto\mathbf M\mathbf x\) remains linear at every
depth. However, for \(L\ge2\) with hidden width \(H\), the factorization
imposes \(\operatorname{rank}(\mathbf M)\le H\); hence the representable
function class coincides with that of the direct \(L=1\) model whenever
\(H\ge\min\{p,C\}\). Even when the representable function classes coincide,
the gradient-flow dynamics depends strongly on how \(\mathbf M\) is
parameterized. We therefore separate three regimes. For \(L=1\), the
end-to-end matrix is optimized directly and the dynamics can be solved
explicitly. For \(L=2\), factorization already creates nontrivial microscopic
spectral dynamics, which can be analyzed in considerable detail. For
\(L>2\), the dynamics is generally less tractable, but depth introduces an
additional mechanism for strongly separating the growth of different
predictive modes.

The two-layer case will be particularly important. Under isotropic inputs,
its microscopic dynamics can be diagonalized in the singular coordinates of
an effective target matrix. This separates two roles that will recur
throughout the analysis: singular values determine the rates at which
different modes are amplified, while singular vectors determine how those
modes affect the classification margins. Unary-operation learning provides
an especially tractable instance of this general theory: its target matrix
has explicit singular vectors and singular values, allowing the spectral
dynamics to be converted into precise statements about classification
accuracy. This leads both to rigorous accuracy wells for heterogeneous unary targets and to a complementary monotonicity result for flat-spectrum permutation targets.

\subsection{Direct linear optimization: \texorpdfstring{\(L=1\)}{L = 1}}
\label{app:linear-L1}

We first consider depth \(L=1\), where the end-to-end matrix
\(\mathbf M\) is optimized directly. Assume that the empirical input
second-moment matrix is isotropic,
\begin{equation}
\mathbf G
=
\frac{1}{N}\sum_{i=1}^N
\mathbf x_i\mathbf x_i^{\top}
=
\gamma \mathbf I_p,
\qquad
\gamma>0,
\label{eq:linear-L1-isotropy}
\end{equation}
and define the effective target matrix
\begin{equation}
\mathbf T
=
\frac{1}{N\gamma}
\sum_{i=1}^N
\mathbf y_i\mathbf x_i^{\top}.
\label{eq:linear-L1-target}
\end{equation}
Expanding the squared loss gives
\begin{equation}
\begin{aligned}
\mathcal L(\mathbf M)
&=
\frac{1}{2N}
\sum_{i=1}^N
\|\mathbf M\mathbf x_i-\mathbf y_i\|_2^2
\\
&=
\frac{\gamma}{2}\|\mathbf M\|_{\mathrm F}^2
-
\gamma\langle \mathbf M,\mathbf T\rangle_{\mathrm F}
+
\mathcal L_{\mathrm{const}}
\\
&=
\frac{\gamma}{2}\|\mathbf M-\mathbf T\|_{\mathrm F}^2
+
\mathcal L(\mathbf T),
\end{aligned}
\label{eq:linear-L1-loss}
\end{equation}
where the final term is independent of \(\mathbf M\). Thus
\(\mathbf T\) is the unique global minimizer of the squared loss over
unconstrained linear predictors.

After the deterministic time rescaling
\begin{equation}
\tau=\gamma t,
\label{eq:linear-L1-time-rescaling}
\end{equation}
gradient flow satisfies
\begin{equation}
\dot{\mathbf M}
=
\mathbf T-\mathbf M,
\qquad
\mathbf M(\tau)
=
e^{-\tau}\mathbf M(0)
+
(1-e^{-\tau})\mathbf T.
\label{eq:linear-L1-explicit-flow}
\end{equation}
Equation~\eqref{eq:linear-L1-explicit-flow} is the standard gradient-flow solution for least-squares regression. Under the isotropy assumption~\eqref{eq:linear-L1-isotropy}, the usual matrix-exponential trajectory reduces to a scalar interpolation between the initialization and the least-squares solution.
Thus direct linear optimization moves the predictor along the line
segment joining its initialization to the least-squares solution
\(\mathbf T\).

While the gradient-flow solution itself is standard, its form under isotropy places the \(L=1\) dynamics in direct connection with the literature on
monotonic linear interpolation (MLI)
\citep{lucas2021monotonic,wang2023plateau}.  Here, however, the
linear interpolation is not a post-hoc path between initialization and a
trained model: equation~\eqref{eq:linear-L1-explicit-flow} shows that it is
the exact gradient-flow trajectory in predictor space.  This additional
structure allows us to obtain pointwise guarantees for the classifier itself,
including quantitative control of both the depth and the time scale of any
possible nonmonotone accuracy transient.

The convergence
\(\mathbf M(\tau)\to\mathbf T\) alone is not sufficient to exclude
nonmonotone accuracy dynamics. In particular, under vanishing
initialization, a nonmonotone transient could in principle be confined
to a correspondingly vanishing time scale. We therefore first identify
a quantity whose dynamics are monotone at all times.

Suppose throughout this subsection that \(\mathbf T\) has a unique
top-scoring class on every training example, and define
\begin{equation}
\widehat c_i
=
\arg\max_{c\in[C]}
(\mathbf T\mathbf x_i)_c.
\label{eq:linear-L1-target-class}
\end{equation}
We measure agreement with the classifier induced by \(\mathbf T\) by
\begin{equation}
\operatorname{Agr}_{\mathbf T}(\mathbf M)
=
\frac{1}{N}
\sum_{i=1}^N
\mathbf 1
\left\{
(\mathbf M\mathbf x_i)_{\widehat c_i}
>
\max_{c\neq\widehat c_i}
(\mathbf M\mathbf x_i)_c
\right\}.
\label{eq:linear-L1-agreement}
\end{equation}
As in the definition of training accuracy, ties are counted as
disagreement.

\begin{lemma}[Monotone agreement with the least-squares classifier]
\label{lem:linear-L1-monotone-agreement}
For every training example \(i\), once the classifier induced by
\(\mathbf M(\tau)\) agrees with the classifier induced by
\(\mathbf T\), this agreement is never subsequently lost.
Consequently, for every \(0\le\tau_1\le\tau_2\),
\begin{equation}
\operatorname{Agr}_{\mathbf T}(\mathbf M(\tau_2))
\ge
\operatorname{Agr}_{\mathbf T}(\mathbf M(\tau_1)),
\label{eq:linear-L1-agreement-monotone}
\end{equation}
and
\begin{equation}
\lim_{\tau\to\infty}
\operatorname{Agr}_{\mathbf T}(\mathbf M(\tau))
=
1.
\label{eq:linear-L1-agreement-limit}
\end{equation}
\end{lemma}

\begin{proof}
Fix a training example \(i\), and suppose that at some time
\(\tau_1\) the classifier induced by \(\mathbf M(\tau_1)\) agrees with
that induced by \(\mathbf T\). Then, for every
\(c\neq\widehat c_i\),
\begin{equation}
(\mathbf M(\tau_1)\mathbf x_i)_{\widehat c_i}
-
(\mathbf M(\tau_1)\mathbf x_i)_c
>
0.
\label{eq:linear-L1-positive-current-margin}
\end{equation}
By the definition of \(\widehat c_i\),
\begin{equation}
(\mathbf T\mathbf x_i)_{\widehat c_i}
-
(\mathbf T\mathbf x_i)_c
>
0.
\label{eq:linear-L1-positive-target-margin}
\end{equation}

For every \(\tau_2\ge\tau_1\), the semigroup property of
\eqref{eq:linear-L1-explicit-flow} gives
\begin{equation}
\mathbf M(\tau_2)
=
e^{-(\tau_2-\tau_1)}
\mathbf M(\tau_1)
+
\left(
1-e^{-(\tau_2-\tau_1)}
\right)
\mathbf T.
\label{eq:linear-L1-semigroup}
\end{equation}
Hence, for each \(c\neq\widehat c_i\),
\begin{equation}
\begin{aligned}
&
(\mathbf M(\tau_2)\mathbf x_i)_{\widehat c_i}
-
(\mathbf M(\tau_2)\mathbf x_i)_c
\\
&\quad=
e^{-(\tau_2-\tau_1)}
\left[
(\mathbf M(\tau_1)\mathbf x_i)_{\widehat c_i}
-
(\mathbf M(\tau_1)\mathbf x_i)_c
\right]
\\
&\qquad+
\left(
1-e^{-(\tau_2-\tau_1)}
\right)
\left[
(\mathbf T\mathbf x_i)_{\widehat c_i}
-
(\mathbf T\mathbf x_i)_c
\right].
\end{aligned}
\label{eq:linear-L1-convex-margin}
\end{equation}
The right-hand side is a convex combination of two strictly positive
quantities and is therefore strictly positive. Thus, once agreement
with \(\mathbf T\) is achieved on example \(i\), it persists for all
later times. Summing the corresponding pointwise indicators proves
\eqref{eq:linear-L1-agreement-monotone}.

Finally, \(\mathbf M(\tau)\to\mathbf T\) as \(\tau\to\infty\).
Since \(\mathbf T\) has a unique top-scoring class on every training
example and the training set is finite, every example eventually has
the same predicted class under \(\mathbf M(\tau)\) and \(\mathbf T\).
This proves \eqref{eq:linear-L1-agreement-limit}.
\end{proof}

The lemma shows that agreement with the terminal least-squares
classifier evolves in only one direction: individual training examples
may enter the agreement set, but can never leave it. In particular,
full agreement is an absorbing state of the classifier dynamics.

\begin{corollary}[Permanent stabilization after full classifier agreement]
\label{cor:linear-L1-full-agreement-stabilization}
Suppose that, at some time \(\tau_0\ge0\),
\begin{equation}
\operatorname{Agr}_{\mathbf T}(\mathbf M(\tau_0))
=
1.
\label{eq:linear-L1-full-agreement}
\end{equation}
Then, for every \(\tau\ge\tau_0\),
\begin{equation}
\operatorname{Agr}_{\mathbf T}(\mathbf M(\tau))
=
1,
\qquad
\Acc(\mathbf M(\tau))
=
\Acc(\mathbf T).
\label{eq:linear-L1-permanent-stabilization}
\end{equation}
Thus, once the current classifier agrees with the least-squares
classifier on every training example, its training accuracy has reached
its terminal value and can never subsequently change.
\end{corollary}

\begin{proof}
By Lemma~\ref{lem:linear-L1-monotone-agreement}, agreement with
\(\mathbf T\) on each individual training example, once achieved, is
never lost. Hence
\(\operatorname{Agr}_{\mathbf T}(\mathbf M(\tau))=1\) for every
\(\tau\ge\tau_0\). The two classifiers therefore make the same
prediction on every training example, which implies
\begin{equation}
\Acc(\mathbf M(\tau))
=
\Acc(\mathbf T)
\end{equation}
for all \(\tau\ge\tau_0\).
\end{proof}

The preceding results show that agreement with the terminal
least-squares classifier evolves monotonically along the trajectory.
Training accuracy itself need not share this monotonicity, since the
classifier induced by \(\mathbf T\) may disagree with the ground-truth
labels on some training examples.  Nevertheless, monotone agreement
with \(\mathbf T\) provides direct control over how much the training
accuracy can decrease.

Indeed, once the prediction on an example agrees with that of
\(\mathbf T\), this prediction can never subsequently change away from
the \(\mathbf T\)-classifier.  Therefore, any loss of correct
predictions between two times can only be attributed to examples on
which the terminal classifier \(\mathbf T\) is itself incorrect.  This
observation yields the following quantitative bound on nonmonotone
accuracy dynamics.

\begin{linkedtheorem}
{Accuracy well bound under direct linear optimization}
{thm:main-linear-L1-accuracy-bound}
\label{thm:linear-L1-accuracy-bound}

Suppose that the least-squares predictor satisfies
\begin{equation}
\Acc(\mathbf T)
=
1-\varepsilon,
\qquad
0\le\varepsilon\le1.
\label{eq:linear-L1-target-accuracy}
\end{equation}
Then, for every pair of times
\(0\le\tau_1\le\tau_2\),
\begin{equation}
\Acc(\mathbf M(\tau_2))
\ge
\Acc(\mathbf M(\tau_1))
-
\varepsilon.
\label{eq:linear-L1-accuracy-drop-bound}
\end{equation}
Equivalently,
\begin{equation}
\sup_{0\le\tau_1\le\tau_2}
\left[
\Acc(\mathbf M(\tau_1))
-
\Acc(\mathbf M(\tau_2))
\right]
\le
\varepsilon.
\label{eq:linear-L1-max-drawdown}
\end{equation}
Thus any accuracy well along the \(L=1\) trajectory has depth at most
the classification error of the terminal least-squares predictor.
\end{linkedtheorem}

\begin{proof}
Partition the training examples according to whether
\(\mathbf T\) classifies them correctly:
\begin{equation}
\mathcal G_{\mathbf T}
=
\left\{
i:\widehat c_i=c_i
\right\},
\qquad
\mathcal B_{\mathbf T}
=
\left\{
i:\widehat c_i\neq c_i
\right\}.
\label{eq:linear-L1-good-bad-sets}
\end{equation}
By assumption,
\begin{equation}
\frac{|\mathcal G_{\mathbf T}|}{N}
=
1-\varepsilon,
\qquad
\frac{|\mathcal B_{\mathbf T}|}{N}
=
\varepsilon.
\label{eq:linear-L1-good-bad-fractions}
\end{equation}

For an example \(i\in\mathcal G_{\mathbf T}\), agreement with the
classifier of \(\mathbf T\) is equivalent to correct classification,
because \(\widehat c_i=c_i\). By
Lemma~\ref{lem:linear-L1-monotone-agreement}, if such an example is
correct at time \(\tau_1\), it remains correct at every later time
\(\tau_2\ge\tau_1\). Therefore every correct-to-incorrect transition
between \(\tau_1\) and \(\tau_2\) must occur on an example belonging
to \(\mathcal B_{\mathbf T}\).

There are only
\begin{equation}
|\mathcal B_{\mathbf T}|
=
\varepsilon N
\label{eq:linear-L1-number-bad}
\end{equation}
such examples. Hence at most \(\varepsilon N\) correct predictions can
be lost between \(\tau_1\) and \(\tau_2\), and therefore
\begin{equation}
N\Acc(\mathbf M(\tau_2))
\ge
N\Acc(\mathbf M(\tau_1))
-
\varepsilon N.
\label{eq:linear-L1-counting-bound}
\end{equation}
This proves
\eqref{eq:linear-L1-accuracy-drop-bound}. Taking the supremum over
\(0\le\tau_1\le\tau_2\) yields
\eqref{eq:linear-L1-max-drawdown}.
\end{proof}

The perfect-classification case gives an especially strong
consequence.

\begin{corollary}[Perfect least-squares classification excludes accuracy wells]
\label{cor:linear-L1-perfect-monotonicity}
If
\begin{equation}
\Acc(\mathbf T)=1,
\label{eq:linear-L1-perfect-target}
\end{equation}
then the training accuracy is monotone nondecreasing:
\begin{equation}
0\le\tau_1\le\tau_2
\qquad\Longrightarrow\qquad
\Acc(\mathbf M(\tau_2))
\ge
\Acc(\mathbf M(\tau_1)).
\label{eq:linear-L1-accuracy-monotone}
\end{equation}
Consequently, direct \(L=1\) gradient flow cannot exhibit an accuracy
well at any time scale.
\end{corollary}

\begin{proof}
The claim follows immediately from
Theorem~\ref{thm:linear-L1-accuracy-bound} by setting
\(\varepsilon=0\).
\end{proof}

The condition \(\Acc(\mathbf T)=1\) should not be identified with
ordinary linear separability. Linear separability guarantees only the
existence of some linear predictor whose predicted classes are correct
on all training examples, whereas \(\mathbf T\) is the particular
least-squares minimizer selected by the \(L=1\) dynamics. A linearly
separable dataset need not satisfy \(\Acc(\mathbf T)=1\). A stronger
sufficient condition is exact linear realizability of the labels: if
there exists a matrix \(\mathbf M_*\) such that
\begin{equation}
\mathbf M_*\mathbf x_i
=
\mathbf y_i
\qquad
\text{for every }i,
\label{eq:linear-L1-exact-realizability}
\end{equation}
then the minimum squared loss is zero, so the least-squares solution
also interpolates all labels and therefore satisfies
\(\Acc(\mathbf T)=1\).

We next turn to the time scale of classifier stabilization in the
vanishing-initialization regime.  Fix an arbitrary matrix
\(\mathbf Z\in\mathbb R^{C\times p}\) and consider the coupled family
\begin{equation}
\mathbf M^{(\sigma)}(0)
=
\sigma\mathbf Z,
\qquad
\sigma>0.
\label{eq:linear-L1-coupled-init-boundary}
\end{equation}
Thus \(\sigma\downarrow0\) changes only the scale of the initialization,
while its direction is held fixed.  In particular, the result applies
to every fixed realization of the Gaussian initialization considered
throughout the paper.

We show that the classifier reaches its terminal state on a vanishing
time scale.  Moreover, throughout this initial boundary layer the
end-to-end predictor remains microscopic, and only a vanishing amount
of squared-loss optimization has taken place.

\begin{theorem}[Boundary-layer stabilization under vanishing initialization]
\label{thm:linear-L1-boundary-layer}
There exists a family of times
\(\tau_*^{(\sigma)}=O(\sigma)\) as \(\sigma\downarrow0\) such that,
for every \(\tau\ge\tau_*^{(\sigma)}\),
\begin{equation}
\operatorname{Agr}_{\mathbf T}
\left(
\mathbf M^{(\sigma)}(\tau)
\right)
=
1,
\qquad
\Acc\left(
\mathbf M^{(\sigma)}(\tau)
\right)
=
\Acc(\mathbf T).
\label{eq:linear-L1-stabilized-after-boundary}
\end{equation}
Moreover,
\begin{equation}
\sup_{0\le\tau\le\tau_*^{(\sigma)}}
\left\|
\mathbf M^{(\sigma)}(\tau)
\right\|_{\mathrm F}
=
O(\sigma),
\qquad
\sup_{0\le\tau\le\tau_*^{(\sigma)}}
\left\|
\mathbf M^{(\sigma)}(\tau)
-
\mathbf M^{(\sigma)}(0)
\right\|_{\mathrm F}
=
O(\sigma).
\label{eq:linear-L1-boundary-predictor-scale}
\end{equation}
At the stabilization time, only an asymptotically vanishing fraction
of the available loss decrease has occurred:
\begin{equation}
\frac{
\mathcal L(\mathbf M^{(\sigma)}(0))
-
\mathcal L(\mathbf M^{(\sigma)}(\tau_*^{(\sigma)}))
}{
\mathcal L(\mathbf M^{(\sigma)}(0))
-
\mathcal L(\mathbf T)
}
=
O(\sigma),
\label{eq:linear-L1-relative-loss-progress}
\end{equation}
and, in particular,
\begin{equation}
\frac{
\mathcal L(\mathbf M^{(\sigma)}(0))
-
\mathcal L(\mathbf M^{(\sigma)}(\tau_*^{(\sigma)}))
}{
\mathcal L(\mathbf M^{(\sigma)}(0))
}
=
O(\sigma).
\label{eq:linear-L1-loss-relative-initial}
\end{equation}
Thus all nontrivial accuracy dynamics are confined to an
\(O(\sigma)\) boundary layer in the rescaled time \(\tau\), equivalently
\(O(\sigma/\gamma)\) in the original gradient-flow time \(t\).
\end{theorem}

\begin{proof}
We first construct a time by which the classifier agrees with
\(\mathbf T\) on every training example.  For every example \(i\) and
every competing class \(c\neq\widehat c_i\), define the target margin
and the corresponding initialization margin by
\begin{equation}
\Delta_{ic}
=
(\mathbf T\mathbf x_i)_{\widehat c_i}
-
(\mathbf T\mathbf x_i)_c
>
0,
\qquad
\zeta_{ic}
=
(\mathbf Z\mathbf x_i)_{\widehat c_i}
-
(\mathbf Z\mathbf x_i)_c.
\label{eq:linear-L1-margins}
\end{equation}
The strict positivity of \(\Delta_{ic}\) follows from the assumed
uniqueness of the top-scoring class under \(\mathbf T\).

Using the explicit gradient-flow trajectory
\eqref{eq:linear-L1-explicit-flow} together with the scaled
initialization
\eqref{eq:linear-L1-coupled-init-boundary}, the corresponding pairwise
margin at time \(\tau\) is
\begin{equation}
\begin{aligned}
&
\left(
\mathbf M^{(\sigma)}(\tau)\mathbf x_i
\right)_{\widehat c_i}
-
\left(
\mathbf M^{(\sigma)}(\tau)\mathbf x_i
\right)_c
\\
&\qquad=
\sigma e^{-\tau}\zeta_{ic}
+
(1-e^{-\tau})\Delta_{ic}
\\
&\qquad=
\sigma e^{-\tau}
\left[
\zeta_{ic}
+
\frac{e^\tau-1}{\sigma}\Delta_{ic}
\right].
\end{aligned}
\label{eq:linear-L1-fast-margin}
\end{equation}
This suggests introducing the fast time variable
\begin{equation}
s(\tau)
=
\frac{e^\tau-1}{\sigma}.
\label{eq:linear-L1-fast-time}
\end{equation}
Since the prefactor \(\sigma e^{-\tau}\) in
\eqref{eq:linear-L1-fast-margin} is strictly positive, the sign of the
pairwise margin is determined by
\begin{equation}
\zeta_{ic}+s(\tau)\Delta_{ic}.
\label{eq:linear-L1-fast-margin-sign}
\end{equation}
Because \(\Delta_{ic}>0\), this quantity increases linearly with the
fast time \(s\).  We therefore define
\begin{equation}
s_*
=
1+
\max_{1\le i\le N}
\max_{c\neq\widehat c_i}
\left(
-\frac{\zeta_{ic}}{\Delta_{ic}}
\right)_+,
\label{eq:linear-L1-fast-stabilization-scale}
\end{equation}
where \((a)_+=\max\{a,0\}\).  The additive constant \(1\) ensures
strict positivity, so that
\begin{equation}
\zeta_{ic}+s_*\Delta_{ic}>0
\label{eq:linear-L1-fast-positive-margin}
\end{equation}
for every \(i\) and every \(c\neq\widehat c_i\).

Choose
\begin{equation}
\tau_*^{(\sigma)}
=
\log(1+\sigma s_*).
\label{eq:linear-L1-boundary-time}
\end{equation}
By the definition of the fast time,
\begin{equation}
s(\tau_*^{(\sigma)})=s_*.
\label{eq:linear-L1-fast-time-at-boundary}
\end{equation}
Hence all pairwise margins relative to the class
\(\widehat c_i\) are strictly positive at
\(\tau_*^{(\sigma)}\), and therefore
\begin{equation}
\operatorname{Agr}_{\mathbf T}
\left(
\mathbf M^{(\sigma)}(\tau_*^{(\sigma)})
\right)
=
1.
\label{eq:linear-L1-full-agreement-at-boundary}
\end{equation}
The permanent-stabilization result in
Corollary~\ref{cor:linear-L1-full-agreement-stabilization} then implies
that, for every \(\tau\ge\tau_*^{(\sigma)}\),
\begin{equation}
\operatorname{Agr}_{\mathbf T}
\left(
\mathbf M^{(\sigma)}(\tau)
\right)
=
1,
\qquad
\Acc\left(
\mathbf M^{(\sigma)}(\tau)
\right)
=
\Acc(\mathbf T).
\label{eq:linear-L1-boundary-stabilization-proof}
\end{equation}
Moreover, \(s_*\) is finite and independent of \(\sigma\), because
\(\mathbf Z\) is fixed and there are only finitely many pairs
\((i,c)\).  Expanding the definition
\eqref{eq:linear-L1-boundary-time} therefore gives
\begin{equation}
\tau_*^{(\sigma)}
=
\sigma s_*+O(\sigma^2)
=
O(\sigma),
\qquad
\sigma\downarrow0.
\label{eq:linear-L1-boundary-time-asymptotic}
\end{equation}

We next control the predictor throughout this boundary layer.  Using
the relation
\begin{equation}
e^\tau=1+\sigma s(\tau)
\label{eq:linear-L1-fast-time-inverse}
\end{equation}
in the explicit gradient-flow trajectory
\eqref{eq:linear-L1-explicit-flow} gives the exact representation
\begin{equation}
\mathbf M^{(\sigma)}(\tau)
=
\frac{\sigma}{1+\sigma s(\tau)}
\left[
\mathbf Z+s(\tau)\mathbf T
\right].
\label{eq:linear-L1-fast-predictor}
\end{equation}
Since \(s(\tau)\) is increasing, for
\(0\le\tau\le\tau_*^{(\sigma)}\) we have
\begin{equation}
0\le s(\tau)\le s_*.
\label{eq:linear-L1-fast-time-range}
\end{equation}
Consequently,
\begin{equation}
\left\|
\mathbf M^{(\sigma)}(\tau)
\right\|_{\mathrm F}
\le
\sigma
\left(
\|\mathbf Z\|_{\mathrm F}
+
s_*\|\mathbf T\|_{\mathrm F}
\right),
\label{eq:linear-L1-predictor-bound}
\end{equation}
uniformly throughout the boundary layer.  Hence
\begin{equation}
\sup_{0\le\tau\le\tau_*^{(\sigma)}}
\left\|
\mathbf M^{(\sigma)}(\tau)
\right\|_{\mathrm F}
=
O(\sigma).
\label{eq:linear-L1-predictor-bound-asymptotic}
\end{equation}

The same fast-time representation also gives
\begin{equation}
\mathbf M^{(\sigma)}(\tau)
-
\mathbf M^{(\sigma)}(0)
=
\frac{
\sigma s(\tau)
}{
1+\sigma s(\tau)
}
\left(
\mathbf T-\sigma\mathbf Z
\right).
\label{eq:linear-L1-displacement-exact}
\end{equation}
Because \(0\le s(\tau)\le s_*\) and both \(\mathbf T\) and
\(\mathbf Z\) are fixed, it follows that
\begin{equation}
\sup_{0\le\tau\le\tau_*^{(\sigma)}}
\left\|
\mathbf M^{(\sigma)}(\tau)
-
\mathbf M^{(\sigma)}(0)
\right\|_{\mathrm F}
=
O(\sigma).
\label{eq:linear-L1-displacement-bound-asymptotic}
\end{equation}
This proves the microscopic predictor bounds
\eqref{eq:linear-L1-boundary-predictor-scale}.

It remains to quantify the progress in squared loss.  The explicit
gradient-flow solution \eqref{eq:linear-L1-explicit-flow} implies the
exact contraction toward the least-squares predictor
\begin{equation}
\mathbf M^{(\sigma)}(\tau)-\mathbf T
=
e^{-\tau}
\left(
\mathbf M^{(\sigma)}(0)-\mathbf T
\right).
\label{eq:linear-L1-distance-contraction}
\end{equation}
Combining this identity with the quadratic loss representation
\eqref{eq:linear-L1-loss} gives
\begin{equation}
\mathcal L(\mathbf M^{(\sigma)}(\tau))
-
\mathcal L(\mathbf T)
=
e^{-2\tau}
\left[
\mathcal L(\mathbf M^{(\sigma)}(0))
-
\mathcal L(\mathbf T)
\right].
\label{eq:linear-L1-loss-contraction}
\end{equation}
Evaluating this identity at the stabilization time
\(\tau_*^{(\sigma)}\) yields
\begin{equation}
\begin{aligned}
&
\frac{
\mathcal L(\mathbf M^{(\sigma)}(0))
-
\mathcal L(\mathbf M^{(\sigma)}(\tau_*^{(\sigma)}))
}{
\mathcal L(\mathbf M^{(\sigma)}(0))
-
\mathcal L(\mathbf T)
}
\\
&\qquad=
1-e^{-2\tau_*^{(\sigma)}}
=
1-\frac{1}{(1+\sigma s_*)^2}
\\
&\qquad=
2\sigma s_*+O(\sigma^2)
=
O(\sigma).
\end{aligned}
\label{eq:linear-L1-loss-progress-exact}
\end{equation}
This proves
\eqref{eq:linear-L1-relative-loss-progress}; in particular, only a
vanishing fraction of the total available loss decrease has occurred
by the time the classifier has stabilized.

Finally, because the labels are one-hot,
\begin{equation}
\mathcal L(\mathbf 0)
=
\frac12.
\label{eq:linear-L1-zero-loss}
\end{equation}
Since
\(\mathbf M^{(\sigma)}(0)=\sigma\mathbf Z\) with fixed
\(\mathbf Z\),
\begin{equation}
\mathcal L(\mathbf M^{(\sigma)}(0))
=
\frac12+O(\sigma),
\label{eq:linear-L1-initial-loss-asymptotic}
\end{equation}
and hence the initial loss is bounded away from zero for all
sufficiently small \(\sigma\).

Moreover, the quadratic loss representation
\eqref{eq:linear-L1-loss} gives
\begin{equation}
\mathcal L(\mathbf 0)-\mathcal L(\mathbf T)
=
\frac{\gamma}{2}
\|\mathbf T\|_{\mathrm F}^2.
\label{eq:linear-L1-total-loss-at-zero}
\end{equation}
The assumed unique top-scoring class under \(\mathbf T\) implies
\(\mathbf T\neq\mathbf 0\), so the right-hand side is strictly
positive.  Thus the total available loss decrease remains bounded as
\(\sigma\downarrow0\), while
\eqref{eq:linear-L1-loss-progress-exact} shows that an \(O(\sigma)\)
fraction of it has occurred by \(\tau_*^{(\sigma)}\).  The absolute
loss decrease is therefore \(O(\sigma)\), and division by the initial
loss yields
\begin{equation}
\frac{
\mathcal L(\mathbf M^{(\sigma)}(0))
-
\mathcal L(\mathbf M^{(\sigma)}(\tau_*^{(\sigma)}))
}{
\mathcal L(\mathbf M^{(\sigma)}(0))
}
=
O(\sigma),
\end{equation}
which proves
\eqref{eq:linear-L1-loss-relative-initial}.
\end{proof}

Combining
Theorems~\ref{thm:linear-L1-accuracy-bound}
and~\ref{thm:linear-L1-boundary-layer} gives a quantitative
description of every possible accuracy well under direct linear
optimization.  If
\(\Acc(\mathbf T)=1-\varepsilon\), then any downward excursion of
training accuracy has depth at most \(\varepsilon\).  Under the
vanishing initialization
\(\mathbf M^{(\sigma)}(0)=\sigma\mathbf Z\), all nontrivial classifier
dynamics, including any such excursion, are confined to an
\(O(\sigma)\) time interval in the rescaled time \(\tau\).
Throughout this boundary layer the predictor remains at scale
\(O(\sigma)\), while only an \(O(\sigma)\) fraction of the available
loss decrease has occurred.

Thus the \(L=1\) dynamics are strongly constrained: agreement with the
terminal least-squares classifier is monotone, the classifier reaches
its terminal state on a vanishing time scale under vanishing
initialization, and any possible decrease in training accuracy is
controlled by the terminal classification error.  In particular, if
\(\Acc(\mathbf T)=1\), training accuracy is monotone and accuracy wells
are impossible.  We next show that factorized linear networks behave
qualitatively differently: the end-to-end predictor no longer follows
a direct convex interpolation toward \(\mathbf T\), and nonmonotone
classifier dynamics can consequently arise.

\subsection{Two-layer linear networks: \texorpdfstring{\(L=2\)}{L = 2}}
\label{app:linear-L2}

For \(L=2\), we write the network as\footnote{
In the general notation introduced above, the end-to-end predictor is
\(\mathbf M=\mathbf W_2\mathbf W_1\).  Throughout this subsection we use
the equivalent parametrization
\(\mathbf W_1=\mathbf U\) and
\(\mathbf W_2=\mathbf V^{\top}\), which simplifies the notation.
}
\begin{equation}
f_{\mathbf U,\mathbf V}(\mathbf x)
=
\mathbf V^{\top}\mathbf U\mathbf x,
\qquad
\mathbf U\in\mathbb R^{H\times p},
\qquad
\mathbf V\in\mathbb R^{H\times C},
\label{eq:linear-L2-predictor}
\end{equation}
with end-to-end predictor
\begin{equation}
\mathbf M
=
\mathbf V^{\top}\mathbf U
\in\mathbb R^{C\times p}.
\label{eq:linear-L2-end-to-end}
\end{equation}

We retain the same isotropic-input assumption as in the \(L=1\)
analysis,
\begin{equation}
\mathbf G
=
\gamma\mathbf I_p,
\qquad
\gamma>0,
\label{eq:linear-L2-isotropy}
\end{equation}
and use the same effective target \(\mathbf T\) defined in
\eqref{eq:linear-L1-target}.  As before, the empirical squared loss
depends on the parameters only through the end-to-end predictor and can
be written as
\begin{equation}
\mathcal L(\mathbf U,\mathbf V)
=
\frac{\gamma}{2}
\left\|
\mathbf V^{\top}\mathbf U-\mathbf T
\right\|_{\mathrm F}^2
+
\mathcal L(\mathbf T).
\label{eq:linear-L2-factorization-loss}
\end{equation}

Thus the objective as a function of the realized predictor
\(\mathbf M\) is exactly the same quadratic objective as for \(L=1\).
What changes is the parametrization: gradient flow now acts on the
weights \(\mathbf U\) and \(\mathbf V\), so the induced trajectory of
the end-to-end predictor
\(\mathbf M=\mathbf V^{\top}\mathbf U\) need no longer follow the
direct convex interpolation toward \(\mathbf T\) found in
\eqref{eq:linear-L1-explicit-flow}.

As in the \(L=1\) analysis, we rescale gradient-flow time according to
\begin{equation}
\tau=\gamma t.
\label{eq:linear-L2-rescaled-time}
\end{equation}
Throughout this subsection a dot denotes differentiation with respect
to \(\tau\).  The gradient flow dynamics of $\mathbf U$ and $\mathbf V$ are then
\begin{equation}
\dot{\mathbf U}
=
\mathbf V
\mathbf T- \mathbf V \mathbf V^{\top}\mathbf U,
\qquad
\dot{\mathbf V}
=
\mathbf U
\mathbf T^{\top}- \mathbf U \mathbf U^{\top}\mathbf V.
\label{eq:linear-L2-flow}
\end{equation}

Let
\begin{equation}
\mathbf T
=
\mathbf P\boldsymbol{\Lambda}\mathbf Q^{\top}
=
\sum_{j=1}^r
s_j\mathbf p_j\mathbf q_j^{\top},
\qquad
s_1\ge\cdots\ge s_r>0,
\qquad
r=\operatorname{rank}(\mathbf T),
\label{eq:linear-L2-compact-svd}
\end{equation}
be a compact singular-value decomposition. Let \(k_\star\) be the
multiplicity of the largest singular value,
\begin{equation}
s_1=\cdots=s_{k_\star}>s_{k_\star+1},
\qquad
s_{r+1}:=0,
\label{eq:linear-L2-top-multiplicity}
\end{equation}
and define the complete top singular block
\begin{equation}
\mathbf T_{k_\star}
:=
\sum_{j=1}^{k_\star}
s_j\mathbf p_j\mathbf q_j^{\top}.
\label{eq:linear-L2-top-block}
\end{equation}
Because the entire singular subspace associated with \(s_1\) is retained,
\(\mathbf T_{k_\star}\) is independent of the choice of singular-vector
basis within that subspace.

We will repeatedly use the two accuracy levels
\begin{equation}
A_{\mathbf T}:=\Acc(\mathbf T),
\qquad
A_{k_\star}:=\Acc(\mathbf T_{k_\star}).
\label{eq:linear-L2-accuracy-levels}
\end{equation}
They describe the classifiers associated with the full effective target
and with its complete top singular block.  The analysis below identifies
three stages of the \(L=2\) dynamics.  In the sequential
small-initialization and large-width limit, we first show that at an
early organization time \(\tau_{\mathrm{org}}\) the network classifier
agrees with that of \(\mathbf T\) with probability tending to one.  At a
later bottleneck time
\(\tau_{\mathrm{bot}}>\tau_{\mathrm{org}}\), spectral concentration
instead makes the classifier agree with that of
\(\mathbf T_{k_\star}\), again with probability tending to one.
Finally, when
\(H\ge\operatorname{rank}(\mathbf T)\), the full Gaussian-initialized
gradient flow converges almost surely to \(\mathbf T\) at late times.
Consequently, the relation between
\(A_{\mathbf T}\) and \(A_{k_\star}\) determines whether these three
stages produce an accuracy well.

\paragraph{Ordering of the two accuracy levels.}
\label{app:ordering}
There is no universal ordering between \(A_{\mathbf T}\) and \(A_{k_\star}\).
The truncated singular-value decomposition is optimal for matrix
approximation in Frobenius or operator norm, but classification accuracy is
determined by the ordering of class scores on individual examples.
Consequently, subleading singular modes may either correct or spoil
top-one decisions.

We give an explicit example in which
\(A_{\mathbf T}<A_{k_\star}\). Let \(p=2\), \(C=3\), and \(N=8\), and take the
isotropic inputs
\begin{equation}
\mathbf x_j
=
\sqrt{2}
\begin{pmatrix}
\cos((j-1)\pi/4)\\
\sin((j-1)\pi/4)
\end{pmatrix},
\qquad
j=1,\ldots,8.
\label{eq:linear-L2-reverse-order-inputs}
\end{equation}
The eight vectors are the equally spaced points on a circle of radius
\(\sqrt2\), and therefore
\begin{equation}
\frac18\sum_{j=1}^8
\mathbf x_j\mathbf x_j^{\top}
=
\mathbf I_2.
\label{eq:linear-L2-reverse-order-isotropy}
\end{equation}
Assign the labels
\begin{equation}
(c_1,\ldots,c_8)
=
(1,1,2,1,2,2,3,2),
\qquad
\mathbf y_j=\mathbf e_{c_j}.
\label{eq:linear-L2-reverse-order-labels}
\end{equation}
Using
\(\mathbf T=(N\gamma)^{-1}\sum_j\mathbf y_j\mathbf x_j^{\top}\),
we obtain
\begin{equation}
\mathbf T
=
\frac18
\begin{pmatrix}
\sqrt2 & 2\\
-\sqrt2 & \sqrt2-2\\
0 & -\sqrt2
\end{pmatrix}.
\label{eq:linear-L2-reverse-order-target}
\end{equation}
Indeed, the three rows are obtained by summing the inputs belonging to each
class:
\begin{equation}
\sum_{j:c_j=1}\mathbf x_j
=
\begin{pmatrix}
\sqrt2\\
2
\end{pmatrix},
\qquad
\sum_{j:c_j=2}\mathbf x_j
=
\begin{pmatrix}
-\sqrt2\\
\sqrt2-2
\end{pmatrix},
\qquad
\sum_{j:c_j=3}\mathbf x_j
=
\begin{pmatrix}
0\\
-\sqrt2
\end{pmatrix}.
\label{eq:linear-L2-reverse-order-row-sums}
\end{equation}

For this matrix,
\begin{equation}
\mathbf T^{\top}\mathbf T
=
\frac1{64}
\begin{pmatrix}
4 & 4\sqrt2-2\\
4\sqrt2-2 & 12-4\sqrt2
\end{pmatrix},
\label{eq:linear-L2-reverse-order-Gram}
\end{equation}
whose two eigenvalues are
\begin{equation}
s_{1,2}^2
=
\frac{
8-2\sqrt2
\pm
2\sqrt{15-8\sqrt2}
}{64}.
\label{eq:linear-L2-reverse-order-singular-values}
\end{equation}
Thus
\begin{equation}
s_1\approx0.37524,
\qquad
s_2\approx0.14425,
\label{eq:linear-L2-reverse-order-singular-values-numeric}
\end{equation}
so the largest singular value is simple and \(k_\star=1\).

Evaluating \(\mathbf T\mathbf x_j\) on the eight training inputs gives the
predicted classes
\begin{equation}
(1,1,1,2,2,2,3,3).
\label{eq:linear-L2-reverse-order-full-predictions}
\end{equation}
Comparing these predictions with
\((1,1,2,1,2,2,3,2)\), the correctly classified examples are
\(j=1,2,5,6,7\). Hence
\begin{equation}
A_{\mathbf T}=\frac58.
\label{eq:linear-L2-reverse-order-full-accuracy}
\end{equation}

The leading rank-one truncation is
\begin{equation}
\mathbf T_1
=
s_1\mathbf p_1\mathbf q_1^{\top}
\approx
\begin{pmatrix}
0.18046 & 0.24731\\
-0.09629 & -0.13196\\
-0.08417 & -0.11536
\end{pmatrix}.
\label{eq:linear-L2-reverse-order-top-block}
\end{equation}
Its predicted classes on the same eight inputs are
\begin{equation}
(1,1,1,1,2,2,2,2).
\label{eq:linear-L2-reverse-order-top-predictions}
\end{equation}
These are correct for
\(j=1,2,4,5,6,8\), and therefore
\begin{equation}
A_{k_\star}=A_1=\frac68=\frac34>A_{\mathbf T}.
\label{eq:linear-L2-reverse-order-accuracy}
\end{equation}
Thus, in this example, adding the subleading singular component decreases
classification accuracy even though it improves the matrix approximation to
the full target.

The opposite ordering,
\begin{equation}
A_{\mathbf T}>A_{k_\star},
\label{eq:linear-L2-well-ordering}
\end{equation}
is the one that produces the accuracy wells studied in this work. To verify
that this ordering is not specific to the synthetic constructions below, we
evaluate it on several standard image datasets after centering and whitening
the training inputs, so that the empirical covariance satisfies the isotropy
assumption of our theory. The results are shown in
Table~\ref{tab:linear-L2-natural-ordering}. In all five datasets the largest
singular value is simple, \(k_\star=1\), and we find
\(A_{\mathbf T}>A_{k_\star}\).

\begin{table}[t]
\centering
\caption{
Accuracy of the full effective target and its complete top singular block on
centered and whitened training data. In all cases \(k_\star=1\).
}
\label{tab:linear-L2-natural-ordering}
\begin{tabular}{lccc}
\toprule
Dataset
&
\(A_{\mathbf T}\)
&
\(A_{k_\star}\)
&
\(A_{\mathbf T}>A_{k_\star}\)
\\
\midrule
MNIST
&
0.859
&
0.209
&
Yes
\\
Fashion-MNIST
&
0.831
&
0.200
&
Yes
\\
KMNIST
&
0.770
&
0.196
&
Yes
\\
CIFAR-10
&
0.487
&
0.180
&
Yes
\\
CIFAR-100
&
0.442
&
0.019
&
Yes
\\
\bottomrule
\end{tabular}
\end{table}

\paragraph{Microscopic linearization.}

We now analyze the vanishing-initialization dynamics generated by
the gradient flow \eqref{eq:linear-L2-flow}.  Write the
initialization as
\begin{equation}
\mathbf U^{(\sigma)}(0)
=
\sigma\mathbf A,
\qquad
\mathbf V^{(\sigma)}(0)
=
\sigma\mathbf B,
\label{eq:linear-L2-scaled-initialization}
\end{equation}
where
\(\mathbf A\in\mathbb R^{H\times p}\) and
\(\mathbf B\in\mathbb R^{H\times C}\) are independent of \(\sigma\).
Thus \(\sigma\) controls only the initialization scale.

Define the normalized exact weights
\begin{equation}
\widehat{\mathbf U}^{(\sigma)}
=
\frac{\mathbf U^{(\sigma)}}{\sigma},
\qquad
\widehat{\mathbf V}^{(\sigma)}
=
\frac{\mathbf V^{(\sigma)}}{\sigma}.
\label{eq:linear-L2-normalized-factors}
\end{equation}
Substituting these variables into the exact gradient-flow equations
\eqref{eq:linear-L2-flow} gives
\begin{equation}
\begin{aligned}
\dot{\widehat{\mathbf U}}^{(\sigma)}
&=
\widehat{\mathbf V}^{(\sigma)}\mathbf T
-
\sigma^2
\widehat{\mathbf V}^{(\sigma)}
\widehat{\mathbf V}^{(\sigma)\top}
\widehat{\mathbf U}^{(\sigma)},
\\
\dot{\widehat{\mathbf V}}^{(\sigma)}
&=
\widehat{\mathbf U}^{(\sigma)}\mathbf T^{\top}
-
\sigma^2
\widehat{\mathbf U}^{(\sigma)}
\widehat{\mathbf U}^{(\sigma)\top}
\widehat{\mathbf V}^{(\sigma)}.
\end{aligned}
\label{eq:linear-L2-normalized-flow}
\end{equation}
The first term in each equation is linear in the normalized weights,
whereas the nonlinear term generated by the factorization is
explicitly suppressed by \(\sigma^2\).  This suggests that, on every
fixed time interval, the normalized dynamics converge as
\(\sigma\downarrow0\) to the linear system
\begin{equation}
\dot{\mathbf U}_0
=
\mathbf V_0\mathbf T,
\qquad
\dot{\mathbf V}_0
=
\mathbf U_0\mathbf T^{\top},
\qquad
\mathbf U_0(0)=\mathbf A,
\qquad
\mathbf V_0(0)=\mathbf B.
\label{eq:linear-L2-linearized-flow}
\end{equation}
Here \((\mathbf U_0,\mathbf V_0)\) denote the limiting normalized
weights.  The following theorem makes this approximation precise
and shows that the corresponding end-to-end predictor and change
in squared loss remain microscopic.

\begin{theorem}[Uniform microscopic linearization for two-layer linear networks]
\label{thm:linear-L2-microscopic-linearization}
Fix \(p,C,H\), an effective target
\(\mathbf T\in\mathbb R^{C\times p}\), a finite time horizon
\(\tau_0<\infty\), and matrices
\(\mathbf A\in\mathbb R^{H\times p}\) and
\(\mathbf B\in\mathbb R^{H\times C}\).
Let
\((\mathbf U^{(\sigma)},\mathbf V^{(\sigma)})\)
denote the exact gradient flow
\eqref{eq:linear-L2-flow} with initialization
\eqref{eq:linear-L2-scaled-initialization}, and let
\((\mathbf U_0,\mathbf V_0)\) solve the limiting linear system
\eqref{eq:linear-L2-linearized-flow}.

For every \(\sigma>0\), the exact gradient flow exists globally.
Moreover, as \(\sigma\downarrow0\), the normalized weights converge
uniformly on \([0,\tau_0]\):
\begin{equation}
\sup_{0\le\tau\le\tau_0}
\left(
\left\|
\widehat{\mathbf U}^{(\sigma)}(\tau)-\mathbf U_0(\tau)
\right\|_{\mathrm F}
+
\left\|
\widehat{\mathbf V}^{(\sigma)}(\tau)-\mathbf V_0(\tau)
\right\|_{\mathrm F}
\right)
\longrightarrow0.
\label{eq:linear-L2-factor-convergence}
\end{equation}

Define the limiting normalized predictor
\begin{equation}
\mathbf M_0(\tau)
=
\mathbf V_0(\tau)^{\top}\mathbf U_0(\tau).
\label{eq:linear-L2-limiting-predictor}
\end{equation}
Then
\begin{equation}
\sup_{0\le\tau\le\tau_0}
\left\|
\frac{\mathbf M^{(\sigma)}(\tau)}{\sigma^2}
-
\mathbf M_0(\tau)
\right\|_{\mathrm F}
\longrightarrow0,
\label{eq:linear-L2-predictor-convergence}
\end{equation}
where
\(\mathbf M^{(\sigma)}
=
\mathbf V^{(\sigma)\top}\mathbf U^{(\sigma)}\).

Furthermore, the unnormalized predictor satisfies
\begin{equation}
\sup_{0\le\tau\le\tau_0}
\left\|
\mathbf M^{(\sigma)}(\tau)
\right\|_{\mathrm{op}}
=
O(\sigma^2),
\label{eq:linear-L2-predictor-microscopic}
\end{equation}
and the change in squared loss obeys
\begin{equation}
\sup_{0\le\tau\le\tau_0}
\left|
\mathcal L
\left(
\mathbf U^{(\sigma)}(\tau),
\mathbf V^{(\sigma)}(\tau)
\right)
-
\mathcal L(\mathbf 0,\mathbf 0)
\right|
=
O(\sigma^2).
\label{eq:linear-L2-loss-microscopic}
\end{equation}
The constants implicit in these estimates may depend on the fixed
problem dimensions, target, initialization matrices, and time horizon.
\end{theorem}

\begin{proof}
Write
\begin{equation}
\mathbf Z^{(\sigma)}
=
\left(
\widehat{\mathbf U}^{(\sigma)},
\widehat{\mathbf V}^{(\sigma)}
\right),
\qquad
\mathbf Z_0
=
\left(
\mathbf U_0,\mathbf V_0
\right),
\label{eq:linear-L2-Z-definition}
\end{equation}
and equip pairs of matrices with the norm
\begin{equation}
\|(\mathbf X,\mathbf Y)\|^2
=
\|\mathbf X\|_{\mathrm F}^2
+
\|\mathbf Y\|_{\mathrm F}^2.
\label{eq:linear-L2-product-norm}
\end{equation}
Define the linear operator
\begin{equation}
\mathcal A(\mathbf X,\mathbf Y)
=
\left(
\mathbf Y\mathbf T,
\mathbf X\mathbf T^{\top}
\right)
\label{eq:linear-L2-linear-operator}
\end{equation}
and the cubic map
\begin{equation}
\mathcal C(\mathbf X,\mathbf Y)
=
\left(
-\mathbf Y\mathbf Y^{\top}\mathbf X,
-\mathbf X\mathbf X^{\top}\mathbf Y
\right).
\label{eq:linear-L2-cubic-map}
\end{equation}
The normalized exact dynamics
\eqref{eq:linear-L2-normalized-flow} and the limiting dynamics
\eqref{eq:linear-L2-linearized-flow} become
\begin{equation}
\dot{\mathbf Z}^{(\sigma)}
=
\mathcal A\mathbf Z^{(\sigma)}
+
\sigma^2
\mathcal C\left(\mathbf Z^{(\sigma)}\right),
\qquad
\dot{\mathbf Z}_0
=
\mathcal A\mathbf Z_0,
\label{eq:linear-L2-normalized-abstract}
\end{equation}
with the common initial condition
\begin{equation}
\mathbf Z^{(\sigma)}(0)
=
\mathbf Z_0(0)
=
(\mathbf A,\mathbf B).
\label{eq:linear-L2-abstract-initialization}
\end{equation}

\emph{Step 1: global existence and uniform bounds.}
The vector field of the exact normalized system in
\eqref{eq:linear-L2-normalized-abstract} is polynomial and hence
locally Lipschitz.  It therefore admits a unique solution on a
maximal forward interval.  We show that this interval is infinite.

Define the normalized factor energy
\begin{equation}
S^{(\sigma)}(\tau)
=
\left\|
\widehat{\mathbf U}^{(\sigma)}(\tau)
\right\|_{\mathrm F}^2
+
\left\|
\widehat{\mathbf V}^{(\sigma)}(\tau)
\right\|_{\mathrm F}^2
=
\|\mathbf Z^{(\sigma)}(\tau)\|^2.
\label{eq:linear-L2-normalized-energy}
\end{equation}
Differentiating along the exact normalized flow gives
\begin{equation}
\begin{aligned}
\frac{d}{d\tau}S^{(\sigma)}(\tau)
&=
4
\left\langle
\mathbf T,
\widehat{\mathbf V}^{(\sigma)}(\tau)^{\top}
\widehat{\mathbf U}^{(\sigma)}(\tau)
\right\rangle_{\mathrm F}
\\
&\quad
-
4\sigma^2
\left\|
\widehat{\mathbf V}^{(\sigma)}(\tau)^{\top}
\widehat{\mathbf U}^{(\sigma)}(\tau)
\right\|_{\mathrm F}^2
\\
&\le 4
\left\langle
\mathbf T,
\widehat{\mathbf V}^{(\sigma)}(\tau)^{\top}
\widehat{\mathbf U}^{(\sigma)}(\tau)
\right\rangle_{\mathrm F} \le
2s_{\max}S^{(\sigma)}(\tau),
\end{aligned}
\label{eq:linear-L2-normalized-energy-bound}
\end{equation}
where
\begin{equation}
s_{\max}
=
\|\mathbf T\|_{\mathrm{op}}.
\label{eq:linear-L2-smax}
\end{equation}
The last inequality follows from
\begin{equation}
\left|
\left\langle
\mathbf T,
\widehat{\mathbf V}^{(\sigma)\top}
\widehat{\mathbf U}^{(\sigma)}
\right\rangle_{\mathrm F}
\right|
\le
s_{\max}
\|\widehat{\mathbf U}^{(\sigma)}\|_{\mathrm F}
\|\widehat{\mathbf V}^{(\sigma)}\|_{\mathrm F}
\le
\frac{s_{\max}}{2}S^{(\sigma)}.
\end{equation}

Gronwall's inequality therefore yields
\begin{equation}
S^{(\sigma)}(\tau)
\le
e^{2s_{\max}\tau}
\left(
\|\mathbf A\|_{\mathrm F}^2
+
\|\mathbf B\|_{\mathrm F}^2
\right).
\label{eq:linear-L2-normalized-energy-gronwall}
\end{equation}
This estimate holds throughout the interval of existence and is
finite on every bounded time interval.  Thus the normalized weights
cannot blow up in finite time.  By the continuation theorem for
finite-dimensional locally Lipschitz ODEs, the exact normalized
solution exists globally for every \(\sigma>0\).

In particular, for any finite \(\tau_0\),
\begin{equation}
\sup_{0\le\tau\le\tau_0}
\|\mathbf Z^{(\sigma)}(\tau)\|
\le
e^{s_{\max}\tau_0}
\|(\mathbf A,\mathbf B)\|.
\label{eq:linear-L2-uniform-Z-bound}
\end{equation}
Crucially, this bound is independent of \(\sigma\).

\emph{Step 2: uniform convergence to the linearized dynamics.}
For every pair \((\mathbf X,\mathbf Y)\),
\begin{equation}
\begin{aligned}
\|\mathcal A(\mathbf X,\mathbf Y)\|^2
&=
\|\mathbf Y\mathbf T\|_{\mathrm F}^2
+
\|\mathbf X\mathbf T^{\top}\|_{\mathrm F}^2
\\
&\le
s_{\max}^2
\left(
\|\mathbf X\|_{\mathrm F}^2
+
\|\mathbf Y\|_{\mathrm F}^2
\right).
\end{aligned}
\label{eq:linear-L2-A-bound}
\end{equation}
Consequently,
\begin{equation}
\|\mathcal A\|_{\mathrm{op}}
\le s_{\max},
\qquad
\left\|
e^{\tau\mathcal A}
\right\|_{\mathrm{op}}
\le
e^{s_{\max}\tau},
\qquad
\tau\ge0.
\label{eq:linear-L2-semigroup-bound}
\end{equation}
The cubic map satisfies
\begin{equation}
\left\|
\mathcal C(\mathbf Z)
\right\|
\le
\|\mathbf Z\|^3.
\label{eq:linear-L2-cubic-bound}
\end{equation}

Variation of constants gives
\begin{equation}
\mathbf Z^{(\sigma)}(\tau)
=
e^{\tau\mathcal A}\mathbf Z^{(\sigma)}(0)
+
\sigma^2
\int_0^\tau
e^{(\tau-u)\mathcal A}
\mathcal C
\left(
\mathbf Z^{(\sigma)}(u)
\right)
\,du.
\label{eq:linear-L2-variation-of-constants}
\end{equation}
Since
\(\mathbf Z_0(\tau)=e^{\tau\mathcal A}\mathbf Z_0(0)\)
and the initial conditions coincide, we obtain
\begin{equation}
\mathbf Z^{(\sigma)}(\tau)-\mathbf Z_0(\tau)
=
\sigma^2
\int_0^\tau
e^{(\tau-u)\mathcal A}
\mathcal C
\left(
\mathbf Z^{(\sigma)}(u)
\right)
\,du.
\label{eq:linear-L2-difference}
\end{equation}

By the energy estimate
\eqref{eq:linear-L2-normalized-energy-gronwall},
\begin{equation}
\|\mathbf Z^{(\sigma)}(u)\|
\le
e^{s_{\max}u}\|(\mathbf A,\mathbf B)\|.
\end{equation}
Combining this with the semigroup and cubic bounds gives
\begin{equation}
\begin{aligned}
\left\|
\mathbf Z^{(\sigma)}(\tau)-\mathbf Z_0(\tau)
\right\|
&\le
\sigma^2
\|(\mathbf A,\mathbf B)\|^3
\int_0^\tau
e^{s_{\max}(\tau-u)}
e^{3s_{\max}u}\,du
\\
&\le
\sigma^2
\tau_0e^{3s_{\max}\tau_0}
\|(\mathbf A,\mathbf B)\|^3
\end{aligned}
\end{equation}
for every \(0\le\tau\le\tau_0\).  Hence
\begin{equation}
\sup_{0\le\tau\le\tau_0}
\left\|
\mathbf Z^{(\sigma)}(\tau)-\mathbf Z_0(\tau)
\right\|
=
O(\sigma^2).
\label{eq:linear-L2-uniform-linearization-error}
\end{equation}
This proves
\eqref{eq:linear-L2-factor-convergence}.

\emph{Step 3: convergence and microscopicity of the predictor.}
By definition,
\begin{equation}
\frac{\mathbf M^{(\sigma)}(\tau)}{\sigma^2}
=
\widehat{\mathbf V}^{(\sigma)}(\tau)^{\top}
\widehat{\mathbf U}^{(\sigma)}(\tau).
\label{eq:linear-L2-normalized-predictor-product}
\end{equation}
To compare this predictor with its limiting counterpart
\(\mathbf M_0=\mathbf V_0^{\top}\mathbf U_0\), we add and subtract
\(\mathbf V_0^{\top}\widehat{\mathbf U}^{(\sigma)}\):
\begin{equation}
\begin{aligned}
\frac{\mathbf M^{(\sigma)}}{\sigma^2}
-
\mathbf M_0
&=
\widehat{\mathbf V}^{(\sigma)\top}
\widehat{\mathbf U}^{(\sigma)}
-
\mathbf V_0^{\top}\mathbf U_0
\\
&=
\left(
\widehat{\mathbf V}^{(\sigma)}-\mathbf V_0
\right)^{\top}
\widehat{\mathbf U}^{(\sigma)}
+
\mathbf V_0^{\top}
\left(
\widehat{\mathbf U}^{(\sigma)}-\mathbf U_0
\right),
\end{aligned}
\end{equation}
where all quantities are evaluated at the same time \(\tau\).
Applying the triangle inequality and submultiplicativity of the
Frobenius norm, we obtain
\begin{equation}
\begin{aligned}
\left\|
\frac{\mathbf M^{(\sigma)}(\tau)}{\sigma^2}
-
\mathbf M_0(\tau)
\right\|_{\mathrm F}
&\le
\left\|
\widehat{\mathbf V}^{(\sigma)}(\tau)-\mathbf V_0(\tau)
\right\|_{\mathrm F}
\left\|
\widehat{\mathbf U}^{(\sigma)}(\tau)
\right\|_{\mathrm F}
\\
&\quad+
\left\|
\mathbf V_0(\tau)
\right\|_{\mathrm F}
\left\|
\widehat{\mathbf U}^{(\sigma)}(\tau)-\mathbf U_0(\tau)
\right\|_{\mathrm F}.
\end{aligned}
\label{eq:linear-L2-predictor-product-bound}
\end{equation}

Both terms on the right-hand side converge uniformly to zero as
\(\sigma\downarrow0\). Indeed, Step~2 established uniform convergence
of each normalized factor to its limiting counterpart, while the
energy estimate \eqref{eq:linear-L2-uniform-Z-bound} provides a
uniform bound on the exact normalized weights. The limiting weights
are likewise bounded on the fixed interval \([0,\tau_0]\).
Consequently, the normalized predictor converges uniformly to
\(\mathbf M_0\), proving
\eqref{eq:linear-L2-predictor-convergence}.

We next show that the unnormalized predictor remains microscopic
throughout the interval \([0,\tau_0]\). Recall that
\(\mathbf M^{(\sigma)}
=
\sigma^2\widehat{\mathbf V}^{(\sigma)\top}
\widehat{\mathbf U}^{(\sigma)}\).
By submultiplicativity of the Frobenius norm and the inequality
\(2ab\le a^2+b^2\), we have, at every time \(\tau\),
\begin{equation}
\begin{aligned}
\|\mathbf M^{(\sigma)}(\tau)\|_{\mathrm F}
&=
\sigma^2
\left\|
\widehat{\mathbf V}^{(\sigma)}(\tau)^{\top}
\widehat{\mathbf U}^{(\sigma)}(\tau)
\right\|_{\mathrm F}
\\
&\le
\sigma^2
\|\widehat{\mathbf V}^{(\sigma)}(\tau)\|_{\mathrm F}
\|\widehat{\mathbf U}^{(\sigma)}(\tau)\|_{\mathrm F}
\\
&\le
\frac{\sigma^2}{2}
\left(
\|\widehat{\mathbf U}^{(\sigma)}(\tau)\|_{\mathrm F}^2
+
\|\widehat{\mathbf V}^{(\sigma)}(\tau)\|_{\mathrm F}^2
\right)
=
\frac{\sigma^2}{2}S^{(\sigma)}(\tau).
\end{aligned}
\end{equation}
Thus the predictor norm is controlled directly by the normalized
factor energy. Applying the uniform energy estimate
\eqref{eq:linear-L2-normalized-energy-gronwall}, we obtain
\begin{equation}
\begin{aligned}
\sup_{0\le\tau\le\tau_0}
\|\mathbf M^{(\sigma)}(\tau)\|_{\mathrm F}
&\le
\frac{\sigma^2}{2}
e^{2s_{\max}\tau_0}
\left(
\|\mathbf A\|_{\mathrm F}^2
+
\|\mathbf B\|_{\mathrm F}^2
\right)
=
O(\sigma^2).
\end{aligned}
\label{eq:linear-L2-predictor-Frobenius-O-sigma2}
\end{equation}
The constant in this bound is independent of \(\sigma\), since the
initialization matrices and the time horizon are fixed.

Finally, the operator norm is bounded by the Frobenius norm.
Consequently,
\begin{equation}
\begin{aligned}
\sup_{0\le\tau\le\tau_0}
\|\mathbf M^{(\sigma)}(\tau)\|_{\mathrm{op}}
&\le
\sup_{0\le\tau\le\tau_0}
\|\mathbf M^{(\sigma)}(\tau)\|_{\mathrm F}
=
O(\sigma^2),
\end{aligned}
\label{eq:linear-L2-predictor-O-sigma2}
\end{equation}
which proves
\eqref{eq:linear-L2-predictor-microscopic}.
In particular, the entire predictor trajectory converges uniformly
to zero on every fixed time interval as \(\sigma\downarrow0\).

\emph{Step 4: microscopicity of the loss change.}
The quadratic loss representation
\eqref{eq:linear-L2-factorization-loss} implies
\begin{equation}
\mathcal L(\mathbf U,\mathbf V)
-
\mathcal L(\mathbf 0,\mathbf 0)
=
\gamma
\left(
\frac12\|\mathbf M\|_{\mathrm F}^2
-
\langle\mathbf M,\mathbf T\rangle_{\mathrm F}
\right).
\label{eq:linear-L2-loss-difference}
\end{equation}
Consequently,
\begin{equation}
\left|
\mathcal L(\mathbf U,\mathbf V)
-
\mathcal L(\mathbf 0,\mathbf 0)
\right|
\le
\gamma
\left(
\frac12\|\mathbf M\|_{\mathrm F}^2
+
\|\mathbf M\|_{\mathrm F}
\|\mathbf T\|_{\mathrm F}
\right).
\label{eq:linear-L2-loss-bound}
\end{equation}
Since
\(\sup_{0\le\tau\le\tau_0}
\|\mathbf M^{(\sigma)}(\tau)\|_{\mathrm F}
=
O(\sigma^2)\)
and \(\mathbf T\) is fixed, the right-hand side is uniformly
\(O(\sigma^2)\).  This proves
\eqref{eq:linear-L2-loss-microscopic} and completes the proof.
\end{proof}

\noindent\emph{Scope of the microscopic approximation.}
The preceding theorem establishes convergence to the linearized
dynamics on every fixed finite time interval as
\(\sigma\downarrow0\), with the problem dimensions held fixed.
It does not establish uniform convergence on time scales that
diverge as \(\sigma\downarrow0\), such as
\(\tau\asymp\log(1/\sigma)\), where nonlinear effects may become
significant. In the sequential small-initialization and large-system limit,
the time horizon may depend on the problem dimensions, since
these remain fixed during the inner \(\sigma\downarrow0\) limit.

We now specialize to the Gaussian initialization used throughout
the subsequent analysis. Specifically, let
\(\mathbf A\in\mathbb R^{H\times p}\) and
\(\mathbf B\in\mathbb R^{H\times C}\) have mutually independent
standard Gaussian entries:
\begin{equation}
A_{hj},B_{hc}
\stackrel{\mathrm{iid}}{\sim}
\mathcal N(0,1).
\label{eq:linear-L2-standard-gaussian-initialization}
\end{equation}
Together with \eqref{eq:linear-L2-scaled-initialization},
this corresponds to independent Gaussian initialization of the
network weights with entrywise variance \(\sigma^2\).
Since the preceding theorem holds for every fixed pair of finite
initialization matrices, all its conclusions hold almost surely
under this Gaussian initialization.

\paragraph{Explicit solution of the microscopic dynamics.}

We now solve the limiting linear system
\eqref{eq:linear-L2-linearized-flow}.  Using the compact singular-value
decomposition
\eqref{eq:linear-L2-compact-svd}, define the coordinates of the limiting
weights in the positive right and left singular subspaces by
\begin{equation}
\mathbf X_0(\tau)
=
\mathbf U_0(\tau)\mathbf Q,
\qquad
\mathbf Y_0(\tau)
=
\mathbf V_0(\tau)\mathbf P.
\label{eq:linear-L2-active-coordinates}
\end{equation}
The limiting equations then decouple across singular modes:
\begin{equation}
\dot{\mathbf X}_0
=
\mathbf Y_0\boldsymbol{\Lambda},
\qquad
\dot{\mathbf Y}_0
=
\mathbf X_0\boldsymbol{\Lambda}.
\label{eq:linear-L2-active-flow}
\end{equation}
The components orthogonal to the positive singular subspaces remain
constant. Indeed, the SVD
\eqref{eq:linear-L2-compact-svd} implies
\(\mathbf T(\mathbf I_p-\mathbf Q\mathbf Q^\top)=0\) and
\(\mathbf T^\top(\mathbf I_C-\mathbf P\mathbf P^\top)=0\).
Multiplying the limiting equations
\eqref{eq:linear-L2-linearized-flow} by the corresponding
orthogonal projectors therefore gives
\begin{equation}
\begin{aligned}
\mathbf U_0(\tau)
\left(
\mathbf I_p-\mathbf Q\mathbf Q^{\top}
\right)
&=
\mathbf A
\left(
\mathbf I_p-\mathbf Q\mathbf Q^{\top}
\right),
\\
\mathbf V_0(\tau)
\left(
\mathbf I_C-\mathbf P\mathbf P^{\top}
\right)
&=
\mathbf B
\left(
\mathbf I_C-\mathbf P\mathbf P^{\top}
\right).
\end{aligned}
\label{eq:linear-L2-null-components}
\end{equation}

Adding and subtracting the equations
\eqref{eq:linear-L2-active-flow} shows that
\(\mathbf X_0+\mathbf Y_0\) and
\(\mathbf X_0-\mathbf Y_0\) evolve independently,
with growth rates \(\boldsymbol{\Lambda}\) and
\(-\boldsymbol{\Lambda}\), respectively. Define
\begin{equation}
\boldsymbol{\Gamma}
=
\frac{\mathbf A\mathbf Q+\mathbf B\mathbf P}{\sqrt2},
\qquad
\boldsymbol{\Xi}
=
\frac{\mathbf A\mathbf Q-\mathbf B\mathbf P}{\sqrt2}.
\label{eq:linear-L2-Gamma-Xi}
\end{equation}
Then
\begin{equation}
\begin{aligned}
\mathbf X_0(\tau)
&=
\frac{1}{\sqrt2}
\left(
\boldsymbol{\Gamma}e^{\boldsymbol{\Lambda}\tau}
+
\boldsymbol{\Xi}e^{-\boldsymbol{\Lambda}\tau}
\right),
\\
\mathbf Y_0(\tau)
&=
\frac{1}{\sqrt2}
\left(
\boldsymbol{\Gamma}e^{\boldsymbol{\Lambda}\tau}
-
\boldsymbol{\Xi}e^{-\boldsymbol{\Lambda}\tau}
\right).
\end{aligned}
\label{eq:linear-L2-explicit-active-flow}
\end{equation}
Thus each singular mode \(j\) contains a growing branch proportional to
\(e^{s_j\tau}\) and a decaying branch proportional to
\(e^{-s_j\tau}\).  Different singular values therefore generate
different microscopic growth rates.

Under the Gaussian initialization
\eqref{eq:linear-L2-standard-gaussian-initialization},
multiplication by matrices with orthonormal columns preserves the
standard Gaussian law.  Hence
\(\mathbf A\mathbf Q\) and \(\mathbf B\mathbf P\) are independent
standard Gaussian matrices in \(\mathbb R^{H\times r}\).
The transformation in
\eqref{eq:linear-L2-Gamma-Xi} is orthogonal, so
\(\boldsymbol{\Gamma}\) and \(\boldsymbol{\Xi}\) are likewise
independent standard Gaussian matrices.

For later use, we record the corresponding predictor within the
positive singular subspaces:
\begin{equation}
\mathbf N_0(\tau)
=
\mathbf P^{\top}
\mathbf M_0(\tau)
\mathbf Q
=
\mathbf Y_0(\tau)^{\top}\mathbf X_0(\tau).
\label{eq:linear-L2-limiting-spectral-predictor}
\end{equation}
Writing
\begin{equation}
\mathbf E(\tau)
=
e^{\boldsymbol{\Lambda}\tau},
\qquad
\mathbf F(\tau)
=
e^{-\boldsymbol{\Lambda}\tau},
\label{eq:linear-L2-E-F}
\end{equation}
the explicit active-coordinate solution
\eqref{eq:linear-L2-explicit-active-flow} gives
\begin{equation}
\begin{aligned}
\mathbf N_0(\tau)
=
\frac12
\Big(
&
\mathbf E(\tau)
\boldsymbol{\Gamma}^{\top}
\boldsymbol{\Gamma}
\mathbf E(\tau)
+
\mathbf E(\tau)
\boldsymbol{\Gamma}^{\top}
\boldsymbol{\Xi}
\mathbf F(\tau)
\\
&
-
\mathbf F(\tau)
\boldsymbol{\Xi}^{\top}
\boldsymbol{\Gamma}
\mathbf E(\tau)
-
\mathbf F(\tau)
\boldsymbol{\Xi}^{\top}
\boldsymbol{\Xi}
\mathbf F(\tau)
\Big).
\end{aligned}
\label{eq:linear-L2-explicit-spectral-predictor}
\end{equation}
This explicit random matrix process will be useful below when we analyze
finite-width spectral interference.

\paragraph{Early target-driven organization and target accuracy.}

The microscopic linearization describes the predictor while its norm and the
change in loss are still vanishing, but it does not by itself determine the
induced classifier. We now show that, at a sufficiently early time, the linearized gradient-flow
dynamics produces a mean predictor aligned with the full effective target
\(\mathbf T\), while, for sufficiently large \(H\), averaging over the hidden units
suppresses the initialization-dependent fluctuations relative to
this mean. Together, these two
effects make the predictor agree with the classifier of \(\mathbf T\).

For each training example, let
\begin{equation}
\widehat c_i
=
\arg\max_{c\in[C]}
(\mathbf T\mathbf x_i)_c
\label{eq:linear-L2-target-predicted-class}
\end{equation}
be the class predicted by \(\mathbf T\). We assume that this maximizer is
unique for every \(i\), and define the minimum decision gap
\begin{equation}
\delta_{\mathbf T}
=
\min_{1\le i\le N}
\left[
(\mathbf T\mathbf x_i)_{\widehat c_i}
-
\max_{c\neq\widehat c_i}
(\mathbf T\mathbf x_i)_c
\right]
>0.
\label{eq:linear-L2-target-decision-gap}
\end{equation}
Recall that
\begin{equation}
A_{\mathbf T}
=
\Acc(\mathbf T)
=
\frac1N
\sum_{i=1}^N
\mathbf 1\{\widehat c_i=c_i\}.
\label{eq:linear-L2-target-accuracy}
\end{equation}
Thus the result below concerns agreement with the classifier induced by
\(\mathbf T\), whether or not that classifier has perfect accuracy.

We consider a sequence of problems indexed by \(H\to\infty\), allowing
\(p,C,N,\mathbf T\), and the dataset to vary with \(H\). Define
\begin{equation}
R
=
\max_{1\le i\le N}\|\mathbf x_i\|_2,
\qquad
s
=
\|\mathbf T\|_{\mathrm{op}},
\qquad
\ell
=
\log(2NCH),
\qquad
\kappa
=
\frac{Rs}{\delta_{\mathbf T}}.
\label{eq:linear-L2-early-parameters}
\end{equation}
The following theorem gives a sufficient width condition under which
the limiting and exact predictors agree with the classifier of
\(\mathbf T\) on every training example.

\begin{theorem}[Early agreement with the target classifier]
\label{thm:linear-L2-early-target-classifier}
Suppose that
\begin{equation}
\kappa^3\frac{\ell}{H}
\xrightarrow[H\to\infty]{}0.
\label{eq:linear-L2-early-width-condition}
\end{equation}
Then there exists a universal constant \(K_0>0\) such that, for every fixed
\(K>K_0\), at the time
\begin{equation}
\tau_{\mathrm{org}}
=
K\frac{R}{\delta_{\mathbf T}}
\sqrt{\frac{\ell}{H}},
\label{eq:linear-L2-organization-time}
\end{equation}
the limiting predictor \(\mathbf M_0\) from
Theorem~\ref{thm:linear-L2-microscopic-linearization} satisfies
\begin{equation}
\mathbb P
\left(
\arg\max_{c\in[C]}
(\mathbf M_0(\tau_{\mathrm{org}})\mathbf x_i)_c
=
\widehat c_i
\quad
\text{for every }i
\right)
\xrightarrow[H\to\infty]{}1.
\label{eq:linear-L2-early-linearized-classifier}
\end{equation}

Consequently, for the exact Gaussian-initialized gradient flow,
\begin{equation}
\lim_{H\to\infty}
\liminf_{\sigma\downarrow0}
\mathbb P
\left(
\arg\max_{c\in[C]}
(\mathbf M^{(\sigma)}(\tau_{\mathrm{org}})\mathbf x_i)_c
=
\widehat c_i
\quad
\text{for every }i
\right)
=
1.
\label{eq:linear-L2-early-exact-classifier}
\end{equation}
In particular,
\begin{equation}
\lim_{H\to\infty}
\liminf_{\sigma\downarrow0}
\mathbb P
\left(
\Acc
\bigl(
\mathbf M^{(\sigma)}(\tau_{\mathrm{org}})
\bigr)
=
A_{\mathbf T}
\right)
=
1.
\label{eq:linear-L2-early-exact-accuracy}
\end{equation}

Moreover, for every fixed member of the large-system sequence,
almost surely as \(\sigma\downarrow0\), the entire trajectory
up to the organization time remains microscopic:
\begin{equation}
\sup_{0\le\tau\le\tau_{\mathrm{org}}}
\left\|
\mathbf M^{(\sigma)}(\tau)
\right\|_{\mathrm{op}}
\longrightarrow0,
\qquad
\sup_{0\le\tau\le\tau_{\mathrm{org}}}
\left|
\mathcal L
\left(
\mathbf U^{(\sigma)}(\tau),
\mathbf V^{(\sigma)}(\tau)
\right)
-
\mathcal L(\mathbf 0,\mathbf 0)
\right|
\longrightarrow0.
\label{eq:linear-L2-early-loss-flat}
\end{equation}
\end{theorem}

\begin{proof}
We first determine the mean predictor of the limiting linearized dynamics.

\smallskip
\noindent
\emph{Step 1: mean microscopic predictor.}
Write the rows of the Gaussian initialization as
\begin{equation}
\mathbf A
=
\begin{pmatrix}
\mathbf a_1^{\top}\\
\vdots\\
\mathbf a_H^{\top}
\end{pmatrix},
\qquad
\mathbf B
=
\begin{pmatrix}
\mathbf b_1^{\top}\\
\vdots\\
\mathbf b_H^{\top}
\end{pmatrix},
\qquad
\mathbf a_h\sim\mathcal N(\mathbf 0,\mathbf I_p),
\qquad
\mathbf b_h\sim\mathcal N(\mathbf 0,\mathbf I_C),
\label{eq:linear-L2-row-initialization}
\end{equation}
independently across \(h\) and between the two families.
Likewise, denote the transposes of the rows of
\(\mathbf U_0(\tau)\) and \(\mathbf V_0(\tau)\) by
\(\mathbf u_h(\tau)\) and \(\mathbf v_h(\tau)\), respectively.
The limiting linearized dynamics
\eqref{eq:linear-L2-linearized-flow} decouple across hidden units:
\begin{equation}
\dot{\mathbf u}_h
=
\mathbf T^{\top}\mathbf v_h,
\qquad
\dot{\mathbf v}_h
=
\mathbf T\mathbf u_h,
\qquad
\mathbf u_h(0)=\mathbf a_h,
\qquad
\mathbf v_h(0)=\mathbf b_h.
\label{eq:linear-L2-single-unit-linearized-flow}
\end{equation}
Consequently, the limiting predictor is the sum of the
contributions from individual hidden units:
\begin{equation}
\mathbf M_0(\tau)
=
\sum_{h=1}^H
\mathbf v_h(\tau)\mathbf u_h(\tau)^{\top}.
\label{eq:linear-L2-M0-hidden-sum}
\end{equation}

We compute its expectation using the explicit solution obtained
in the preceding subsection. Recall that
\(\boldsymbol{\Gamma}\) and \(\boldsymbol{\Xi}\), defined in
\eqref{eq:linear-L2-Gamma-Xi}, are independent standard Gaussian
matrices in \(\mathbb R^{H\times r}\). Therefore,
\begin{equation}
\mathbb E[
\boldsymbol{\Gamma}^{\top}\boldsymbol{\Gamma}
]
=
\mathbb E[
\boldsymbol{\Xi}^{\top}\boldsymbol{\Xi}
]
=
H\mathbf I_r,
\qquad
\mathbb E[
\boldsymbol{\Gamma}^{\top}\boldsymbol{\Xi}
]
=
\mathbf 0.
\end{equation}
Taking expectations in the explicit spectral predictor
\eqref{eq:linear-L2-explicit-spectral-predictor},
the mixed terms vanish, and we obtain
\begin{equation}
\begin{aligned}
\mathbb E[\mathbf N_0(\tau)]
&=
\frac{H}{2}
\left(
\mathbf E(\tau)^2-\mathbf F(\tau)^2
\right)
\\
&=
\frac{H}{2}
\left(
e^{2\boldsymbol{\Lambda}\tau}
-
e^{-2\boldsymbol{\Lambda}\tau}
\right)
\\
&=
H\sinh(2\boldsymbol{\Lambda}\tau),
\end{aligned}
\end{equation}
where the hyperbolic sine acts entrywise on the diagonal matrix
\(2\boldsymbol{\Lambda}\tau\).

The components of the limiting weights orthogonal to the
positive singular subspaces are given by
\eqref{eq:linear-L2-null-components}. Under Gaussian
initialization, these components are centered and independent
of the active components, so their contributions to the mean
predictor vanish. Thus
\(\mathbb E[\mathbf M_0(\tau)]
=
\mathbf P\,\mathbb E[\mathbf N_0(\tau)]\,\mathbf Q^\top\).

Since the hidden-unit contributions in
\eqref{eq:linear-L2-M0-hidden-sum} are independent and identically
distributed, the preceding calculation also gives
\begin{equation}
\mathbb E
\left[
\mathbf v_h(\tau)\mathbf u_h(\tau)^{\top}
\right]
=
\mathbf P
\sinh(2\boldsymbol{\Lambda}\tau)
\mathbf Q^{\top}.
\label{eq:linear-L2-single-unit-mean}
\end{equation}
We therefore define the normalized mean predictor by
\begin{equation}
\overline{\mathbf M}(\tau)
=
\mathbf P
\sinh(2\boldsymbol{\Lambda}\tau)
\mathbf Q^{\top}
=
\sum_{j=1}^r
\sinh(2s_j\tau)
\mathbf p_j\mathbf q_j^{\top}.
\label{eq:linear-L2-wide-microscopic-predictor}
\end{equation}
Summing the identical mean contributions of all \(H\) hidden units,
we conclude that
\begin{equation}
\mathbb E[\mathbf M_0(\tau)]
=
H\overline{\mathbf M}(\tau).
\label{eq:linear-L2-M0-mean}
\end{equation}

For the concentration analysis below, it will also be convenient
to express the dynamics of each hidden unit in block form.
Introduce the symmetric matrix
\begin{equation}
\mathbf S
=
\begin{pmatrix}
\mathbf 0 & \mathbf T^{\top}\\
\mathbf T & \mathbf 0
\end{pmatrix}.
\label{eq:linear-L2-block-generator}
\end{equation}
The single-unit system
\eqref{eq:linear-L2-single-unit-linearized-flow}
then has the solution
\begin{equation}
\begin{pmatrix}
\mathbf u_h(\tau)\\
\mathbf v_h(\tau)
\end{pmatrix}
=
e^{\tau\mathbf S}
\begin{pmatrix}
\mathbf a_h\\
\mathbf b_h
\end{pmatrix}.
\label{eq:linear-L2-single-unit-block-solution}
\end{equation}

\smallskip
\noindent
\emph{Step 2: early alignment of the mean predictor with \(\mathbf T\).}
For \(z\ge0\),
\begin{equation}
0
\le
\sinh z-z
\le
\frac{z^3}{6}e^z.
\label{eq:linear-L2-sinh-bound}
\end{equation}
Using the singular-value decomposition of \(\mathbf T\) and
\eqref{eq:linear-L2-wide-microscopic-predictor}, we obtain
\begin{equation}
\begin{aligned}
\left\|
\overline{\mathbf M}(\tau)-2\tau\mathbf T
\right\|_{\mathrm{op}}
&=
\max_{1\le j\le r}
\left|
\sinh(2s_j\tau)-2s_j\tau
\right|
\\
&\le
\max_{1\le j\le r}
\frac{(2s_j\tau)^3}{6}e^{2s_j\tau}
\\
&\le
\frac43 s^3\tau^3e^{2s\tau},
\end{aligned}
\label{eq:linear-L2-wide-target-error}
\end{equation}
where the equality follows from the orthonormality of the
singular vectors, and the inequalities use
\eqref{eq:linear-L2-sinh-bound} and \(s_j\le s\).

For each training example \(i\) and each competing class
\(c\neq\widehat c_i\), define
\begin{equation}
\mathbf d_{ic}
=
\mathbf e_{\widehat c_i}-\mathbf e_c,
\qquad
m_{ic}
=
\mathbf d_{ic}^{\top}
\mathbf T\mathbf x_i.
\label{eq:linear-L2-pairwise-target-margin}
\end{equation}
The corresponding pairwise classification margin of the mean
predictor is
\begin{equation}
\mathbf d_{ic}^{\top}
\overline{\mathbf M}(\tau)\mathbf x_i
=
\left(\overline{\mathbf M}(\tau)\mathbf x_i\right)_{\widehat c_i}
-
\left(\overline{\mathbf M}(\tau)\mathbf x_i\right)_c.
\end{equation}
Thus, to show that the mean predictor agrees with the classifier
of \(\mathbf T\), it suffices to establish that this margin is strictly positive for every training example and every competing class.

By the definition of \(\delta_{\mathbf T}\) and the orthonormality of the standard basis vectors, respectively, we have
\begin{equation}
m_{ic}\ge\delta_{\mathbf T},
\qquad
\|\mathbf d_{ic}\|_2=\sqrt2.
\label{eq:linear-L2-pairwise-target-margin-bound}
\end{equation}
We now bound the pairwise margin of the mean predictor.
The correction to the target-aligned term may have either sign,
so we bound it from below using the operator norm:
\begin{equation}
\begin{aligned}
\mathbf d_{ic}^{\top}
\overline{\mathbf M}(\tau)
\mathbf x_i
&=
\mathbf d_{ic}^{\top}
\left(2\tau\mathbf T
\right)
\mathbf x_i
+
\mathbf d_{ic}^{\top}
\left(
\overline{\mathbf M}(\tau)-2\tau\mathbf T
\right)
\mathbf x_i
\\
&\ge
2\tau m_{ic}
-
\|\mathbf d_{ic}\|_2
\left\|
\overline{\mathbf M}(\tau)-2\tau\mathbf T
\right\|_{\mathrm{op}}
\|\mathbf x_i\|_2
\\
&\ge
2\tau\delta_{\mathbf T}
-
\frac{4\sqrt2}{3}
Rs^3\tau^3e^{2s\tau}.
\end{aligned}
\label{eq:linear-L2-wide-margin-lower-bound}
\end{equation}

We now set \(\tau=\tau_{\mathrm{org}}\). Since
\begin{equation}
\delta_{\mathbf T}
\le
\mathbf d_{ic}^{\top}
\mathbf T\mathbf x_i
\le
\sqrt2\,Rs,
\label{eq:linear-L2-target-gap-upper-bound}
\end{equation}
we have
\begin{equation}
\kappa
=
\frac{Rs}{\delta_{\mathbf T}}
\ge
\frac1{\sqrt2}.
\label{eq:linear-L2-kappa-lower-bound}
\end{equation}
Together with the width condition 
\eqref{eq:linear-L2-early-width-condition} this implies
\begin{equation}
\frac{\ell}{H}
\xrightarrow[H\to\infty]{}0.
\label{eq:linear-L2-log-over-H}
\end{equation}

We first verify that the organization time is sufficiently
small relative to the inverse singular-value scale.
Indeed,
\begin{equation}
\left(
\kappa\sqrt{\frac{\ell}{H}}
\right)^3
=
\left(
\kappa^3\frac{\ell}{H}
\right)
\sqrt{\frac{\ell}{H}}
\xrightarrow[H\to\infty]{}0,
\label{eq:linear-L2-early-spectral-time-proof}
\end{equation}
because both factors on the right-hand side vanish.
Consequently, recalling the definition of
\(\tau_{\mathrm{org}}\) in
\eqref{eq:linear-L2-organization-time}, we obtain
\begin{equation}
s\tau_{\mathrm{org}}
=
K\kappa\sqrt{\frac{\ell}{H}}
\xrightarrow[H\to\infty]{}0,
\label{eq:linear-L2-early-spectral-time}
\end{equation}
since \(K\) is fixed independently of \(H\).

We now show that the approximation error in the pairwise
margin bound \eqref{eq:linear-L2-wide-margin-lower-bound}
is negligible compared with the leading target-aligned
term. At \(\tau=\tau_{\mathrm{org}}\), the ratio of the
error scale to \(\tau_{\mathrm{org}}\delta_{\mathbf T}\)
is
\begin{equation}
\frac{
Rs^3\tau_{\mathrm{org}}^3
e^{2s\tau_{\mathrm{org}}}
}{
\tau_{\mathrm{org}}\delta_{\mathbf T}
}
=
K^2\kappa^3\frac{\ell}{H}
e^{2s\tau_{\mathrm{org}}}
\xrightarrow[H\to\infty]{}0.
\label{eq:linear-L2-wide-margin-relative-error}
\end{equation}
Here the convergence follows from the width condition
\eqref{eq:linear-L2-early-width-condition} and
\(s\tau_{\mathrm{org}}\to0\), which ensures that the
exponential factor tends to one.

Thus the correction to the target-aligned margin is
asymptotically negligible. In particular, for all
sufficiently large \(H\), the bound
\eqref{eq:linear-L2-wide-margin-lower-bound} yields
\begin{equation}
\mathbf d_{ic}^{\top}
\overline{\mathbf M}(\tau_{\mathrm{org}})
\mathbf x_i
\ge
2\tau_{\mathrm{org}}\delta_{\mathbf T}
-
\frac{4\sqrt2}{3}
Rs^3\tau_{\mathrm{org}}^3
e^{2s\tau_{\mathrm{org}}}
\ge
\frac32
\tau_{\mathrm{org}}\delta_{\mathbf T},
\label{eq:linear-L2-wide-margin-positive}
\end{equation}
uniformly over all training examples \(i\) and competing
classes \(c\neq\widehat c_i\).
The fixed factor \(3/2\) leaves a positive margin that
will accommodate the initialization-dependent fluctuations
in the next step.

\smallskip
\noindent
\emph{Step 3: uniform concentration around the mean.}
We now show that the initialization-dependent fluctuations of
\(\mathbf M_0\) around its mean are sufficiently small to preserve
the positive classification margins established in Step~2.

Fix a training example \(i\) and a competing class
\(c\neq\widehat c_i\), and abbreviate
\(\mathbf d=\mathbf d_{ic}\) and \(\mathbf x=\mathbf x_i\).
Recall that the limiting predictor is the sum of independent
hidden-unit contributions,
\eqref{eq:linear-L2-M0-hidden-sum}, and that its expectation is
\(H\overline{\mathbf M}(\tau)\).
Thus the fluctuation of the pairwise classification margin
around its mean can be written as
\begin{equation}
\mathbf d^{\top}
\left(
\mathbf M_0(\tau)
-
H\overline{\mathbf M}(\tau)
\right)
\mathbf x
=
\sum_{h=1}^H Z_h,
\label{eq:linear-L2-margin-fluctuation-sum}
\end{equation}
where
\begin{equation}
Z_h
=
\left(
\mathbf d^{\top}\mathbf v_h(\tau)
\right)
\left(
\mathbf u_h(\tau)^{\top}\mathbf x
\right)
-
\mathbb E\left[
\left(
\mathbf d^{\top}\mathbf v_h(\tau)
\right)
\left(
\mathbf u_h(\tau)^{\top}\mathbf x
\right)
\right].
\label{eq:linear-L2-Zh}
\end{equation}
The variables \(Z_h\) are centered and independent because
the Gaussian initializations are independent across hidden
units and the limiting dynamics evolve independently for
each unit.

To control these fluctuations, recall from
\eqref{eq:linear-L2-single-unit-block-solution} that the
single-unit dynamics are obtained by applying
\(e^{\tau\mathbf S}\) to a standard Gaussian vector.
Since \(\mathbf S\) is symmetric and
\(\|\mathbf S\|_{\mathrm{op}}=s\), we have
\begin{equation}
\|e^{\tau\mathbf S}\|_{\mathrm{op}}
\le e^{s\tau}.
\label{eq:linear-L2-block-semigroup-bound}
\end{equation}
Consequently, the two scalar variables
\(\mathbf d^\top\mathbf v_h(\tau)\) and
\(\mathbf u_h(\tau)^\top\mathbf x\) are jointly centered
Gaussian, with standard deviations bounded by
\(e^{s\tau}\|\mathbf d\|_2\) and
\(e^{s\tau}\|\mathbf x\|_2\), respectively.

The product of two centered Gaussian variables is
sub-exponential, even when the variables are correlated,
and the same holds after subtracting its expectation.
Together with the preceding standard deviation bounds,
this allows us to apply Bernstein's inequality for
independent centered sub-exponential variables
\citep[Lemma~2.8.6 and Theorem~2.9.1]{vershynin2026high}.
Consequently, there exists a universal constant \(C_1>0\)
such that, for every \(t\ge1\),
\begin{equation}
\mathbb P\Bigg(
\left|
\mathbf d^{\top}
\left(
\mathbf M_0(\tau)-H\overline{\mathbf M}(\tau)
\right)
\mathbf x
\right|
>
C_1e^{2s\tau}R
\left(
\sqrt{Ht}+t
\right)
\Bigg)
\le
2e^{-t}.
\label{eq:linear-L2-margin-Bernstein}
\end{equation}

We next make this concentration bound uniform over all
training examples and competing classes.
Taking \(t=\ell=\log(2NCH)\) and applying a union bound
over the at most \(N(C-1)\) pairwise comparisons, we obtain
an event \(\mathcal E_H\) satisfying
\begin{equation}
\mathbb P(\mathcal E_H^c)
\le
2N(C-1)e^{-\ell}
\le
\frac1H,
\label{eq:linear-L2-margin-union-probability}
\end{equation}
such that, on \(\mathcal E_H\), the fluctuations at the
organization time obey
\begin{equation}
\max_{\substack{1\le i\le N\\c\neq\widehat c_i}}
\left|
\mathbf d_{ic}^{\top}
\left(
\mathbf M_0(\tau_{\mathrm{org}})
-
H\overline{\mathbf M}(\tau_{\mathrm{org}})
\right)
\mathbf x_i
\right|
\le
C_1e^{2s\tau_{\mathrm{org}}}R
\left(
\sqrt{H\ell}+\ell
\right).
\label{eq:linear-L2-uniform-margin-fluctuation}
\end{equation}

It remains to compare these random fluctuations with the
positive deterministic margins obtained in Step~2.
Recall that the latter are bounded below by
\(\frac32 H\tau_{\mathrm{org}}\delta_{\mathbf T}\).
By the definition of the organization time,
\begin{equation}
H\tau_{\mathrm{org}}\delta_{\mathbf T}
=
KR\sqrt{H\ell}.
\label{eq:linear-L2-organization-signal-size}
\end{equation}
Dividing the fluctuation bound by this reference signal
scale gives
\begin{equation}
\frac{
C_1e^{2s\tau_{\mathrm{org}}}R
\left(\sqrt{H\ell}+\ell\right)
}{
H\tau_{\mathrm{org}}\delta_{\mathbf T}
}
=
\frac{C_1}{K}
e^{2s\tau_{\mathrm{org}}}
\left(
1+\sqrt{\frac{\ell}{H}}
\right)
=
\frac{C_1}{K}+o(1).
\label{eq:linear-L2-noise-to-signal}
\end{equation}
Here we used the conclusions of Step~2 that
\(s\tau_{\mathrm{org}}\to0\) and \(\ell/H\to0\),
with \(K\) held fixed as \(H\to\infty\).

We can now choose the universal threshold \(K_0\) appearing
in the theorem. Set \(K_0=2C_1\).
For every fixed \(K>K_0\), the preceding ratio is strictly
smaller than \(1/2\) for all sufficiently large \(H\).
Therefore, on \(\mathcal E_H\),
\begin{equation}
\max_{\substack{1\le i\le N\\c\neq\widehat c_i}}
\left|
\mathbf d_{ic}^{\top}
\left(
\mathbf M_0(\tau_{\mathrm{org}})
-
H\overline{\mathbf M}(\tau_{\mathrm{org}})
\right)
\mathbf x_i
\right|
\le
\frac12
H\tau_{\mathrm{org}}\delta_{\mathbf T}.
\label{eq:linear-L2-noise-smaller-than-signal}
\end{equation}

Finally, combining this fluctuation bound with the
deterministic margin bound
\eqref{eq:linear-L2-wide-margin-positive}, we obtain,
simultaneously for every training example \(i\) and
competing class \(c\neq\widehat c_i\),
\begin{equation}
\begin{aligned}
\mathbf d_{ic}^{\top}
\mathbf M_0(\tau_{\mathrm{org}})
\mathbf x_i
&=
H\mathbf d_{ic}^{\top}
\overline{\mathbf M}(\tau_{\mathrm{org}})
\mathbf x_i
+
\mathbf d_{ic}^{\top}
\left(
\mathbf M_0(\tau_{\mathrm{org}})
-
H\overline{\mathbf M}(\tau_{\mathrm{org}})
\right)
\mathbf x_i
\\
&\ge
\frac32
H\tau_{\mathrm{org}}\delta_{\mathbf T}
-
\frac12
H\tau_{\mathrm{org}}\delta_{\mathbf T}
=
H\tau_{\mathrm{org}}\delta_{\mathbf T}
>0.
\end{aligned}
\label{eq:linear-L2-final-positive-margin}
\end{equation}
Thus, on \(\mathcal E_H\), every pairwise classification
margin of \(\mathbf M_0(\tau_{\mathrm{org}})\) is positive.
Hence the limiting predictor agrees with the classifier of
\(\mathbf T\) on the entire training set.
Since \(\mathbb P(\mathcal E_H)\ge1-1/H\to1\),
the desired classifier agreement holds with probability
tending to one as \(H\to\infty\).

\smallskip
\noindent
\emph{Step 4: transfer to the exact gradient flow.}
For each fixed member of the outer large-system sequence,
\(\tau_{\mathrm{org}}\) is a fixed finite time during the inner limit
\(\sigma\downarrow0\). Hence
Theorem~\ref{thm:linear-L2-microscopic-linearization} gives
\begin{equation}
\frac{
\mathbf M^{(\sigma)}(\tau_{\mathrm{org}})
}{
\sigma^2
}
\longrightarrow
\mathbf M_0(\tau_{\mathrm{org}})
\qquad
\text{in Frobenius norm}
\label{eq:linear-L2-transfer-at-organization-time}
\end{equation}
almost surely as \(\sigma\downarrow0\).

On the event established above, every relevant pairwise decision gap of
\(\mathbf M_0(\tau_{\mathrm{org}})\) is strictly positive. Since multiplying
all scores by the positive scalar \(\sigma^2\) does not change their
ordering, convergence in
\eqref{eq:linear-L2-transfer-at-organization-time} implies that, for all
sufficiently small \(\sigma\), the exact predictor makes the same prediction
\(\widehat c_i\) on every training example. Taking the inner
\(\sigma\downarrow0\) limit and then \(H\to\infty\) proves
\eqref{eq:linear-L2-early-exact-classifier} and
\eqref{eq:linear-L2-early-exact-accuracy}.

Finally, for each fixed member of the large-system sequence,
the organization time \(\tau_{\mathrm{org}}\) is finite and
independent of \(\sigma\). Applying
Theorem~\ref{thm:linear-L2-microscopic-linearization}
on \([0,\tau_{\mathrm{org}}]\), the uniform predictor
and loss estimates
\eqref{eq:linear-L2-predictor-microscopic}
and \eqref{eq:linear-L2-loss-microscopic}
directly yield
\eqref{eq:linear-L2-early-loss-flat}.
This completes the proof.
\end{proof}

The proof reveals how early organization emerges from the
competition between a coherent target-aligned signal and
initialization-dependent fluctuations. At the organization time,
the reference signal scale is
\begin{equation}
H\tau_{\mathrm{org}}\delta_{\mathbf T}
=
KR\sqrt{H\ell},
\qquad
\ell=\log(2NCH),
\label{eq:linear-L2-early-signal-scale}
\end{equation}
whereas the uniform fluctuation bound is at most
\((C_1+o(1))R\sqrt{H\ell}\) with probability tending to one.
Thus, although both contributions have the same width dependence,
choosing a sufficiently large fixed \(K\) ensures that the
coherent signal dominates the fluctuations. Meanwhile,
\(s\tau_{\mathrm{org}}\to0\), so the mean predictor remains
asymptotically proportional to the full target \(\mathbf T\),
before substantial spectral reweighting takes place.
This is the target-driven organization stage.

The width condition
\eqref{eq:linear-L2-early-width-condition} has a simple
interpretation when the normalized inverse decision margin
\(\kappa=Rs/\delta_{\mathbf T}\) remains bounded. In this case,
it suffices that \(\log(NCH)=o(H)\), which holds, in
particular, when \(N\) and \(C\) grow at most polynomially
with \(H\). If additionally \(R/\delta_{\mathbf T}=\Theta(1)\)
and \(N,C\) grow at most polynomially with \(H\),
the organization time satisfies
\begin{equation}
\tau_{\mathrm{org}}
\asymp
\sqrt{\frac{\log H}{H}}.
\label{eq:linear-L2-organization-time-scaling}
\end{equation}
Hence classifier agreement can emerge at a vanishing
organization time, while, in the sequential
small-initialization limit, the predictor norm and loss
change remain vanishingly small.

\paragraph{Spectral concentration and the predictive bottleneck.}

The preceding theorem shows that the early linearized dynamics can reproduce
the classifier of the full effective target \(\mathbf T\) while both the
end-to-end predictor norm and the change in loss vanish as
\(\sigma\downarrow0\). We now show that this target-driven organization need
not persist. Since singular modes with larger singular values grow faster,
the microscopic mean predictor progressively concentrates onto the complete
top singular block \(\mathbf T_{k_\star}\). Under suitable width and margin
conditions, the finite-width predictor inherits the classifier of this block,
producing the predictive bottleneck.

Recall from \eqref{eq:linear-L2-wide-microscopic-predictor} that
\begin{equation}
\overline{\mathbf M}(\tau)
=
\sum_{j=1}^r
\sinh(2s_j\tau)\,
\mathbf p_j\mathbf q_j^{\top}.
\label{eq:linear-L2-wide-spectral-expansion}
\end{equation}
We first record the deterministic concentration of this mean predictor.

\begin{lemma}[Spectral concentration onto the top singular block]
\label{lem:linear-L2-spectral-concentration}
Let \(k_\star\) be defined by
\eqref{eq:linear-L2-top-multiplicity}, and set
\begin{equation}
\rho_\star(\tau)
:=
\frac{
\sinh(2s_{k_\star+1}\tau)
}{
\sinh(2s_1\tau)
},
\qquad
s_{r+1}:=0.
\label{eq:linear-L2-top-block-rho}
\end{equation}
Then, for every \(\tau>0\),
\begin{equation}
\left\|
\frac{
\overline{\mathbf M}(\tau)
}{
\sinh(2s_1\tau)
}
-
\frac{\mathbf T_{k_\star}}{s_1}
\right\|_{\mathrm{op}}
=
\rho_\star(\tau).
\label{eq:linear-L2-top-block-concentration-bound}
\end{equation}
Equivalently,
\begin{equation}
\left\|
\frac{
s_1\overline{\mathbf M}(\tau)
}{
\sinh(2s_1\tau)
}
-
\mathbf T_{k_\star}
\right\|_{\mathrm{op}}
=
s_1\rho_\star(\tau).
\label{eq:linear-L2-top-block-concentration-scaled}
\end{equation}
In particular,
\begin{equation}
\rho_\star(\tau)
\xrightarrow[\tau\to\infty]{}0,
\label{eq:linear-L2-top-block-rho-limit}
\end{equation}
and hence
\begin{equation}
\frac{
s_1\overline{\mathbf M}(\tau)
}{
\sinh(2s_1\tau)
}
\xrightarrow[\tau\to\infty]{}
\mathbf T_{k_\star}
\qquad
\text{in operator norm}.
\label{eq:linear-L2-top-block-limit}
\end{equation}
\end{lemma}

\begin{proof}
Since
\(s_1=\cdots=s_{k_\star}\),
dividing \eqref{eq:linear-L2-wide-spectral-expansion} by
\(\sinh(2s_1\tau)\) gives
\begin{equation}
\frac{
\overline{\mathbf M}(\tau)
}{
\sinh(2s_1\tau)
}
-
\frac{\mathbf T_{k_\star}}{s_1}
=
\sum_{j=k_\star+1}^r
\frac{
\sinh(2s_j\tau)
}{
\sinh(2s_1\tau)
}
\mathbf p_j\mathbf q_j^{\top}.
\label{eq:linear-L2-top-block-remainder}
\end{equation}
The left and right singular vectors in the sum form orthonormal families, so
the operator norm of the remainder equals its largest coefficient. Since
\(s_{k_\star+1}\ge s_j\) for every \(j>k_\star\), this coefficient is
\(\rho_\star(\tau)\), proving
\eqref{eq:linear-L2-top-block-concentration-bound}. Multiplying by \(s_1\)
gives \eqref{eq:linear-L2-top-block-concentration-scaled}.

By definition of \(k_\star\),
\(s_{k_\star+1}<s_1\) whenever \(k_\star<r\), so
\(\rho_\star(\tau)\to0\). If \(k_\star=r\), then
\(s_{k_\star+1}=0\) and \(\rho_\star(\tau)=0\) identically. This proves the
remaining statements.
\end{proof}

The lemma concerns the mean predictor. We next show that the
finite-width microscopic predictor has the same classifier as
\(\mathbf T_{k_\star}\) at a suitable time. For each training example,
define
\begin{equation}
\widehat c_i^{(\star)}
=
\arg\max_{c\in[C]}
(\mathbf T_{k_\star}\mathbf x_i)_c,
\label{eq:linear-L2-top-block-predicted-class}
\end{equation}
and assume that the maximizer is unique for every \(i\). Define the minimum
decision gap
\begin{equation}
\delta_{k_\star}
=
\min_{1\le i\le N}
\left[
(\mathbf T_{k_\star}\mathbf x_i)_{\widehat c_i^{(\star)}}
-
\max_{c\neq\widehat c_i^{(\star)}}
(\mathbf T_{k_\star}\mathbf x_i)_c
\right]
>0.
\label{eq:linear-L2-top-block-decision-gap}
\end{equation}

\begin{theorem}[Microscopic top-block predictive bottleneck]
\label{thm:linear-L2-top-block-bottleneck}
Consider a large-system sequence indexed by \(H\to\infty\),
and let \(\tau_{\mathrm{org}}\) be the early organization time
from Theorem~\ref{thm:linear-L2-early-target-classifier}.
Suppose that there exists a sequence of finite bottleneck
times \(\tau_{\mathrm{bot}}=\tau_{\mathrm{bot}}(H)\) such that,
for all sufficiently large \(H\),
\begin{equation}
\tau_{\mathrm{bot}}
>
\tau_{\mathrm{org}},
\qquad
s_1\tau_{\mathrm{bot}}\ge1,
\label{eq:linear-L2-bottleneck-time-order}
\end{equation}
and
\begin{equation}
\frac{Rs_1}{\delta_{k_\star}}
\rho_\star(\tau_{\mathrm{bot}})
\xrightarrow[H\to\infty]{}0,
\qquad
\frac{Rs_1}{\delta_{k_\star}}
\sqrt{\frac{\ell}{H}}
\xrightarrow[H\to\infty]{}0,
\label{eq:linear-L2-bottleneck-conditions}
\end{equation}
where
\begin{equation}
R=\max_i\|\mathbf x_i\|_2,
\qquad
\ell=\log(2NCH).
\label{eq:linear-L2-bottleneck-parameters}
\end{equation}

Then the limiting predictor satisfies
\begin{equation}
\mathbb P
\left(
\arg\max_{c\in[C]}
(\mathbf M_0(\tau_{\mathrm{bot}})\mathbf x_i)_c
=
\widehat c_i^{(\star)}
\quad
\text{for every }i
\right)
\xrightarrow[H\to\infty]{}1.
\label{eq:linear-L2-bottleneck-linearized-classifier}
\end{equation}

Consequently, for the exact Gaussian-initialized gradient flow,
\begin{equation}
\lim_{H\to\infty}
\liminf_{\sigma\downarrow0}
\mathbb P
\left(
\arg\max_{c\in[C]}
(\mathbf M^{(\sigma)}(\tau_{\mathrm{bot}})\mathbf x_i)_c
=
\widehat c_i^{(\star)}
\quad
\text{for every }i
\right)
=
1.
\label{eq:linear-L2-bottleneck-exact-classifier}
\end{equation}
In particular,
\begin{equation}
\lim_{H\to\infty}
\liminf_{\sigma\downarrow0}
\mathbb P
\left(
\Acc
\bigl(
\mathbf M^{(\sigma)}(\tau_{\mathrm{bot}})
\bigr)
=
A_{k_\star}
\right)
=
1.
\label{eq:linear-L2-bottleneck-exact-accuracy}
\end{equation}

Moreover, for each fixed member of the large-system sequence,
almost surely as \(\sigma\downarrow0\), the entire trajectory
up to the bottleneck time remains microscopic:
\begin{equation}
\sup_{0\le\tau\le\tau_{\mathrm{bot}}}
\left\|
\mathbf M^{(\sigma)}(\tau)
\right\|_{\mathrm{op}}
\longrightarrow0,
\qquad
\sup_{0\le\tau\le\tau_{\mathrm{bot}}}
\left|
\mathcal L
\left(
\mathbf U^{(\sigma)}(\tau),
\mathbf V^{(\sigma)}(\tau)
\right)
-
\mathcal L(\mathbf 0,\mathbf 0)
\right|
\longrightarrow0.
\label{eq:linear-L2-bottleneck-loss-flat}
\end{equation}
\end{theorem}

\begin{proof}
We first establish that the mean predictor has the same classifier as
\(\mathbf T_{k_\star}\) at the bottleneck time. We then show that the
initialization-dependent fluctuations are too small to alter any of
these predictions.

\smallskip
\noindent
\emph{Step 1: classification by the spectrally concentrated mean.}
For each training example \(i\) and each competing class
\(c\neq\widehat c_i^{(\star)}\), define
\begin{equation}
\mathbf d_{ic}^{(\star)}
=
\mathbf e_{\widehat c_i^{(\star)}}-\mathbf e_c.
\label{eq:linear-L2-top-block-comparison-vector}
\end{equation}
The quantity
\((\mathbf d_{ic}^{(\star)})^\top
\overline{\mathbf M}(\tau)\mathbf x_i\)
is the corresponding pairwise classification margin of the
mean predictor. By the definition of \(\delta_{k_\star}\),
\begin{equation}
(\mathbf d_{ic}^{(\star)})^\top
\mathbf T_{k_\star}\mathbf x_i
\ge
\delta_{k_\star},
\qquad
\|\mathbf d_{ic}^{(\star)}\|_2=\sqrt2.
\label{eq:linear-L2-top-block-pairwise-gap}
\end{equation}

Denote the amplitude of the leading singular modes by
\begin{equation}
a_\star(\tau)
=
\sinh(2s_1\tau).
\label{eq:linear-L2-top-block-amplitude}
\end{equation}
By Lemma~\ref{lem:linear-L2-spectral-concentration},
the mean predictor admits the decomposition
\begin{equation}
\overline{\mathbf M}(\tau)
=
a_\star(\tau)
\left(
\frac{\mathbf T_{k_\star}}{s_1}
+
\mathbf E_\star(\tau)
\right),
\qquad
\|\mathbf E_\star(\tau)\|_{\mathrm{op}}
=
\rho_\star(\tau).
\label{eq:linear-L2-top-block-mean-decomposition}
\end{equation}
The correction \(\mathbf E_\star(\tau)\) contains the
contributions of the lower singular modes, normalized
by the leading amplitude.

For every training example and competing class,
\begin{equation}
\begin{aligned}
(\mathbf d_{ic}^{(\star)})^\top
\overline{\mathbf M}(\tau)\mathbf x_i
&=
a_\star(\tau)
\left(
\frac{
(\mathbf d_{ic}^{(\star)})^\top
\mathbf T_{k_\star}\mathbf x_i
}{s_1}
+
(\mathbf d_{ic}^{(\star)})^\top
\mathbf E_\star(\tau)\mathbf x_i
\right)
\\
&\ge
a_\star(\tau)
\left(
\frac{\delta_{k_\star}}{s_1}
-
\sqrt2\,R\rho_\star(\tau)
\right).
\end{aligned}
\label{eq:linear-L2-bottleneck-mean-margin}
\end{equation}

The first condition in
\eqref{eq:linear-L2-bottleneck-conditions}
implies that, for all sufficiently large \(H\),
\begin{equation}
\sqrt2\,R\rho_\star(\tau_{\mathrm{bot}})
\le
\frac{\delta_{k_\star}}{4s_1}.
\label{eq:linear-L2-bottleneck-remainder-small}
\end{equation}
Therefore, at the bottleneck time,
\begin{equation}
(\mathbf d_{ic}^{(\star)})^\top
\overline{\mathbf M}(\tau_{\mathrm{bot}})
\mathbf x_i
\ge
\frac34
a_\star(\tau_{\mathrm{bot}})
\frac{\delta_{k_\star}}{s_1}
\label{eq:linear-L2-bottleneck-mean-positive}
\end{equation}
uniformly over all training examples and competing
classes. In particular, the mean predictor has the
same classifier as \(\mathbf T_{k_\star}\), with
strictly positive pairwise margins.

\smallskip
\noindent
\emph{Step 2: uniform control of the finite-width fluctuations.}
We now compare the random fluctuations of the limiting
predictor with the positive deterministic margins obtained
in Step~1.

Recall that
\begin{equation}
\mathbb E[\mathbf M_0(\tau)]
=
H\overline{\mathbf M}(\tau).
\end{equation}
Thus, for any training example \(i\) and competing class
\(c\neq\widehat c_i^{(\star)}\), the random deviation
of the pairwise margin from its expectation is
\begin{equation}
(\mathbf d_{ic}^{(\star)})^\top
\left(
\mathbf M_0(\tau)
-
H\overline{\mathbf M}(\tau)
\right)
\mathbf x_i.
\end{equation}

At the bottleneck time, define the maximum absolute
fluctuation over all such comparisons by
\begin{equation}
\mathsf N_H
:=
\max_{\substack{1\le i\le N\\
c\neq\widehat c_i^{(\star)}}}
\left|
(\mathbf d_{ic}^{(\star)})^\top
\left(
\mathbf M_0(\tau_{\mathrm{bot}})
-
H\overline{\mathbf M}(\tau_{\mathrm{bot}})
\right)
\mathbf x_i
\right|.
\label{eq:linear-L2-bottleneck-noise-definition}
\end{equation}
Our goal is to show that \(\mathsf N_H\) is small
compared with the deterministic top-block margin.

The concentration argument from Step~3 of the proof of
Theorem~\ref{thm:linear-L2-early-target-classifier}
applies without modification. Indeed, each pairwise
fluctuation is a sum of independent centered hidden-unit
contributions, each being a centered product of jointly
Gaussian variables. The same sub-exponential Bernstein
bound therefore applies with \(s=s_1\).

Applying \eqref{eq:linear-L2-margin-Bernstein} at
\(\tau=\tau_{\mathrm{bot}}\), taking
\(t=\ell=\log(2NCH)\), and using a union bound over
the at most \(N(C-1)\) comparisons, we obtain an event
\(\mathcal F_H\) satisfying
\begin{equation}
\mathbb P(\mathcal F_H^c)
\le
2N(C-1)e^{-\ell}
\le
\frac1H.
\label{eq:linear-L2-bottleneck-good-event-probability}
\end{equation}
On this event, the maximum fluctuation obeys
\begin{equation}
\mathsf N_H
\le
C_1e^{2s_1\tau_{\mathrm{bot}}}R
\left(
\sqrt{H\ell}+\ell
\right),
\label{eq:linear-L2-bottleneck-uniform-fluctuation}
\end{equation}
where \(C_1>0\) is a universal constant.

We must compare this estimate with the positive
deterministic margin from Step~1. Since the expectation
of \(\mathbf M_0\) is \(H\overline{\mathbf M}\),
equation~\eqref{eq:linear-L2-bottleneck-mean-positive}
shows that the mean pairwise margins of
\(\mathbf M_0(\tau_{\mathrm{bot}})\) are at least
\(3/4\) times the deterministic scale
\begin{equation}
\mathsf S_H
:=
H a_\star(\tau_{\mathrm{bot}})
\frac{\delta_{k_\star}}{s_1}
>0.
\label{eq:linear-L2-bottleneck-signal-scale}
\end{equation}
Thus it suffices to prove that
\(\mathsf N_H/\mathsf S_H\to0\) on events whose
probabilities tend to one.

Using the definitions of \(\mathsf N_H\) and
\(\mathsf S_H\), the concentration estimate
\eqref{eq:linear-L2-bottleneck-uniform-fluctuation}
gives, on \(\mathcal F_H\),
\begin{equation}
\begin{aligned}
\frac{\mathsf N_H}{\mathsf S_H}
&\le
\frac{
C_1e^{2s_1\tau_{\mathrm{bot}}}R
\left(\sqrt{H\ell}+\ell\right)
}{
H\sinh(2s_1\tau_{\mathrm{bot}})
\delta_{k_\star}/s_1
}
\\
&=
C_1
\frac{
e^{2s_1\tau_{\mathrm{bot}}}
}{
\sinh(2s_1\tau_{\mathrm{bot}})
}
\frac{Rs_1}{\delta_{k_\star}}
\left(
\sqrt{\frac{\ell}{H}}
+
\frac{\ell}{H}
\right).
\end{aligned}
\end{equation}

The assumption \(s_1\tau_{\mathrm{bot}}\ge1\)
ensures that the exponential growth factor in the
concentration estimate is comparable to the leading
spectral amplitude:
\begin{equation}
\frac{
e^{2s_1\tau_{\mathrm{bot}}}
}{
\sinh(2s_1\tau_{\mathrm{bot}})
}
=
\frac{2}{1-e^{-4s_1\tau_{\mathrm{bot}}}}
\le
\frac{2}{1-e^{-4}}
=:C_2.
\label{eq:linear-L2-exp-sinh-ratio}
\end{equation}
Consequently, on \(\mathcal F_H\),
\begin{equation}
\boxed{
\frac{\mathsf N_H}{\mathsf S_H}
\le
C_1C_2
\frac{Rs_1}{\delta_{k_\star}}
\left(
\sqrt{\frac{\ell}{H}}
+
\frac{\ell}{H}
\right).
}
\label{eq:linear-L2-bottleneck-noise-ratio}
\end{equation}

We next verify that the right-hand side vanishes
under the assumptions of the theorem.
Since
\(\|\mathbf T_{k_\star}\|_{\mathrm{op}}=s_1\)
and each comparison vector has norm \(\sqrt2\),
the definition of the decision gap gives
\begin{equation}
\delta_{k_\star}
\le
\sqrt2\,Rs_1,
\qquad
\frac{Rs_1}{\delta_{k_\star}}
\ge
\frac1{\sqrt2}.
\end{equation}
The second condition in
\eqref{eq:linear-L2-bottleneck-conditions}
states that
\begin{equation}
\frac{Rs_1}{\delta_{k_\star}}
\sqrt{\frac{\ell}{H}}
\longrightarrow0.
\end{equation}
Because \(Rs_1/\delta_{k_\star}\) is bounded
below by a positive constant, this condition
also implies \(\ell/H\to0\). In particular,
\begin{equation}
\frac{Rs_1}{\delta_{k_\star}}
\frac{\ell}{H}
=
\left(
\frac{Rs_1}{\delta_{k_\star}}
\sqrt{\frac{\ell}{H}}
\right)
\sqrt{\frac{\ell}{H}}
\longrightarrow0.
\end{equation}
Hence the right-hand side of
\eqref{eq:linear-L2-bottleneck-noise-ratio}
converges to zero.

It follows that, for all sufficiently large \(H\),
on \(\mathcal F_H\),
\begin{equation}
\mathsf N_H
\le
\frac14\mathsf S_H.
\label{eq:linear-L2-bottleneck-noise-small}
\end{equation}

We can now combine the deterministic mean-margin
estimate with the fluctuation bound. For every training
example \(i\) and every competing class
\(c\neq\widehat c_i^{(\star)}\), on \(\mathcal F_H\),
\begin{equation}
\begin{aligned}
(\mathbf d_{ic}^{(\star)})^\top
\mathbf M_0(\tau_{\mathrm{bot}})\mathbf x_i
&=
H(\mathbf d_{ic}^{(\star)})^\top
\overline{\mathbf M}(\tau_{\mathrm{bot}})
\mathbf x_i
+
(\mathbf d_{ic}^{(\star)})^\top
\left(
\mathbf M_0(\tau_{\mathrm{bot}})
-
H\overline{\mathbf M}(\tau_{\mathrm{bot}})
\right)
\mathbf x_i
\\
&\ge
\frac34\mathsf S_H-\mathsf N_H
\ge
\frac12\mathsf S_H
>0.
\end{aligned}
\label{eq:linear-L2-bottleneck-final-margin}
\end{equation}

Thus every pairwise margin of the limiting predictor
is strictly positive relative to the class
\(\widehat c_i^{(\star)}\), simultaneously for all
training examples, on \(\mathcal F_H\).
Since
\(\mathbb P(\mathcal F_H)\ge1-1/H\to1\),
the classifier of
\(\mathbf M_0(\tau_{\mathrm{bot}})\)
agrees with that of \(\mathbf T_{k_\star}\)
with probability tending to one. This proves
\eqref{eq:linear-L2-bottleneck-linearized-classifier}.

\smallskip
\noindent
\emph{Step 3: transfer to the exact gradient flow.}
Although \(\tau_{\mathrm{bot}}\) may grow with \(H\),
it is a fixed finite time for every fixed member
of the large-system sequence during the inner
limit \(\sigma\downarrow0\).

Theorem~\ref{thm:linear-L2-microscopic-linearization},
applied on \([0,\tau_{\mathrm{bot}}]\), therefore gives
\begin{equation}
\frac{
\mathbf M^{(\sigma)}(\tau_{\mathrm{bot}})
}{
\sigma^2
}
\xrightarrow[\sigma\downarrow0]{}
\mathbf M_0(\tau_{\mathrm{bot}})
\qquad
\text{almost surely in Frobenius norm}.
\label{eq:linear-L2-bottleneck-transfer}
\end{equation}

For all sufficiently large \(H\), on the event
\(\mathcal F_H\), every pairwise margin of the
limiting predictor is strictly positive by
\eqref{eq:linear-L2-bottleneck-final-margin}.
Since the training set is finite for each fixed
\(H\), the preceding convergence implies that,
almost surely on \(\mathcal F_H\), all corresponding
margins of
\(\sigma^{-2}\mathbf M^{(\sigma)}(\tau_{\mathrm{bot}})\)
are also strictly positive for every sufficiently
small \(\sigma\).

Multiplication by the positive scalar \(\sigma^{-2}\)
does not affect classification. Consequently,
Fatou's lemma gives
\begin{equation}
\begin{aligned}
&
\liminf_{\sigma\downarrow0}
\mathbb P\left(
\arg\max_{c\in[C]}
(\mathbf M^{(\sigma)}(\tau_{\mathrm{bot}})
\mathbf x_i)_c
=
\widehat c_i^{(\star)}
\quad\text{for every }i
\right)
\\
&\qquad\ge
\mathbb P(\mathcal F_H)
\ge
1-\frac1H.
\end{aligned}
\end{equation}
Taking \(H\to\infty\) proves
\eqref{eq:linear-L2-bottleneck-exact-classifier}.
The accuracy statement
\eqref{eq:linear-L2-bottleneck-exact-accuracy}
follows immediately.

Finally, for each fixed member of the sequence,
the bottleneck time is finite and independent of
\(\sigma\). Applying
Theorem~\ref{thm:linear-L2-microscopic-linearization}
on \([0,\tau_{\mathrm{bot}}]\), the uniform predictor
and loss estimates
\eqref{eq:linear-L2-predictor-microscopic}
and \eqref{eq:linear-L2-loss-microscopic}
directly imply
\eqref{eq:linear-L2-bottleneck-loss-flat}.
This completes the proof.
\end{proof}

The two classifier regimes can be separated on explicit time scales
when the problem geometry and the spectral gap below the top singular
block remain controlled. Suppose that \(N\) and \(C\) grow at most
polynomially with \(H\), and that
\begin{equation}
R=O(1),
\qquad
s_1=O(1),
\qquad
\delta_{\mathbf T}^{-1}=O(1),
\qquad
\delta_{k_\star}^{-1}=O(1),
\label{eq:linear-L2-standard-bottleneck-scaling}
\end{equation}
while, for some constant \(\Delta>0\),
\begin{equation}
s_1-s_{k_\star+1}
\ge
\Delta
\label{eq:linear-L2-uniform-spectral-gap}
\end{equation}
uniformly along the large-system sequence.

Under these assumptions, \(\ell=\Theta(\log H)\) and
the width condition of
Theorem~\ref{thm:linear-L2-early-target-classifier}
is satisfied. The early organization time therefore scales as
\begin{equation}
\tau_{\mathrm{org}}
\asymp
\sqrt{\frac{\log H}{H}}.
\label{eq:linear-L2-organization-time-order}
\end{equation}
For the bottleneck, we may choose
\begin{equation}
\tau_{\mathrm{bot}}
=
\zeta\log H
\label{eq:linear-L2-logarithmic-bottleneck-time}
\end{equation}
with any fixed \(\zeta>0\). Indeed, the uniform spectral gap gives
\begin{equation}
\rho_\star(\tau_{\mathrm{bot}})
=
e^{-2(s_1-s_{k_\star+1})\tau_{\mathrm{bot}}}
\frac{
1-e^{-4s_{k_\star+1}\tau_{\mathrm{bot}}}
}{
1-e^{-4s_1\tau_{\mathrm{bot}}}
}
\le
H^{-2\zeta\Delta}
\xrightarrow[H\to\infty]{}0.
\label{eq:linear-L2-top-block-rho-log-time}
\end{equation}
Moreover, the second bottleneck condition holds because
\(Rs_1/\delta_{k_\star}=O(1)\) and
\(\sqrt{\ell/H}\to0\).
Since \(s_1\ge\Delta\), we also have
\(s_1\tau_{\mathrm{bot}}\ge1\) for all sufficiently large
\(H\). Thus all conditions of
Theorem~\ref{thm:linear-L2-top-block-bottleneck}
are satisfied, with
\(\tau_{\mathrm{org}}<\tau_{\mathrm{bot}}\).

Consequently, the predictor first reproduces the classifier
of the full target \(\mathbf T\), and at the later bottleneck
time reproduces that of the complete top singular block
\(\mathbf T_{k_\star}\). Combining the two theorems, these
classifier agreements hold simultaneously with probability
tending to one in the sequential limit
\(\sigma\downarrow0\) followed by \(H\to\infty\).
In particular,
\begin{equation}
\Acc\!\left(
\mathbf M^{(\sigma)}(\tau_{\mathrm{org}})
\right)
=
A_{\mathbf T},
\qquad
\Acc\!\left(
\mathbf M^{(\sigma)}(\tau_{\mathrm{bot}})
\right)
=
A_{k_\star}
\label{eq:linear-L2-two-microscopic-accuracy-levels}
\end{equation}
with probability tending to one in this limit.

Hence an accuracy decrease between the two stages occurs
precisely when
\begin{equation}
A_{\mathbf T}>A_{k_\star}.
\label{eq:linear-L2-accuracy-drop-condition}
\end{equation}
This ordering is a classification property of the dataset
and target, rather than a consequence of the singular-value
decomposition alone.

\paragraph{Recovery to the target classifier.}

The preceding results concern the microscopic small-initialization regime:
the classifier first agrees with the full effective target
\(\mathbf T\), and at a later microscopic time spectral concentration makes
it agree with the top singular block \(\mathbf T_{k_\star}\). We now leave
the microscopic approximation and consider the full gradient flow at
arbitrarily late times. In the realizable two-layer setting, the
end-to-end predictor converges to \(\mathbf T\). The positive decision gap
of \(\mathbf T\) then implies permanent recovery of its classifier.

\begin{proposition}[Eventual recovery of the target classifier]
\label{prop:linear-L2-target-recovery}
Under the standing isotropy assumption
\(\mathbf G=\gamma \mathbf I_p\), suppose that
\begin{equation}
H\ge\operatorname{rank}(\mathbf T),
\label{eq:linear-L2-realizability}
\end{equation}
and that the effective target \(\mathbf T\) has the strictly positive
decision gap \(\delta_{\mathbf T}\) defined in
\eqref{eq:linear-L2-target-decision-gap}. Then, for every fixed
\(\sigma>0\), the Gaussian-initialized gradient flow satisfies
\begin{equation}
\mathbf M^{(\sigma)}(\tau)
\xrightarrow[\tau\to\infty]{}
\mathbf T
\qquad
\text{almost surely in Frobenius norm}.
\label{eq:linear-L2-end-to-end-recovery}
\end{equation}

Consequently, almost surely there exists a finite random time
\(\tau_{\mathrm{rec}}^{(\sigma)}\) such that
\begin{equation}
\arg\max_{c\in[C]}
\left(
\mathbf M^{(\sigma)}(\tau)\mathbf x_i
\right)_c
=
\widehat c_i
\qquad
\text{for every }i
\text{ and every }
\tau\ge\tau_{\mathrm{rec}}^{(\sigma)}.
\label{eq:linear-L2-eventual-classifier-recovery}
\end{equation}
In particular,
\begin{equation}
\Acc
\left(
\mathbf M^{(\sigma)}(\tau)
\right)
=
A_{\mathbf T}
\qquad
\text{for every }
\tau\ge\tau_{\mathrm{rec}}^{(\sigma)}.
\label{eq:linear-L2-eventual-accuracy-recovery}
\end{equation}
\end{proposition}

\begin{proof}
We first establish convergence of the end-to-end predictor
to \(\mathbf T\). Recall that the input data matrix
\(\mathbf X\in\mathbb R^{p\times N}\) satisfies
\begin{equation}
\mathbf X\mathbf X^\top
=
N\gamma\mathbf I_p
\label{eq:linear-L2-input-full-rank}
\end{equation}
by the standing isotropy assumption. Since \(\gamma>0\),
the matrix \(\mathbf X\mathbf X^\top\) has full rank.

We may therefore apply the global convergence result
of \citet[Theorem~6.12(b)]{bah2022learning}, which states
that the gradient flow of a two-layer linear network
converges to a global minimizer for almost every
initialization. This result applies to our dynamics
because they differ from the gradient flow of the
unnormalized empirical squared loss only by a positive
rescaling of time.

It remains to show that every global minimizer has
end-to-end predictor \(\mathbf T\). Recall from
\eqref{eq:linear-L2-factorization-loss} that
\begin{equation}
\mathcal L(\mathbf U,\mathbf V)
=
\frac{\gamma}{2}
\left\|
\mathbf V^\top\mathbf U-\mathbf T
\right\|_{\mathrm F}^2
+
\mathcal L_{\mathrm{const}}.
\label{eq:linear-L2-recovery-factorization-loss}
\end{equation}
The width condition
\(H\ge\operatorname{rank}(\mathbf T)\)
ensures that the network can represent \(\mathbf T\)
exactly, since every matrix of rank at most \(H\)
admits a factorization of the form
\(\mathbf V^\top\mathbf U\) with the prescribed
weight dimensions.
Consequently, the global minimum of
\eqref{eq:linear-L2-recovery-factorization-loss}
is \(\mathcal L_{\mathrm{const}}\), attained precisely
when
\begin{equation}
\mathbf V^\top\mathbf U
=
\mathbf T.
\label{eq:linear-L2-global-minimizer-target}
\end{equation}

The convergence theorem applies outside a set of initializations
of Lebesgue measure zero. For every fixed \(\sigma>0\),
our Gaussian initialization has a density with respect to
Lebesgue measure on the full parameter space, and therefore
belongs to this exceptional set with probability zero.
Since every global minimizer has end-to-end predictor
\(\mathbf T\), we conclude that
\(\mathbf M^{(\sigma)}(\tau)\to\mathbf T\)
in Frobenius norm almost surely as \(\tau\to\infty\),
proving \eqref{eq:linear-L2-end-to-end-recovery}.

We now show that convergence of the predictor implies
permanent recovery of the target classifier. As in the
preceding classification arguments, for every training
example \(i\) and competing class \(c\neq\widehat c_i\),
the pairwise decision margin satisfies
\begin{equation}
\begin{aligned}
\mathbf d_{ic}^\top\mathbf M\mathbf x_i
&=
\mathbf d_{ic}^\top\mathbf T\mathbf x_i
+
\mathbf d_{ic}^\top
(\mathbf M-\mathbf T)\mathbf x_i
\\
&\ge
\delta_{\mathbf T}
-
\sqrt2\,R
\|\mathbf M-\mathbf T\|_{\mathrm{op}}.
\end{aligned}
\label{eq:linear-L2-recovery-margin-bound}
\end{equation}
Here we used the positive target decision gap,
\(\|\mathbf d_{ic}\|_2=\sqrt2\), and
\(\|\mathbf x_i\|_2\le R\).

Since convergence in Frobenius norm implies convergence
in operator norm, almost surely there exists a finite
random time \(\tau_{\mathrm{rec}}^{(\sigma)}\) such that
\begin{equation}
\left\|
\mathbf M^{(\sigma)}(\tau)-\mathbf T
\right\|_{\mathrm{op}}
<
\frac{\delta_{\mathbf T}}{2\sqrt2\,R}
\qquad
\text{for every }
\tau\ge\tau_{\mathrm{rec}}^{(\sigma)}.
\label{eq:linear-L2-recovery-neighborhood}
\end{equation}
The preceding margin bound then gives
\begin{equation}
\mathbf d_{ic}^\top
\mathbf M^{(\sigma)}(\tau)\mathbf x_i
>
\frac{\delta_{\mathbf T}}{2}
>0
\label{eq:linear-L2-recovery-positive-margin}
\end{equation}
simultaneously for every training example and every
competing class, at all such times.

Thus the exact predictor permanently agrees with the
classifier of \(\mathbf T\) after
\(\tau_{\mathrm{rec}}^{(\sigma)}\), proving
\eqref{eq:linear-L2-eventual-classifier-recovery} and
\eqref{eq:linear-L2-eventual-accuracy-recovery}.
\end{proof}

Unlike the target-driven organization and the predictive bottleneck,
recovery is not a statement about the microscopic limiting dynamics.
For every fixed \(\sigma>0\),
Proposition~\ref{prop:linear-L2-target-recovery} concerns the full
gradient flow as \(\tau\to\infty\). In particular, the recovery time
\(\tau_{\mathrm{rec}}^{(\sigma)}\) may depend on \(\sigma\) and may
diverge as \(\sigma\downarrow0\).

We now combine the preceding results into a three-stage description.
Consider a sequence of two-layer problems indexed by \(H\to\infty\),
where \(N,p,C\), the dataset, and the effective target \(\mathbf T\)
may vary with \(H\). Assume the standing isotropy condition and
\begin{equation}
R:=\max_i\|\mathbf x_i\|_2=O(1),
\qquad
\|\mathbf T\|_{\mathrm{op}}=O(1),
\qquad
\log(2NCH)=o(H),
\qquad
H\ge\operatorname{rank}(\mathbf T).
\label{eq:linear-L2-three-stage-basic-assumptions}
\end{equation}
Suppose further that the top singular block is separated
from the remaining modes by a uniform spectral gap
\begin{equation}
s_1-s_{k_\star+1}\ge\Delta>0,
\label{eq:linear-L2-three-stage-spectral-gap}
\end{equation}
where \(\Delta\) is independent of \(H\) and \(s_{r+1}:=0\),
and that both target decision gaps remain uniformly positive:
\begin{equation}
\inf_H\delta_{\mathbf T}>0,
\qquad
\inf_H\delta_{k_\star}>0.
\label{eq:linear-L2-three-stage-decision-gaps}
\end{equation}

\begin{linkedtheorem}
{Three-stage accuracy dynamics for \(L=2\)}
{thm:main-linear-L2-accuracy-dynamics}
\label{thm:linear-L2-three-stage-accuracy-dynamics}
Under assumptions
\eqref{eq:linear-L2-three-stage-basic-assumptions}--
\eqref{eq:linear-L2-three-stage-decision-gaps},
there exist deterministic times
\(\tau_{\mathrm{org}}<\tau_{\mathrm{bot}}\),
which may depend on \(N,p,C,H\) and the dataset
\(\{(\mathbf x_i,\mathbf y_i)\}_{i=1}^N\),
but not on \(\sigma\) or the random initialization,
such that
\begin{equation}
\lim_{H\to\infty}
\liminf_{\sigma\downarrow0}
\mathbb P\Big(
\Acc\bigl(
\mathbf M^{(\sigma)}(\tau_{\mathrm{org}})
\bigr)=A_{\mathbf T},
\Acc\bigl(
\mathbf M^{(\sigma)}(\tau_{\mathrm{bot}})
\bigr)=A_{k_\star}
\Big)=1.
\label{eq:linear-L2-three-stage-microscopic-accuracies}
\end{equation}

Moreover, for every fixed member of the large-system
sequence, the entire trajectory up to the bottleneck time
remains microscopic, almost surely as \(\sigma\downarrow0\):
\begin{equation}
\sup_{0\le\tau\le\tau_{\mathrm{bot}}}
\|\mathbf M^{(\sigma)}(\tau)\|_{\mathrm{op}}
\xrightarrow[\sigma\downarrow0]{}0,
\qquad
\sup_{0\le\tau\le\tau_{\mathrm{bot}}}
\left|
\mathcal L(\tau)-\mathcal L(0)
\right|
\xrightarrow[\sigma\downarrow0]{}0.
\label{eq:linear-L2-three-stage-microscopic-flat-loss}
\end{equation}

For every fixed member of the sequence and every fixed
\(\sigma>0\), almost surely there also exists a finite
random time
\(\tau_{\mathrm{rec}}^{(\sigma)}>\tau_{\mathrm{bot}}\)
such that
\begin{equation}
\arg\max_{c\in[C]}
\left(
\mathbf M^{(\sigma)}(\tau)\mathbf x_i
\right)_c
=
\widehat c_i
\qquad
\text{for every }i
\text{ and every }\tau\ge\tau_{\mathrm{rec}}^{(\sigma)}.
\label{eq:linear-L2-three-stage-permanent-classifier-recovery}
\end{equation}
Consequently, the accuracy permanently equals \(A_{\mathbf T}\)
after this time.

In particular, if \(A_{\mathbf T}>A_{k_\star}\)
for all sufficiently large \(H\), then
\begin{equation}
\lim_{H\to\infty}
\liminf_{\sigma\downarrow0}
\mathbb P\Big(
\Acc\bigl(
\mathbf M^{(\sigma)}(\tau_{\mathrm{org}})
\bigr)
>
\Acc\bigl(
\mathbf M^{(\sigma)}(\tau_{\mathrm{bot}})
\bigr)
\Big)=1.
\label{eq:linear-L2-three-stage-strict-drop}
\end{equation}
Thus the predictor exhibits an early target-level classifier,
a lower-accuracy bottleneck, and eventual permanent recovery.
\end{linkedtheorem}

\begin{proof}
We verify that the assumptions of the preceding organization
and bottleneck theorems are satisfied, and then combine their
conclusions with the recovery proposition.

First, recall that the early-organization theorem uses
\begin{equation}
\ell=\log(2NCH),
\qquad
\kappa=
\frac{R\|\mathbf T\|_{\mathrm{op}}}{\delta_{\mathbf T}}.
\label{eq:linear-L2-three-stage-proof-parameters}
\end{equation}
The assumed bounds on \(R\), \(\|\mathbf T\|_{\mathrm{op}}\),
and \(\delta_{\mathbf T}^{-1}\) imply
\(\kappa=O(1)\). Since \(\ell=o(H)\), we have
\(\kappa^3\ell/H\to0\), as required by
Theorem~\ref{thm:linear-L2-early-target-classifier}.
We may therefore fix any \(K>K_0\) and choose
\begin{equation}
\tau_{\mathrm{org}}
=
K\frac{R}{\delta_{\mathbf T}}
\sqrt{\frac{\ell}{H}}.
\label{eq:linear-L2-three-stage-organization-time}
\end{equation}
The standing assumptions ensure that
\(\tau_{\mathrm{org}}\xrightarrow[H\to\infty]{}0\).
At this time, Theorem~\ref{thm:linear-L2-early-target-classifier} guarantees
agreement with the classifier of \(\mathbf T\), with
probability tending to one in the sequential limit.

For the bottleneck, fix any \(\zeta>0\) and choose
\begin{equation}
\tau_{\mathrm{bot}}=\zeta\log H.
\label{eq:linear-L2-three-stage-bottleneck-time}
\end{equation}
The uniform spectral gap
\(s_1-s_{k_\star+1}\ge\Delta>0\)
implies \(s_1\ge\Delta\). Hence
\(s_1\tau_{\mathrm{bot}}\ge1\) for all sufficiently
large \(H\).

Moreover, the spectral concentration ratio satisfies
\begin{equation}
\begin{aligned}
\rho_\star(\tau_{\mathrm{bot}})
&=
e^{-2(s_1-s_{k_\star+1})\tau_{\mathrm{bot}}}
\frac{
1-e^{-4s_{k_\star+1}\tau_{\mathrm{bot}}}
}{
1-e^{-4s_1\tau_{\mathrm{bot}}}
}
\\
&\le
H^{-2\zeta\Delta}
\xrightarrow[H\to\infty]{}0,
\end{aligned}
\label{eq:linear-L2-three-stage-rho-decay}
\end{equation}
where the inequality uses
\(0\le s_{k_\star+1}<s_1\).
Together with
\(Rs_1/\delta_{k_\star}=O(1)\) and
\(\ell/H\to0\), this verifies both conditions in
\eqref{eq:linear-L2-bottleneck-conditions}.
Furthermore,
\(\tau_{\mathrm{org}}\to0\) while
\(\tau_{\mathrm{bot}}\to\infty\), so
\(\tau_{\mathrm{org}}<\tau_{\mathrm{bot}}\)
for all sufficiently large \(H\).
Thus Theorem~\ref{thm:linear-L2-top-block-bottleneck}
applies and guarantees agreement with the classifier of
\(\mathbf T_{k_\star}\) at \(\tau_{\mathrm{bot}}\).

The two classifier-agreement events concern the same
Gaussian initialization. Applying a union bound to their
failure probabilities shows that both agreements hold
simultaneously with probability tending to one in the
sequential limit. This proves
\eqref{eq:linear-L2-three-stage-microscopic-accuracies}.

To establish the uniform microscopic behavior, note that
\(\tau_{\mathrm{bot}}\), although potentially growing with
\(H\), is finite for each fixed member of the sequence.
Theorem~\ref{thm:linear-L2-microscopic-linearization},
applied on \([0,\tau_{\mathrm{bot}}]\), therefore gives
the uniform predictor-norm and loss limits in
\eqref{eq:linear-L2-three-stage-microscopic-flat-loss}
almost surely as \(\sigma\downarrow0\).

Finally, the width condition
\(H\ge\operatorname{rank}(\mathbf T)\) and the positive
decision gap \(\delta_{\mathbf T}\) allow us to apply
Proposition~\ref{prop:linear-L2-target-recovery}.
For every fixed member of the sequence and every fixed
\(\sigma>0\), the exact gradient flow almost surely
recovers the classifier of \(\mathbf T\) permanently
after some finite random time. Enlarging this time
if necessary ensures
\(\tau_{\mathrm{rec}}^{(\sigma)}>\tau_{\mathrm{bot}}\),
proving the permanent recovery statement.

If \(A_{\mathbf T}>A_{k_\star}\) for all sufficiently
large \(H\), the simultaneous classifier agreements
imply a strict accuracy decrease from
\(\tau_{\mathrm{org}}\) to \(\tau_{\mathrm{bot}}\),
with probability tending to one in the sequential limit.
This proves
\eqref{eq:linear-L2-three-stage-strict-drop}
and completes the proof.
\end{proof}

\paragraph{Unary-operation targets.}

We now specialize the two-layer theory to unary-operation learning.
This setting is particularly tractable because the target matrix admits
an explicit singular-value decomposition determined entirely by the
output occupancies. Consequently, we can analyze the growth and
competition of singular modes in detail and derive explicit
classification-accuracy profiles, including cases where the general
top-block margin assumptions are violated.

Let \(f:[p]\to[p]\) be a unary operation, and consider the dataset
\begin{equation}
N=C=p,
\qquad
\mathbf x_i=\mathbf e_i,
\qquad
\mathbf y_i=\mathbf e_{f(i)},
\qquad
i\in[p],
\label{eq:linear-L2-unary-data}
\end{equation}
where \(\{\mathbf e_i\}_{i=1}^p\) denotes the standard basis of
\(\mathbb R^p\). This corresponds to learning the complete input-output
table of \(f\), with both inputs and outputs represented by one-hot
vectors.

The empirical input covariance is
\begin{equation}
\mathbf G
=
\frac1p
\sum_{i=1}^p
\mathbf e_i\mathbf e_i^{\top}
=
\frac1p\mathbf I_p,
\label{eq:linear-L2-unary-covariance}
\end{equation}
so the isotropy parameter in
\eqref{eq:linear-L2-isotropy} is
\begin{equation}
\gamma=\frac1p.
\label{eq:linear-L2-unary-gamma}
\end{equation}
Since the inputs form the standard basis, the effective
least-squares target is simply the matrix whose \(i\)-th column
is the corresponding label vector:
\begin{equation}
\mathbf T
=
\frac{1}{N\gamma}
\sum_{i=1}^N
\mathbf y_i\mathbf x_i^{\top}
=
\sum_{i=1}^p
\mathbf e_{f(i)}\mathbf e_i^{\top},
\qquad
T_{ci}
=
\mathbf 1\{f(i)=c\}.
\label{eq:linear-L2-unary-target}
\end{equation}

By construction,
\begin{equation}
\mathbf T\mathbf e_i
=
\mathbf e_{f(i)}
\label{eq:linear-L2-unary-target-action}
\end{equation}
for every \(i\in[p]\). Thus \(\mathbf T\) classifies every
training example correctly. The correct class has score one,
while every competing class has score zero, so
\begin{equation}
A_{\mathbf T}=1,
\qquad
\delta_{\mathbf T}=1.
\label{eq:linear-L2-unary-target-accuracy-gap}
\end{equation}
As throughout the preceding two-layer analysis, time is
expressed in the rescaled variable \(\tau=\gamma t=t/p\).

To characterize the singular structure of \(\mathbf T\),
for each output \(c\in[p]\) define its fiber and occupancy by
\begin{equation}
\mathcal C_c
=
\{i\in[p]:f(i)=c\},
\qquad
n_c
=
|\mathcal C_c|,
\label{eq:linear-L2-unary-occupancies}
\end{equation}
and let
\begin{equation}
\mathcal O
=
\{c\in[p]:n_c>0\}
\label{eq:linear-L2-unary-occupied-outputs}
\end{equation}
be the set of occupied outputs. As we show next, these
occupancies determine the entire nonzero singular spectrum
of \(\mathbf T\), providing the basis for our subsequent
analysis of spectral interference and accuracy dynamics.

\begin{lemma}[Singular structure of a unary target]
\label{lem:linear-L2-unary-svd}
For every \(c\in\mathcal O\), define
\begin{equation}
\mathbf p_c
=
\mathbf e_c,
\qquad
\mathbf q_c
=
\frac{1}{\sqrt{n_c}}
\sum_{i\in\mathcal C_c}
\mathbf e_i.
\label{eq:linear-L2-unary-singular-vectors}
\end{equation}
Then
\begin{equation}
\mathbf T
=
\sum_{c\in\mathcal O}
\sqrt{n_c}\,
\mathbf p_c\mathbf q_c^{\top}
\label{eq:linear-L2-unary-svd}
\end{equation}
is a compact singular-value decomposition of \(\mathbf T\). Consequently,
the nonzero singular values are
\begin{equation}
s_c
=
\sqrt{n_c},
\qquad
c\in\mathcal O,
\label{eq:linear-L2-unary-singular-values}
\end{equation}
and
\begin{equation}
\operatorname{rank}(\mathbf T)
=
|\mathcal O|.
\label{eq:linear-L2-unary-rank}
\end{equation}
\end{lemma}

\begin{proof}
The vectors
\(\{\mathbf p_c\}_{c\in\mathcal O}\)
are distinct standard basis vectors and hence form an orthonormal family.
The vectors
\(\{\mathbf q_c\}_{c\in\mathcal O}\)
are also orthonormal because the fibers
\(\{\mathcal C_c\}_{c\in\mathcal O}\)
are pairwise disjoint. Finally,
\begin{equation}
\sum_{c\in\mathcal O}
\sqrt{n_c}\,
\mathbf p_c\mathbf q_c^{\top}=
\sum_{c\in\mathcal O}
\mathbf e_c
\sum_{i\in\mathcal C_c}
\mathbf e_i^{\top}
=
\sum_{i=1}^p
\mathbf e_{f(i)}\mathbf e_i^{\top}
=
\mathbf T.
\label{eq:linear-L2-unary-svd-verification}
\end{equation}
Therefore
\eqref{eq:linear-L2-unary-svd}
is a compact singular-value decomposition, and the stated singular values
and rank follow immediately.
\end{proof}

At initialization, the classifier is at chance level. Indeed, fix
\(i\in[p]\). Conditional on
\(\mathbf U^{(\sigma)}(0)\mathbf e_i\), the \(p\) scores
\begin{equation}
\left(
\mathbf M^{(\sigma)}(0)\mathbf e_i
\right)_c
=
\left(
\mathbf V^{(\sigma)}(0)^{\top}
\mathbf U^{(\sigma)}(0)\mathbf e_i
\right)_c,
\qquad
c\in[p],
\label{eq:linear-L2-unary-initial-scores}
\end{equation}
are independent and identically distributed centered Gaussian random
variables. Their common variance is positive almost surely, so the maximizing
class is almost surely unique and uniformly distributed on \([p]\).
Consequently, for every \(\sigma>0\),
\begin{equation}
\mathbb E
\left[
\Acc\left(\mathbf M^{(\sigma)}(0)\right)
\right]
=
\frac1p.
\label{eq:linear-L2-unary-initial-accuracy}
\end{equation}
Therefore, for every fixed \(\varepsilon>0\), Markov's inequality gives
\begin{equation}
\mathbb P
\left(
\Acc\left(\mathbf M^{(\sigma)}(0)\right)
>
\varepsilon
\right)
\le
\frac{1}{\varepsilon p},
\label{eq:linear-L2-unary-initial-accuracy-Markov}
\end{equation}
and hence
\begin{equation}
\Acc\left(\mathbf M^{(\sigma)}(0)\right)
\longrightarrow0
\qquad
\text{in probability as }p\to\infty.
\label{eq:linear-L2-unary-initial-accuracy-limit}
\end{equation}

Despite this vanishing initial accuracy, the early target-driven
organization established above rapidly recovers the entire unary
target classifier.

We consider a sequence of unary-operation problems with
\(p,H\to\infty\), allowing the operation \(f:[p]\to[p]\)
and its occupancies to vary along the sequence. Define the
maximal output occupancy by
\begin{equation}
n_{\max}
=
\max_{c\in[p]}n_c.
\label{eq:linear-L2-unary-max-occupancy}
\end{equation}
We assume that the width grows sufficiently rapidly relative
to the maximal occupancy:
\begin{equation}
n_{\max}^{3/2}
\frac{\log(2p^2H)}{H}
\longrightarrow0.
\label{eq:linear-L2-unary-early-width-condition}
\end{equation}
Under this condition, the general early-organization result
yields the following conclusion.

\begin{proposition}[Early perfect classification for unary targets]
\label{prop:linear-L2-unary-early-perfect}
There exists a universal constant \(K_0>0\) such that,
for every fixed \(K>K_0\), at the deterministic time
\begin{equation}
\tau_{\mathrm{org}}
=
K
\sqrt{\frac{\log(2p^2H)}{H}},
\label{eq:linear-L2-unary-organization-time}
\end{equation}
the exact gradient flow satisfies
\begin{equation}
\lim_{p,H\to\infty}
\liminf_{\sigma\downarrow0}
\mathbb P
\left(
\Acc
\left(
\mathbf M^{(\sigma)}(\tau_{\mathrm{org}})
\right)
=
1
\right)
=
1.
\label{eq:linear-L2-unary-early-perfect}
\end{equation}
Moreover, for each fixed \(p,H\), almost surely as
\(\sigma\downarrow0\), the entire trajectory up to
\(\tau_{\mathrm{org}}\) remains microscopic:
\begin{equation}
\sup_{0\le\tau\le\tau_{\mathrm{org}}}
\left\|
\mathbf M^{(\sigma)}(\tau)
\right\|_{\mathrm{op}}
\longrightarrow0,
\label{eq:linear-L2-unary-early-microscopic-predictor}
\end{equation}
and
\begin{equation}
\sup_{0\le\tau\le\tau_{\mathrm{org}}}
\left|
\mathcal L
\left(
\mathbf U^{(\sigma)}(\tau),
\mathbf V^{(\sigma)}(\tau)
\right)
-
\mathcal L
\left(
\mathbf U^{(\sigma)}(0),
\mathbf V^{(\sigma)}(0)
\right)
\right|
\longrightarrow0.
\label{eq:linear-L2-unary-early-microscopic}
\end{equation}
Thus perfect accuracy is reached within an interval
throughout which the predictor norm and the change
in loss vanish in the small-initialization limit.
\end{proposition}

\begin{proof}
We prove the proposition by specializing
Theorem~\ref{thm:linear-L2-early-target-classifier}
to the unary-operation dataset.

Since the inputs are standard basis vectors, their maximal norm
is \(R=1\). Moreover, as established in
\eqref{eq:linear-L2-unary-target-accuracy-gap},
the effective target \(\mathbf T\) perfectly classifies
all training examples with decision gap
\(\delta_{\mathbf T}=1\), so \(A_{\mathbf T}=1\).
The explicit singular-value decomposition from
Lemma~\ref{lem:linear-L2-unary-svd} also gives
\(\|\mathbf T\|_{\mathrm{op}}=\sqrt{n_{\max}}\).

Consequently, the parameters controlling the early-organization
theorem specialize to
\begin{equation}
\kappa
=
\frac{R\|\mathbf T\|_{\mathrm{op}}}
{\delta_{\mathbf T}}
=
\sqrt{n_{\max}},
\qquad
\ell
=
\log(2NCH)
=
\log(2p^2H).
\label{eq:linear-L2-unary-kappa}
\end{equation}
In particular, the sufficient width condition of
Theorem~\ref{thm:linear-L2-early-target-classifier}
becomes
\begin{equation}
\kappa^3\frac{\ell}{H}
=
n_{\max}^{3/2}
\frac{\log(2p^2H)}{H}
\longrightarrow0,
\label{eq:linear-L2-unary-width-specialization}
\end{equation}
which holds by our assumption
\eqref{eq:linear-L2-unary-early-width-condition}.

The organization time prescribed by
Theorem~\ref{thm:linear-L2-early-target-classifier} is
\begin{equation}
K\frac{R}{\delta_{\mathbf T}}
\sqrt{\frac{\ell}{H}}
=
K\sqrt{\frac{\log(2p^2H)}{H}}
=
\tau_{\mathrm{org}},
\label{eq:linear-L2-unary-organization-time-specialization}
\end{equation}
exactly as defined in
\eqref{eq:linear-L2-unary-organization-time}.
Since \(A_{\mathbf T}=1\),
Theorem~\ref{thm:linear-L2-early-target-classifier}
therefore guarantees perfect classification at this time
with probability tending to one in the sequential limit
\(\sigma\downarrow0\) followed by \(p,H\to\infty\).
This proves \eqref{eq:linear-L2-unary-early-perfect}.

Finally, for each fixed \(p,H\), the organization time
\(\tau_{\mathrm{org}}\) is finite and independent of
\(\sigma\). We may therefore apply
Theorem~\ref{thm:linear-L2-microscopic-linearization}
on the entire interval \([0,\tau_{\mathrm{org}}]\).
Its uniform small-initialization conclusions imply that
the predictor norm and the change in loss vanish uniformly
throughout this interval, almost surely as
\(\sigma\downarrow0\). This establishes the claimed
microscopic behavior and completes the proof.
\end{proof}

Combining the initial accuracy calculation
\eqref{eq:linear-L2-unary-initial-accuracy-limit}
with
Proposition~\ref{prop:linear-L2-unary-early-perfect},
the unary model therefore has asymptotically vanishing accuracy at
initialization but reaches perfect accuracy at
\(\tau_{\mathrm{org}}\), entirely within the microscopic,
asymptotically constant-loss phase.

What happens after the early perfect-classification stage is more subtle.
To relate the unary geometry to the general two-layer theory, define the set
of outputs with maximal occupancy by
\begin{equation}
\mathcal O_{\max}
=
\{c\in\mathcal O:n_c=n_{\max}\}.
\label{eq:linear-L2-unary-maximal-occupancy-set}
\end{equation}
By Lemma~\ref{lem:linear-L2-unary-svd}, the largest singular value is
\begin{equation}
s_1
=
\sqrt{n_{\max}},
\qquad
k_\star
=
|\mathcal O_{\max}|,
\label{eq:linear-L2-unary-top-multiplicity}
\end{equation}
and the complete top singular block from
\eqref{eq:linear-L2-top-block} is
\begin{equation}
\begin{aligned}
\mathbf T_{k_\star}
&=
\sum_{c\in\mathcal O_{\max}}
\sqrt{n_{\max}}\,
\mathbf p_c\mathbf q_c^{\top}
\\
&=
\sum_{c\in\mathcal O_{\max}}
\mathbf e_c
\sum_{i\in\mathcal C_c}
\mathbf e_i^{\top}.
\end{aligned}
\label{eq:linear-L2-unary-top-block}
\end{equation}
If the occupied outputs have heterogeneous occupancies, then there exists
\(c\in\mathcal O\setminus\mathcal O_{\max}\). For every
\(i\in\mathcal C_c\),
\begin{equation}
\mathbf T_{k_\star}\mathbf e_i
=
\mathbf 0.
\label{eq:linear-L2-unary-top-block-zero}
\end{equation}
Thus all output scores of \(\mathbf T_{k_\star}\) are tied on every
submaximal fiber, and its minimum decision gap is zero:
\begin{equation}
\delta_{k_\star}
=
0.
\label{eq:linear-L2-unary-top-block-zero-gap}
\end{equation}
Consequently,
Theorem~\ref{thm:linear-L2-top-block-bottleneck} does not determine the
classifier on these examples. In the opposite case, when all occupied
outputs have the same occupancy, the entire nonzero singular spectrum is
flat and
\begin{equation}
k_\star
=
\operatorname{rank}(\mathbf T),
\qquad
\mathbf T_{k_\star}
=
\mathbf T,
\label{eq:linear-L2-unary-flat-spectrum-top-block}
\end{equation}
so deterministic concentration onto the top singular block does not alter
the classifier at all.

Unary targets therefore expose a classification-degenerate case of the general spectral picture. For heterogeneous occupancies, the top singular block has zero decision margin on all submaximal fibers, so its classifier is not uniquely determined there. Importantly, the deterministic mean predictor itself continues to classify every example correctly at every positive time. Nevertheless, the singular values \(s_c=\sqrt{n_c}\) grow at different microscopic rates, allowing random finite-width cross-mode contributions associated with faster modes to eventually overwhelm the slower correct modes. We analyze this spectral interference next.

\paragraph{Heterogeneous unary targets: spectral interference.}

We now investigate how the perfect classifier established in
Proposition~\ref{prop:linear-L2-unary-early-perfect} can subsequently lose accuracy. The mechanism arises from the different growth rates of singular modes associated with heterogeneous output occupancies.

Under the microscopic dynamics, the correct classification score is reinforced by the singular mode associated with the target output, while incorrect scores also receive random contributions from other singular modes. When a competing output has a larger occupancy, its associated mode grows faster and can eventually overwhelm the correct score through finite-width fluctuations. We show below that this spectral interference produces classification transitions on logarithmic time scales.

We first consider a sequence of unary targets with a fixed
number \(\mathcal K\ge2\) of distinct positive occupancy levels
\begin{equation}
1\le m_1<\cdots<m_{\mathcal K},
\label{eq:linear-L2-unary-occupancy-levels}
\end{equation}
and suppose that every occupied output belongs to one of
the strata
\begin{equation}
\mathcal O_k
=
\{c\in[p]:n_c=m_k\},
\qquad
k\in[\mathcal K].
\label{eq:linear-L2-unary-occupancy-strata}
\end{equation}
Thus all outputs within the \(k\)-th stratum share the
singular value \(\sqrt{m_k}\). We assume that, as
\(p,H\to\infty\),
\begin{equation}
\frac{H}{p}
\longrightarrow
\alpha\in(0,\infty),
\qquad
\frac{|\mathcal O_k|}{p}
\longrightarrow
\rho_k\in(0,\infty),
\qquad
k\in[\mathcal K].
\label{eq:linear-L2-unary-stratum-asymptotics}
\end{equation}
In particular, each stratum contains an extensive
number of output classes.

The competition between the correct singular mode and the
faster-growing modes leads to occupancy-dependent interference
thresholds. For \(k<\mathcal K\), define
\begin{equation}
\zeta_k
=
\frac{1}{
2(\sqrt{m_{\mathcal K}}-\sqrt{m_k})
},
\qquad
\zeta_{\mathcal K}=+\infty.
\label{eq:linear-L2-unary-interference-threshold}
\end{equation}
The following lemma establishes these thresholds rigorously using the explicit microscopic dynamics.

\begin{lemma}[Spectral interference and degeneracy bottleneck]
\label{lem:linear-L2-unary-interference-threshold}
Under the assumptions
\eqref{eq:linear-L2-unary-stratum-asymptotics},
fix \(\zeta>0\) and let
\(\tau_\zeta=\zeta\log H\).
Then, with probability tending to one as
\(p,H\to\infty\), the following statements
hold simultaneously for all relevant training examples:

\begin{enumerate}
\item
For every \(c\in\mathcal O_{\mathcal K}\) and
\(i\in\mathcal C_c\), the correct output \(c\)
has strictly the largest classification score.
Thus the maximal-occupancy stratum remains
correctly classified at every fixed
logarithmic time scale.

\item
For every \(k<\mathcal K\), \(c\in\mathcal O_{k}\),
and \(i\in\mathcal C_c\), the correct output
\(c\) has strictly the largest score if
\(\zeta<\zeta_k\).
If \(\zeta\ge\zeta_k\), the predicted output
instead belongs to the maximal-occupancy stratum:
\begin{equation}
\arg\max_{r\in[p]}
\left(
\mathbf M_0(\tau_\zeta)\mathbf e_i
\right)_r
\in\mathcal O_{\mathcal K}.
\label{eq:linear-L2-unary-degenerate-prediction}
\end{equation}
In particular, these examples are misclassified
at and above their interference threshold
\(\zeta_k\).
\end{enumerate}

Consequently, for every fixed
\(\zeta\ge\zeta_{\mathcal K-1}\), all predicted outputs
belong to \(\mathcal O_{\mathcal K}\) with probability
tending to one.
\end{lemma}

\begin{proof}
\emph{Proof strategy.}
We first solve the microscopic dynamics in the singular
basis of the unary target and derive explicit expressions
for all classification scores. We then establish uniform
Gaussian estimates that distinguish the correct diagonal
scores, of leading order \(H\), from the random cross-mode
contributions, of order \(\sqrt{H\log p}\) before their
exponential growth factors are included.
Comparing these contributions proves correct classification
below the interference thresholds.
For the converse, we exploit the extensive
maximal-occupancy stratum: its many independent singular
modes guarantee a sufficiently large positive interference
term for every lower-occupancy class, including exactly
at the critical threshold. We then show that this term
dominates all scores outside the maximal-occupancy stratum,
establishing the degeneracy bottleneck.

\smallskip
\noindent
\emph{Classification scores in the singular basis.}
Recall that \(\mathbf U_0(\tau)\) and
\(\mathbf V_0(\tau)\) solve
\eqref{eq:linear-L2-linearized-flow}, with independent
standard Gaussian initial matrices \(\mathbf A,\mathbf B\)
as in
\eqref{eq:linear-L2-standard-gaussian-initialization}.

By Lemma~\ref{lem:linear-L2-unary-svd}, each occupied
output \(c\in\mathcal O\) corresponds to a singular mode
with singular value \(s_c=\sqrt{n_c}\), right singular
vector \(\mathbf q_c\), and left singular vector
\(\mathbf e_c\). Define
\begin{equation}
\boldsymbol{\gamma}_c
=
\frac{\mathbf A\mathbf q_c+\mathbf B\mathbf e_c}{\sqrt2},
\qquad
\boldsymbol{\xi}_c
=
\frac{\mathbf A\mathbf q_c-\mathbf B\mathbf e_c}{\sqrt2}.
\label{eq:linear-L2-unary-gamma-xi}
\end{equation}
The orthonormality of the singular vectors and
Gaussianity of the initialization imply that
\(\{\boldsymbol{\gamma}_c,\boldsymbol{\xi}_c\}_{c\in\mathcal O}\)
are mutually independent standard Gaussian vectors
in \(\mathbb R^H\).

Projecting the linearized dynamics onto the singular
vectors, and using
\(\mathbf T\mathbf q_c=s_c\mathbf e_c\) and
\(\mathbf T^\top\mathbf e_c=s_c\mathbf q_c\), gives
\begin{equation}
\frac{d}{d\tau}
\left(
\mathbf U_0\mathbf q_c
\pm
\mathbf V_0\mathbf e_c
\right)
=
\pm s_c
\left(
\mathbf U_0\mathbf q_c
\pm
\mathbf V_0\mathbf e_c
\right).
\label{eq:linear-L2-unary-mode-evolution}
\end{equation}
Solving these decoupled equations with the Gaussian
initial conditions yields
\begin{equation}
\begin{aligned}
\mathbf U_0(\tau)\mathbf q_c
&=
\frac{
e^{s_c\tau}\boldsymbol{\gamma}_c
+
e^{-s_c\tau}\boldsymbol{\xi}_c
}{\sqrt2},
\\
\mathbf V_0(\tau)\mathbf e_c
&=
\frac{
e^{s_c\tau}\boldsymbol{\gamma}_c
-
e^{-s_c\tau}\boldsymbol{\xi}_c
}{\sqrt2}.
\end{aligned}
\label{eq:linear-L2-unary-active-coordinates}
\end{equation}
Thus every active singular mode contains an exponentially
growing and an exponentially decaying Gaussian component.

Fix an input \(i\in\mathcal C_c\), so that its correct
output is \(c=f(i)\). Decompose the input as
\begin{equation}
\mathbf e_i
=
\frac{\mathbf q_c}{\sqrt{n_c}}+\mathbf h_i,
\qquad
\mathbf h_i
=
\mathbf e_i-\frac{\mathbf q_c}{\sqrt{n_c}}.
\label{eq:linear-L2-unary-input-decomposition}
\end{equation}
Since
\(\mathbf T\mathbf e_i=\mathbf e_c\) and
\(\mathbf T\mathbf q_c=\sqrt{n_c}\mathbf e_c\),
we have \(\mathbf h_i\in\ker\mathbf T\).
The nullspace component therefore remains constant:
\begin{equation}
\mathbf U_0(\tau)\mathbf e_i
=
\frac{\mathbf U_0(\tau)\mathbf q_c}{\sqrt{n_c}}
+
\mathbf A\mathbf h_i.
\label{eq:linear-L2-unary-hidden-score}
\end{equation}

For an occupied output \(r\in\mathcal O\), the
classification score consequently satisfies
\begin{equation}
\left(\mathbf M_0(\tau)\mathbf e_i\right)_r
=
\left(\mathbf V_0(\tau)\mathbf e_r\right)^\top
\mathbf U_0(\tau)\mathbf e_i
=
\frac{
\left(\mathbf V_0(\tau)\mathbf e_r\right)^\top
\mathbf U_0(\tau)\mathbf q_c
}{\sqrt{n_c}}
+
\left(\mathbf V_0(\tau)\mathbf e_r\right)^\top
\mathbf A\mathbf h_i.
\label{eq:linear-L2-unary-occupied-score}
\end{equation}
The first contribution involves interactions between
active singular modes, while the second comes from
the input's nullspace component.

Substituting
\eqref{eq:linear-L2-unary-active-coordinates}
gives the explicit expression
\begin{equation}
\begin{aligned}
\left(\mathbf M_0(\tau)\mathbf e_i\right)_r
=
\frac{1}{2\sqrt{n_c}}\Bigl(
&
e^{(s_r+s_c)\tau}
\boldsymbol{\gamma}_r^\top\boldsymbol{\gamma}_c
+
e^{(s_r-s_c)\tau}
\boldsymbol{\gamma}_r^\top\boldsymbol{\xi}_c
\\
&-
e^{(s_c-s_r)\tau}
\boldsymbol{\xi}_r^\top\boldsymbol{\gamma}_c
-
e^{-(s_r+s_c)\tau}
\boldsymbol{\xi}_r^\top\boldsymbol{\xi}_c
\Bigr)
\\
&+
\left(\mathbf V_0(\tau)\mathbf e_r\right)^\top
\mathbf A\mathbf h_i.
\end{aligned}
\label{eq:linear-L2-unary-cross-score}
\end{equation}
For \(r\neq c\), this contains cross-mode
contributions whose leading exponential growth
rate is \(s_r+s_c\).

For the correct output \(r=c\), the two mixed terms
cancel exactly, giving
\begin{equation}
\left(\mathbf M_0(\tau)\mathbf e_i\right)_c
=
\frac{1}{2\sqrt{n_c}}
\left(
e^{2s_c\tau}\|\boldsymbol{\gamma}_c\|_2^2
-
e^{-2s_c\tau}\|\boldsymbol{\xi}_c\|_2^2
\right)
+
\left(\mathbf V_0(\tau)\mathbf e_c\right)^\top
\mathbf A\mathbf h_i.
\label{eq:linear-L2-unary-diagonal-score}
\end{equation}

Finally, if \(r\notin\mathcal O\), then
\(\mathbf T^\top\mathbf e_r=\mathbf0\).
The linearized equation
\(\dot{\mathbf V}_0=\mathbf U_0\mathbf T^\top\)
therefore implies
\(\mathbf V_0(\tau)\mathbf e_r=\mathbf B\mathbf e_r\).
Using the input decomposition and
\eqref{eq:linear-L2-unary-active-coordinates},
we obtain
\begin{equation}
\left(\mathbf M_0(\tau)\mathbf e_i\right)_r
=
(\mathbf B\mathbf e_r)^\top
\mathbf U_0(\tau)\mathbf e_i
=
(\mathbf B\mathbf e_r)^\top
\left(
\frac{
e^{s_c\tau}\boldsymbol{\gamma}_c
+
e^{-s_c\tau}\boldsymbol{\xi}_c
}{\sqrt{2n_c}}
+
\mathbf A\mathbf h_i
\right).
\label{eq:linear-L2-unary-unoccupied-score}
\end{equation}
Unlike an occupied competitor, an unoccupied output
has no exponentially growing output-side mode.
Its score therefore contains at most the growth
factor \(e^{s_c\tau}\).

\smallskip
\noindent
\emph{Uniform Gaussian estimates.}
We now establish estimates that hold simultaneously
over all training examples and competing outputs.
The key distinction is between squared Gaussian norms,
which concentrate around \(H\), and inner products
of independent Gaussian vectors, which have typical
size \(\sqrt H\). Uniform control introduces an
additional factor \(\sqrt{\log p}\).

Since there are at most \(p\) occupied outputs,
concentration of the \(\chi_H^2\) distribution
and a union bound give
\begin{equation}
\max_{c\in\mathcal O}
\left|
\frac{\|\boldsymbol{\gamma}_c\|_2^2}{H}-1
\right|
+
\max_{c\in\mathcal O}
\left|
\frac{\|\boldsymbol{\xi}_c\|_2^2}{H}-1
\right|
=
O_{\mathbb P}
\left(
\sqrt{\frac{\log p}{H}}
\right)
=
o_{\mathbb P}(1).
\label{eq:linear-L2-unary-uniform-gaussian-norms}
\end{equation}
Here and below we use
\(H/p\to\alpha\in(0,\infty)\), so that
\(H\asymp p\) and \(\log p/H\to0\).

To control the cross-mode terms, recall that
products of independent centered Gaussian variables
are sub-exponential. Bernstein's inequality
\citep[Theorem~2.9.1]{vershynin2026high}
therefore gives, for independent centered Gaussian
vectors \(\mathbf g,\mathbf h\in\mathbb R^H\)
whose covariance matrices are bounded by
\(\mathbf I_H\),
\begin{equation}
\mathbb P\left(
|\mathbf g^\top\mathbf h|
>
C\left(\sqrt{Hu}+u\right)
\right)
\le2e^{-u},
\qquad u\ge1,
\label{eq:linear-L2-unary-gaussian-product-tail}
\end{equation}
with a universal constant \(C>0\).

In \eqref{eq:linear-L2-unary-cross-score},
each incorrect occupied score involves four
inner products between independent Gaussian vectors
associated with distinct outputs \(r\neq c\).
There are at most \(4p(p-1)\) such products.
Applying \eqref{eq:linear-L2-unary-gaussian-product-tail}
with \(u=4\log p\) and taking a union bound yields
\begin{equation}
\max_{\substack{
r,c\in\mathcal O\\r\neq c
}}
\max_{\substack{
\mathbf a\in\{
\boldsymbol{\gamma}_r,\boldsymbol{\xi}_r\}\\
\mathbf b\in\{
\boldsymbol{\gamma}_c,\boldsymbol{\xi}_c\}
}}
|\mathbf a^\top\mathbf b|
=
O_{\mathbb P}
\left(
\sqrt{H\log p}
\right).
\label{eq:linear-L2-unary-cross-inner-product-bound}
\end{equation}
Indeed, the union-bound failure probability is at most
\(8p^{-2}\) for a sufficiently large constant in the
implicit bound. Independence between different
input-output pairs is not required.

We also control the nullspace contributions.
Since \(\mathbf h_i\perp\mathbf q_r\) for every
occupied output \(r\), Gaussianity implies that
\(\mathbf A\mathbf h_i\) is independent of
\(\{\boldsymbol{\gamma}_r,\boldsymbol{\xi}_r\}_{r\in\mathcal O}\).
Moreover,
\(\|\mathbf h_i\|_2^2=1-1/n_{f(i)}\le1\), so
\(\operatorname{Cov}(\mathbf A\mathbf h_i)
\preceq\mathbf I_H\).

For each occupied output \(r\), the explicit solution
\eqref{eq:linear-L2-unary-active-coordinates} gives
\begin{equation}
e^{-s_r\tau}\mathbf V_0(\tau)\mathbf e_r
=
\frac{
\boldsymbol{\gamma}_r
-
e^{-2s_r\tau}\boldsymbol{\xi}_r
}{\sqrt2}.
\label{eq:linear-L2-unary-rescaled-output-coordinate}
\end{equation}
For every \(\tau\ge0\), this is a centered Gaussian
vector with covariance
\((1+e^{-4s_r\tau})\mathbf I_H/2
\preceq\mathbf I_H\).
It is independent of \(\mathbf A\mathbf h_i\).

Thus Bernstein's inequality
\eqref{eq:linear-L2-unary-gaussian-product-tail}
and a union bound over at most \(p^2\)
input-output pairs give
\begin{equation}
\max_{\substack{
r\in\mathcal O\\i\in[p]
}}
e^{-s_r\tau_\zeta}
\left|
\left(
\mathbf V_0(\tau_\zeta)\mathbf e_r
\right)^\top
\mathbf A\mathbf h_i
\right|
=
O_{\mathbb P}
\left(
\sqrt{H\log p}
\right).
\label{eq:linear-L2-unary-null-bound}
\end{equation}

For unoccupied outputs, the Gaussian vector
\(\mathbf B\mathbf e_r\) is independent of
\(\mathbf U_0(\tau)\mathbf e_i\).
To apply the same bound, we verify that the
rescaled hidden representation has bounded covariance.
For \(i\in\mathcal C_c\),
\eqref{eq:linear-L2-unary-hidden-score} and
\eqref{eq:linear-L2-unary-active-coordinates}
give
\begin{equation}
\operatorname{Cov}\left(
e^{-s_c\tau}
\mathbf U_0(\tau)\mathbf e_i
\right)
=
\left[
\frac{1+e^{-4s_c\tau}}{2n_c}
+
\left(1-\frac1{n_c}\right)e^{-2s_c\tau}
\right]\mathbf I_H
\preceq\mathbf I_H.
\label{eq:linear-L2-unary-rescaled-hidden-covariance}
\end{equation}
Here the active and nullspace components are
independent, and the covariance bound holds
for all \(\tau\ge0\).

Applying \eqref{eq:linear-L2-unary-gaussian-product-tail}
to the independent vectors
\(\mathbf B\mathbf e_r\) and
\(e^{-s_c\tau_\zeta}\mathbf U_0(\tau_\zeta)\mathbf e_i\),
followed by a union bound, yields
\begin{equation}
\max_{\substack{
i\in[p]\\r\notin\mathcal O
}}
e^{-s_{f(i)}\tau_\zeta}
\left|
\left(
\mathbf M_0(\tau_\zeta)\mathbf e_i
\right)_r
\right|
=
O_{\mathbb P}
\left(
\sqrt{H\log p}
\right).
\label{eq:linear-L2-unary-unoccupied-score-bound}
\end{equation}
These uniform estimates will be used both to
establish the interference thresholds and to
identify the predicted outputs after the transitions.

\smallskip
\noindent
\emph{Correct classification below the threshold.}
Fix \(k\in[\mathcal K]\), and write
\(s=\sqrt{m_k}\) and \(s_*=\sqrt{m_{\mathcal K}}\).
Consider any \(c\in\mathcal O_k\) and
\(i\in\mathcal C_c\), so that \(s_c=s\).

We first normalize the correct score by its
leading scale \(H e^{2s\tau_\zeta}/(2s)\).
From \eqref{eq:linear-L2-unary-diagonal-score},
\begin{equation}
\frac{2s}{H e^{2s\tau_\zeta}}
\left(
\mathbf M_0(\tau_\zeta)\mathbf e_i
\right)_c
=
\frac{\|\boldsymbol{\gamma}_c\|_2^2}{H}
-
e^{-4s\tau_\zeta}
\frac{\|\boldsymbol{\xi}_c\|_2^2}{H}
+
\frac{2s}{H e^{2s\tau_\zeta}}
\left(
\mathbf V_0(\tau_\zeta)\mathbf e_c
\right)^\top
\mathbf A\mathbf h_i.
\label{eq:linear-L2-unary-normalized-correct-score}
\end{equation}

By the uniform norm estimate
\eqref{eq:linear-L2-unary-uniform-gaussian-norms},
the first term converges to one uniformly.
The second vanishes as \(H\to\infty\), since
\(e^{-4s\tau_\zeta}=H^{-4s\zeta}\to0\).
Finally, the nullspace bound
\eqref{eq:linear-L2-unary-null-bound}
shows that the third term is uniformly
\(O_{\mathbb P}(
\sqrt{\log p/H}\,e^{-s\tau_\zeta})
=o_{\mathbb P}(1)\).

Consequently,
\begin{equation}
\max_{\substack{
c\in\mathcal O_k\\i\in\mathcal C_c
}}
\left|
\frac{2s}{H e^{2s\tau_\zeta}}
\left(
\mathbf M_0(\tau_\zeta)\mathbf e_i
\right)_c
-1
\right|
=
o_{\mathbb P}(1).
\label{eq:linear-L2-unary-correct-score-asymptotic}
\end{equation}
In particular, every correct score in the
\(k\)-th stratum is positive and asymptotic
to \(H e^{2s\tau_\zeta}/(2s)\), uniformly
with probability tending to one.

We next consider incorrect occupied outputs
\(r\in\mathcal O\setminus\{c\}\).
By \eqref{eq:linear-L2-unary-cross-score},
all four exponential factors are bounded by
\(e^{(s_r+s)\tau_\zeta}\).
Using the cross-mode estimate
\eqref{eq:linear-L2-unary-cross-inner-product-bound}
and the nullspace estimate
\eqref{eq:linear-L2-unary-null-bound},
we obtain
\begin{equation}
\max_{\substack{
c\in\mathcal O_k,\ i\in\mathcal C_c\\
r\in\mathcal O,\ r\neq c
}}
e^{-(s_r+s)\tau_\zeta}
\left|
\left(
\mathbf M_0(\tau_\zeta)\mathbf e_i
\right)_r
\right|
=
O_{\mathbb P}
\left(
\sqrt{H\log p}
\right).
\label{eq:linear-L2-unary-incorrect-score-upper-bound}
\end{equation}
Here the nullspace contribution is included
because its growth factor is at most
\(e^{s_r\tau_\zeta}\), which is no larger than
\(e^{(s_r+s)\tau_\zeta}\).

Since \(s_r\le s_*\), dividing by the leading
correct-score scale gives
\begin{equation}
\max_{\substack{
c\in\mathcal O_k,\ i\in\mathcal C_c\\
r\in\mathcal O,\ r\neq c
}}
\frac{
\left|
\left(
\mathbf M_0(\tau_\zeta)\mathbf e_i
\right)_r
\right|
}{
\frac{H}{2s}e^{2s\tau_\zeta}
}
=
O_{\mathbb P}
\left(
\sqrt{\frac{\log p}{H}}\,
e^{(s_*-s)\tau_\zeta}
\right)
=
O_{\mathbb P}
\left(
\sqrt{\log H}\,
H^{-1/2+\zeta(s_*-s)}
\right).
\label{eq:linear-L2-unary-interference-power}
\end{equation}
For \(k<\mathcal K\), the exponent of \(H\) is
strictly negative precisely when
\begin{equation}
\zeta
<
\frac{1}{2(s_*-s)}
=
\zeta_k.
\end{equation}
Thus every incorrect occupied score is negligible
relative to the correct score below the threshold.

For unoccupied outputs, the uniform estimate
\eqref{eq:linear-L2-unary-unoccupied-score-bound}
similarly implies
\begin{equation}
\max_{\substack{
c\in\mathcal O_k,\ i\in\mathcal C_c\\
r\notin\mathcal O
}}
\frac{
\left|
\left(
\mathbf M_0(\tau_\zeta)\mathbf e_i
\right)_r
\right|
}{
\frac{H}{2s}e^{2s\tau_\zeta}
}
=
O_{\mathbb P}
\left(
\sqrt{\log H}\,
H^{-1/2-s\zeta}
\right)
=
o_{\mathbb P}(1).
\label{eq:linear-L2-unary-unoccupied-relative-bound}
\end{equation}
Unlike occupied competitors, unoccupied outputs
are therefore negligible at every fixed
\(\zeta>0\).

Combining these bounds with
\eqref{eq:linear-L2-unary-correct-score-asymptotic},
we conclude that all examples in the \(k\)-th
stratum are correctly classified with probability
tending to one whenever \(\zeta<\zeta_k\).

For \(k=\mathcal K\), we have \(s=s_*\), so the bound
\eqref{eq:linear-L2-unary-interference-power}
is \(O_{\mathbb P}(\sqrt{\log H/H})\).
Hence the maximal-occupancy stratum remains
correctly classified for every fixed \(\zeta>0\).
Since \(\mathcal K\) is fixed, a union bound makes these
conclusions simultaneous across all strata.
This proves the correct-classification claims
of the lemma.

\smallskip
\noindent
\emph{Degeneracy at and above the threshold.}
Fix \(k<\mathcal K\), and retain
\(s=\sqrt{m_k}\) and \(s_*=\sqrt{m_{\mathcal K}}\).
Define their positive spectral gap by
\begin{equation}
\Delta_k
=
s_*-s
=
\sqrt{m_{\mathcal K}}-\sqrt{m_k}
>0.
\label{eq:linear-L2-unary-spectral-gap-k}
\end{equation}
We show that, whenever \(\zeta\ge\zeta_k\),
each correct output \(c\in\mathcal O_k\)
has a competitor in \(\mathcal O_{\mathcal K}\) whose
score exceeds every score outside \(\mathcal O_{\mathcal K}\),
simultaneously for all \(i\in\mathcal C_c\).

For \(c\in\mathcal O_k\), \(i\in\mathcal C_c\),
and \(r\in\mathcal O_{\mathcal K}\), the leading term in
\eqref{eq:linear-L2-unary-cross-score} is
proportional to
\(e^{(s_*+s)\tau_\zeta}
\boldsymbol{\gamma}_r^\top\boldsymbol{\gamma}_c\).
To verify that the remaining terms cannot cancel
a sufficiently positive leading contribution,
we use the previously established estimates.

Indeed, after factoring out
\(e^{(s_*+s)\tau_\zeta}\), the other three
cross-mode terms carry additional factors
\(e^{-2s\tau_\zeta}\),
\(e^{-2s_*\tau_\zeta}\), and
\(e^{-2(s_*+s)\tau_\zeta}\), respectively.
The nullspace term has an additional factor
\(e^{-s\tau_\zeta}\), by
\eqref{eq:linear-L2-unary-null-bound}.
Therefore, combining
\eqref{eq:linear-L2-unary-cross-inner-product-bound}
and \eqref{eq:linear-L2-unary-null-bound} gives
\begin{equation}
\max_{\substack{
c\in\mathcal O_k,\ i\in\mathcal C_c\\
r\in\mathcal O_{\mathcal K}
}}
\Bigg|
\frac{
2s
\left(
\mathbf M_0(\tau_\zeta)\mathbf e_i
\right)_r
}{
\sqrt H\,e^{(s_*+s)\tau_\zeta}
}
-
\frac{
\boldsymbol{\gamma}_r^\top
\boldsymbol{\gamma}_c
}{\sqrt H}
\Bigg|
=
O_{\mathbb P}
\left(
\sqrt{\log H}\,H^{-s\zeta}
\right)
=
o_{\mathbb P}(1).
\label{eq:linear-L2-unary-interference-remainder}
\end{equation}
Here \(s\ge1\) and \(\zeta>0\) are fixed.
Crucially, this estimate is uniform over
all competitors \(r\in\mathcal O_{\mathcal K}\), including
those selected according to the initialization.

We now establish the existence of a sufficiently
positive interference term.
Fix \(c\in\mathcal O_k\).
Since \(k<\mathcal K\), the Gaussian vectors
\(\{\boldsymbol{\gamma}_r\}_{r\in\mathcal O_{\mathcal K}}\)
are independent of \(\boldsymbol{\gamma}_c\)
and of one another.
Conditional on \(\boldsymbol{\gamma}_c\),
\begin{equation}
\left\{
\frac{
\boldsymbol{\gamma}_r^\top
\boldsymbol{\gamma}_c
}{
\|\boldsymbol{\gamma}_c\|_2
}
\right\}_{r\in\mathcal O_{\mathcal K}}
\;\Bigg|\;
\boldsymbol{\gamma}_c
\stackrel{\mathrm{iid}}{\sim}
\mathcal N(0,1).
\label{eq:linear-L2-unary-conditional-interference-gaussians}
\end{equation}

By the extensive-stratum assumption and
\(H/p\to\alpha\),
\begin{equation}
\frac{|\mathcal O_{\mathcal K}|}{H}
\longrightarrow
\frac{\rho_{\mathcal K}}{\alpha}>0.
\label{eq:linear-L2-unary-max-stratum-size}
\end{equation}
Thus every correct class in the \(k\)-th stratum
has \(\Theta(H)\) independent candidate competitors.
Moreover,
\eqref{eq:linear-L2-unary-uniform-gaussian-norms}
implies that
\(\|\boldsymbol{\gamma}_c\|_2\ge\sqrt H/2\)
simultaneously for all \(c\in\mathcal O_k\)
with probability tending to one.

Fix an arbitrary constant \(L>0\).
Conditional on any \(\boldsymbol{\gamma}_c\)
satisfying this norm bound, each competitor has
probability at least
\(\overline\Phi(2L)>0\) of satisfying
\(\boldsymbol{\gamma}_r^\top
\boldsymbol{\gamma}_c\ge L\sqrt H\),
where \(\overline\Phi\) denotes the standard
Gaussian upper-tail function.

Conditional independence and a union bound
over at most \(p\) correct outputs give
\begin{equation}
\begin{aligned}
&\mathbb P\Bigl(
\exists c\in\mathcal O_k:
\max_{r\in\mathcal O_{\mathcal K}}
\boldsymbol{\gamma}_r^\top
\boldsymbol{\gamma}_c
<
L\sqrt H
\Bigr)
\\
&\quad\le
\mathbb P\left(
\min_{c\in\mathcal O_k}
\|\boldsymbol{\gamma}_c\|_2
<
\frac{\sqrt H}{2}
\right)
+
p\left(
1-\overline\Phi(2L)
\right)^{|\mathcal O_{\mathcal K}|}
\longrightarrow0.
\end{aligned}
\label{eq:linear-L2-unary-positive-interferer}
\end{equation}
The first probability vanishes by
\eqref{eq:linear-L2-unary-uniform-gaussian-norms};
the second decays exponentially in \(H\)
because \(|\mathcal O_{\mathcal K}|\asymp H\) and \(p=O(H)\).

Consequently, with probability tending to one,
every \(c\in\mathcal O_k\) admits a competitor
\(r(c)\in\mathcal O_{\mathcal K}\) satisfying
\begin{equation}
\boldsymbol{\gamma}_{r(c)}^\top
\boldsymbol{\gamma}_c
\ge L\sqrt H.
\end{equation}
The constant \(L>0\) can be chosen arbitrarily
large, provided it does not depend on \(H\).

Applying the uniform remainder estimate
\eqref{eq:linear-L2-unary-interference-remainder}
to these selected competitors gives
\begin{equation}
\min_{\substack{
c\in\mathcal O_k\\i\in\mathcal C_c
}}
\frac{
2s
\left(
\mathbf M_0(\tau_\zeta)\mathbf e_i
\right)_{r(c)}
}{
\sqrt H\,e^{(s_*+s)\tau_\zeta}
}
\ge
L-o_{\mathbb P}(1).
\end{equation}
Meanwhile, the uniform correct-score estimate
\eqref{eq:linear-L2-unary-correct-score-asymptotic}
shows that
\begin{equation}
\max_{\substack{
c\in\mathcal O_k\\i\in\mathcal C_c
}}
\left|
\frac{
2s
\left(
\mathbf M_0(\tau_\zeta)\mathbf e_i
\right)_c
}{
H e^{2s\tau_\zeta}
}
-1
\right|
=
o_{\mathbb P}(1).
\end{equation}
In particular, all correct scores are positive
with probability tending to one.

Comparing the two score scales, we obtain
\begin{equation}
\min_{\substack{
c\in\mathcal O_k\\i\in\mathcal C_c
}}
\frac{
\left(
\mathbf M_0(\tau_\zeta)\mathbf e_i
\right)_{r(c)}
}{
\left(
\mathbf M_0(\tau_\zeta)\mathbf e_i
\right)_c
}
\ge
\left(L-o_{\mathbb P}(1)\right)
\frac{e^{(s_*-s)\tau_\zeta}}{\sqrt H}
=
\left(L-o_{\mathbb P}(1)\right)
H^{-1/2+\zeta\Delta_k}.
\label{eq:linear-L2-unary-leading-interference-ratio}
\end{equation}

If \(\zeta>\zeta_k=1/(2\Delta_k)\),
the exponent \(-1/2+\zeta\Delta_k\) is positive,
so the lower bound diverges for any fixed \(L>0\).
At the critical value \(\zeta=\zeta_k\),
the power of \(H\) equals zero, but the preceding
construction works for every fixed \(L>0\).
Choosing \(L>1\) makes the lower bound in
\eqref{eq:linear-L2-unary-leading-interference-ratio}
strictly larger than one with probability
tending to one.
Thus the selected maximal-occupancy competitor
defeats the correct output whenever
\(\zeta\ge\zeta_k\).

It remains to show that no other output outside
\(\mathcal O_{\mathcal K}\) can defeat this competitor.
First consider an incorrect occupied output
\(r\in\mathcal O\setminus\mathcal O_{\mathcal K}\),
with \(r\neq c\).
Since all such outputs satisfy
\(s_r\le\sqrt{m_{\mathcal K-1}}<s_*\),
the uniform incorrect-score bound
\eqref{eq:linear-L2-unary-incorrect-score-upper-bound}
implies
\begin{equation}
\max_{\substack{
c\in\mathcal O_k,\ i\in\mathcal C_c\\
r\in\mathcal O\setminus\mathcal O_{\mathcal K},\ r\neq c
}}
\frac{
2s
\left|
\left(
\mathbf M_0(\tau_\zeta)\mathbf e_i
\right)_r
\right|
}{
\sqrt H\,e^{(s_*+s)\tau_\zeta}
}
=
O_{\mathbb P}
\left(
\sqrt{\log H}\,
H^{-\zeta(s_*-\sqrt{m_{\mathcal K-1}})}
\right)
=
o_{\mathbb P}(1).
\label{eq:linear-L2-unary-submaximal-competitor-bound}
\end{equation}
The strict inequality
\(s_*>\sqrt{m_{\mathcal K-1}}\) ensures that
the right-hand side vanishes for every
fixed \(\zeta>0\).

For unoccupied outputs, applying
\eqref{eq:linear-L2-unary-unoccupied-score-bound}
with the same normalization gives
\begin{equation}
\max_{\substack{
c\in\mathcal O_k,\ i\in\mathcal C_c\\
r\notin\mathcal O
}}
\frac{
2s
\left|
\left(
\mathbf M_0(\tau_\zeta)\mathbf e_i
\right)_r
\right|
}{
\sqrt H\,e^{(s_*+s)\tau_\zeta}
}
=
O_{\mathbb P}
\left(
\sqrt{\log H}\,H^{-s_*\zeta}
\right)
=
o_{\mathbb P}(1).
\label{eq:linear-L2-unary-unoccupied-competitor-bound}
\end{equation}
Thus all incorrect scores outside
\(\mathcal O_{\mathcal K}\) are negligible relative
to the maximal-stratum interference scale.

On the other hand, for any fixed \(L>1\),
the selected competitor \(r(c)\in\mathcal O_{\mathcal K}\)
has normalized score at least
\(L-o_{\mathbb P}(1)\), uniformly over all
examples in the \(k\)-th stratum.
Together with
\eqref{eq:linear-L2-unary-leading-interference-ratio},
\eqref{eq:linear-L2-unary-submaximal-competitor-bound},
and \eqref{eq:linear-L2-unary-unoccupied-competitor-bound},
this shows that, whenever \(\zeta\ge\zeta_k\),
the selected competitor has strictly larger
score than every output outside \(\mathcal O_{\mathcal K}\),
with probability tending to one.

Consequently, the predicted output belongs to
\(\mathcal O_{\mathcal K}\) for every example in the
\(k\)-th stratum. Since \(c\notin\mathcal O_{\mathcal K}\),
these examples are necessarily misclassified.

Finally, the interference thresholds satisfy
\(\zeta_1<\cdots<\zeta_{\mathcal K-1}\).
For every fixed \(\zeta\ge\zeta_{\mathcal K-1}\),
the degeneracy conclusion applies to all
submaximal strata, while examples in
\(\mathcal O_{\mathcal K}\) remain correctly classified
by the first part of the proof.
Hence all predicted outputs belong to
\(\mathcal O_{\mathcal K}\) with probability tending
to one. Since \(\mathcal K\) is fixed, the conclusions
hold simultaneously across all strata
by a union bound. This completes the proof.
\end{proof}

The stratum-wise transitions established in
Lemma~\ref{lem:linear-L2-unary-interference-threshold}
yield an explicit microscopic accuracy profile.
To express it, define
\begin{equation}
\pi_k
=
m_k\rho_k,
\qquad
k\in[\mathcal K].
\label{eq:linear-L2-unary-stratum-weights}
\end{equation}
Since each output in \(\mathcal O_k\) is the
correct class for exactly \(m_k\) inputs,
\(\pi_k\) is the asymptotic fraction of training
examples belonging to the \(k\)-th stratum.
In particular,
\(\sum_{k=1}^{\mathcal K}\pi_k=1\).

\begin{theorem}[Accuracy profile for heterogeneous unary targets]
\label{thm:linear-L2-unary-heterogeneous-profile}
Under the assumptions
\eqref{eq:linear-L2-unary-stratum-asymptotics},
the limiting accuracy profile is
\begin{equation}
\mathcal A_{\mathrm{int}}(\zeta)
=
\pi_{\mathcal K}
+
\sum_{k=1}^{\mathcal K-1}
\pi_k
\mathbf 1\{\zeta<\zeta_k\}.
\label{eq:linear-L2-unary-interference-profile}
\end{equation}
For every fixed \(\zeta>0\),
\begin{equation}
\Acc\left(
\mathbf M_0(\zeta\log H)
\right)
\xrightarrow[p,H\to\infty]{\mathbb P}
\mathcal A_{\mathrm{int}}(\zeta).
\label{eq:linear-L2-unary-linearized-profile}
\end{equation}
Moreover, for the exact gradient flow,
\begin{equation}
\Acc\left(
\mathbf M^{(\sigma)}(\zeta\log H)
\right)
\longrightarrow
\mathcal A_{\mathrm{int}}(\zeta)
\label{eq:linear-L2-unary-exact-profile}
\end{equation}
in probability in the sequential limit
\(\sigma\downarrow0\) followed by \(p,H\to\infty\).
\end{theorem}

\begin{proof}
By Lemma~\ref{lem:linear-L2-unary-interference-threshold},
for every fixed \(\zeta>0\), with probability tending
to one as \(p,H\to\infty\), all examples in the
\(k\)-th stratum are correctly classified if
\(k=\mathcal K\) or \(\zeta<\zeta_k\), and misclassified
otherwise.

Since each output in \(\mathcal O_k\) is the target
of exactly \(m_k\) training examples, the fraction
of examples belonging to this stratum satisfies
\begin{equation}
\frac{m_k|\mathcal O_k|}{p}
\xrightarrow[p,H\to\infty]{}
m_k\rho_k
=
\pi_k.
\label{eq:linear-L2-unary-stratum-empirical-mass}
\end{equation}
Summing the fractions of correctly classified
examples over the \(\mathcal K\) strata, which is fixed,
gives the profile
\eqref{eq:linear-L2-unary-interference-profile}
and establishes
\eqref{eq:linear-L2-unary-linearized-profile}.

We now transfer this result to the exact gradient
flow. Fix \(p,H\) and \(\zeta>0\), so that
\(\tau_\zeta=\zeta\log H\) is finite.
By Theorem~\ref{thm:linear-L2-microscopic-linearization},
\begin{equation}
\frac{
\mathbf M^{(\sigma)}(\tau_\zeta)
}{\sigma^2}
\xrightarrow[\sigma\downarrow0]{\mathrm{a.s.}}
\mathbf M_0(\tau_\zeta)
\qquad
\text{in Frobenius norm}.
\label{eq:linear-L2-unary-profile-transfer}
\end{equation}

For fixed \(p,H\), the linearized predictor has
no classification ties almost surely.
Indeed, the linearized flow is an invertible
linear transformation of the Gaussian initial
matrices \((\mathbf A,\mathbf B)\).
Consequently,
\((\mathbf U_0(\tau_\zeta),
\mathbf V_0(\tau_\zeta))\)
has a nondegenerate Gaussian distribution.
For any input \(i\) and distinct outputs
\(r,c\), the score difference
\begin{equation}
\begin{aligned}
&
\left(
\mathbf M_0(\tau_\zeta)\mathbf e_i
\right)_r
-
\left(
\mathbf M_0(\tau_\zeta)\mathbf e_i
\right)_c
\\
&=
\left(
\mathbf V_0(\tau_\zeta)
(\mathbf e_r-\mathbf e_c)
\right)^\top
\mathbf U_0(\tau_\zeta)\mathbf e_i
\end{aligned}
\end{equation}
is a nonzero polynomial in these Gaussian
variables and therefore vanishes with
probability zero. Since there are finitely
many input-output pairs, all classification
scores are pairwise distinct almost surely.

The almost-sure convergence
\eqref{eq:linear-L2-unary-profile-transfer}
therefore preserves all score comparisons for
sufficiently small \(\sigma\).
Since multiplication by \(\sigma^2>0\)
does not change the predicted class, we obtain
\begin{equation}
\Acc\left(
\mathbf M^{(\sigma)}(\tau_\zeta)
\right)
\xrightarrow[\sigma\downarrow0]{\mathrm{a.s.}}
\Acc\left(
\mathbf M_0(\tau_\zeta)
\right).
\end{equation}
Taking the subsequent large-system limit and
using \eqref{eq:linear-L2-unary-linearized-profile}
proves \eqref{eq:linear-L2-unary-exact-profile}.
\end{proof}

Theorem~\ref{thm:linear-L2-unary-heterogeneous-profile},
together with
Lemma~\ref{lem:linear-L2-unary-interference-threshold},
establishes a microscopic accuracy staircase accompanied
by a degeneracy of the classifier. Following the initial
perfect-accuracy stage, examples from lower-occupancy
strata lose correct classification before those from
higher-occupancy strata. At each interference threshold
\(\zeta_k\), the corresponding examples become
misclassified into the maximal-occupancy stratum
\(\mathcal O_{\mathcal K}\), while examples belonging to
\(\mathcal O_{\mathcal K}\) remain correctly classified.
After the final transition, all predicted outputs
belong to \(\mathcal O_{\mathcal K}\), and the limiting accuracy
equals \(\pi_{\mathcal K}\).

Crucially, this degeneracy is driven by finite-width
cross-mode fluctuations rather than the deterministic
mean predictor, which correctly classifies every unary
input at all positive times. The dominant singular block
has zero classification margin on submaximal fibers,
allowing fluctuations from faster-growing modes to
overwhelm their correct scores. Thus the accuracy
bottleneck reflects a collapse of the predicted
label set toward the maximally occupied outputs,
even though the mean predictor retains the correct
classification structure.

Indeed, taking expectations in the explicit
classification-score formulas
\eqref{eq:linear-L2-unary-cross-score}--%
\eqref{eq:linear-L2-unary-unoccupied-score},
we obtain, for every \(i\in\mathcal C_c\),
\begin{equation}
\mathbb E\left[
\mathbf M_0(\tau)\mathbf e_i
\right]
=
\frac{H}{\sqrt{n_c}}
\sinh(2\sqrt{n_c}\tau)\,\mathbf e_c.
\end{equation}
Thus the mean microscopic predictor correctly
classifies every unary input for all \(\tau>0\),
despite the degeneracy of individual realizations.

\paragraph{Random unary targets.}
\label{app:linear-L2-random-unary}

We now draw the unary target itself at random:
\begin{equation}
f(1),\ldots,f(p)
\stackrel{\mathrm{iid}}{\sim}
\operatorname{Unif}[p],
\qquad
\frac{H}{p}
\longrightarrow
\alpha
\in
(0,\infty).
\label{eq:linear-L2-random-unary-model}
\end{equation}
The occupancies
\(\{n_c\}_{c=1}^p\)
then follow the classical balls-into-bins model with \(p\) balls and
\(p\) bins. For \(m\ge0\), let
\begin{equation}
N_m
=
\left|
\{c\in[p]:n_c=m\}
\right|
\label{eq:linear-L2-random-unary-Nm}
\end{equation}
be the number of outputs with occupancy \(m\).

\begin{lemma}[Occupancy statistics of a random unary target]
\label{lem:linear-L2-random-unary-occupancies}
For every fixed integer \(m\ge0\),
\begin{equation}
\frac{N_m}{p}
\longrightarrow
\frac{e^{-1}}{m!}
\qquad
\text{in probability}.
\label{eq:linear-L2-random-unary-Poisson-occupancies}
\end{equation}
Consequently, for every fixed \(m\ge1\), the fraction of
training examples whose target output has occupancy \(m\)
satisfies
\begin{equation}
\frac{mN_m}{p}
\xrightarrow{\mathbb P}
\pi_m
:=
\frac{e^{-1}}{(m-1)!}.
\label{eq:linear-L2-random-unary-size-biased-occupancies}
\end{equation}
Moreover,
\begin{equation}
\frac{
\operatorname{rank}(\mathbf T)
}{
p
}
\longrightarrow
1-e^{-1}
\qquad
\text{in probability},
\label{eq:linear-L2-random-unary-rank}
\end{equation}
and
\begin{equation}
n_{\max}
=
O_{\mathbb P}
\left(
\frac{\log p}{\log\log p}
\right).
\label{eq:linear-L2-random-unary-max-occupancy}
\end{equation}
\end{lemma}

\begin{proof}
For each fixed \(m\ge0\), the expected occupancy count satisfies
\begin{equation}
\frac{\mathbb E[N_m]}{p}
=
\binom{p}{m}
\left(\frac1p\right)^m
\left(1-\frac1p\right)^{p-m}
\xrightarrow[p\to\infty]{}
\frac{e^{-1}}{m!}.
\end{equation}
Moreover, \(N_m\) is a function of the \(p\) independent
random variables \(f(1),\ldots,f(p)\).
Changing a single value \(f(i)\) decreases the occupancy
of one output by one and increases that of another by one.
Consequently, at most two outputs can change their
contribution to \(N_m\), so the total change in \(N_m\)
is bounded by two.

McDiarmid's bounded-differences inequality therefore gives,
for every \(\varepsilon>0\),
\begin{equation}
\mathbb P\left(
\left|
\frac{N_m}{p}
-
\frac{\mathbb E[N_m]}{p}
\right|
>
\varepsilon
\right)
\le
2\exp\left(-\frac{\varepsilon^2p}{2}\right)
\xrightarrow[p\to\infty]{}0.
\end{equation}
Together with the convergence of
\(\mathbb E[N_m]/p\), this proves
\eqref{eq:linear-L2-random-unary-Poisson-occupancies}.
Equation
\eqref{eq:linear-L2-random-unary-size-biased-occupancies}
follows immediately from
\eqref{eq:linear-L2-random-unary-Poisson-occupancies}
by multiplying by \(m\).

By Lemma~\ref{lem:linear-L2-unary-svd}, and since
\(N_0=p-|\mathcal O|\), we have
\begin{equation}
\operatorname{rank}(\mathbf T)
=
|\mathcal O|
=
p-N_0.
\label{eq:linear-L2-random-unary-rank-N0}
\end{equation}
So
\eqref{eq:linear-L2-random-unary-rank}
follows from
\eqref{eq:linear-L2-random-unary-Poisson-occupancies}
with \(m=0\).

The maximum-load estimate
\eqref{eq:linear-L2-random-unary-max-occupancy}
is a classical balls-into-bins result; see
\citet{raab1998balls}.
\end{proof}

The random unary model first exhibits the early target-driven
organization established for general unary operations.
Indeed, Lemma~\ref{lem:linear-L2-random-unary-occupancies}
implies that
\begin{equation}
n_{\max}^{3/2}
\frac{\log(2p^2H)}{H}
\longrightarrow0
\qquad
\text{in probability}.
\end{equation}
Consequently,
Proposition~\ref{prop:linear-L2-unary-early-perfect}
applies with probability tending to one over the random target. At the organization time
\(\tau_{\mathrm{org}}
=K\sqrt{\log(2p^2H)/H}\),
for sufficiently large fixed \(K\), the exact gradient flow
achieves perfect classification in the sequential
small-initialization and large-system limit.

We now investigate the subsequent loss of this perfect
accuracy. Unlike the finite-stratum setting considered
above, random unary targets have a growing number of
distinct occupancy levels, so the preceding staircase
result cannot be applied directly. Instead, we isolate
the interference mechanism that requires only an
extensive collection of faster-growing singular modes.
For \(m\ge0\), write
\begin{equation}
\mathcal O^{(m)}
=
\{c\in[p]:n_c=m\}.
\label{eq:linear-L2-random-unary-occupancy-set}
\end{equation}

\begin{lemma}[Interference from an extensive faster stratum]
\label{lem:linear-L2-unary-extensive-faster-stratum}
Fix integers \(1\le m<n\), and consider a sequence of unary targets
satisfying
\begin{equation}
\frac{H}{p}
\longrightarrow
\alpha
\in(0,\infty),
\qquad
\frac{|\mathcal O^{(n)}|}{p}
\longrightarrow
\rho_n>0.
\label{eq:linear-L2-unary-extensive-faster-assumptions}
\end{equation}
Fix
\begin{equation}
\zeta
\ge
\frac{1}{2(\sqrt n-\sqrt m)}
\label{eq:linear-L2-unary-faster-stratum-threshold}
\end{equation}
and let
\begin{equation}
\tau=\zeta\log H.
\label{eq:linear-L2-unary-faster-stratum-time}
\end{equation}
Then, with probability tending to one as \(p,H\to\infty\),
every training example \(i\in\mathcal C_c\) with
\(c\in\mathcal O^{(m)}\) satisfies
\begin{equation}
\max_{r\in\mathcal O^{(n)}}
\left(\mathbf M_0(\tau)\mathbf e_i\right)_r
>
\left(\mathbf M_0(\tau)\mathbf e_i\right)_c.
\label{eq:linear-L2-unary-faster-stratum-dominance}
\end{equation}
In particular, all such examples are misclassified by
the microscopic predictor \(\mathbf M_0\).

The same conclusions hold for the exact gradient flow
with probability tending to one in the sequential limit
\(\sigma\downarrow0\) followed by \(p,H\to\infty\).
\end{lemma}

\begin{proof}
Write \(s=\sqrt m\), \(s'=\sqrt n\), and
\(\tau=\zeta\log H\).
We adapt the interference argument from
Lemma~\ref{lem:linear-L2-unary-interference-threshold},
replacing its maximal-occupancy stratum by
\(\mathcal O^{(n)}\).

First, the explicit cross-mode score formula
\eqref{eq:linear-L2-unary-cross-score},
together with the uniform cross-mode Gaussian bound
\eqref{eq:linear-L2-unary-cross-inner-product-bound}
and the uniform nullspace bound
\eqref{eq:linear-L2-unary-null-bound},
gives
\begin{equation}
\max_{\substack{
c\in\mathcal O^{(m)},\ i\in\mathcal C_c\\
r\in\mathcal O^{(n)}
}}
\left|
\frac{
2s(\mathbf M_0(\tau)\mathbf e_i)_r
}{
\sqrt H\,e^{(s'+s)\tau}
}
-
\frac{
\boldsymbol{\gamma}_r^\top\boldsymbol{\gamma}_c
}{
\sqrt H
}
\right|
=
O_{\mathbb P}
\left(
\sqrt{\log p}\,H^{-s\zeta}
\right)
=
o_{\mathbb P}(1).
\label{eq:linear-L2-unary-faster-stratum-remainder}
\end{equation}
This is the same estimate as
\eqref{eq:linear-L2-unary-interference-remainder}.
Indeed, after factoring out the leading exponential
\(e^{(s'+s)\tau}\), the three subleading
cross-mode terms in
\eqref{eq:linear-L2-unary-cross-score}
carry the factors
\(e^{-2s\tau}\), \(e^{-2s'\tau}\), and
\(e^{-2(s+s')\tau}\), while the nullspace term
is suppressed by \(e^{-s\tau}\).
The stated bound follows because \(s'>s\ge1\).

Next, the uniform correct-score asymptotic
\eqref{eq:linear-L2-unary-correct-score-asymptotic},
derived from the diagonal-score formula
\eqref{eq:linear-L2-unary-diagonal-score},
the Gaussian norm concentration
\eqref{eq:linear-L2-unary-uniform-gaussian-norms},
and the nullspace bound
\eqref{eq:linear-L2-unary-null-bound},
applies unchanged to the fixed occupancy level \(m\).
Thus, uniformly over
\(c\in\mathcal O^{(m)}\) and \(i\in\mathcal C_c\),
\begin{equation}
(\mathbf M_0(\tau)\mathbf e_i)_c
=
\frac{H}{2s}e^{2s\tau}
\left(1+o_{\mathbb P}(1)\right).
\label{eq:linear-L2-unary-faster-stratum-correct-score}
\end{equation}

It remains to find a sufficiently positive
interfering score. Since \(m<n\), the Gaussian
vectors \(\boldsymbol{\gamma}_c\) and
\(\{\boldsymbol{\gamma}_r\}_{r\in\mathcal O^{(n)}}\)
are independent, by their construction in
\eqref{eq:linear-L2-unary-gamma-xi}.
Consequently, the conditional Gaussian identity
\eqref{eq:linear-L2-unary-conditional-interference-gaussians}
applies with \(\mathcal O^{(n)}\) in place of
the maximal stratum.

Moreover, assumption
\eqref{eq:linear-L2-unary-extensive-faster-assumptions}
implies
\begin{equation}
\frac{|\mathcal O^{(n)}|}{H}
\longrightarrow
\frac{\rho_n}{\alpha}>0.
\end{equation}
We therefore have \(\Theta(H)\) independent
competitors. The conditional Gaussian maximum
argument established in
\eqref{eq:linear-L2-unary-positive-interferer},
using the uniform norm concentration
\eqref{eq:linear-L2-unary-uniform-gaussian-norms},
shows that, for any fixed \(L>1\), with
probability tending to one every
\(c\in\mathcal O^{(m)}\) admits
\(r(c)\in\mathcal O^{(n)}\) satisfying
\begin{equation}
\boldsymbol{\gamma}_{r(c)}^\top
\boldsymbol{\gamma}_c
\ge L\sqrt H.
\end{equation}

Combining this inequality with the uniform
remainder estimate
\eqref{eq:linear-L2-unary-faster-stratum-remainder}
and the correct-score asymptotic
\eqref{eq:linear-L2-unary-faster-stratum-correct-score},
we obtain, exactly as in the interference-ratio
comparison
\eqref{eq:linear-L2-unary-leading-interference-ratio},
\begin{equation}
\min_{\substack{
c\in\mathcal O^{(m)}\\
i\in\mathcal C_c
}}
\frac{
(\mathbf M_0(\tau)\mathbf e_i)_{r(c)}
}{
(\mathbf M_0(\tau)\mathbf e_i)_c
}
\ge
\left(L-o_{\mathbb P}(1)\right)
H^{-1/2+\zeta(s'-s)}.
\label{eq:linear-L2-unary-faster-stratum-relative-scale}
\end{equation}
Under
\eqref{eq:linear-L2-unary-faster-stratum-threshold},
the exponent is nonnegative. Hence the
right-hand side is strictly greater than one
with probability tending to one, including
at equality, since \(L>1\).
This proves
\eqref{eq:linear-L2-unary-faster-stratum-dominance}.

Finally, for each fixed \(p,H\), the time
\eqref{eq:linear-L2-unary-faster-stratum-time}
is finite. The normalized predictor convergence
in
Theorem~\ref{thm:linear-L2-microscopic-linearization}
therefore transfers all strict score inequalities
to the exact gradient flow for sufficiently small
\(\sigma\). Taking the sequential limits proves
the remaining assertion.
\end{proof}

We can now combine the early target-driven organization with the random
occupancy statistics and the interference mechanism to obtain the
random-unary accuracy dynamics.

\begin{linkedtheorem}
{Random unary operation}
{thm:main-random-unary}
\label{thm:linear-L2-random-unary}
Consider the random unary model
\eqref{eq:linear-L2-random-unary-model}.
For every \(\alpha\in(0,\infty)\), there exist deterministic times
\begin{equation}
\tau_{\mathrm{org}}
<
\tau_{\mathrm{bot}},
\label{eq:linear-L2-random-unary-time-order}
\end{equation}
which may depend on \(p,H\), but not on \(\sigma\), the random
target, or the random initialization, such that, in the sequential limit
\(\sigma\downarrow0\) followed by
\(p,H\to\infty\) with \(H/p\to\alpha\),
\begin{equation}
\Acc\left(\mathbf M^{(\sigma)}(0)\right)
\longrightarrow0,
\qquad
\Acc\left(\mathbf M^{(\sigma)}(\tau_{\mathrm{org}})\right)
\longrightarrow1,
\qquad
\Acc\left(\mathbf M^{(\sigma)}(\tau_{\mathrm{bot}})\right)
\longrightarrow0
\label{eq:linear-L2-random-unary-microscopic-profile}
\end{equation}
in probability.

The organization time may be chosen as
\begin{equation}
\tau_{\mathrm{org}}
=
K\sqrt{\frac{\log(2p^2H)}{H}}
\label{eq:linear-L2-random-unary-organization-time}
\end{equation}
for any sufficiently large fixed universal constant \(K\).

Moreover, for each fixed \(p,H\) and fixed realization of
the random target, almost surely over the Gaussian initialization
as \(\sigma\downarrow0\), the entire trajectory up to the
bottleneck remains microscopic:
\begin{equation}
\sup_{0\le\tau\le\tau_{\mathrm{bot}}}
\left\|\mathbf M^{(\sigma)}(\tau)\right\|_{\mathrm{op}}
\longrightarrow0,
\qquad
\sup_{0\le\tau\le\tau_{\mathrm{bot}}}
\left|
\mathcal L\left(
\mathbf U^{(\sigma)}(\tau),
\mathbf V^{(\sigma)}(\tau)
\right)
-
\mathcal L\left(
\mathbf U^{(\sigma)}(0),
\mathbf V^{(\sigma)}(0)
\right)
\right|
\longrightarrow0.
\label{eq:linear-L2-random-unary-microscopic-flat-loss}
\end{equation}

If, in addition,
\begin{equation}
\alpha>1-e^{-1},
\label{eq:linear-L2-random-unary-realizable-alpha}
\end{equation}
then
\begin{equation}
\mathbb P_f
\left(
H\ge\operatorname{rank}(\mathbf T)
\right)
\longrightarrow1,
\label{eq:linear-L2-random-unary-realizable-probability}
\end{equation}
where \(\mathbb P_f\) denotes probability over the random target.

On this realizability event, for every fixed \(\sigma>0\),
almost surely over the Gaussian initialization, there exists
a finite, possibly \(\sigma\)- and realization-dependent
recovery time
\(\tau_{\mathrm{rec}}^{(\sigma)}>\tau_{\mathrm{bot}}\)
such that
\begin{equation}
\Acc\left(\mathbf M^{(\sigma)}(\tau)\right)
=
1
\qquad
\text{for every }\tau\ge\tau_{\mathrm{rec}}^{(\sigma)}.
\label{eq:linear-L2-random-unary-permanent-recovery}
\end{equation}
\end{linkedtheorem}

\begin{proof}
The initial-accuracy statement in
\eqref{eq:linear-L2-random-unary-microscopic-profile}
follows from earlier calculation
\eqref{eq:linear-L2-unary-initial-accuracy-limit}.

\smallskip
\noindent
\emph{Early target-driven organization.}
By the random occupancy statistics in
Lemma~\ref{lem:linear-L2-random-unary-occupancies},
the maximal occupancy satisfies
\begin{equation}
n_{\max}
=
O_{\mathbb P}
\left(
\frac{\log p}{\log\log p}
\right).
\label{eq:linear-L2-random-unary-proof-max-occupancy}
\end{equation}
Since \(H/p\to\alpha\in(0,\infty)\), this implies
\begin{equation}
n_{\max}^{3/2}
\frac{\log(2p^2H)}{H}
\longrightarrow0
\qquad
\text{in probability}.
\label{eq:linear-L2-random-unary-width-condition-probability}
\end{equation}
Thus the width condition of
Proposition~\ref{prop:linear-L2-unary-early-perfect}
holds with probability tending to one over the random target.

Applying that proposition with the deterministic
organization time
\begin{equation}
\tau_{\mathrm{org}}
=
K\sqrt{\frac{\log(2p^2H)}{H}},
\label{eq:linear-L2-random-unary-proof-organization-time}
\end{equation}
where \(K\) is any sufficiently large fixed universal
constant, we obtain
\begin{equation}
\Acc\left(
\mathbf M^{(\sigma)}(\tau_{\mathrm{org}})
\right)
\longrightarrow1
\label{eq:linear-L2-random-unary-early-perfect-limit}
\end{equation}
in probability in the sequential limit
\(\sigma\downarrow0\) followed by \(p,H\to\infty\).
Moreover,
\begin{equation}
\tau_{\mathrm{org}}
\longrightarrow0.
\label{eq:linear-L2-random-unary-organization-time-limit}
\end{equation}

\smallskip
\noindent
\emph{Microscopic accuracy bottleneck.}
We show that the deterministic choice
\begin{equation}
\tau_{\mathrm{bot}}
=
\log H
\label{eq:linear-L2-random-unary-down-time}
\end{equation}
satisfies the required vanishing-accuracy conclusion.

Fix an arbitrary integer \(J\ge1\).
We first establish misclassification of all examples
whose target output has occupancy at most \(J\).

Choose the fixed faster occupancy level
\begin{equation}
n=4J.
\label{eq:linear-L2-random-unary-faster-level-choice}
\end{equation}
By the Poisson occupancy law
\eqref{eq:linear-L2-random-unary-Poisson-occupancies},
\begin{equation}
\frac{|\mathcal O^{(4J)}|}{p}
=
\frac{N_{4J}}{p}
\longrightarrow
\frac{e^{-1}}{(4J)!}
>
0
\qquad
\text{in probability}.
\label{eq:linear-L2-random-unary-faster-stratum-density}
\end{equation}
Therefore, with probability tending to one over
the random target, the occupancy-\(4J\) stratum
contains a positive fraction of all outputs.

For every \(1\le m\le J\), we have
\begin{equation}
\sqrt{4J}-\sqrt m
\ge
\sqrt J
\ge1.
\end{equation}
Consequently,
\begin{equation}
1
>
\frac{1}{2(\sqrt{4J}-\sqrt m)}.
\label{eq:linear-L2-random-unary-all-lower-thresholds}
\end{equation}
Thus the interference threshold
\eqref{eq:linear-L2-unary-faster-stratum-threshold}
is satisfied with \(\zeta=1\) for every
occupancy \(1\le m\le J\).

Conditioning on the random target,
Lemma~\ref{lem:linear-L2-unary-extensive-faster-stratum}
therefore implies that, at time
\(\tau_{\mathrm{bot}}=\log H\), every training
example whose target output has occupancy
\(1\le m\le J\) is misclassified with probability
tending to one in the sequential limit.

Indeed, the conditional Gaussian estimates in
that lemma are uniform over target configurations
for which the chosen faster stratum contains
at least a fixed positive fraction of the outputs.
By
\eqref{eq:linear-L2-random-unary-faster-stratum-density},
this condition holds with probability tending
to one over the random target.
Since \(J\) is fixed, a union bound makes the
misclassification conclusions simultaneous
for all \(1\le m\le J\).

On this high-probability event, only examples
whose target output has occupancy greater than
\(J\) can be correctly classified. Therefore,
\begin{equation}
\Acc\left(
\mathbf M^{(\sigma)}(\tau_{\mathrm{bot}})
\right)
\le
\frac1p\sum_{m>J}mN_m.
\label{eq:linear-L2-random-unary-tail-bound}
\end{equation}

Since every input belongs to exactly one
occupied fiber,
\begin{equation}
\sum_{m\ge1}mN_m=p.
\label{eq:linear-L2-random-unary-total-input-count}
\end{equation}
Using the size-biased occupancy convergence
\eqref{eq:linear-L2-random-unary-size-biased-occupancies},
we obtain, for every fixed \(J\),
\begin{equation}
\begin{aligned}
\frac1p\sum_{m>J}mN_m
&=
1-\sum_{m=1}^{J}\frac{mN_m}{p}
\\
&\longrightarrow
1-e^{-1}\sum_{m=1}^{J}\frac1{(m-1)!}
=
e^{-1}\sum_{j=J}^{\infty}\frac1{j!}
=:r_J
\end{aligned}
\label{eq:linear-L2-random-unary-tail-mass}
\end{equation}
in probability.

The limiting tail mass satisfies
\begin{equation}
r_J\longrightarrow0
\qquad
\text{as }J\to\infty.
\label{eq:linear-L2-random-unary-tail-mass-zero}
\end{equation}

Now fix any \(\varepsilon>0\).
Choose a sufficiently large but fixed \(J\)
such that \(r_J<\varepsilon/2\).
The misclassification result for occupancies
\(1\le m\le J\), together with the tail convergence
\eqref{eq:linear-L2-random-unary-tail-mass},
then gives
\begin{equation}
\lim_{\substack{p,H\to\infty\\H/p\to\alpha}}
\limsup_{\sigma\downarrow0}
\mathbb P\left(
\Acc\left(
\mathbf M^{(\sigma)}(\log H)
\right)
>
\varepsilon
\right)
=
0.
\label{eq:linear-L2-random-unary-interference-collapse}
\end{equation}
Since \(\varepsilon>0\) was arbitrary, this proves
\begin{equation}
\Acc\left(
\mathbf M^{(\sigma)}(\tau_{\mathrm{bot}})
\right)
\longrightarrow0
\end{equation}
in probability in the claimed sequential limit.

Finally,
\begin{equation}
\tau_{\mathrm{bot}}
=
\log H
\longrightarrow\infty,
\label{eq:linear-L2-random-unary-interference-time-limit}
\end{equation}
whereas
\(\tau_{\mathrm{org}}\to0\) by
\eqref{eq:linear-L2-random-unary-organization-time-limit}.
Hence
\begin{equation}
\tau_{\mathrm{org}}
<
\tau_{\mathrm{bot}}
\label{eq:linear-L2-random-unary-order-proof}
\end{equation}
for all sufficiently large \(p,H\).

\smallskip
\noindent
\emph{Uniformly microscopic dynamics and flat loss.}
Fix \(p,H\) and a realization of the random target.
Although \(\tau_{\mathrm{bot}}=\log H\) diverges
in the large-system limit, it is a finite
deterministic time during the inner limit
\(\sigma\downarrow0\).

Theorem~\ref{thm:linear-L2-microscopic-linearization}
establishes uniform convergence of the normalized
exact predictor to the linearized microscopic
predictor on every fixed finite time interval.
Applying it on
\([0,\tau_{\mathrm{bot}}]\) gives
\begin{equation}
\sup_{0\le\tau\le\tau_{\mathrm{bot}}}
\left\|
\sigma^{-2}\mathbf M^{(\sigma)}(\tau)
-
\mathbf M_0(\tau)
\right\|_{\mathrm{op}}
\longrightarrow0
\label{eq:linear-L2-random-unary-uniform-linearization}
\end{equation}
almost surely as \(\sigma\downarrow0\).

For fixed \(p,H\), the microscopic trajectory
\(\mathbf M_0(\tau)\) is continuous on the compact
interval \([0,\tau_{\mathrm{bot}}]\) and therefore
has bounded operator norm.
Multiplying
\eqref{eq:linear-L2-random-unary-uniform-linearization}
by \(\sigma^2\) consequently yields
\begin{equation}
\sup_{0\le\tau\le\tau_{\mathrm{bot}}}
\left\|
\mathbf M^{(\sigma)}(\tau)
\right\|_{\mathrm{op}}
\longrightarrow0
\label{eq:linear-L2-random-unary-uniform-microscopic}
\end{equation}
almost surely.

The squared-loss functional depends continuously
on the predictor \(\mathbf M\).
For fixed \(p,H\), convergence in operator norm
implies convergence in Frobenius norm.
Thus
\eqref{eq:linear-L2-random-unary-uniform-microscopic}
also gives
\begin{equation}
\sup_{0\le\tau\le\tau_{\mathrm{bot}}}
\left|
\mathcal L\left(
\mathbf U^{(\sigma)}(\tau),
\mathbf V^{(\sigma)}(\tau)
\right)
-
\mathcal L\left(
\mathbf U^{(\sigma)}(0),
\mathbf V^{(\sigma)}(0)
\right)
\right|
\longrightarrow0
\label{eq:linear-L2-random-unary-uniform-flat-loss}
\end{equation}
almost surely as \(\sigma\downarrow0\).

Together,
\eqref{eq:linear-L2-random-unary-uniform-microscopic}
and
\eqref{eq:linear-L2-random-unary-uniform-flat-loss}
establish
\eqref{eq:linear-L2-random-unary-microscopic-flat-loss}
over the entire interval
\([0,\tau_{\mathrm{bot}}]\).

\smallskip
\noindent
\emph{Final recovery.}
By the rank asymptotic
\eqref{eq:linear-L2-random-unary-rank}
from
Lemma~\ref{lem:linear-L2-random-unary-occupancies},
\begin{equation}
\frac{\operatorname{rank}(\mathbf T)}{p}
\longrightarrow
1-e^{-1}
\qquad
\text{in probability}.
\label{eq:linear-L2-random-unary-proof-rank-limit}
\end{equation}

Suppose
\(\alpha>1-e^{-1}\), as required by
\eqref{eq:linear-L2-random-unary-realizable-alpha}.
Choose \(\varepsilon>0\) such that
\begin{equation}
1-e^{-1}+\varepsilon<\alpha.
\end{equation}
Since \(H/p\to\alpha\), the rank convergence
\eqref{eq:linear-L2-random-unary-proof-rank-limit}
implies
\begin{equation}
\mathbb P_f\left(
H\ge\operatorname{rank}(\mathbf T)
\right)
\longrightarrow1.
\label{eq:linear-L2-random-unary-proof-realizability}
\end{equation}
This proves
\eqref{eq:linear-L2-random-unary-realizable-probability}.

Every unary target has decision gap
\(\delta_{\mathbf T}=1\).
On the realizability event
\eqref{eq:linear-L2-random-unary-proof-realizability},
Proposition~\ref{prop:linear-L2-target-recovery}
therefore applies.
For every fixed \(\sigma>0\), almost surely
over the Gaussian initialization, the exact
gradient flow eventually induces the classifier
of \(\mathbf T\) permanently.

Since \(\Acc(\mathbf T)=1\), there exists a finite
random recovery time after which the training
accuracy remains equal to one.
Enlarging this time if necessary, we may choose
\begin{equation}
\tau_{\mathrm{rec}}^{(\sigma)}
>
\tau_{\mathrm{bot}}.
\label{eq:linear-L2-random-unary-recovery-after-interference}
\end{equation}
This proves
\eqref{eq:linear-L2-random-unary-permanent-recovery}
and completes the proof.
\end{proof}

The theorem separates two distinct roles of width. Any fixed
\(\alpha>0\) is sufficient for the microscopic target-driven organization
and the subsequent interference-induced collapse: the classifier first
becomes perfect and later loses essentially all accuracy while both the
predictor and the change in loss remain microscopic in the
small-initialization limit. The stronger condition
\(\alpha>1-e^{-1}\) is needed only for permanent late-time recovery,
because it makes the random unary target realizable with probability tending
to one.

\paragraph{Permutation targets: permanent perfect accuracy.}

We finally consider unary targets given by permutations
\(f:[p]\to[p]\). Every output then has occupancy one:
\begin{equation}
n_c=1,
\qquad c\in[p].
\label{eq:linear-L2-permutation-occupancies}
\end{equation}
By Lemma~\ref{lem:linear-L2-unary-svd}, the target has
rank \(p\) and a completely flat singular spectrum:
\begin{equation}
s_1=\cdots=s_p=1,
\qquad
k_\star=p,
\qquad
\mathbf T_{k_\star}=\mathbf T.
\label{eq:linear-L2-permutation-flat-spectrum}
\end{equation}
Thus the top singular block coincides with the full target,
and the faster-mode interference responsible for the accuracy
collapse of heterogeneous unary targets is absent.
We will prove that, in this case, training accuracy is
asymptotically nondecreasing and reaches one permanently,
even when \(H<p\) and exact target fitting is impossible.

We first reduce an arbitrary permutation to the identity.
Let \(\mathbf P_f\in\mathbb R^{p\times p}\) denote its
permutation matrix:
\begin{equation}
\mathbf T=\mathbf P_f,
\qquad
\mathbf P_f\mathbf e_i=\mathbf e_{f(i)}.
\label{eq:linear-L2-permutation-matrix}
\end{equation}
Define the transformed output factor and predictor by
\begin{equation}
\widetilde{\mathbf V}
=
\mathbf V\mathbf P_f,
\qquad
\widetilde{\mathbf M}
=
\widetilde{\mathbf V}^{\top}\mathbf U
=
\mathbf P_f^{\top}\mathbf M.
\label{eq:linear-L2-permutation-output-rotation}
\end{equation}
Orthogonality of \(\mathbf P_f\) gives
\begin{equation}
\|\mathbf M-\mathbf P_f\|_{\mathrm F}
=
\|\widetilde{\mathbf M}-\mathbf I_p\|_{\mathrm F}.
\label{eq:linear-L2-permutation-loss-invariance}
\end{equation}
This transformation preserves the gradient-flow equations
and training accuracy, while isotropic Gaussian initialization
is invariant under the output permutation.
It therefore suffices to consider
\begin{equation}
\mathbf T=\mathbf I_p.
\label{eq:linear-L2-permutation-identity-target}
\end{equation}

For the identity target, the gradient flow becomes
\begin{equation}
\dot{\mathbf U}
=
\mathbf V(\mathbf I_p-\mathbf V^{\top}\mathbf U),
\qquad
\dot{\mathbf V}
=
\mathbf U(\mathbf I_p-\mathbf U^{\top}\mathbf V).
\label{eq:linear-L2-permutation-flow}
\end{equation}
Introduce the symmetric and antisymmetric factor combinations
\begin{equation}
\mathbf W_+
=
\frac{\mathbf U+\mathbf V}{2},
\qquad
\mathbf W_-
=
\frac{\mathbf U-\mathbf V}{2}.
\label{eq:linear-L2-permutation-plus-minus}
\end{equation}
Under Gaussian initialization,
\begin{equation}
\mathbf U^{(\sigma)}(0)=\sigma\mathbf A,
\qquad
\mathbf V^{(\sigma)}(0)=\sigma\mathbf B,
\label{eq:linear-L2-permutation-AB-initialization}
\end{equation}
where \(\mathbf A,\mathbf B\in\mathbb R^{H\times p}\)
are independent standard Gaussian matrices, define
\(\mathbf X=(\mathbf A+\mathbf B)/\sqrt2\) and
\(\mathbf Y=(\mathbf A-\mathbf B)/\sqrt2\).
Then \(\mathbf X,\mathbf Y\) are independent standard Gaussian
matrices and
\begin{equation}
\mathbf W_+^{(\sigma)}(0)
=
\frac{\sigma}{\sqrt2}\mathbf X,
\qquad
\mathbf W_-^{(\sigma)}(0)
=
\frac{\sigma}{\sqrt2}\mathbf Y.
\label{eq:linear-L2-permutation-XY-initialization}
\end{equation}
For the identity target, their columns are precisely
the growing and decaying Gaussian mode vectors
\(\boldsymbol{\gamma}_c,\boldsymbol{\xi}_c\)
defined in
\eqref{eq:linear-L2-unary-gamma-xi}.

The linearized dynamics satisfy
\begin{equation}
\dot{\mathbf W}_+=\mathbf W_+,
\qquad
\dot{\mathbf W}_-=-\mathbf W_-,
\label{eq:linear-L2-permutation-linearized-plus-minus}
\end{equation}
so the symmetric component grows exponentially while
the antisymmetric component decays. We use the growing
scale
\begin{equation}
\varepsilon=\frac{H\sigma^2}{2},
\qquad
q(\tau)=\varepsilon e^{2\tau}
\label{eq:linear-L2-permutation-q}
\end{equation}
to distinguish the early approximately linear regime
\(q\ll1\) from the subsequent nonlinear evolution.

Suppose
\begin{equation}
\frac Hp\longrightarrow\alpha\in(0,\infty).
\label{eq:linear-L2-permutation-aspect-ratio}
\end{equation}
Since \(n_{\max}=1\), the width condition of
Proposition~\ref{prop:linear-L2-unary-early-perfect}
holds for every \(\alpha>0\). Hence, in the sequential
small-initialization and large-system limit, perfect
classification is achieved at
\begin{equation}
\tau_{\mathrm{org}}
=
K\sqrt{\frac{\log(2p^2H)}{H}},
\label{eq:linear-L2-permutation-organization-time}
\end{equation}
for any sufficiently large fixed universal constant \(K\).
This occurs while the predictor and loss change remain
vanishing throughout \([0,\tau_{\mathrm{org}}]\).
Moreover, for each fixed \(p,H\),
\begin{equation}
q(\tau_{\mathrm{org}})
=
\frac{H\sigma^2}{2}e^{2\tau_{\mathrm{org}}}
\longrightarrow0
\qquad
\text{as }\sigma\downarrow0.
\label{eq:linear-L2-permutation-organization-small-q}
\end{equation}
Thus perfect accuracy emerges inside the small-\(q\) regime.

To establish permanent perfect accuracy, we combine two
controls: strict growth of all correct-class margins in
the small-\(q\) regime, and uniform approximation of the
later trajectory by an explicit balanced solution with
positive classification margins.

We begin with the Gaussian geometry of the initialization.
Define the normalized Gram matrices
\begin{equation}
\mathbf G_+
=
\frac{\mathbf X^{\top}\mathbf X}{H},
\qquad
\mathbf G_-
=
\frac{\mathbf Y^{\top}\mathbf Y}{H},
\qquad
\mathbf G_{\pm}
=
\frac{\mathbf X^{\top}\mathbf Y}{H}.
\label{eq:linear-L2-permutation-Gram-matrices}
\end{equation}

\begin{lemma}[Gaussian initialization geometry]
\label{lem:linear-L2-permutation-Gaussian-geometry}
Under the aspect-ratio assumption
\eqref{eq:linear-L2-permutation-aspect-ratio},
there exist a constant \(C_\alpha<\infty\)
and a deterministic sequence
\begin{equation}
\eta_p
=
C_\alpha\sqrt{\frac{\log p}{H}}
\longrightarrow0
\label{eq:linear-L2-permutation-eta-p}
\end{equation}
such that, with probability tending to one,
\begin{equation}
\max_{i\in[p]}
\left|
(\mathbf G_+)_{ii}-1
\right|
\le
\eta_p,
\qquad
\max_{i\in[p]}
\left|
(\mathbf G_-)_{ii}-1
\right|
\le
\eta_p,
\label{eq:linear-L2-permutation-diagonal-Gram}
\end{equation}
and
\begin{equation}
\max_{i\neq j}
\left|
(\mathbf G_+)_{ij}
\right|
\le
\eta_p,
\qquad
\max_{i\neq j}
\left|
(\mathbf G_-)_{ij}
\right|
\le
\eta_p,
\qquad
\max_{i,j}
\left|
(\mathbf G_{\pm})_{ij}
\right|
\le
\eta_p.
\label{eq:linear-L2-permutation-offdiagonal-Gram}
\end{equation}
Moreover,
\begin{equation}
\frac{
\|\mathbf X\|_{\mathrm{op}}
}{
\sqrt H
}
+
\frac{
\|\mathbf Y\|_{\mathrm{op}}
}{
\sqrt H
}
\le
C_\alpha
\label{eq:linear-L2-permutation-operator-norm-geometry}
\end{equation}
with probability tending to one. In particular, after increasing
\(C_\alpha\) if necessary,
\begin{equation}
\|\mathbf G_+\|_{\mathrm{op}}
+
\|\mathbf G_-\|_{\mathrm{op}}
+
\|\mathbf G_{\pm}\|_{\mathrm{op}}
\le
C_\alpha^2
\label{eq:linear-L2-permutation-Gram-operator-bounds}
\end{equation}
with probability tending to one.
\end{lemma}

\begin{proof}
For the identity target, the columns of
\(\mathbf X\) and \(\mathbf Y\) coincide with
the independent standard Gaussian mode vectors
\(\boldsymbol{\gamma}_c\) and
\(\boldsymbol{\xi}_c\) defined in
\eqref{eq:linear-L2-unary-gamma-xi}.

We first establish the entrywise Gram bounds.
For a standard Gaussian vector
\(\mathbf g\in\mathbb R^H\), the squared norm satisfies
\begin{equation}
\|\mathbf g\|_2^2-H
=
\sum_{h=1}^H(g_h^2-1).
\end{equation}
The summands \(g_h^2-1\) are independent, centered,
and sub-exponential, since squares of Gaussian variables
are sub-exponential and centering preserves this property.
We may therefore apply Bernstein's inequality for
independent centered sub-exponential variables
\citep[Lemma~2.8.6 and Theorem~2.9.1]{vershynin2026high}
to obtain
\begin{equation}
\mathbb P\left(
\left|
\frac{\|\mathbf g\|_2^2}{H}-1
\right|>t
\right)
\le
2e^{-cHt^2},
\qquad 0<t<1,
\label{eq:linear-L2-permutation-chi-square-tail}
\end{equation}
for a universal constant \(c>0\).
Applying this bound to the \(2p\) columns of
\(\mathbf X,\mathbf Y\), followed by a union bound,
establishes the diagonal estimates
\eqref{eq:linear-L2-permutation-diagonal-Gram}
with
\(\eta_p=C_\alpha\sqrt{\log p/H}\)
for a sufficiently large constant \(C_\alpha\).
This is the same argument used to establish
\eqref{eq:linear-L2-unary-uniform-gaussian-norms}.

For the off-diagonal entries, products of independent
Gaussian coordinates are likewise centered and
sub-exponential. Applying the same Bernstein inequality,
as quantified in
\eqref{eq:linear-L2-unary-gaussian-product-tail},
with \(u=4\log p\), followed by a union bound over
\(O(p^2)\) pairs, gives
\begin{equation}
\max_{\substack{
\mathbf a,\mathbf b\text{ columns of }\mathbf X,\mathbf Y\\
\mathbf a\neq\mathbf b
}}
|\mathbf a^\top\mathbf b|
=
O_{\mathbb P}\left(\sqrt{H\log p}\right).
\end{equation}
This recovers the cross-mode estimate
\eqref{eq:linear-L2-unary-cross-inner-product-bound}
and additionally includes the same-index mixed
products between columns of \(\mathbf X\) and
\(\mathbf Y\), which are also independent.
Consequently,
\eqref{eq:linear-L2-permutation-offdiagonal-Gram}
holds with the same \(\eta_p\), after increasing
\(C_\alpha\) if necessary.

For the operator norms, the standard Gaussian
matrix estimate gives
\begin{equation}
\mathbb P\left(
\|\mathbf X\|_{\mathrm{op}}
>
\sqrt H+\sqrt p+u
\right)
\le e^{-u^2/2},
\qquad u\ge0,
\label{eq:linear-L2-permutation-Gaussian-operator-tail}
\end{equation}
and the same estimate holds for \(\mathbf Y\).
Taking \(u=\sqrt H\) and using
\(p/H\to\alpha^{-1}\) proves
\eqref{eq:linear-L2-permutation-operator-norm-geometry}
with probability tending to one.

Finally, by the definitions of the Gram matrices
\eqref{eq:linear-L2-permutation-Gram-matrices}
and submultiplicativity of the operator norm,
\begin{equation}
\begin{aligned}
\|\mathbf G_+\|_{\mathrm{op}}
+
\|\mathbf G_-\|_{\mathrm{op}}
+
\|\mathbf G_{\pm}\|_{\mathrm{op}}
&\le
\frac{
\|\mathbf X\|_{\mathrm{op}}^2
+
\|\mathbf Y\|_{\mathrm{op}}^2
+
\|\mathbf X\|_{\mathrm{op}}
\|\mathbf Y\|_{\mathrm{op}}
}{H}
\\
&\le
\left(
\frac{
\|\mathbf X\|_{\mathrm{op}}
+
\|\mathbf Y\|_{\mathrm{op}}
}{\sqrt H}
\right)^2
\le C_\alpha^2.
\end{aligned}
\end{equation}
This establishes
\eqref{eq:linear-L2-permutation-Gram-operator-bounds}
and completes the proof.
\end{proof}

The Gaussian geometry established above ensures that the
Gram matrices \(\mathbf G_+\) and \(\mathbf G_-\) have
diagonal entries close to one and uniformly small
off-diagonal entries. We now show that, in a fixed
small-\(q\) regime, the derivative of the predictor
is governed by these matrices, with nonlinear corrections
of order \(q^2\). Consequently, all correct-class
margins increase simultaneously.

\begin{lemma}[Small-\(q\) margin monotonicity]
\label{lem:linear-L2-permutation-small-q-monotonicity}
Consider the identity target \(\mathbf T=\mathbf I_p\)
along a large-system sequence
\begin{equation}
p,H\longrightarrow\infty,
\qquad
\frac Hp\longrightarrow\alpha\in(0,\infty).
\label{eq:linear-L2-permutation-small-q-aspect-ratio}
\end{equation}
There exists a constant
\begin{equation}
q_{\mathrm{sm}}=q_{\mathrm{sm}}(\alpha)>0
\label{eq:linear-L2-permutation-q-small}
\end{equation}
and events \(\mathcal E_{p,H}\), depending only on
the Gaussian matrices \(\mathbf X,\mathbf Y\), such that
\begin{equation}
\mathbb P(\mathcal E_{p,H})
\longrightarrow1.
\label{eq:linear-L2-permutation-small-q-event}
\end{equation}

On \(\mathcal E_{p,H}\), for all sufficiently large
\(p,H\), the following holds simultaneously for
every \(\sigma>0\) satisfying
\begin{equation}
\varepsilon
=
\frac{H\sigma^2}{2}
\le q_{\mathrm{sm}}.
\label{eq:linear-L2-permutation-small-q-initial-scale}
\end{equation}

For every \(i\in[p]\) and \(c\neq i\), define the
correct-versus-incorrect classification margin
\begin{equation}
\Delta_{ci}(\tau)
=
M_{ii}^{(\sigma)}(\tau)
-
M_{ci}^{(\sigma)}(\tau).
\label{eq:linear-L2-permutation-margin}
\end{equation}
Then
\begin{equation}
\frac{d}{d\tau}\Delta_{ci}(\tau)
\ge q(\tau)>0
\label{eq:linear-L2-permutation-margin-increasing}
\end{equation}
simultaneously for all \(i\in[p]\), \(c\neq i\),
and all \(\tau\ge0\) satisfying
\begin{equation}
q(\tau)
=
\varepsilon e^{2\tau}
\le q_{\mathrm{sm}}.
\label{eq:linear-L2-permutation-small-q-range}
\end{equation}

Consequently, the training accuracy
\begin{equation}
\tau\longmapsto
\Acc\bigl(\mathbf M^{(\sigma)}(\tau)\bigr)
\label{eq:linear-L2-permutation-small-q-accuracy}
\end{equation}
is nondecreasing throughout the interval
on which \(q(\tau)\le q_{\mathrm{sm}}\).
\end{lemma}

\begin{proof}
\emph{Proof strategy.}
We compare the exact factor dynamics with their linearization
around the origin throughout the small-\(q\) regime.
The linearized dynamics yield a predictor whose
correct-class margins have strictly positive derivatives,
because the Gaussian Gram matrices are diagonally dominant
entrywise. We show that the nonlinear corrections to the
factors are \(O(q^{3/2})\), which produces an \(O(q^2)\)
error in the predictor derivative. Since the leading
margin derivatives are of order \(q\), these corrections
cannot change their sign when \(q\) is sufficiently small.
All estimates will be uniform over the admissible
initialization scales and times.

We work on the high-probability event
\(\mathcal E_{p,H}\) supplied by
Lemma~\ref{lem:linear-L2-permutation-Gaussian-geometry}.
For all sufficiently large \(p,H\), this event satisfies
\begin{equation}
\eta_p\le\frac18,
\qquad
\frac{\|\mathbf X\|_{\mathrm{op}}}{\sqrt H}
+
\frac{\|\mathbf Y\|_{\mathrm{op}}}{\sqrt H}
\le C,
\label{eq:linear-L2-permutation-good-event}
\end{equation}
where \(C=C_\alpha<\infty\).
Throughout the proof, we suppress the superscript
\((\sigma)\) on the exact trajectory.

\smallskip
\noindent
\emph{Step 1: approximation of the factors.}

Recall the symmetric and antisymmetric factor
combinations
\eqref{eq:linear-L2-permutation-plus-minus}.
Define the corresponding parts of the predictor by
\begin{equation}
\mathbf M_{\mathrm{s}}
=
\frac{\mathbf M+\mathbf M^\top}{2},
\qquad
\mathbf M_{\mathrm{a}}
=
\frac{\mathbf M^\top-\mathbf M}{2}.
\label{eq:linear-L2-permutation-M-symmetric-antisymmetric}
\end{equation}
The identity-target flow
\eqref{eq:linear-L2-permutation-flow}
then becomes
\begin{equation}
\dot{\mathbf W}_+
=
\mathbf W_+
-
\mathbf W_+\mathbf M_{\mathrm{s}}
-
\mathbf W_-\mathbf M_{\mathrm{a}},
\label{eq:linear-L2-permutation-Wplus-flow}
\end{equation}
and
\begin{equation}
\dot{\mathbf W}_-
=
-\mathbf W_-
+
\mathbf W_+\mathbf M_{\mathrm{a}}
+
\mathbf W_-\mathbf M_{\mathrm{s}}.
\label{eq:linear-L2-permutation-Wminus-flow}
\end{equation}

Set
\begin{equation}
w(\tau)
=
\|\mathbf W_+(\tau)\|_{\mathrm{op}}
+
\|\mathbf W_-(\tau)\|_{\mathrm{op}}.
\label{eq:linear-L2-permutation-factor-size}
\end{equation}
Since
\begin{equation}
\|\mathbf M_{\mathrm{s}}\|_{\mathrm{op}},
\,
\|\mathbf M_{\mathrm{a}}\|_{\mathrm{op}}
\le
\|\mathbf M\|_{\mathrm{op}}
\le w(\tau)^2,
\label{eq:linear-L2-permutation-M-bound}
\end{equation}
the nonlinear terms in each of
\eqref{eq:linear-L2-permutation-Wplus-flow}
and
\eqref{eq:linear-L2-permutation-Wminus-flow}
have operator norm at most \(w(\tau)^3\).

We now compare the exact factors with the solutions of their
linearized dynamics around the origin. Dropping the nonlinear
terms in
\eqref{eq:linear-L2-permutation-Wplus-flow}
and
\eqref{eq:linear-L2-permutation-Wminus-flow}
gives the decoupled linear system
\eqref{eq:linear-L2-permutation-linearized-plus-minus}.
Denote its solutions by
\(\overline{\mathbf W}_+(\tau)\) and
\(\overline{\mathbf W}_-(\tau)\), with the same initial
conditions as the exact factors.

Solving these linear equations and using the Gaussian
initialization
\eqref{eq:linear-L2-permutation-XY-initialization},
we obtain
\begin{equation}
\begin{aligned}
\overline{\mathbf W}_+(\tau)
&=
e^\tau\mathbf W_+(0)
=
\frac{\sigma e^\tau}{\sqrt2}\mathbf X
=
\sqrt{\frac{q(\tau)}{H}}\mathbf X,
\\
\overline{\mathbf W}_-(\tau)
&=
e^{-\tau}\mathbf W_-(0)
=
\frac{\sigma e^{-\tau}}{\sqrt2}\mathbf Y
=
\sqrt{\frac{r(\tau)}{H}}\mathbf Y,
\end{aligned}
\label{eq:linear-L2-permutation-linearized-W}
\end{equation}
where \(q(\tau)=\varepsilon e^{2\tau}\) was defined in
\eqref{eq:linear-L2-permutation-q}, and we have introduced
the analogous decaying scale
\begin{equation}
r(\tau)
=
\varepsilon e^{-2\tau}
\le q(\tau),
\qquad \tau\ge0.
\label{eq:linear-L2-permutation-r}
\end{equation}
Thus \(q\) and \(r\) characterize the growing and
decaying components of the linearized dynamics,
respectively. 

We now use a bootstrap argument to control the exact factors
throughout the small-\(q\) regime. The idea is to assume
temporarily that their norms remain of order \(\sqrt q\),
use this assumption to bound the nonlinear corrections
to the linearized trajectories, and then show that the
resulting estimates improve the original assumption.

First, the Gaussian operator-norm bound
\eqref{eq:linear-L2-permutation-good-event}
and the linearized solutions
\eqref{eq:linear-L2-permutation-linearized-W}
give
\begin{equation}
\|\overline{\mathbf W}_+(\tau)\|_{\mathrm{op}}
+
\|\overline{\mathbf W}_-(\tau)\|_{\mathrm{op}}
\le C\sqrt{q(\tau)}.
\label{eq:linear-L2-permutation-linearized-W-bound}
\end{equation}
We aim to establish the analogous bound for the exact
factors,
\begin{equation}
w(\tau)\le D\sqrt{q(\tau)},
\qquad D=2C,
\label{eq:linear-L2-permutation-factor-bootstrap}
\end{equation}
whenever \(q(\tau)\le q_{\mathrm{sm}}\), for a sufficiently
small constant \(q_{\mathrm{sm}}>0\).

To this end, suppose temporarily that
\eqref{eq:linear-L2-permutation-factor-bootstrap}
holds for every \(0\le u\le\tau\).
By the cubic bound on the nonlinear terms in
\eqref{eq:linear-L2-permutation-Wplus-flow}
and
\eqref{eq:linear-L2-permutation-Wminus-flow},
their operator norms are then bounded by
\(D^3q(u)^{3/2}\).
Applying variation of constants to these equations
and subtracting the linearized solutions
\eqref{eq:linear-L2-permutation-linearized-W}
therefore yields
\begin{equation}
\begin{aligned}
\left\|
\mathbf W_+(\tau)
-
\overline{\mathbf W}_+(\tau)
\right\|_{\mathrm{op}}
&\le
D^3\int_0^\tau
e^{\tau-u}q(u)^{3/2}\,du
\\
&\le
\frac{D^3}{2}q(\tau)^{3/2},
\end{aligned}
\label{eq:linear-L2-permutation-Wplus-error}
\end{equation}
and similarly
\begin{equation}
\begin{aligned}
\left\|
\mathbf W_-(\tau)
-
\overline{\mathbf W}_-(\tau)
\right\|_{\mathrm{op}}
&\le
D^3\int_0^\tau
e^{-(\tau-u)}q(u)^{3/2}\,du
\\
&\le
\frac{D^3}{4}q(\tau)^{3/2}.
\end{aligned}
\label{eq:linear-L2-permutation-Wminus-error}
\end{equation}
Here the integrals were evaluated using
\(q(u)=\varepsilon e^{2u}\), as defined in
\eqref{eq:linear-L2-permutation-q}.

Combining these error estimates with the linearized
bound
\eqref{eq:linear-L2-permutation-linearized-W-bound}
gives
\begin{equation}
w(\tau)
\le
C\sqrt{q(\tau)}
+
\frac{3D^3}{4}q(\tau)^{3/2}.
\label{eq:linear-L2-permutation-bootstrap-improvement}
\end{equation}
Since \(D=2C\), choosing
\begin{equation}
q_{\mathrm{sm}}
\le
\frac{1}{12C^2}
\label{eq:linear-L2-permutation-bootstrap-q-small}
\end{equation}
ensures that, whenever \(q(\tau)\le q_{\mathrm{sm}}\),
the right-hand side is at most
\(\frac32C\sqrt{q(\tau)}\), strictly smaller than
the assumed upper bound \(D\sqrt{q(\tau)}\).

Thus the bootstrap assumption implies a strictly
stronger estimate. Moreover, the initial bound holds
strictly, since
\eqref{eq:linear-L2-permutation-good-event}
gives \(w(0)\le C\sqrt{\varepsilon}\).
By continuity and a first-exit argument,
\eqref{eq:linear-L2-permutation-factor-bootstrap}
therefore holds throughout the interval
\(q(\tau)\le q_{\mathrm{sm}}\).
The exact gradient flow exists globally by
Theorem~\ref{thm:linear-L2-microscopic-linearization}.

Having closed the bootstrap, we conclude that the
factor approximation errors
\eqref{eq:linear-L2-permutation-Wplus-error}
and
\eqref{eq:linear-L2-permutation-Wminus-error}
are \(O(q^{3/2})\) throughout this interval,
uniformly over all admissible initialization scales
\(\sigma\).

\smallskip
\noindent
\emph{Step 2: approximation of the predictor derivative.}

We next transfer the factor estimates to
the derivative of the end-to-end predictor.
Differentiating
\(\mathbf M=\mathbf V^\top\mathbf U\)
along
\eqref{eq:linear-L2-permutation-flow}
gives
\begin{equation}
\begin{aligned}
\dot{\mathbf M}
&=
(\mathbf I_p-\mathbf M)\mathbf U^\top\mathbf U
+
\mathbf V^\top\mathbf V(\mathbf I_p-\mathbf M)
\\
&=
2\mathbf W_+^\top\mathbf W_+
+
2\mathbf W_-^\top\mathbf W_-
-
\mathbf M\mathbf U^\top\mathbf U
-
\mathbf V^\top\mathbf V\mathbf M.
\end{aligned}
\label{eq:linear-L2-permutation-Mdot}
\end{equation}

The leading terms are quadratic in the factors.
By submultiplicativity,
\begin{equation}
\begin{aligned}
&
\left\|
\mathbf W_\pm^\top\mathbf W_\pm
-
\overline{\mathbf W}_\pm^\top
\overline{\mathbf W}_\pm
\right\|_{\mathrm{op}}
\\
&\qquad\le
\left(
\|\mathbf W_\pm\|_{\mathrm{op}}
+
\|\overline{\mathbf W}_\pm\|_{\mathrm{op}}
\right)
\left\|
\mathbf W_\pm-\overline{\mathbf W}_\pm
\right\|_{\mathrm{op}}.
\end{aligned}
\label{eq:linear-L2-permutation-quadratic-factor-difference}
\end{equation}
The factor bounds
\eqref{eq:linear-L2-permutation-linearized-W-bound}
and
\eqref{eq:linear-L2-permutation-factor-bootstrap},
together with the \(O(q^{3/2})\) errors
\eqref{eq:linear-L2-permutation-Wplus-error}
and
\eqref{eq:linear-L2-permutation-Wminus-error},
therefore yield
\begin{equation}
\sum_{\nu\in\{+,-\}}
\left\|
\mathbf W_\nu^\top\mathbf W_\nu
-
\overline{\mathbf W}_\nu^\top
\overline{\mathbf W}_\nu
\right\|_{\mathrm{op}}
\le
\frac34(D+C)D^3q^2.
\label{eq:linear-L2-permutation-quadratic-Gram-errors}
\end{equation}

Moreover,
\eqref{eq:linear-L2-permutation-factor-bootstrap}
implies
\(\|\mathbf U\|_{\mathrm{op}},
\|\mathbf V\|_{\mathrm{op}}\le D\sqrt q\).
Thus the explicitly nonlinear terms in
\eqref{eq:linear-L2-permutation-Mdot}
satisfy
\begin{equation}
\left\|
\mathbf M\mathbf U^\top\mathbf U
+
\mathbf V^\top\mathbf V\mathbf M
\right\|_{\mathrm{op}}
\le
2D^4q^2.
\label{eq:linear-L2-permutation-Mdot-nonlinear-bound}
\end{equation}

Finally, the linearized factors in
\eqref{eq:linear-L2-permutation-linearized-W}
satisfy
\begin{equation}
2\overline{\mathbf W}_+^\top
\overline{\mathbf W}_+
=
2q\mathbf G_+,
\qquad
2\overline{\mathbf W}_-^\top
\overline{\mathbf W}_-
=
2r\mathbf G_-.
\label{eq:linear-L2-permutation-linearized-Grams}
\end{equation}
Combining these identities with
\eqref{eq:linear-L2-permutation-Mdot}
and the error estimates above, we arrive at
\begin{equation}
\dot{\mathbf M}
=
2q\mathbf G_+
+
2r\mathbf G_-
+
\mathbf E,
\qquad
\|\mathbf E\|_{\mathrm{op}}
\le C_1q^2,
\label{eq:linear-L2-permutation-Mdot-Gram}
\end{equation}
where
\begin{equation}
C_1
=
\frac32(D+C)D^3+2D^4
<\infty
\label{eq:linear-L2-permutation-C1}
\end{equation}
depends only on \(\alpha\).

Thus the predictor derivative is approximated
by the derivative of the linearized predictor,
with a uniform \(O(q^2)\) error.
This is the main estimate needed for monotonicity.

\smallskip
\noindent
\emph{Step 3: strict growth of classification margins.}

Fix \(i\in[p]\) and \(c\neq i\).
By the margin definition
\eqref{eq:linear-L2-permutation-margin}
and the predictor derivative
\eqref{eq:linear-L2-permutation-Mdot-Gram},
\begin{equation}
\begin{aligned}
\dot{\Delta}_{ci}
&=
2q\left[
(\mathbf G_+)_{ii}-(\mathbf G_+)_{ci}
\right]
\\
&\quad+
2r\left[
(\mathbf G_-)_{ii}-(\mathbf G_-)_{ci}
\right]
+
(\mathbf e_i-\mathbf e_c)^\top
\mathbf E\mathbf e_i.
\end{aligned}
\label{eq:linear-L2-permutation-margin-derivative}
\end{equation}

The Gaussian geometry estimates
\eqref{eq:linear-L2-permutation-diagonal-Gram}
and
\eqref{eq:linear-L2-permutation-offdiagonal-Gram}
imply, on \(\mathcal E_{p,H}\),
\begin{equation}
(\mathbf G_\nu)_{ii}
-
(\mathbf G_\nu)_{ci}
\ge
1-2\eta_p
\ge
\frac34,
\qquad
\nu\in\{+,-\}.
\label{eq:linear-L2-permutation-Gram-margin}
\end{equation}
Thus both leading contributions to
\eqref{eq:linear-L2-permutation-margin-derivative}
are nonnegative, and the first is at least
\(3q/2\).

The nonlinear error satisfies
\begin{equation}
\left|
(\mathbf e_i-\mathbf e_c)^\top
\mathbf E\mathbf e_i
\right|
\le
\sqrt2\,\|\mathbf E\|_{\mathrm{op}}
\le
\sqrt2 C_1q^2.
\label{eq:linear-L2-permutation-margin-error}
\end{equation}
Consequently,
\begin{equation}
\dot{\Delta}_{ci}(\tau)
\ge
\frac32q(\tau)
-
\sqrt2 C_1q(\tau)^2.
\label{eq:linear-L2-permutation-margin-positive-bound}
\end{equation}

We choose \(q_{\mathrm{sm}}\) sufficiently small
that, in addition to
\eqref{eq:linear-L2-permutation-bootstrap-q-small},
\begin{equation}
\sqrt2 C_1q_{\mathrm{sm}}
\le
\frac12.
\label{eq:linear-L2-permutation-q-small-margin-condition}
\end{equation}
Then
\begin{equation}
\dot{\Delta}_{ci}(\tau)
\ge
q(\tau)>0
\qquad
\text{whenever }q(\tau)\le q_{\mathrm{sm}},
\label{eq:linear-L2-permutation-margin-final-positive}
\end{equation}
simultaneously for every \(i\in[p]\) and \(c\neq i\).

All estimates hold on the same event
\(\mathcal E_{p,H}\), independently of \(\sigma\).
Hence they apply simultaneously to every
initialization scale satisfying
\eqref{eq:linear-L2-permutation-small-q-initial-scale},
and the probability of this event tends to one
by
Lemma~\ref{lem:linear-L2-permutation-Gaussian-geometry}.

Since every correct-versus-incorrect margin
is strictly increasing, an example that is
correctly classified cannot subsequently become
misclassified while \(q\le q_{\mathrm{sm}}\).
Therefore training accuracy is nondecreasing
throughout the small-\(q\) interval.
This proves
\eqref{eq:linear-L2-permutation-small-q-accuracy}
and completes the proof.
\end{proof}

\begin{lemma}[Early microscopic organization and small-\(q\) persistence]
\label{lem:linear-L2-permutation-early-organization}
Consider the identity target \(\mathbf T=\mathbf I_p\)
along a large-system sequence
\(p,H\to\infty\) with
\(H/p\to\alpha\in(0,\infty)\).
Let
\begin{equation}
\delta=q_{\mathrm{sm}}(\alpha)>0
\label{eq:linear-L2-permutation-early-delta}
\end{equation}
be the constant from
Lemma~\ref{lem:linear-L2-permutation-small-q-monotonicity},
and define the endpoint of the small-\(q\) interval by
\begin{equation}
\tau_\sigma(\delta)
=
\frac12\log\frac{\delta}{\varepsilon},
\qquad
\varepsilon=\frac{H\sigma^2}{2}.
\label{eq:linear-L2-permutation-small-q-endpoint}
\end{equation}
Let
\begin{equation}
\tau_{\mathrm{org}}
=
K\sqrt{\frac{\log(2p^2H)}{H}},
\label{eq:linear-L2-permutation-early-organization-time}
\end{equation}
where \(K\) is a sufficiently large universal constant.

For each fixed \(p,H\), we have
\begin{equation}
\tau_{\mathrm{org}}
<
\tau_\sigma(\delta)
\label{eq:linear-L2-permutation-early-time-inclusion}
\end{equation}
for all sufficiently small \(\sigma\).
Moreover, in the sequential limit
\(\sigma\downarrow0\) followed by
\(p,H\to\infty\), with probability tending to one,
the training accuracy is nondecreasing throughout
\begin{equation}
0\le\tau\le\tau_\sigma(\delta),
\label{eq:linear-L2-permutation-early-monotone-interval}
\end{equation}
and satisfies
\begin{equation}
\Acc\bigl(\mathbf M^{(\sigma)}(\tau)\bigr)
=
1
\qquad
\text{for every }
\tau_{\mathrm{org}}
\le\tau\le\tau_\sigma(\delta).
\label{eq:linear-L2-permutation-early-perfect-interval}
\end{equation}

Furthermore, for every fixed \(p,H\), the predictor
remains microscopic throughout the organization interval:
\begin{equation}
\sup_{0\le\tau\le\tau_{\mathrm{org}}}
\left\|
\mathbf M^{(\sigma)}(\tau)
\right\|_{\mathrm{op}}
\longrightarrow0
\qquad
\text{as }\sigma\downarrow0,
\label{eq:linear-L2-permutation-early-microscopic}
\end{equation}
and the corresponding loss change vanishes uniformly:
\begin{equation}
\sup_{0\le\tau\le\tau_{\mathrm{org}}}
\left|
\mathcal L\left(
\mathbf U^{(\sigma)}(\tau),
\mathbf V^{(\sigma)}(\tau)
\right)
-
\mathcal L\left(
\mathbf U^{(\sigma)}(0),
\mathbf V^{(\sigma)}(0)
\right)
\right|
\longrightarrow0.
\label{eq:linear-L2-permutation-early-flat-loss}
\end{equation}
The last two convergences hold almost surely
with respect to the Gaussian initialization.
\end{lemma}

\begin{proof}
We combine the small-\(q\) monotonicity result
with the earlier perfect-classification proposition.

First, for each fixed \(p,H\), the organization time
\(\tau_{\mathrm{org}}\) is finite and independent of
\(\sigma\). By
\eqref{eq:linear-L2-permutation-q},
\begin{equation}
q(\tau_{\mathrm{org}})
=
\frac{H\sigma^2}{2}e^{2\tau_{\mathrm{org}}}
\longrightarrow0
\qquad
\text{as }\sigma\downarrow0.
\label{eq:linear-L2-permutation-early-q-zero}
\end{equation}
Since \(\delta>0\), this implies
\(\tau_{\mathrm{org}}<\tau_\sigma(\delta)\)
for all sufficiently small \(\sigma\), proving
\eqref{eq:linear-L2-permutation-early-time-inclusion}.

On the high-probability event of
Lemma~\ref{lem:linear-L2-permutation-small-q-monotonicity},
the margin estimate
\eqref{eq:linear-L2-permutation-margin-increasing}
holds simultaneously for all classes, admissible
initialization scales, and times satisfying
\(q(\tau)\le\delta\).
Since \(q(\tau)\) is increasing, this condition is
equivalent to
\(0\le\tau\le\tau_\sigma(\delta)\).
Hence training accuracy is nondecreasing throughout
\eqref{eq:linear-L2-permutation-early-monotone-interval}.

Next, since \(n_{\max}=1\), the width condition in
Proposition~\ref{prop:linear-L2-unary-early-perfect}
is satisfied under the assumed aspect-ratio scaling.
That proposition implies that, for sufficiently large
universal \(K\),
\begin{equation}
\mathbb P\left(
\Acc\bigl(
\mathbf M^{(\sigma)}(\tau_{\mathrm{org}})
\bigr)=1
\right)
\longrightarrow1
\label{eq:linear-L2-permutation-early-perfect-event}
\end{equation}
in the prescribed sequential limit.

On the intersection of this event with the
small-\(q\) monotonicity event, accuracy equals one
at \(\tau_{\mathrm{org}}\) and cannot subsequently
decrease before \(\tau_\sigma(\delta)\).
This proves
\eqref{eq:linear-L2-permutation-early-perfect-interval}.
Both events have probability tending to one,
so their intersection does as well.

Finally, for each fixed \(p,H\), apply
Theorem~\ref{thm:linear-L2-microscopic-linearization}
on the finite interval \([0,\tau_{\mathrm{org}}]\).
The uniform predictor estimate
\eqref{eq:linear-L2-predictor-microscopic}
directly yields
\eqref{eq:linear-L2-permutation-early-microscopic}.
Likewise, the uniform loss estimate
\eqref{eq:linear-L2-loss-microscopic},
applied also at \(\tau=0\), gives
\eqref{eq:linear-L2-permutation-early-flat-loss}.
This completes the proof.
\end{proof}

The preceding two lemmas establish that, with probability tending
to one, training accuracy is nondecreasing throughout the
small-\(q\) regime, reaches one at the microscopic organization time
\(\tau_{\mathrm{org}}\), and remains perfect until
\(q(\tau)=q_{\mathrm{sm}}\). However, these results do not exclude a
subsequent deterioration of the classifier. On longer timescales,
the nonlinear terms in the gradient flow become significant,
and the small-\(q\) approximation no longer applies.
To establish permanent perfect accuracy, we must therefore
control the trajectory beyond this regime, including at
arbitrarily late times.

The key observation is that, near the origin, the antisymmetric
component \(\mathbf W_-\) is a decaying direction, whereas the
symmetric component \(\mathbf W_+\) grows. This suggests that,
after shifting time to keep the growing scale \(q\) fixed as
\(\sigma\downarrow0\), the dynamics should approach the balanced
submanifold \(\mathbf U=\mathbf V\). On this submanifold, the
gradient flow admits an explicit solution. Writing
\begin{equation}
\mathbf C
=
\frac1H\mathbf X^\top\mathbf X,
\label{eq:linear-L2-permutation-C}
\end{equation}
its predictor is
\begin{equation}
\mathbf M_{\mathrm{bal}}(q)
=
q\mathbf C
\left(
\mathbf I_p+q\mathbf C
\right)^{-1},
\qquad q>0.
\label{eq:linear-L2-permutation-balanced-predictor}
\end{equation}
The following lemma makes this limiting description rigorous.
Crucially, the convergence is uniform not only on compact
\(q\)-intervals but over the entire tail \(q\ge\delta>0\).
This stronger conclusion will allow us to rule out a
classification collapse at arbitrarily late times.

\begin{lemma}[Global reduction to the balanced trajectory]
\label{lem:linear-L2-permutation-balanced-reduction}
Fix \(p,H\in\mathbb N\) and fixed matrices
\(\mathbf X,\mathbf Y\in\mathbb R^{H\times p}\), with
\begin{equation}
\operatorname{rank}(\mathbf X)
=
\min\{H,p\}.
\label{eq:linear-L2-permutation-X-full-rank}
\end{equation}
For each \(\sigma>0\), initialize
\begin{equation}
\mathbf U^{(\sigma)}(0)
=
\frac{\sigma}{\sqrt2}
\left(
\mathbf X+\mathbf Y
\right),
\qquad
\mathbf V^{(\sigma)}(0)
=
\frac{\sigma}{\sqrt2}
\left(
\mathbf X-\mathbf Y
\right),
\label{eq:linear-L2-permutation-fixed-XY-init}
\end{equation}
and let
\begin{equation}
\mathbf M^{(\sigma)}(\tau)
=
\mathbf V^{(\sigma)}(\tau)^\top
\mathbf U^{(\sigma)}(\tau)
\label{eq:linear-L2-permutation-fixed-XY-M}
\end{equation}
evolve under the identity-target flow
\eqref{eq:linear-L2-permutation-flow}.

Set
\begin{equation}
\varepsilon
=
\frac{H\sigma^2}{2},
\qquad
\tau_\sigma(q)
=
\frac12\log\frac{q}{\varepsilon},
\qquad
q\ge\varepsilon,
\label{eq:linear-L2-permutation-q-time-map}
\end{equation}
and define the predictor in \(q\)-coordinates by
\begin{equation}
\widehat{\mathbf M}_\sigma(q)
=
\mathbf M^{(\sigma)}
\left(
\tau_\sigma(q)
\right).
\label{eq:linear-L2-permutation-rescaled-predictor}
\end{equation}
Then, for every fixed \(\delta>0\),
\begin{equation}
\lim_{\sigma\downarrow0}
\sup_{q\ge\delta}
\left\|
\widehat{\mathbf M}_\sigma(q)
-
\mathbf M_{\mathrm{bal}}(q)
\right\|_{\mathrm F}
=
0.
\label{eq:linear-L2-permutation-global-balanced-reduction}
\end{equation}
Consequently,
\begin{equation}
\lim_{\sigma\downarrow0}
\sup_{q\ge\delta}
\left\|
\widehat{\mathbf M}_\sigma(q)
-
\mathbf M_{\mathrm{bal}}(q)
\right\|_{\max}
=
0.
\label{eq:linear-L2-permutation-global-balanced-reduction-max}
\end{equation}
Here the suprema are understood for sufficiently small
\(\sigma\) such that \(\varepsilon<\delta\).

Since a standard Gaussian \(H\times p\) matrix has rank
\(\min\{H,p\}\) almost surely, the conclusion applies
almost surely to the Gaussian initialization considered above.
\end{lemma}

\begin{proof}
\emph{Proof strategy.}
We establish the uniform approximation in three steps.
First, we identify an explicit trajectory on the invariant
balanced submanifold \(\mathbf U=\mathbf V\).
Second, we show that the exact trajectory approaches this
balanced solution on every compact \(q\)-interval bounded
away from zero. This is achieved by matching the two
trajectories at a vanishing intermediate scale and
propagating the resulting approximation using
Gronwall's inequality.
Finally, we extend the approximation to arbitrarily
late times. For this, we establish a local gradient
inequality near the limiting balanced equilibrium,
which ensures that trajectories entering a sufficiently
small neighborhood remain close to it forever.
The last step is essential because convergence on
compact \(q\)-intervals alone does not imply uniform
convergence over the entire infinite-time tail.

Throughout the proof, \(p,H,\mathbf X,\mathbf Y\)
are fixed. Constants in the estimates may depend on
these quantities but are independent of \(\sigma\).

\medskip
\noindent
\emph{Step 1: the balanced reference trajectory.}

We first establish global existence of the exact gradient flow.
For the identity target, the loss is
\begin{equation}
\mathcal L(\mathbf U,\mathbf V)
=
\frac{1}{2p}
\left\|
\mathbf V^\top\mathbf U-\mathbf I_p
\right\|_{\mathrm F}^2.
\label{eq:linear-L2-permutation-identity-loss}
\end{equation}
The dynamics
\eqref{eq:linear-L2-permutation-flow}
are gradient flow for this loss in rescaled time:
\begin{equation}
\dot{\mathbf U}
=
-p\nabla_{\mathbf U}\mathcal L,
\qquad
\dot{\mathbf V}
=
-p\nabla_{\mathbf V}\mathcal L.
\label{eq:linear-L2-permutation-loss-gradient-flow}
\end{equation}
Consequently,
\begin{equation}
\frac{d}{d\tau}\mathcal L
=
-p\left(
\|\nabla_{\mathbf U}\mathcal L\|_{\mathrm F}^2
+
\|\nabla_{\mathbf V}\mathcal L\|_{\mathrm F}^2
\right)
\le0.
\label{eq:linear-L2-permutation-loss-dissipation}
\end{equation}
Since the loss is nonincreasing, the predictor
\(\mathbf M=\mathbf V^\top\mathbf U\)
remains bounded in Frobenius norm on the maximal
existence interval.

To control the factors themselves, recall that the
balance difference
\begin{equation}
\mathbf D
=
\mathbf U\mathbf U^\top
-
\mathbf V\mathbf V^\top
\label{eq:linear-L2-permutation-balance-invariant}
\end{equation}
is conserved along gradient flow. Indeed, differentiating
both factor Gram matrices using
\eqref{eq:linear-L2-permutation-flow} gives
\begin{equation}
\begin{aligned}
\frac{d}{d\tau}
(\mathbf U\mathbf U^\top)
&=
\mathbf V(\mathbf I_p-\mathbf M)\mathbf U^\top
+
\mathbf U(\mathbf I_p-\mathbf M^\top)\mathbf V^\top
\\
&=
\frac{d}{d\tau}
(\mathbf V\mathbf V^\top).
\end{aligned}
\label{eq:linear-L2-permutation-balance-invariant-proof}
\end{equation}

Writing
\(\mathbf G_U=\mathbf U\mathbf U^\top\) and
\(\mathbf G_V=\mathbf V\mathbf V^\top\), we obtain
\begin{equation}
\begin{aligned}
\|\mathbf G_U+\mathbf G_V\|_{\mathrm F}^2
&=
\|\mathbf G_U-\mathbf G_V\|_{\mathrm F}^2
+
4\operatorname{Tr}(\mathbf G_U\mathbf G_V)
\\
&=
\|\mathbf D\|_{\mathrm F}^2
+
4\|\mathbf M\|_{\mathrm F}^2.
\end{aligned}
\label{eq:linear-L2-permutation-factor-bound}
\end{equation}
The right-hand side is bounded by conservation of
\(\mathbf D\) and dissipation of the loss.
Since \(\mathbf G_U,\mathbf G_V\succeq0\),
\begin{equation}
\begin{aligned}
\|\mathbf U\|_{\mathrm F}^2
+
\|\mathbf V\|_{\mathrm F}^2
&=
\operatorname{Tr}(\mathbf G_U+\mathbf G_V)
\\
&\le
\sqrt H\,
\|\mathbf G_U+\mathbf G_V\|_{\mathrm F}.
\end{aligned}
\label{eq:linear-L2-permutation-factor-norm-bound}
\end{equation}
Thus both factors remain bounded, and the standard
continuation theorem for finite-dimensional ODEs
rules out finite-time blow-up. The exact gradient
flow therefore exists for all \(\tau\ge0\).

We now identify the balanced reference trajectory.
The submanifold
\begin{equation}
\mathbf U=\mathbf V=\mathbf W
\label{eq:linear-L2-permutation-balanced-submanifold}
\end{equation}
is invariant under the identity-target dynamics
\eqref{eq:linear-L2-permutation-flow}.
Indeed, substituting
\(\mathbf U=\mathbf V=\mathbf W\)
into both factor equations gives the same evolution:
\begin{equation}
\dot{\mathbf W}
=
\mathbf W
\left(
\mathbf I_p-\mathbf W^\top\mathbf W
\right).
\label{eq:linear-L2-permutation-balanced-factor-flow}
\end{equation}
Since the growing scale \(q\) defined in
\eqref{eq:linear-L2-permutation-q}
satisfies
\begin{equation}
\frac{dq}{d\tau}=2q,
\label{eq:linear-L2-permutation-q-derivative}
\end{equation}
the balanced dynamics in \(q\)-coordinates become
\begin{equation}
2q\frac{d\mathbf W}{dq}
=
\mathbf W
\left(
\mathbf I_p-\mathbf W^\top\mathbf W
\right).
\label{eq:linear-L2-permutation-balanced-q-flow}
\end{equation}

We claim that an explicit solution of this equation is
\begin{equation}
\mathbf W_*(q)
=
\sqrt{\frac qH}\,
\mathbf X
\left(
\mathbf I_p+q\mathbf C
\right)^{-1/2},
\label{eq:linear-L2-permutation-balanced-factor}
\end{equation}
where
\(\mathbf C=\mathbf X^\top\mathbf X/H\)
was defined in
\eqref{eq:linear-L2-permutation-C}.
To verify this, first note that \(\mathbf C\)
commutes with its matrix functions, so
\begin{equation}
\begin{aligned}
\mathbf W_*(q)^\top\mathbf W_*(q)
&=
(\mathbf I_p+q\mathbf C)^{-1/2}
q\mathbf C
(\mathbf I_p+q\mathbf C)^{-1/2}
\\
&=
q\mathbf C
(\mathbf I_p+q\mathbf C)^{-1}
\\
&=
\mathbf M_{\mathrm{bal}}(q),
\end{aligned}
\label{eq:linear-L2-permutation-balanced-factor-product}
\end{equation}
recovering the balanced predictor
\eqref{eq:linear-L2-permutation-balanced-predictor}.

Next, differentiating
\eqref{eq:linear-L2-permutation-balanced-factor}
with respect to \(q\) gives
\begin{equation}
\begin{aligned}
2q\frac{d\mathbf W_*}{dq}
&=
\mathbf W_*
-
\mathbf W_*
q\mathbf C
(\mathbf I_p+q\mathbf C)^{-1}
\\
&=
\mathbf W_*
\left(
\mathbf I_p-\mathbf M_{\mathrm{bal}}(q)
\right)
\\
&=
\mathbf W_*
\left(
\mathbf I_p-\mathbf W_*^\top\mathbf W_*
\right).
\end{aligned}
\label{eq:linear-L2-permutation-balanced-factor-verification}
\end{equation}
Thus \(\mathbf W_*(q)\) solves the balanced factor
equation
\eqref{eq:linear-L2-permutation-balanced-q-flow},
and
\((\mathbf W_*(q),\mathbf W_*(q))\)
defines an exact trajectory of the original
gradient flow restricted to the balanced submanifold.

Finally, using the expansion
\begin{equation}
(\mathbf I_p+q\mathbf C)^{-1/2}
=
\mathbf I_p-\frac q2\mathbf C+O(q^2),
\qquad q\downarrow0,
\end{equation}
we obtain
\begin{equation}
\mathbf W_*(q)
=
\sqrt{\frac qH}\,\mathbf X
+
O(q^{3/2}).
\label{eq:linear-L2-permutation-balanced-small-q}
\end{equation}
The leading term coincides exactly with the
growing symmetric linearized factor
\(\overline{\mathbf W}_+(\tau)\)
in
\eqref{eq:linear-L2-permutation-linearized-W},
after expressing that solution in terms of
\(q=\varepsilon e^{2\tau}\).
Thus the balanced trajectory has precisely the
small-\(q\) behavior selected by the growing
linearized mode. In the next step, we make this
connection rigorous by proving convergence of the
exact factors to the balanced trajectory.

\medskip
\noindent
\emph{Step 2: convergence on compact \(q\)-intervals.}

We now show that the exact factors converge to the
balanced trajectory on every compact interval
\(0<\delta\le q\le Q<\infty\), after expressing both
trajectories in terms of the growing scale \(q\).

Recall from
\eqref{eq:linear-L2-permutation-q-time-map}
that
\(\tau_\sigma(q)=\frac12\log(q/\varepsilon)\)
is the inverse of \(q(\tau)=\varepsilon e^{2\tau}\).
Since \(\varepsilon=H\sigma^2/2\), we may write
\begin{equation}
\tau_\sigma(q)
=
\log\frac1{\sigma}
+
\frac12\log\frac{2q}{H}.
\label{eq:linear-L2-permutation-time-shift}
\end{equation}
Thus, for each fixed \(q>0\), the corresponding
observation time diverges as \(\sigma\downarrow0\).
The purpose of this reparametrization is to follow
the nonlinear evolution on the timescale
\(\tau\sim\log(1/\sigma)\), where the growing
component becomes order one.

Importantly, differences between observation times
at two fixed scales are independent of \(\sigma\):
\begin{equation}
\tau_\sigma(q_2)-\tau_\sigma(q_1)
=
\frac12\log\frac{q_2}{q_1}.
\label{eq:linear-L2-permutation-time-difference}
\end{equation}
This identity will allow us to apply Gronwall's
inequality directly to trajectories expressed
in \(q\)-coordinates.

We proceed in three parts: control the exact dynamics
at small \(q\), match them to the balanced trajectory
at a vanishing intermediate scale, and propagate
the comparison to arbitrary fixed \(q\)-intervals.

\smallskip
\noindent
\emph{Small-\(q\) control of the exact dynamics.}

Introduce the stacked factor variable
\begin{equation}
\mathbf Z
=
\begin{pmatrix}
\mathbf U\\
\mathbf V
\end{pmatrix}.
\label{eq:linear-L2-permutation-stacked-variable}
\end{equation}
The identity-target flow
\eqref{eq:linear-L2-permutation-flow}
can be written as
\begin{equation}
\dot{\mathbf Z}
=
\mathcal A\mathbf Z+\mathcal N(\mathbf Z),
\qquad
\mathcal A
=
\begin{pmatrix}
\mathbf 0 & \mathbf I_H\\
\mathbf I_H & \mathbf 0
\end{pmatrix},
\label{eq:linear-L2-permutation-stacked-flow}
\end{equation}
where
\begin{equation}
\mathcal N(\mathbf Z)
=
-
\begin{pmatrix}
\mathbf V\mathbf M\\
\mathbf U\mathbf M^\top
\end{pmatrix},
\qquad
\mathbf M=\mathbf V^\top\mathbf U.
\label{eq:linear-L2-permutation-stacked-nonlinearity}
\end{equation}
The nonlinearity is cubic, so there exists
\(K_0<\infty\) such that
\begin{equation}
\|\mathcal N(\mathbf Z)\|_{\mathrm F}
\le
K_0\|\mathbf Z\|_{\mathrm F}^3.
\label{eq:linear-L2-permutation-cubic-bound}
\end{equation}
Moreover, since \(\mathcal A\) is symmetric with
eigenvalues \(+1\) and \(-1\),
\begin{equation}
\|e^{\tau\mathcal A}\|_{\mathrm{op}}
=
e^\tau,
\qquad \tau\ge0.
\label{eq:linear-L2-permutation-linear-semigroup}
\end{equation}

We first establish a uniform bound on the exact
factors while \(q\) is sufficiently small.
Set
\begin{equation}
z(\tau)
=
\|\mathbf Z(\tau)\|_{\mathrm F}.
\label{eq:linear-L2-permutation-z}
\end{equation}
Variation of constants in
\eqref{eq:linear-L2-permutation-stacked-flow},
together with
\eqref{eq:linear-L2-permutation-cubic-bound}
and
\eqref{eq:linear-L2-permutation-linear-semigroup},
gives
\begin{equation}
z(\tau)
\le
e^\tau z(0)
+
K_0\int_0^\tau
e^{\tau-u}z(u)^3\,du.
\label{eq:linear-L2-permutation-small-q-bootstrap-start}
\end{equation}
By the initialization
\eqref{eq:linear-L2-permutation-fixed-XY-init},
\begin{equation}
z(0)=c_0\sqrt{\varepsilon},
\qquad
c_0^2
=
\frac2H
\left(
\|\mathbf X\|_{\mathrm F}^2
+
\|\mathbf Y\|_{\mathrm F}^2
\right).
\label{eq:linear-L2-permutation-z0}
\end{equation}

We apply a bootstrap argument.
Define the rescaled factor norm
\begin{equation}
a(\tau)
=
e^{-\tau}z(\tau).
\label{eq:linear-L2-permutation-rescaled-z}
\end{equation}
Dividing
\eqref{eq:linear-L2-permutation-small-q-bootstrap-start}
by \(e^\tau\) yields
\begin{equation}
a(\tau)
\le
z(0)
+
K_0\int_0^\tau
e^{2u}a(u)^3\,du.
\label{eq:linear-L2-permutation-r-bootstrap}
\end{equation}

Suppose temporarily that
\(a(u)\le2z(0)\) for every \(0\le u\le\tau\).
Then
\begin{equation}
\begin{aligned}
K_0\int_0^\tau e^{2u}a(u)^3\,du
&\le
8K_0z(0)^3
\int_0^\tau e^{2u}\,du
\\
&=
4K_0z(0)^3(e^{2\tau}-1)
\\
&\le
4K_0c_0^2q(\tau)z(0),
\end{aligned}
\label{eq:linear-L2-permutation-local-bootstrap-improvement}
\end{equation}
where we used
\(z(0)^2=c_0^2\varepsilon\) and
\(q(\tau)=\varepsilon e^{2\tau}\).

Choose a fixed \(q_{\mathrm{loc}}>0\),
depending only on the fixed
\(p,H,\mathbf X,\mathbf Y\), such that
\begin{equation}
4K_0c_0^2q_{\mathrm{loc}}
<
\frac12.
\label{eq:linear-L2-permutation-local-q-choice}
\end{equation}
Whenever \(q(\tau)\le q_{\mathrm{loc}}\),
the bootstrap assumption therefore implies
\begin{equation}
a(\tau)
\le
\frac32z(0)
<
2z(0).
\label{eq:linear-L2-permutation-local-bootstrap-strict}
\end{equation}

To close the bootstrap, suppose that \(a(\tau)\)
reaches \(2z(0)\) for the first time before
\(q=q_{\mathrm{loc}}\). On the entire interval
preceding this first hitting time, the bootstrap
assumption holds, so
\eqref{eq:linear-L2-permutation-local-bootstrap-strict}
applies at the hitting time itself. This contradicts
\(a(\tau)=2z(0)\).
Since \(a(0)=z(0)<2z(0)\), such a first hitting
time cannot exist.

Consequently,
\begin{equation}
z(\tau)
\le
2e^\tau z(0)
=
2c_0\sqrt{q(\tau)}
\qquad
\text{whenever }q(\tau)\le q_{\mathrm{loc}}.
\label{eq:linear-L2-permutation-small-q-factor-bound}
\end{equation}
The global existence established in Step 1
ensures that this interval is fully covered.

Substituting this bound back into the
variation-of-constants formula gives
\begin{equation}
\begin{aligned}
&
\left\|
\mathbf Z(\tau)
-
e^{\tau\mathcal A}\mathbf Z(0)
\right\|_{\mathrm F}
\\
&\qquad\le
8K_0c_0^3
\int_0^\tau
e^{\tau-u}q(u)^{3/2}\,du
\\
&\qquad\le
4K_0c_0^3q(\tau)^{3/2}.
\end{aligned}
\label{eq:linear-L2-permutation-linearization-error}
\end{equation}
Thus the exact factors differ from their
linearized evolution by \(O(q^{3/2})\)
throughout the small-\(q\) interval.

\smallskip
\noindent
\emph{Matching at an intermediate scale.}

We next compare the exact factors with the
balanced solution \(\mathbf W_*(q)\) from
\eqref{eq:linear-L2-permutation-balanced-factor}.

Recall that the symmetric and antisymmetric
linearized factors are given by
\eqref{eq:linear-L2-permutation-linearized-W}.
Combining them using
\eqref{eq:linear-L2-permutation-plus-minus}
and expressing the result in \(q\)-coordinates
gives
\begin{equation}
\mathbf U_{\mathrm{lin}}(q)
=
\sqrt{\frac qH}\,\mathbf X
+
\frac{\varepsilon}{\sqrt{qH}}\,\mathbf Y,
\label{eq:linear-L2-permutation-U-linear-q}
\end{equation}
and
\begin{equation}
\mathbf V_{\mathrm{lin}}(q)
=
\sqrt{\frac qH}\,\mathbf X
-
\frac{\varepsilon}{\sqrt{qH}}\,\mathbf Y.
\label{eq:linear-L2-permutation-V-linear-q}
\end{equation}
The first term in each expression is the growing
symmetric mode, whereas the second is the decaying
antisymmetric mode.

We choose the intermediate scale
\begin{equation}
q_0=\sqrt{\varepsilon}.
\label{eq:linear-L2-permutation-intermediate-q}
\end{equation}
For sufficiently small \(\sigma\), it satisfies
\(\varepsilon<q_0<q_{\mathrm{loc}}\).
Moreover, \(q_0\downarrow0\), while
\begin{equation}
\tau_\sigma(q_0)
=
\frac14\log\frac1{\varepsilon}
\longrightarrow\infty
\qquad
\text{as }\sigma\downarrow0.
\label{eq:linear-L2-permutation-intermediate-time}
\end{equation}
Thus the matching takes place at a time that
diverges as initialization vanishes, but while
the growing factor amplitude is still small.

Define the stacked balanced trajectory by
\begin{equation}
\mathbf Z_*(q)
=
\begin{pmatrix}
\mathbf W_*(q)\\
\mathbf W_*(q)
\end{pmatrix},
\label{eq:linear-L2-permutation-balanced-stacked-factor}
\end{equation}
and measure the factor discrepancy by
\begin{equation}
E_\sigma(q)
=
\left\|
\begin{pmatrix}
\mathbf U^{(\sigma)}(\tau_\sigma(q))\\
\mathbf V^{(\sigma)}(\tau_\sigma(q))
\end{pmatrix}
-
\mathbf Z_*(q)
\right\|_{\mathrm F}.
\label{eq:linear-L2-permutation-factor-error}
\end{equation}

At \(q=q_0\), the nonlinear approximation error
\eqref{eq:linear-L2-permutation-linearization-error}
is \(O(q_0^{3/2})=O(\varepsilon^{3/4})\).
The antisymmetric contributions in
\eqref{eq:linear-L2-permutation-U-linear-q}
and
\eqref{eq:linear-L2-permutation-V-linear-q}
have Frobenius norm
\begin{equation}
\frac{\varepsilon}{\sqrt{q_0H}}
\|\mathbf Y\|_{\mathrm F}
=
O(\varepsilon^{3/4}).
\label{eq:linear-L2-permutation-decaying-match-error}
\end{equation}
Finally, the balanced small-\(q\) expansion
\eqref{eq:linear-L2-permutation-balanced-small-q}
shows that
\(\mathbf W_*(q_0)\) differs from the common
growing term
\(\sqrt{q_0/H}\mathbf X\)
by \(O(\varepsilon^{3/4})\).

By the triangle inequality, these three estimates
give
\begin{equation}
E_\sigma(q_0)
=
O(\varepsilon^{3/4}).
\label{eq:linear-L2-permutation-balanced-match}
\end{equation}
Thus, at the intermediate scale \(q_0\), the exact
and balanced factors are already close.

\smallskip
\noindent
\emph{Propagation to a fixed positive scale.}

We now propagate this approximation from
\(q_0\downarrow0\) to the fixed scale
\(q_{\mathrm{loc}}>0\).

Let \(\mathcal F(\mathbf Z)\) denote the polynomial
vector field in
\eqref{eq:linear-L2-permutation-stacked-flow}.
Its derivative at the origin is \(\mathcal A\),
whose operator norm equals one.
Since the remaining terms are cubic, there exists
a neighborhood of the origin in which
\(\mathcal F\) is \(2\)-Lipschitz.

By decreasing \(q_{\mathrm{loc}}\), if necessary,
the exact factor bound
\eqref{eq:linear-L2-permutation-small-q-factor-bound}
and the balanced factor formula
\eqref{eq:linear-L2-permutation-balanced-factor}
ensure that both trajectories remain in this
neighborhood whenever \(q\le q_{\mathrm{loc}}\).
Consequently, their difference satisfies the
usual Gronwall estimate
\begin{equation}
E_\sigma(q)
\le
E_\sigma(q_0)
\exp\left(
2[
\tau_\sigma(q)-\tau_\sigma(q_0)
]
\right),
\qquad
q_0\le q\le q_{\mathrm{loc}}.
\end{equation}
Using the time-difference identity
\eqref{eq:linear-L2-permutation-time-difference},
this becomes
\begin{equation}
E_\sigma(q)
\le
E_\sigma(q_0)\frac{q}{q_0},
\qquad
q_0\le q\le q_{\mathrm{loc}}.
\label{eq:linear-L2-permutation-small-q-Gronwall}
\end{equation}
In particular, by
\eqref{eq:linear-L2-permutation-balanced-match},
\begin{equation}
E_\sigma(q_{\mathrm{loc}})
\le
O(\varepsilon^{3/4})
\frac{q_{\mathrm{loc}}}{\sqrt{\varepsilon}}
=
O(\varepsilon^{1/4})
\longrightarrow0.
\label{eq:linear-L2-permutation-entry-convergence}
\end{equation}
This establishes convergence at the fixed
positive scale \(q_{\mathrm{loc}}\).

\smallskip
\noindent
\emph{Propagation over arbitrary compact intervals.}

It remains to extend the approximation beyond
\(q_{\mathrm{loc}}\), where the factor dynamics
need no longer be close to their linearization.

Fix any finite
\(Q>q_{\mathrm{loc}}\).
The balanced trajectory
\(\mathbf Z_*(q)\) is continuous on
\([q_{\mathrm{loc}},Q]\), and hence bounded.
Choose
\begin{equation}
R_Q
=
1+
\sup_{q_{\mathrm{loc}}\le q\le Q}
\|\mathbf Z_*(q)\|_{\mathrm F}.
\label{eq:linear-L2-permutation-compact-radius}
\end{equation}
Since the vector field \(\mathcal F\) is
polynomial, it is Lipschitz on the closed
ball of radius \(R_Q\). Let \(L_Q<\infty\)
be a corresponding Lipschitz constant.

We first show that, for sufficiently small \(\sigma\),
the exact trajectory remains within a fixed neighborhood
of the balanced trajectory on
\([q_{\mathrm{loc}},Q]\).
This preliminary bound ensures that a common Lipschitz
constant applies throughout the interval. We will then
use Gronwall's inequality to obtain the stronger conclusion
that the distance between the trajectories converges
uniformly to zero.

By
\eqref{eq:linear-L2-permutation-entry-convergence},
we have
\(E_\sigma(q_{\mathrm{loc}})\to0\).
Suppose, for contradiction, that for arbitrarily small
\(\sigma\), the distance \(E_\sigma(q)\) reaches one
somewhere on \([q_{\mathrm{loc}},Q]\).
Let \(q_*\) be the first such point:
\begin{equation}
E_\sigma(q_*)=1.
\label{eq:linear-L2-permutation-compact-first-exit}
\end{equation}
Up to \(q_*\), both trajectories lie in the ball
of radius \(R_Q\), where the vector field is
\(L_Q\)-Lipschitz. Gronwall's inequality and
the time-difference identity
\eqref{eq:linear-L2-permutation-time-difference}
therefore give
\begin{equation}
\begin{aligned}
1
=
E_\sigma(q_*)
&\le
E_\sigma(q_{\mathrm{loc}})
\exp\left(
L_Q[
\tau_\sigma(q_*)-\tau_\sigma(q_{\mathrm{loc}})
]
\right)
\\
&=
E_\sigma(q_{\mathrm{loc}})
\left(
\frac{q_*}{q_{\mathrm{loc}}}
\right)^{L_Q/2}
\\
&\le
E_\sigma(q_{\mathrm{loc}})
\left(
\frac{Q}{q_{\mathrm{loc}}}
\right)^{L_Q/2}.
\end{aligned}
\label{eq:linear-L2-permutation-compact-Gronwall}
\end{equation}
The last expression tends to zero as
\(\sigma\downarrow0\), since \(Q\),
\(q_{\mathrm{loc}}\), and \(L_Q\) are fixed.
For sufficiently small \(\sigma\), this contradicts
the left-hand side being one.

Hence no first exit occurs, and both trajectories
remain in the region where the Lipschitz bound
is valid throughout \([q_{\mathrm{loc}},Q]\).
We may therefore apply the same Gronwall estimate
on the entire interval, obtaining
\begin{equation}
\begin{aligned}
\sup_{q_{\mathrm{loc}}\le q\le Q}E_\sigma(q)
&\le
E_\sigma(q_{\mathrm{loc}})
\left(
\frac{Q}{q_{\mathrm{loc}}}
\right)^{L_Q/2}
\\
&=
O(\varepsilon^{1/4})
\longrightarrow0,
\end{aligned}
\label{eq:linear-L2-permutation-compact-factor-convergence}
\end{equation}
where we used
\eqref{eq:linear-L2-permutation-entry-convergence}.

We now combine this estimate with the earlier
small-\(q\) comparison.
Fix arbitrary \(0<\delta<Q<\infty\);
we may enlarge \(Q\) if necessary so that
\(Q>q_{\mathrm{loc}}\).
For sufficiently small \(\sigma\), the matching
scale \(q_0=\sqrt{\varepsilon}\) satisfies
\(q_0<\delta\).

If \(\delta<q_{\mathrm{loc}}\), the small-\(q\)
Gronwall bound
\eqref{eq:linear-L2-permutation-small-q-Gronwall}
implies
\begin{equation}
\sup_{\delta\le q\le q_{\mathrm{loc}}}
E_\sigma(q)
\le
E_\sigma(q_0)
\frac{q_{\mathrm{loc}}}{q_0}
=
O(\varepsilon^{1/4}).
\label{eq:linear-L2-permutation-lower-compact-factor-convergence}
\end{equation}
The remaining interval is controlled by
\eqref{eq:linear-L2-permutation-compact-factor-convergence}.
If \(\delta\ge q_{\mathrm{loc}}\), the latter
estimate alone suffices.

In either case,
\begin{equation}
\sup_{\delta\le q\le Q}
E_\sigma(q)
\longrightarrow0
\qquad
\text{as }\sigma\downarrow0.
\label{eq:linear-L2-permutation-compact-total-factor-convergence}
\end{equation}

Finally, we transfer factor convergence to
predictor convergence. By
\eqref{eq:linear-L2-permutation-rescaled-predictor},
\eqref{eq:linear-L2-permutation-factor-error},
and
\eqref{eq:linear-L2-permutation-balanced-factor-product},
the difference between the predictors is a
difference of products of the corresponding
factors. In particular,
\begin{equation}
\begin{aligned}
&
\left\|
\widehat{\mathbf M}_\sigma(q)
-
\mathbf M_{\mathrm{bal}}(q)
\right\|_{\mathrm F}
\\
&\quad\le
\left\|
\mathbf V^{(\sigma)}(\tau_\sigma(q))
-
\mathbf W_*(q)
\right\|_{\mathrm F}
\left\|
\mathbf U^{(\sigma)}(\tau_\sigma(q))
\right\|_{\mathrm F}
\\
&\qquad+
\|\mathbf W_*(q)\|_{\mathrm F}
\left\|
\mathbf U^{(\sigma)}(\tau_\sigma(q))
-
\mathbf W_*(q)
\right\|_{\mathrm F}.
\end{aligned}
\label{eq:linear-L2-permutation-compact-product-bound}
\end{equation}
The balanced factors are bounded on
\([\delta,Q]\), and the exact factors are
uniformly bounded there for sufficiently
small \(\sigma\) by
\eqref{eq:linear-L2-permutation-compact-total-factor-convergence}.
Therefore
\begin{equation}
\sup_{\delta\le q\le Q}
\left\|
\widehat{\mathbf M}_\sigma(q)
-
\mathbf M_{\mathrm{bal}}(q)
\right\|_{\mathrm F}
\longrightarrow0.
\label{eq:linear-L2-permutation-compact-predictor-convergence}
\end{equation}

This completes the proof of convergence on
arbitrary compact \(q\)-intervals bounded away
from zero. The remaining step is to extend
this approximation uniformly to \(q\to\infty\).

\medskip
\noindent
\emph{Step 3: uniform control of the infinite-time tail.}

Step 2 establishes convergence to the balanced trajectory
uniformly on every fixed compact interval
\([\delta,Q]\), as stated in
\eqref{eq:linear-L2-permutation-compact-total-factor-convergence}.
However, the required smallness of \(\sigma\) may depend
on \(Q\). Consequently, this result alone does not exclude
a departure from the balanced trajectory at scales
\(q=q_\sigma\to\infty\) as \(\sigma\downarrow0\).

To establish convergence uniformly over the entire
infinite-time tail, we exploit the fact that the balanced
trajectory approaches a global minimizer of the loss.
We will prove local stability near this limiting equilibrium:
once the exact trajectory enters a sufficiently small
neighborhood, it cannot subsequently depart from it.

We first identify the limiting equilibrium.
Let
\begin{equation}
r=\min\{H,p\},
\label{eq:linear-L2-permutation-rank-r}
\end{equation}
and take a compact singular-value decomposition
\begin{equation}
\mathbf X
=
\mathbf L\boldsymbol{\Sigma}\mathbf R^\top,
\label{eq:linear-L2-permutation-X-SVD}
\end{equation}
where
\(\mathbf L\in\mathbb R^{H\times r}\) and
\(\mathbf R\in\mathbb R^{p\times r}\)
have orthonormal columns, and
\(\boldsymbol{\Sigma}\in\mathbb R^{r\times r}\)
is positive diagonal.
The positivity of its diagonal entries follows
from the full-rank assumption
\eqref{eq:linear-L2-permutation-X-full-rank}.

Substituting this decomposition into the balanced
factor formula
\eqref{eq:linear-L2-permutation-balanced-factor}
gives
\begin{equation}
\mathbf W_*(q)
=
\mathbf L
\operatorname{diag}
\left(
\frac{\sqrt{q/H}\,\Sigma_{kk}}
{\sqrt{1+(q/H)\Sigma_{kk}^2}}
\right)_{k=1}^r
\mathbf R^\top.
\label{eq:linear-L2-permutation-balanced-SVD}
\end{equation}
Since every diagonal coefficient tends to one,
\begin{equation}
\mathbf W_*(q)
\longrightarrow
\mathbf W_\infty
:=
\mathbf L\mathbf R^\top
\qquad
\text{as }q\to\infty.
\label{eq:linear-L2-permutation-balanced-factor-limit}
\end{equation}
Consequently, by
\eqref{eq:linear-L2-permutation-balanced-factor-product},
\begin{equation}
\mathbf M_{\mathrm{bal}}(q)
\longrightarrow
\mathbf P_X
:=
\mathbf W_\infty^\top\mathbf W_\infty
=
\mathbf R\mathbf R^\top.
\label{eq:linear-L2-permutation-balanced-predictor-limit}
\end{equation}
Thus \(\mathbf P_X\) is the orthogonal projector
onto the row space of \(\mathbf X\).
When \(H\ge p\), this projector equals
\(\mathbf I_p\); when \(H<p\), it has rank \(H\).
In both cases, all \(r\) singular values of
\(\mathbf W_\infty\) are equal to one.

\smallskip
\noindent
\emph{Local stability of the limiting equilibrium.}

We now establish the following property:
for every \(\eta>0\), there exists \(\rho>0\)
such that, if at some time \(\tau_0\),
\begin{equation}
\|\mathbf U(\tau_0)-\mathbf W_\infty\|_{\mathrm F}
+
\|\mathbf V(\tau_0)-\mathbf W_\infty\|_{\mathrm F}
<
\rho,
\label{eq:linear-L2-permutation-tail-entry}
\end{equation}
then
\begin{equation}
\|\mathbf V(\tau)^\top\mathbf U(\tau)
-\mathbf P_X\|_{\mathrm F}
<
\eta
\qquad
\text{for every }\tau\ge\tau_0.
\label{eq:linear-L2-permutation-tail-stability}
\end{equation}

Our approach is to establish a local gradient inequality
near \((\mathbf W_\infty,\mathbf W_\infty)\).
Although the gradient determines the instantaneous
speed of the factors, the loss dissipation rate
depends on its squared norm. A lower bound on
the gradient relative to the remaining excess loss
therefore yields an upper bound on the total
distance traveled by the factors.
We will use this bound to prevent the trajectory
from leaving a sufficiently small neighborhood.

Recall the identity-target loss
\eqref{eq:linear-L2-permutation-identity-loss}.
Since the predictor
\(\mathbf M=\mathbf V^\top\mathbf U\)
has rank at most \(r=\min\{H,p\}\),
the minimum achievable loss is
\begin{equation}
\mathcal L_{\min}
=
\frac{p-r}{2p}.
\label{eq:linear-L2-permutation-minimum-loss}
\end{equation}
Indeed, the optimal rank-\(r\) approximation
to \(\mathbf I_p\) has squared Frobenius error
\(p-r\).
By
\eqref{eq:linear-L2-permutation-balanced-predictor-limit},
the limiting balanced factors attain this minimum:
\begin{equation}
\mathcal L(
\mathbf W_\infty,
\mathbf W_\infty
)
=
\mathcal L_{\min}.
\label{eq:linear-L2-permutation-limiting-minimum}
\end{equation}

We next establish the local gradient inequality.
Here \(\nabla\mathcal L\) denotes the full gradient
with respect to both factor matrices, equipped
with the joint Frobenius norm
\begin{equation}
\|\nabla\mathcal L\|^2
=
\|\nabla_{\mathbf U}\mathcal L\|_{\mathrm F}^2
+
\|\nabla_{\mathbf V}\mathcal L\|_{\mathrm F}^2.
\label{eq:linear-L2-permutation-full-gradient-norm}
\end{equation}
The partial gradients are
\begin{equation}
\nabla_{\mathbf U}\mathcal L
=
\frac1p\mathbf V(\mathbf M-\mathbf I_p),
\qquad
\nabla_{\mathbf V}\mathcal L
=
\frac1p\mathbf U(\mathbf M^\top-\mathbf I_p).
\label{eq:linear-L2-permutation-loss-partial-gradients}
\end{equation}

We will show that, in a sufficiently small
neighborhood of
\((\mathbf W_\infty,\mathbf W_\infty)\),
\begin{equation}
\|\nabla\mathcal L\|^2
\ge
\frac{1}{2p}
\left(
\mathcal L-\mathcal L_{\min}
\right).
\label{eq:linear-L2-permutation-local-PL}
\end{equation}
It suffices to establish this lower bound for
\(\|\nabla_{\mathbf U}\mathcal L\|_{\mathrm F}^2\),
since the full squared gradient norm contains
this term and the additional nonnegative contribution
\(\|\nabla_{\mathbf V}\mathcal L\|_{\mathrm F}^2\).

We treat the overcomplete and undercomplete
regimes separately.

\smallskip
\noindent
\emph{Case \(H\ge p\).}

In this case, \(\mathbf W_\infty\) has full
column rank, with all singular values equal to one.
Consequently, by continuity of singular values,
in a sufficiently small neighborhood of
\(\mathbf W_\infty\), the matrix \(\mathbf V\)
also has full column rank and satisfies
\begin{equation}
s_{\min}(\mathbf V)
\ge
\frac12,
\label{eq:linear-L2-permutation-local-singular-over}
\end{equation}
where \(s_{\min}\) denotes the smallest singular value.

Using
\eqref{eq:linear-L2-permutation-loss-partial-gradients},
we obtain
\begin{equation}
\begin{aligned}
\|\nabla_{\mathbf U}\mathcal L\|_{\mathrm F}^2
&=
\frac1{p^2}
\|\mathbf V(\mathbf M-\mathbf I_p)\|_{\mathrm F}^2
\\
&\ge
\frac{s_{\min}(\mathbf V)^2}{p^2}
\|\mathbf M-\mathbf I_p\|_{\mathrm F}^2
\\
&=
\frac{2s_{\min}(\mathbf V)^2}{p}
(\mathcal L-\mathcal L_{\min})
\\
&\ge
\frac1{2p}
(\mathcal L-\mathcal L_{\min}),
\end{aligned}
\label{eq:linear-L2-permutation-PL-overcomplete}
\end{equation}
where \(\mathcal L_{\min}=0\) for \(H\ge p\).
This proves
\eqref{eq:linear-L2-permutation-local-PL}
in the overcomplete regime.

\smallskip
\noindent
\emph{Case \(H<p\).}

Here \(\mathbf W_\infty\) has full row rank,
with all \(H\) singular values equal to one.
Thus, in a sufficiently small neighborhood,
\(\mathbf V\) remains full row rank and satisfies
\begin{equation}
s_{\min}(\mathbf V)
\ge
\frac12.
\label{eq:linear-L2-permutation-local-singular-under}
\end{equation}

Define the orthogonal projector onto the row
space of \(\mathbf V\):
\begin{equation}
\mathbf P_V
=
\mathbf V^\top
(\mathbf V\mathbf V^\top)^{-1}
\mathbf V.
\label{eq:linear-L2-permutation-PV}
\end{equation}
Since \(\mathbf M=\mathbf V^\top\mathbf U\),
we have
\(\mathbf P_V\mathbf M=\mathbf M\).
Therefore
\(\mathbf M-\mathbf P_V\)
has all its columns in the row space of
\(\mathbf V\), whereas
\(\mathbf I_p-\mathbf P_V\)
has all its columns in the orthogonal
complement of that space.
These two matrices are thus orthogonal in
the Frobenius inner product.

Using
\begin{equation}
\mathbf M-\mathbf I_p
=
(\mathbf M-\mathbf P_V)
-
(\mathbf I_p-\mathbf P_V),
\end{equation}
we obtain
\begin{equation}
\begin{aligned}
\|\mathbf M-\mathbf I_p\|_{\mathrm F}^2
&=
\|\mathbf M-\mathbf P_V\|_{\mathrm F}^2
+
\|\mathbf I_p-\mathbf P_V\|_{\mathrm F}^2
\\
&=
\|\mathbf M-\mathbf P_V\|_{\mathrm F}^2
+
p-H.
\end{aligned}
\label{eq:linear-L2-permutation-undercomplete-projection-decomposition}
\end{equation}
The last equality uses the fact that
\(\mathbf I_p-\mathbf P_V\)
is an orthogonal projector of rank \(p-H\).

Since
\(\mathcal L_{\min}=(p-H)/(2p)\),
the decomposition gives
\begin{equation}
\mathcal L-\mathcal L_{\min}
=
\frac1{2p}
\|\mathbf M-\mathbf P_V\|_{\mathrm F}^2.
\label{eq:linear-L2-permutation-undercomplete-energy-gap}
\end{equation}

Furthermore,
\(\mathbf V(\mathbf I_p-\mathbf P_V)=0\),
so
\eqref{eq:linear-L2-permutation-loss-partial-gradients}
yields
\begin{equation}
\nabla_{\mathbf U}\mathcal L
=
\frac1p
\mathbf V(\mathbf M-\mathbf P_V).
\label{eq:linear-L2-permutation-undercomplete-gradient}
\end{equation}
Every column of \(\mathbf M-\mathbf P_V\)
belongs to the row space of \(\mathbf V\).
On this subspace, multiplication by
\(\mathbf V\) has smallest singular value
\(s_{\min}(\mathbf V)\).
Therefore,
\begin{equation}
\begin{aligned}
\|\nabla_{\mathbf U}\mathcal L\|_{\mathrm F}^2
&\ge
\frac{s_{\min}(\mathbf V)^2}{p^2}
\|\mathbf M-\mathbf P_V\|_{\mathrm F}^2
\\
&=
\frac{2s_{\min}(\mathbf V)^2}{p}
(\mathcal L-\mathcal L_{\min})
\\
&\ge
\frac1{2p}
(\mathcal L-\mathcal L_{\min}).
\end{aligned}
\label{eq:linear-L2-permutation-PL-undercomplete}
\end{equation}
This proves
\eqref{eq:linear-L2-permutation-local-PL}
also when \(H<p\).

Thus, in both dimension regimes, the full
gradient satisfies the local inequality
\eqref{eq:linear-L2-permutation-local-PL}
in a sufficiently small neighborhood of
\((\mathbf W_\infty,\mathbf W_\infty)\).

\smallskip
\noindent
\emph{Stability from the gradient inequality.}

We now use
\eqref{eq:linear-L2-permutation-local-PL}
to establish the stability property
\eqref{eq:linear-L2-permutation-tail-stability}.
The key observation is that the gradient inequality
controls the total distance traveled by the factors,
not merely the decrease of the loss.

Recall the stacked factor variable \(\mathbf Z\)
from
\eqref{eq:linear-L2-permutation-stacked-variable},
and define
\begin{equation}
\mathbf Z_\infty
=
\begin{pmatrix}
\mathbf W_\infty\\
\mathbf W_\infty
\end{pmatrix}.
\label{eq:linear-L2-permutation-tail-factor-space}
\end{equation}
We use the joint Frobenius norm
\begin{equation}
\|\mathbf Z\|_{\mathrm F}^2
=
\|\mathbf U\|_{\mathrm F}^2
+
\|\mathbf V\|_{\mathrm F}^2.
\label{eq:linear-L2-permutation-tail-factor-norm}
\end{equation}
Define the nonnegative excess loss along
the trajectory by
\begin{equation}
e(\tau)
=
\mathcal L(\mathbf Z(\tau))
-
\mathcal L_{\min}.
\label{eq:linear-L2-permutation-excess-loss}
\end{equation}

By the rescaled gradient-flow identity
\eqref{eq:linear-L2-permutation-loss-gradient-flow},
\begin{equation}
\dot{\mathbf Z}
=
-p\nabla\mathcal L(\mathbf Z).
\label{eq:linear-L2-permutation-tail-gradient-flow}
\end{equation}
Consequently,
\begin{equation}
\dot e(\tau)
=
-p\|\nabla\mathcal L(\mathbf Z(\tau))\|^2
\le0.
\label{eq:linear-L2-permutation-excess-loss-dissipation}
\end{equation}
On the neighborhood where
\eqref{eq:linear-L2-permutation-local-PL}
holds, we also have
\begin{equation}
\|\nabla\mathcal L(\mathbf Z(\tau))\|
\ge
\sqrt{\frac{e(\tau)}{2p}}.
\label{eq:linear-L2-permutation-gradient-lower-bound}
\end{equation}

Combining these estimates, whenever \(e(\tau)>0\),
\begin{equation}
\begin{aligned}
-\frac{d}{d\tau}\sqrt{e(\tau)}
&=
\frac{
p\|\nabla\mathcal L(\mathbf Z(\tau))\|^2
}{
2\sqrt{e(\tau)}
}
\\
&\ge
\frac{
p\|\nabla\mathcal L(\mathbf Z(\tau))\|
}{
2\sqrt{2p}
}
=
\frac{
\|\dot{\mathbf Z}(\tau)\|_{\mathrm F}
}{
2\sqrt{2p}
}.
\end{aligned}
\label{eq:linear-L2-permutation-sqrt-energy-decay}
\end{equation}
Integrating over \([\tau_0,\tau]\) gives
\begin{equation}
\begin{aligned}
\int_{\tau_0}^{\tau}
\|\dot{\mathbf Z}(u)\|_{\mathrm F}\,du
&\le
2\sqrt{2p}
\left(
\sqrt{e(\tau_0)}-\sqrt{e(\tau)}
\right)
\\
&\le
2\sqrt{2p}\sqrt{e(\tau_0)}.
\end{aligned}
\label{eq:linear-L2-permutation-local-path-length}
\end{equation}
If the excess loss reaches zero, the trajectory
reaches a global minimizer and becomes stationary,
so the bound remains valid by continuity.

Crucially, the right-hand side is independent
of the final time \(\tau\). Thus, as long as
the local gradient inequality applies, the total
subsequent displacement of the factors is bounded
solely in terms of their excess loss at \(\tau_0\).

We now use this bound to prove local stability.
Fix an arbitrary tolerance \(\eta>0\).
Since the gradient inequality holds near
\(\mathbf Z_\infty\), and since the predictor map
\((\mathbf U,\mathbf V)\mapsto\mathbf V^\top\mathbf U\)
is continuous with value \(\mathbf P_X\) at
\(\mathbf Z_\infty\), we may choose \(R>0\)
such that throughout the ball
\begin{equation}
\|\mathbf Z-\mathbf Z_\infty\|_{\mathrm F}<R,
\label{eq:linear-L2-permutation-local-ball}
\end{equation}
both
\eqref{eq:linear-L2-permutation-local-PL}
and
\begin{equation}
\|\mathbf V^\top\mathbf U-\mathbf P_X\|_{\mathrm F}
<
\eta
\label{eq:linear-L2-permutation-local-product-control}
\end{equation}
hold.

Next, recall from
\eqref{eq:linear-L2-permutation-limiting-minimum}
that the excess loss vanishes at
\(\mathbf Z_\infty\).
By continuity of the loss, we can therefore
choose \(0<\rho<R/2\) sufficiently small that
\begin{equation}
\|\mathbf Z-\mathbf Z_\infty\|_{\mathrm F}<\rho
\quad\Longrightarrow\quad
2\sqrt{2p}
\sqrt{
\mathcal L(\mathbf Z)-\mathcal L_{\min}
}
<
\frac R2.
\label{eq:linear-L2-permutation-entry-energy-small}
\end{equation}

Suppose now that the exact trajectory satisfies
\eqref{eq:linear-L2-permutation-tail-entry}
at some time \(\tau_0\).
Since the joint Frobenius distance is bounded by
the sum of the two factor distances, this places
\(\mathbf Z(\tau_0)\) inside the smaller ball
of radius \(\rho\).

We claim that the trajectory remains inside the
larger ball of radius \(R\) for all subsequent times.
Indeed, suppose for contradiction that there exists
a first exit time \(\tau_*>\tau_0\), at which
\(\|\mathbf Z(\tau_*)-\mathbf Z_\infty\|_{\mathrm F}=R\).
Before this time, the trajectory stays inside the
larger ball, so the local gradient inequality
holds and the path-length estimate
\eqref{eq:linear-L2-permutation-local-path-length}
applies. By continuity, it also applies at
the first exit time. Therefore,
\begin{equation}
\begin{aligned}
\|\mathbf Z(\tau_*)-\mathbf Z_\infty\|_{\mathrm F}
&\le
\|\mathbf Z(\tau_0)-\mathbf Z_\infty\|_{\mathrm F}
+
\int_{\tau_0}^{\tau_*}
\|\dot{\mathbf Z}(u)\|_{\mathrm F}\,du
\\
&<
\rho+\frac R2
<
R.
\end{aligned}
\label{eq:linear-L2-permutation-no-exit}
\end{equation}
This contradicts the definition of \(\tau_*\).
Hence no first exit can occur.

Since global existence was established in Step 1,
the trajectory remains inside the larger ball
for every \(\tau\ge\tau_0\).
The predictor estimate
\eqref{eq:linear-L2-permutation-local-product-control}
then holds at all subsequent times, proving
\eqref{eq:linear-L2-permutation-tail-stability}.

\smallskip
\noindent
\emph{Completion of the uniform tail comparison.}

We now combine the compact-interval convergence from
Step 2 with the local stability property established
above. The key point is that local stability applies
only near the limiting equilibrium
\((\mathbf W_\infty,\mathbf W_\infty)\).
For an arbitrary fixed \(\delta>0\), the balanced
trajectory at \(q=\delta\) need not be close to this
equilibrium, so we cannot apply local stability
starting directly from \(q=\delta\).

Instead, we will choose a sufficiently large but
fixed \(Q\), where the balanced trajectory is close
to its limiting equilibrium. The compact-interval
convergence from Step 2 will then imply that the
exact trajectory is also close to this equilibrium
at \(q=Q\). Local stability will control all later
times, while Step 2 will control the remaining
interval \([\delta,Q]\).

Fix \(\delta>0\) and an arbitrary tolerance
\(\eta>0\).
Let \(\rho>0\) be the stability radius in
\eqref{eq:linear-L2-permutation-tail-entry}
corresponding to the predictor tolerance
\(\eta/3\). Thus, once the sum of the two factor
distances from \(\mathbf W_\infty\) is smaller
than \(\rho\), the predictor remains within
\(\eta/3\) of \(\mathbf P_X\) for all subsequent
times, by
\eqref{eq:linear-L2-permutation-tail-stability}.

We first choose the point at which this stability
argument will begin.
By the convergence of the balanced factors
\eqref{eq:linear-L2-permutation-balanced-factor-limit},
we can choose a finite
\(Q>\max\{\delta,q_{\mathrm{loc}}\}\)
sufficiently large that
\begin{equation}
2\|\mathbf W_*(Q)-\mathbf W_\infty\|_{\mathrm F}
<
\frac{\rho}{2}.
\label{eq:linear-L2-permutation-large-Q-factor-choice}
\end{equation}
Increasing \(Q\) further if necessary, the
balanced predictor limit
\eqref{eq:linear-L2-permutation-balanced-predictor-limit}
also ensures
\begin{equation}
\sup_{q\ge Q}
\|\mathbf M_{\mathrm{bal}}(q)-\mathbf P_X\|_{\mathrm F}
<
\frac{\eta}{3}.
\label{eq:linear-L2-permutation-large-Q-predictor-choice}
\end{equation}
Importantly, \(Q\) depends on the fixed matrices
\(\mathbf X,\mathbf Y\) and the chosen tolerances,
but not on \(\sigma\).

We next show that the exact trajectory enters
the stability neighborhood at this fixed \(Q\).
Recall the factor discrepancy \(E_\sigma(q)\)
defined in
\eqref{eq:linear-L2-permutation-factor-error}.
The compact-interval convergence
\eqref{eq:linear-L2-permutation-compact-total-factor-convergence}
implies, in particular, that
\begin{equation}
E_\sigma(Q)
\longrightarrow0
\qquad
\text{as }\sigma\downarrow0.
\label{eq:linear-L2-permutation-EQ-zero}
\end{equation}
Since \(Q\) and \(\rho\) are fixed independently
of \(\sigma\), we may take \(\sigma\) sufficiently
small that
\begin{equation}
E_\sigma(Q)
<
\frac{\rho}{2\sqrt2}.
\label{eq:linear-L2-permutation-EQ-small}
\end{equation}

At time \(\tau_\sigma(Q)\), the exact factors
are therefore close to \(\mathbf W_\infty\).
Indeed, inserting the balanced factors
\(\mathbf W_*(Q)\) and applying the triangle
inequality gives
\begin{equation}
\begin{aligned}
&
\|\mathbf U^{(\sigma)}(\tau_\sigma(Q))
-\mathbf W_\infty\|_{\mathrm F}
+
\|\mathbf V^{(\sigma)}(\tau_\sigma(Q))
-\mathbf W_\infty\|_{\mathrm F}
\\
&\quad\le
\|\mathbf U^{(\sigma)}(\tau_\sigma(Q))
-\mathbf W_*(Q)\|_{\mathrm F}
\\
&\qquad+
\|\mathbf V^{(\sigma)}(\tau_\sigma(Q))
-\mathbf W_*(Q)\|_{\mathrm F}
+
2\|\mathbf W_*(Q)-\mathbf W_\infty\|_{\mathrm F}
\\
&\quad\le
\sqrt2 E_\sigma(Q)
+
2\|\mathbf W_*(Q)-\mathbf W_\infty\|_{\mathrm F}
\\
&\quad<
\frac{\rho}{2}
+
\frac{\rho}{2}
=
\rho.
\end{aligned}
\label{eq:linear-L2-permutation-tail-entry-at-Q}
\end{equation}
Here the second inequality is Cauchy--Schwarz,
relating the sum of the two factor errors
to their joint Frobenius norm.

The two contributions in
\eqref{eq:linear-L2-permutation-tail-entry-at-Q}
are controlled differently: the balanced factor
distance is made small by choosing \(Q\) large,
whereas the exact-to-balanced discrepancy
\(E_\sigma(Q)\) is made small by subsequently
choosing \(\sigma\) small.
This establishes the entry condition
\eqref{eq:linear-L2-permutation-tail-entry}
at time \(\tau_\sigma(Q)\).

We may now apply the local stability property
\eqref{eq:linear-L2-permutation-tail-stability}.
It guarantees that the exact predictor remains
close to \(\mathbf P_X\) for every subsequent
time, including arbitrarily late times:
\begin{equation}
\sup_{q\ge Q}
\left\|
\widehat{\mathbf M}_\sigma(q)
-
\mathbf P_X
\right\|_{\mathrm F}
\le
\frac{\eta}{3}.
\label{eq:linear-L2-permutation-tail-close-PX}
\end{equation}
The balanced predictor is also close to
\(\mathbf P_X\) throughout this tail, by
\eqref{eq:linear-L2-permutation-large-Q-predictor-choice}.
Hence the triangle inequality yields
\begin{equation}
\begin{aligned}
&
\sup_{q\ge Q}
\left\|
\widehat{\mathbf M}_\sigma(q)
-
\mathbf M_{\mathrm{bal}}(q)
\right\|_{\mathrm F}
\\
&\qquad\le
\sup_{q\ge Q}
\left\|
\widehat{\mathbf M}_\sigma(q)
-
\mathbf P_X
\right\|_{\mathrm F}
+
\sup_{q\ge Q}
\left\|
\mathbf M_{\mathrm{bal}}(q)
-
\mathbf P_X
\right\|_{\mathrm F}
\\
&\qquad\le
\frac{2\eta}{3}
<
\eta.
\end{aligned}
\label{eq:linear-L2-permutation-tail-balanced-convergence}
\end{equation}
Unlike the compact-interval estimates in Step 2,
this bound holds uniformly over the entire
infinite interval \([Q,\infty)\).
It is precisely here that local stability
provides additional control.

It remains to treat the interval \([\delta,Q]\),
which is not covered by the stability argument.
Indeed, on this interval the balanced trajectory
need not be close to \(\mathbf P_X\), so the
preceding comparison through the limiting
equilibrium would not suffice.

However, \(Q\) is fixed and finite.
Therefore the compact-interval predictor
convergence established in Step 2,
\eqref{eq:linear-L2-permutation-compact-predictor-convergence},
applies:
\begin{equation}
\sup_{\delta\le q\le Q}
\left\|
\widehat{\mathbf M}_\sigma(q)
-
\mathbf M_{\mathrm{bal}}(q)
\right\|_{\mathrm F}
\longrightarrow0
\qquad
\text{as }\sigma\downarrow0.
\label{eq:linear-L2-permutation-compact-final}
\end{equation}
Thus, by decreasing \(\sigma\) further if necessary,
we can make this supremum smaller than \(\eta\),
while preserving all the preceding estimates
on \([Q,\infty)\).

Combining the two intervals, we conclude that
for every \(\delta>0\) and \(\eta>0\), there exists
\(\sigma_0>0\) such that for all
\(0<\sigma<\sigma_0\),
\begin{equation}
\begin{aligned}
&
\sup_{q\ge\delta}
\left\|
\widehat{\mathbf M}_\sigma(q)
-
\mathbf M_{\mathrm{bal}}(q)
\right\|_{\mathrm F}
\\
&\quad=
\max\left\{
\sup_{\delta\le q\le Q}
\left\|
\widehat{\mathbf M}_\sigma(q)
-
\mathbf M_{\mathrm{bal}}(q)
\right\|_{\mathrm F},
\right.
\left.
\sup_{q\ge Q}
\left\|
\widehat{\mathbf M}_\sigma(q)
-
\mathbf M_{\mathrm{bal}}(q)
\right\|_{\mathrm F}
\right\}
<
\eta.
\end{aligned}
\end{equation}
Since \(\eta>0\) was arbitrary, this proves
the uniform convergence
\eqref{eq:linear-L2-permutation-global-balanced-reduction}.

Finally, the elementary inequality
\begin{equation}
\|\mathbf A\|_{\max}
\le
\|\mathbf A\|_{\mathrm F}
\label{eq:linear-L2-permutation-max-Frobenius}
\end{equation}
implies
\eqref{eq:linear-L2-permutation-global-balanced-reduction-max},
completing the proof.
\end{proof}

The global balanced reduction
\eqref{eq:linear-L2-permutation-global-balanced-reduction-max}
shows that, for fixed \(p,H\), the exact predictor
converges uniformly to \(\mathbf M_{\mathrm{bal}}(q)\)
over \(q\ge\delta\) as \(\sigma\downarrow0\).
To transfer perfect classification from this reference
trajectory to the exact dynamics, we need a classification
margin that remains positive uniformly in \(q\) and
does not vanish as \(p,H\to\infty\).
Together with the earlier small-\(q\) monotonicity
and organization results, such a margin will establish
permanent perfect accuracy.

The required margin cannot be deduced from closeness
of \(\mathbf C\) to the identity: when \(H/p\) approaches
a finite positive constant, its eigenvalues need not
be close to one, and \(\mathbf C\) may even be
rank-deficient. Instead, we exploit the orthogonal
invariance of the Gaussian ensemble. Conditional
on the eigenvalues of \(\mathbf C\), its eigenvectors
have an isotropic orientation. This implies that,
for each fixed \(q\), the diagonal entries of
\(\mathbf M_{\mathrm{bal}}(q)\) concentrate around
their common average, while the off-diagonal
entries are small.

To make these estimates uniform over the entire
unbounded range \(q\ge\delta\), we use the monotonicity
of \(\mathbf M_{\mathrm{bal}}(q)\) in the
positive-semidefinite order, together with a finite
grid constructed from its normalized trace.
Finally, we show that the average diagonal entry
is bounded away from zero uniformly for \(q\ge\delta\).
These properties yield the following positive
classification margin.

\begin{lemma}[Uniform accuracy margin of the balanced trajectory]
\label{lem:linear-L2-permutation-balanced-margin}
Let \(p,H\to\infty\) with
\begin{equation}
\frac Hp
\longrightarrow
\alpha\in(0,\infty).
\label{eq:linear-L2-permutation-balanced-margin-aspect-ratio}
\end{equation}
Let \(\mathbf X\in\mathbb R^{H\times p}\)
have independent standard Gaussian entries,
and define
\begin{equation}
\mathbf C
=
\frac1H\mathbf X^\top\mathbf X,
\qquad
\mathbf M_{\mathrm{bal}}(q)
=
q\mathbf C(\mathbf I_p+q\mathbf C)^{-1}.
\label{eq:linear-L2-permutation-balanced-margin-definitions}
\end{equation}
For every fixed \(\delta>0\), there exists
a constant \(c_{\alpha,\delta}>0\),
depending only on \(\alpha\) and \(\delta\),
such that, with probability tending to one
as \(p,H\to\infty\),
\begin{equation}
\inf_{q\ge\delta}
\min_{i\in[p]}
\min_{c\in[p]\setminus\{i\}}
\left[
(\mathbf M_{\mathrm{bal}}(q))_{ii}
-
(\mathbf M_{\mathrm{bal}}(q))_{ci}
\right]
\ge
c_{\alpha,\delta}.
\label{eq:linear-L2-permutation-balanced-uniform-margin}
\end{equation}
In particular, with probability tending to one,
\(\mathbf M_{\mathrm{bal}}(q)\) achieves perfect
classification simultaneously for all \(p\)
canonical-basis inputs and every \(q\ge\delta\).
\end{lemma}

\begin{proof}
We prove the result in three steps.
First, we show that the normalized trace of
\(\mathbf M_{\mathrm{bal}}(q)\), which is its
average diagonal entry, is bounded away from zero
uniformly for \(q\ge\delta\).
Second, we use orthogonal invariance of the
Gaussian Gram matrix to show that, uniformly in \(q\),
all diagonal entries are close to this average
and all off-diagonal entries are small.
Finally, these two properties give a positive
classification margin.

The main technical point is uniformity over the
unbounded interval \(q\in[0,\infty)\).
For each fixed \(q\), entrywise concentration follows
from concentration of quadratic forms of Haar vectors.
To extend this estimate to all \(q\) simultaneously,
we exploit the monotonicity of
\(\mathbf M_{\mathrm{bal}}(q)\)
in the positive-semidefinite order.
This allows us to construct a finite grid on
\([0,\infty]\), with successive increments of the
normalized trace uniformly small.
Concentration on this grid then controls the
entire trajectory.

\smallskip
\noindent
\emph{Step 1: a positive lower bound on the average diagonal.}

By orthogonal invariance of the Gaussian Wishart law,
we may write
\begin{equation}
\mathbf C
=
\mathbf O\boldsymbol{\Lambda}\mathbf O^\top,
\qquad
\boldsymbol{\Lambda}
=
\operatorname{diag}(\lambda_1,\ldots,\lambda_p),
\qquad
0\le\lambda_1\le\cdots\le\lambda_p,
\label{eq:linear-L2-permutation-C-eigendecomposition}
\end{equation}
where, conditional on the ordered eigenvalues,
\(\mathbf O\) is Haar distributed on the orthogonal
group and independent of
\(\boldsymbol{\Lambda}\).
If \(H<p\), an independent Haar basis may be chosen
inside the nullspace corresponding to the repeated
zero eigenvalue.

For \(q\ge0\), define
\begin{equation}
d_j(q)
=
\frac{q\lambda_j}{1+q\lambda_j},
\qquad
\overline d(q)
=
\frac1p\sum_{j=1}^p d_j(q)
=
\frac1p\operatorname{Tr}
\mathbf M_{\mathrm{bal}}(q).
\label{eq:linear-L2-permutation-djq}
\end{equation}
Then
\begin{equation}
\mathbf M_{\mathrm{bal}}(q)
=
\mathbf O
\operatorname{diag}
\bigl(d_1(q),\ldots,d_p(q)\bigr)
\mathbf O^\top.
\label{eq:linear-L2-permutation-balanced-spectral}
\end{equation}
Since \(0\le d_j(q)\le1\) and each \(d_j(q)\)
is nondecreasing in \(q\), we have
\begin{equation}
\mathbf 0
\preceq
\mathbf M_{\mathrm{bal}}(q_1)
\preceq
\mathbf M_{\mathrm{bal}}(q_2)
\preceq
\mathbf I_p,
\qquad
0\le q_1\le q_2.
\label{eq:linear-L2-permutation-balanced-Loewner}
\end{equation}
In particular, \(\overline d(q)\) is continuous,
nondecreasing, and takes values in \([0,1]\).

By Lemma~\ref{lem:linear-L2-permutation-Gaussian-geometry},
there exists \(L_\alpha<\infty\), depending only
on \(\alpha\), such that with probability
tending to one,
\begin{equation}
\|\mathbf C\|_{\mathrm{op}}
\le L_\alpha,
\qquad
\frac1p\operatorname{Tr}\mathbf C
\ge\frac12.
\label{eq:linear-L2-permutation-C-basic-bounds}
\end{equation}
Indeed,
\(\|\mathbf C\|_{\mathrm{op}}
=\|\mathbf X\|_{\mathrm{op}}^2/H\),
and the trace estimate follows from concentration
of the normalized diagonal Gram entries.

On this event, for every \(q\ge\delta\),
\begin{equation}
\begin{aligned}
\overline d(q)
&\ge
\overline d(\delta)
=
\frac1p\sum_{j=1}^p
\frac{\delta\lambda_j}{1+\delta\lambda_j}
\\
&\ge
\frac{\delta}{1+\delta L_\alpha}
\frac1p\sum_{j=1}^p\lambda_j
\\
&\ge
\frac{\delta}{2(1+\delta L_\alpha)}
=:m_{\alpha,\delta}>0.
\end{aligned}
\label{eq:linear-L2-permutation-average-diagonal-lower}
\end{equation}
Thus the average diagonal entry is uniformly
positive throughout \(q\ge\delta\).

\smallskip
\noindent
\emph{Step 2: uniform concentration of individual entries.}

We next establish that the diagonal entries of
\(\mathbf M_{\mathrm{bal}}(q)\) are uniformly close
to their common average, while all off-diagonal
entries are uniformly small:
\begin{equation}
\sup_{q\ge0}
\left[
\max_i
\left|
(\mathbf M_{\mathrm{bal}}(q))_{ii}
-\overline d(q)
\right|
+
\max_{i\neq c}
\left|
(\mathbf M_{\mathrm{bal}}(q))_{ci}
\right|
\right]
=
o_{\mathbb P}(1).
\label{eq:linear-L2-permutation-uniform-entry-delocalization}
\end{equation}

The proof proceeds by first controlling quadratic forms
of \(\mathbf M_{\mathrm{bal}}(q)\) at a fixed \(q\),
using concentration on the sphere. The difficulty is
then to obtain a single high-probability event on
which these bounds hold simultaneously for every
\(q\ge0\).

To overcome this difficulty, we construct a finite
grid in \(q\), but measure its spacing through the
normalized trace \(\overline d(q)\), rather than
through \(q\) itself. This distinction is important:
although \(q\) ranges over an unbounded interval,
\(\overline d(q)\) is nondecreasing and bounded
between zero and one. We can therefore choose only
\(O(1/\omega_p)\) grid points so that the trace
increases by at most \(\omega_p\) between consecutive
points, including the final interval extending to
\(q=\infty\).

We will first establish concentration at all grid
points by a union bound. The positive-semidefinite
monotonicity
\eqref{eq:linear-L2-permutation-balanced-Loewner}
will then allow us to control every intermediate
\(q\) by the two neighboring grid points.
Finally, we choose particular unit vectors whose
quadratic forms determine the individual matrix
entries.

\smallskip
\noindent
\emph{Concentration of quadratic forms.}

We use concentration of Lipschitz functions on
the sphere; see
\citet[Theorem~5.1.3 and Exercise~5.5]
{vershynin2026high}.
Specifically, if \(\mathbf z\) is uniformly
distributed on \(\mathbb S^{p-1}\) and
\(\mathbf D\) is a deterministic diagonal
matrix with entries in \([0,1]\), then there
exists a universal constant \(c>0\) such that
for every \(0<t<1\),
\begin{equation}
\mathbb P\left(
\left|
\mathbf z^\top\mathbf D\mathbf z
-
\frac1p\operatorname{Tr}\mathbf D
\right|>t
\right)
\le
4e^{-cpt^2}.
\label{eq:linear-L2-permutation-Haar-quadratic-tail}
\end{equation}
Indeed, the function
\(f(\mathbf z)=\mathbf z^\top\mathbf D\mathbf z\)
is \(2\)-Lipschitz on the unit sphere, since
for unit vectors \(\mathbf z,\mathbf w\),
\begin{equation}
|f(\mathbf z)-f(\mathbf w)|
=
|(\mathbf z-\mathbf w)^\top
\mathbf D(\mathbf z+\mathbf w)|
\le
2\|\mathbf z-\mathbf w\|_2.
\end{equation}
Moreover, rotational symmetry gives
\begin{equation}
\mathbb E f(\mathbf z)
=
\frac1p\operatorname{Tr}\mathbf D.
\end{equation}
Thus spherical concentration yields
\eqref{eq:linear-L2-permutation-Haar-quadratic-tail},
after adjusting the universal constant.

Recall the spectral representation
\eqref{eq:linear-L2-permutation-balanced-spectral}.
Conditional on \(\boldsymbol{\Lambda}\),
the eigenvector matrix \(\mathbf O\) may be chosen
Haar-distributed on \(O(p)\), meaning uniformly
distributed over orthogonal matrices.
In particular, for every fixed unit vector
\(\mathbf v\in\mathbb R^p\),
the vector \(\mathbf O^\top\mathbf v\)
is uniformly distributed on
\(\mathbb S^{p-1}\), independently of
\(\boldsymbol{\Lambda}\).

Writing
\begin{equation}
\mathbf D(q)
=
\operatorname{diag}
\bigl(d_1(q),\ldots,d_p(q)\bigr),
\end{equation}
we have
\begin{equation}
\mathbf v^\top
\mathbf M_{\mathrm{bal}}(q)\mathbf v
=
\mathbf v^\top
\mathbf O\mathbf D(q)\mathbf O^\top
\mathbf v
=
(\mathbf O^\top\mathbf v)^\top
\mathbf D(q)
(\mathbf O^\top\mathbf v).
\end{equation}
Conditional on the eigenvalues,
\(\mathbf D(q)\) is deterministic, its diagonal
entries lie in \([0,1]\), and
\begin{equation}
\frac1p\operatorname{Tr}\mathbf D(q)
=
\overline d(q).
\end{equation}
Therefore
\eqref{eq:linear-L2-permutation-Haar-quadratic-tail}
implies, for every fixed \(q\ge0\) and
every fixed unit vector \(\mathbf v\),
\begin{equation}
\mathbb P\left(
\left|
\mathbf v^\top
\mathbf M_{\mathrm{bal}}(q)\mathbf v
-\overline d(q)
\right|>t
\,\middle|\,
\boldsymbol{\Lambda}
\right)
\le
4e^{-cpt^2}.
\label{eq:linear-L2-permutation-Haar-fixed-q}
\end{equation}

We will also use this estimate at the limiting
point \(q=\infty\).
By the spectral formula
\eqref{eq:linear-L2-permutation-djq},
\begin{equation}
d_j(\infty)
:=
\lim_{q\to\infty}d_j(q)
=
\mathbf 1_{\{\lambda_j>0\}}.
\end{equation}
Thus
\(\mathbf M_{\mathrm{bal}}(\infty)\)
denotes the spectral limit
\(\mathbf P_X\) from
\eqref{eq:linear-L2-permutation-balanced-predictor-limit}.
Its spectral weights also belong to \([0,1]\),
so
\eqref{eq:linear-L2-permutation-Haar-fixed-q}
continues to hold at \(q=\infty\).

\smallskip
\noindent
\emph{Construction of the trace-adapted grid.}

Set
\begin{equation}
\omega_p
=
K\sqrt{\frac{\log p}{p}},
\label{eq:linear-L2-permutation-omega-p}
\end{equation}
where \(K>0\) is a sufficiently large
universal constant.
For large \(p\), we have \(0<\omega_p<1\)
and \(\omega_p\to0\) as \(p \to \infty\).

We now construct a grid adapted to the growth
of \(\overline d(q)\).
Since each \(d_j(q)\) is continuous and
nondecreasing, the same is true of
\(\overline d(q)\).
Moreover,
\begin{equation}
\overline d(0)=0,
\qquad
\overline d(\infty)
:=
\lim_{q\to\infty}\overline d(q)
=
\frac{\operatorname{rank}(\mathbf C)}p
\le1.
\end{equation}
For the Gaussian matrix under consideration,
\(\operatorname{rank}(\mathbf C)=\min\{H,p\}\)
almost surely.
In particular, \(\overline d(\infty)>0\),
and \(\overline d(q)\) is strictly increasing
for finite \(q\ge0\).

Let
\begin{equation}
N
=
\left\lceil
\frac{\overline d(\infty)}{\omega_p}
\right\rceil.
\end{equation}
For \(k=1,\ldots,N-1\), choose the unique
finite \(q_k>0\) satisfying
\begin{equation}
\overline d(q_k)=k\omega_p.
\end{equation}
Such points exist because
\begin{equation}
k\omega_p
<
\overline d(\infty),
\qquad
1\le k\le N-1,
\end{equation}
and \(\overline d(q)\) is continuous and
strictly increasing from zero to its limit.

Finally, set
\begin{equation}
q_0=0,
\qquad
q_N=\infty.
\end{equation}
We obtain the grid
\begin{equation}
0=q_0<q_1<\cdots<q_{N-1}<q_N=\infty.
\end{equation}
By construction,
\begin{equation}
0\le
\overline d(q_{k+1})-\overline d(q_k)
\le
\omega_p,
\qquad
0\le k<N.
\label{eq:linear-L2-permutation-trace-grid}
\end{equation}
Indeed, the increment is exactly
\(\omega_p\) for \(k<N-1\), while the
last increment satisfies
\begin{equation}
\overline d(\infty)-\overline d(q_{N-1})
=
\overline d(\infty)-(N-1)\omega_p
\le
\omega_p,
\end{equation}
by the definition of \(N\).

Importantly, we do not require the differences
\(q_{k+1}-q_k\) to be small, or even finite.
We require only that the normalized trace
change by at most \(\omega_p\) over each
interval. In particular, although the last
interval \([q_{N-1},\infty)\) is unbounded,
the total increase of \(\overline d(q)\)
over that interval is at most \(\omega_p\).

Furthermore,
\begin{equation}
N
\le
\left\lceil\frac1{\omega_p}\right\rceil,
\label{eq:linear-L2-permutation-grid-size}
\end{equation}
because \(\overline d(\infty)\le1\).
Thus the grid contains only
\(O(1/\omega_p)\) points, independently
of the range of \(q\).

Importantly, the grid depends only on
the eigenvalues \(\boldsymbol{\Lambda}\),
through the function \(\overline d(q)\).
It does not depend on the eigenvector
matrix \(\mathbf O\).
Therefore, conditional on
\(\boldsymbol{\Lambda}\), the grid points
are deterministic, and the concentration
estimate
\eqref{eq:linear-L2-permutation-Haar-fixed-q}
applies to each of them.

\smallskip
\noindent
\emph{Controlling matrix entries through quadratic forms.}

Before applying the concentration estimate
simultaneously at the grid points, we specify
which quadratic forms need to be controlled.

The estimate
\eqref{eq:linear-L2-permutation-Haar-fixed-q}
concerns a quantity of the form
\(\mathbf v^\top\mathbf M_{\mathrm{bal}}(q)\mathbf v\),
where \(\mathbf v\) is a unit vector.
Our goal, however, concerns individual
matrix entries. We therefore choose vectors
whose quadratic forms directly determine
those entries.

For a diagonal entry, the choice is immediate:
taking \(\mathbf v=\mathbf e_i\), where
\(\mathbf e_i\) is the \(i\)-th standard
coordinate vector, gives
\begin{equation}
\mathbf e_i^\top
\mathbf M_{\mathrm{bal}}(q)\mathbf e_i
=
(\mathbf M_{\mathrm{bal}}(q))_{ii}.
\end{equation}

For an off-diagonal entry with \(i\neq c\),
we use the two unit vectors
\begin{equation}
\mathbf v_+^{(i,c)}
=
\frac{\mathbf e_i+\mathbf e_c}{\sqrt2},
\qquad
\mathbf v_-^{(i,c)}
=
\frac{\mathbf e_i-\mathbf e_c}{\sqrt2}.
\end{equation}
Since \(\mathbf M_{\mathrm{bal}}(q)\)
is symmetric, expanding their quadratic forms
gives
\begin{equation}
\begin{aligned}
(\mathbf v_+^{(i,c)})^\top
\mathbf M_{\mathrm{bal}}(q)
\mathbf v_+^{(i,c)}
&=
\frac12
\left[
M_{ii}(q)+M_{cc}(q)+2M_{ci}(q)
\right],
\\
(\mathbf v_-^{(i,c)})^\top
\mathbf M_{\mathrm{bal}}(q)
\mathbf v_-^{(i,c)}
&=
\frac12
\left[
M_{ii}(q)+M_{cc}(q)-2M_{ci}(q)
\right],
\end{aligned}
\end{equation}
where \(M_{ij}(q)\) abbreviates
\((\mathbf M_{\mathrm{bal}}(q))_{ij}\)
in these two identities.

Subtracting these expressions cancels
the diagonal terms and isolates the
off-diagonal entry:
\begin{equation}
(\mathbf M_{\mathrm{bal}}(q))_{ci}
=
\frac12\Big[
(\mathbf v_+^{(i,c)})^\top
\mathbf M_{\mathrm{bal}}(q)
\mathbf v_+^{(i,c)}
-
(\mathbf v_-^{(i,c)})^\top
\mathbf M_{\mathrm{bal}}(q)
\mathbf v_-^{(i,c)}
\Big].
\label{eq:linear-L2-permutation-polarization}
\end{equation}
This elementary identity is the
polarization identity for symmetric
bilinear forms.

Consequently, it is sufficient to control
quadratic forms for the finite set
\begin{equation}
\mathcal V_p
=
\left\{
\mathbf e_i:i\in[p]
\right\}
\cup
\left\{
\frac{\mathbf e_i+\mathbf e_c}{\sqrt2},
\frac{\mathbf e_i-\mathbf e_c}{\sqrt2}
:
1\le i<c\le p
\right\}.
\label{eq:linear-L2-permutation-test-vectors}
\end{equation}
This set contains exactly
\begin{equation}
p+2\binom p2=p^2
\end{equation}
unit vectors.

The significance of this choice is that
a uniform concentration estimate for
\(\mathbf v^\top\mathbf M_{\mathrm{bal}}(q)\mathbf v\)
over \(\mathbf v\in\mathcal V_p\)
will imply uniform control of every
matrix entry, without requiring a
separate concentration theorem for
off-diagonal entries.

\smallskip
\noindent
\emph{Concentration simultaneously at all grid points.}

Conditional on \(\boldsymbol{\Lambda}\),
both the grid
\(\{q_0,\ldots,q_N\}\) and the test-vector
set \(\mathcal V_p\) are deterministic.
We may therefore apply
\eqref{eq:linear-L2-permutation-Haar-fixed-q}
with \(t=\omega_p\) to each pair
\((q_k,\mathbf v)\).

The concentration inequality does not
require these quadratic forms to be
independent. A union bound gives
\begin{equation}
\mathbb P\left(
\max_{\substack{0\le k\le N\\
\mathbf v\in\mathcal V_p}}
\left|
\mathbf v^\top
\mathbf M_{\mathrm{bal}}(q_k)\mathbf v
-\overline d(q_k)
\right|
>
\omega_p
\,\middle|\,
\boldsymbol{\Lambda}
\right)
\le
4p^2
\left(
\left\lceil\frac1{\omega_p}\right\rceil+1
\right)
e^{-cp\omega_p^2}.
\label{eq:linear-L2-permutation-grid-union-bound}
\end{equation}
Here we used
\(|\mathcal V_p|=p^2\) and the grid-size
bound
\eqref{eq:linear-L2-permutation-grid-size}.

Since
\begin{equation}
p\omega_p^2=K^2\log p,
\end{equation}
the exponential factor equals
\(p^{-cK^2}\), whereas the number of
tests grows only polynomially in \(p\).
Hence, by choosing \(K\) sufficiently large,
the right-hand side of
\eqref{eq:linear-L2-permutation-grid-union-bound}
tends to zero.

This bound is uniform over
\(\boldsymbol{\Lambda}\), so it remains
valid after averaging over the eigenvalues.
We conclude that, with probability tending
to one, the inequalities
\begin{equation}
\left|
\mathbf v^\top
\mathbf M_{\mathrm{bal}}(q_k)\mathbf v
-\overline d(q_k)
\right|
\le
\omega_p
\end{equation}
hold simultaneously for every grid point
\(q_k\) and every test vector
\(\mathbf v\in\mathcal V_p\).

\smallskip
\noindent
\emph{Extension from the grid to every \(q\ge0\).}

We now explain why concentration at these
finitely many points controls the entire
continuous and unbounded trajectory.

Fix any \(q\ge0\).
There is a grid interval
\([q_k,q_{k+1}]\) containing \(q\);
the final interval is understood to be
\([q_{N-1},\infty)\).
Fix also any
\(\mathbf v\in\mathcal V_p\).

By the positive-semidefinite monotonicity
\eqref{eq:linear-L2-permutation-balanced-Loewner},
\begin{equation}
\mathbf v^\top
\mathbf M_{\mathrm{bal}}(q_k)\mathbf v
\le
\mathbf v^\top
\mathbf M_{\mathrm{bal}}(q)\mathbf v
\le
\mathbf v^\top
\mathbf M_{\mathrm{bal}}(q_{k+1})\mathbf v.
\label{eq:linear-L2-permutation-grid-Loewner-sandwich}
\end{equation}
The same inequalities hold when
\(q_{k+1}=\infty\), using the limiting
matrix
\(\mathbf M_{\mathrm{bal}}(\infty)\).

On the high-probability event established
above, the upper endpoint satisfies
\begin{equation}
\mathbf v^\top
\mathbf M_{\mathrm{bal}}(q_{k+1})\mathbf v
\le
\overline d(q_{k+1})+\omega_p.
\end{equation}
Combining this with
\eqref{eq:linear-L2-permutation-grid-Loewner-sandwich}
and subtracting \(\overline d(q)\), we obtain
\begin{equation}
\begin{aligned}
\mathbf v^\top
\mathbf M_{\mathrm{bal}}(q)\mathbf v
-\overline d(q)
&\le
\overline d(q_{k+1})-\overline d(q)
+\omega_p
\\
&\le
\overline d(q_{k+1})-\overline d(q_k)
+\omega_p
\le
2\omega_p.
\end{aligned}
\end{equation}
The last inequality follows from the
trace-grid construction
\eqref{eq:linear-L2-permutation-trace-grid}.

Combining the upper bound and a similar lower bound,
\begin{equation}
\left|
\mathbf v^\top
\mathbf M_{\mathrm{bal}}(q)\mathbf v
-\overline d(q)
\right|
\le
2\omega_p.
\label{eq:linear-L2-permutation-grid-interpolation}
\end{equation}

This conclusion holds for every \(q\ge0\)
and every \(\mathbf v\in\mathcal V_p\)
on the same high-probability event.
In particular, no new probabilistic
estimate is needed between grid points:
the extension is entirely deterministic.

The argument also explains why the infinite
endpoint is harmless. Although \(q\) can
increase arbitrarily far on the final
interval, the entire variation of
\(\overline d(q)\) over that interval is
at most \(\omega_p\), and the quadratic
forms are trapped between their values
at its two endpoints.

\smallskip
\noindent
\emph{Uniform bounds on diagonal and off-diagonal entries.}

We now return to the matrix entries.
Taking \(\mathbf v=\mathbf e_i\) in
\eqref{eq:linear-L2-permutation-grid-interpolation}
immediately yields
\begin{equation}
\sup_{q\ge0}
\max_i
\left|
(\mathbf M_{\mathrm{bal}}(q))_{ii}
-\overline d(q)
\right|
\le
2\omega_p.
\label{eq:linear-L2-permutation-uniform-diagonal}
\end{equation}

For off-diagonal entries, recall the vectors
\(\mathbf v_\pm^{(i,c)}\) and the polarization
identity
\eqref{eq:linear-L2-permutation-polarization}.
Writing the difference of the two quadratic
forms after subtracting their common reference
value \(\overline d(q)\), we obtain
\begin{equation}
\begin{aligned}
&
\left|
(\mathbf M_{\mathrm{bal}}(q))_{ci}
\right|
\\
&\quad\le
\frac12
\left|
(\mathbf v_+^{(i,c)})^\top
\mathbf M_{\mathrm{bal}}(q)
\mathbf v_+^{(i,c)}
-\overline d(q)
\right|
\\
&\qquad+
\frac12
\left|
(\mathbf v_-^{(i,c)})^\top
\mathbf M_{\mathrm{bal}}(q)
\mathbf v_-^{(i,c)}
-\overline d(q)
\right|
\le
2\omega_p,
\end{aligned}
\end{equation}
where the last inequality uses
\eqref{eq:linear-L2-permutation-grid-interpolation}
for both test vectors.
Consequently,
\begin{equation}
\sup_{q\ge0}
\max_{i\neq c}
\left|
(\mathbf M_{\mathrm{bal}}(q))_{ci}
\right|
\le
2\omega_p.
\label{eq:linear-L2-permutation-uniform-offdiagonal}
\end{equation}

The bounds
\eqref{eq:linear-L2-permutation-uniform-diagonal}
and
\eqref{eq:linear-L2-permutation-uniform-offdiagonal}
hold simultaneously on a single event whose
probability tends to one.
Since
\(\omega_p=K\sqrt{\log p/p}\to0\),
their sum is at most \(4\omega_p=o(1)\).
This establishes
\eqref{eq:linear-L2-permutation-uniform-entry-delocalization}
and completes Step 2.

\smallskip
\noindent
\emph{Step 3: conclusion of the uniform margin bound.}

The spectral bounds
\eqref{eq:linear-L2-permutation-C-basic-bounds}
and the uniform concentration event from
Step 2 each hold with probability tending
to one, and hence so does their intersection.

On this intersection, for every \(q\ge\delta\),
every \(i\in[p]\), and every \(c\neq i\),
\begin{equation}
\begin{aligned}
(\mathbf M_{\mathrm{bal}}(q))_{ii}
-
(\mathbf M_{\mathrm{bal}}(q))_{ci}
&\ge
\overline d(q)-4\omega_p
\\
&\ge
m_{\alpha,\delta}-4\omega_p.
\end{aligned}
\label{eq:linear-L2-permutation-balanced-margin-final-bound}
\end{equation}
The first inequality follows from
\eqref{eq:linear-L2-permutation-uniform-diagonal}
and
\eqref{eq:linear-L2-permutation-uniform-offdiagonal},
and the second from
\eqref{eq:linear-L2-permutation-average-diagonal-lower}.

Since \(m_{\alpha,\delta}>0\) is independent
of \(p,H\) and \(\omega_p\to0\), for all
sufficiently large \(p,H\),
\begin{equation}
m_{\alpha,\delta}-4\omega_p
\ge
\frac{m_{\alpha,\delta}}2.
\end{equation}
Thus we may take
\begin{equation}
c_{\alpha,\delta}
=
\frac{m_{\alpha,\delta}}2
=
\frac{\delta}
{4(1+\delta L_\alpha)}
>0.
\label{eq:linear-L2-permutation-margin-constant}
\end{equation}
This proves
\eqref{eq:linear-L2-permutation-balanced-uniform-margin}.

Since the correct-class score exceeds every
incorrect-class score by a strictly positive
amount, the balanced predictor achieves
perfect training accuracy simultaneously for
every \(q\ge\delta\), completing the proof.
\end{proof}

We can now combine the early-time analysis with the
global balanced reduction to establish permanent perfect
classification for permutation targets.

The two arguments control complementary parts of the
training trajectory. Lemma~\ref{lem:linear-L2-permutation-early-organization}
shows that accuracy is nondecreasing in the small-\(q\)
regime and reaches one at the microscopic organization
time. On the other hand,
Lemmas~\ref{lem:linear-L2-permutation-balanced-reduction}
and~\ref{lem:linear-L2-permutation-balanced-margin}
show that, after vanishing initialization, the exact
predictor remains uniformly close to a reference
trajectory with a strictly positive classification
margin for every \(q\ge\delta>0\).

Crucially, we may choose \(\delta\) strictly below
the small-\(q\) threshold \(q_{\mathrm{sm}}\).
The two controls then overlap, ensuring that
perfect accuracy persists through the transition
from the early approximately linear dynamics
to the nonlinear balanced regime, and at every
subsequent time.

\begin{linkedtheorem}
{Permutation target: monotone accuracy and permanent perfect classification}
{thm:main-permutation-unary}
\label{thm:linear-L2-permutation}

Let \(f:[p]\to[p]\) be any permutation, and consider
the two-layer linear gradient flow with Gaussian
initialization of scale \(\sigma\).
Suppose that
\begin{equation}
\frac Hp
\longrightarrow
\alpha\in(0,\infty)
\label{eq:linear-L2-permutation-theorem-aspect-ratio}
\end{equation}
as \(p,H\to\infty\).
Define the organization time
\begin{equation}
\tau_{\mathrm{org}}
=
K
\sqrt{
\frac{\log(2p^2H)}{H}
},
\label{eq:linear-L2-permutation-theorem-organization-time}
\end{equation}
where \(K\) is a sufficiently large universal constant.

Then, in the sequential limit
\(\sigma\downarrow0\) followed by
\(p,H\to\infty\) with \(H/p\to\alpha\),
the training accuracy is nondecreasing
throughout the entire gradient-flow trajectory,
with probability tending to one:
\begin{equation}
\mathbb P\left(
\Acc\bigl(\mathbf M^{(\sigma)}(\tau_1)\bigr)
\le
\Acc\bigl(\mathbf M^{(\sigma)}(\tau_2)\bigr)
\quad
\text{for all }
0\le\tau_1\le\tau_2
\right)
\longrightarrow1.
\label{eq:linear-L2-permutation-global-accuracy-monotonicity}
\end{equation}
Moreover, perfect accuracy is attained by
\(\tau_{\mathrm{org}}\) and persists permanently:
\begin{equation}
\mathbb P\left(
\Acc\bigl(\mathbf M^{(\sigma)}(\tau)\bigr)
=
1
\quad
\text{for all }
\tau\ge\tau_{\mathrm{org}}
\right)
\longrightarrow1.
\label{eq:linear-L2-permutation-permanent-perfect}
\end{equation}

The organization stage is microscopic in predictor
space. More precisely, for every fixed \(p,H\),
as \(\sigma\downarrow0\),
\begin{equation}
\sup_{0\le\tau\le\tau_{\mathrm{org}}}
\left\|
\mathbf M^{(\sigma)}(\tau)
\right\|_{\mathrm{op}}
\longrightarrow0,
\label{eq:linear-L2-permutation-theorem-microscopic-predictor}
\end{equation}
and
\begin{equation}
\begin{aligned}
\sup_{0\le\tau\le\tau_{\mathrm{org}}}
\Big|
&
\mathcal L\left(
\mathbf U^{(\sigma)}(\tau),
\mathbf V^{(\sigma)}(\tau)
\right)
\\
&-
\mathcal L\left(
\mathbf U^{(\sigma)}(0),
\mathbf V^{(\sigma)}(0)
\right)
\Big|
\longrightarrow0.
\end{aligned}
\label{eq:linear-L2-permutation-theorem-flat-loss}
\end{equation}
These two limits hold almost surely when the
Gaussian matrices underlying the initialization
are held fixed as \(\sigma\downarrow0\).

Finally, the initial training accuracy satisfies
\begin{equation}
\Acc\left(
\mathbf M^{(\sigma)}(0)
\right)
\longrightarrow0
\label{eq:linear-L2-permutation-theorem-initial-accuracy}
\end{equation}
in probability as \(p,H\to\infty\), independently
of the positive initialization scale \(\sigma\).

Thus, permutation targets exhibit a microscopic
rise from vanishing to perfect training accuracy,
while the subsequent nonlinear evolution preserves
perfect classification for all time.
\end{linkedtheorem}

\begin{proof}
By the permutation reduction preceding
\eqref{eq:linear-L2-permutation-identity-target},
it suffices to consider the identity target
\(\mathbf T=\mathbf I_p\).

Fix
\begin{equation}
\delta
=
\frac12 q_{\mathrm{sm}}(\alpha)>0,
\qquad
\tau_\sigma(\delta)
=
\frac12\log\frac{\delta}{H\sigma^2/2},
\label{eq:linear-L2-permutation-proof-delta}
\end{equation}
where \(q_{\mathrm{sm}}(\alpha)\) is the
small-\(q\) threshold from
Lemma~\ref{lem:linear-L2-permutation-small-q-monotonicity}.
We show that the early-time and late-time
classification guarantees overlap, thereby
covering the entire trajectory.

\smallskip
\noindent
\emph{Early organization and small-\(q\) monotonicity.}

By
Lemma~\ref{lem:linear-L2-permutation-small-q-monotonicity},
with probability tending to one as
\(p,H\to\infty\), every classification margin
\begin{equation}
\Delta_{ci}(\tau)
=
M_{ii}(\tau)-M_{ci}(\tau),
\qquad c\neq i,
\label{eq:linear-L2-permutation-proof-margin}
\end{equation}
is strictly increasing whenever
\(q(\tau)\le q_{\mathrm{sm}}(\alpha)\).
In particular, training accuracy is
nondecreasing throughout this regime.

Moreover, since permutation targets have
\(n_{\max}=1\),
Proposition~\ref{prop:linear-L2-unary-early-perfect}
guarantees perfect accuracy at
\(\tau_{\mathrm{org}}\), with probability
tending to one in the prescribed sequential
limit.

For every fixed \(p,H\),
\begin{equation}
q(\tau_{\mathrm{org}})
=
\frac{H\sigma^2}{2}
e^{2\tau_{\mathrm{org}}}
\longrightarrow0
\qquad
\text{as }\sigma\downarrow0.
\label{eq:linear-L2-permutation-proof-q-org-zero}
\end{equation}
Thus, for sufficiently small \(\sigma\),
\begin{equation}
\tau_{\mathrm{org}}
<
\tau_\sigma(\delta)
<
\tau_\sigma(q_{\mathrm{sm}}).
\label{eq:linear-L2-permutation-proof-org-before-delta}
\end{equation}

Consequently, on the intersection of the
early-perfect and small-\(q\) monotonicity
events, accuracy is nondecreasing from
initialization and remains equal to one
throughout
\begin{equation}
\tau_{\mathrm{org}}
\le\tau\le
\tau_\sigma(q_{\mathrm{sm}}).
\label{eq:linear-L2-permutation-proof-perfect-small-q}
\end{equation}

\smallskip
\noindent
\emph{Permanent accuracy in the balanced regime.}

By
Lemma~\ref{lem:linear-L2-permutation-balanced-margin},
there exists \(c_{\alpha,\delta}>0\),
depending only on \(\alpha,\delta\), such that,
with probability tending to one as
\(p,H\to\infty\),
\begin{equation}
(\mathbf M_{\mathrm{bal}}(q))_{ii}
-
(\mathbf M_{\mathrm{bal}}(q))_{ci}
\ge
c_{\alpha,\delta}
\label{eq:linear-L2-permutation-proof-balanced-margin}
\end{equation}
simultaneously for all \(q\ge\delta\),
\(i\in[p]\), and \(c\neq i\).

Since a standard Gaussian
\(\mathbf X\in\mathbb R^{H\times p}\)
has rank \(\min\{H,p\}\) almost surely,
Lemma~\ref{lem:linear-L2-permutation-balanced-reduction}
applies for each fixed \(p,H\).
In particular,
\begin{equation}
\sup_{q\ge\delta}
\left\|
\widehat{\mathbf M}_\sigma(q)
-
\mathbf M_{\mathrm{bal}}(q)
\right\|_{\max}
\longrightarrow0
\qquad
\text{as }\sigma\downarrow0
\label{eq:linear-L2-permutation-proof-balanced-reduction}
\end{equation}
almost surely.

Therefore, on the balanced-margin event,
for sufficiently small \(\sigma\), the
supremum in
\eqref{eq:linear-L2-permutation-proof-balanced-reduction}
is smaller than \(c_{\alpha,\delta}/4\).
For every \(q\ge\delta\) and \(c\neq i\),
the exact classification margin then satisfies
\begin{equation}
\begin{aligned}
&
(\widehat{\mathbf M}_\sigma(q))_{ii}
-
(\widehat{\mathbf M}_\sigma(q))_{ci}
\\
&\quad\ge
(\mathbf M_{\mathrm{bal}}(q))_{ii}
-
(\mathbf M_{\mathrm{bal}}(q))_{ci}
-
2\left\|
\widehat{\mathbf M}_\sigma(q)
-
\mathbf M_{\mathrm{bal}}(q)
\right\|_{\max}
\ge
\frac{c_{\alpha,\delta}}2>0.
\end{aligned}
\label{eq:linear-L2-permutation-proof-tail-positive-margin}
\end{equation}
Hence the exact predictor has perfect
accuracy for every
\begin{equation}
\tau\ge\tau_\sigma(\delta).
\label{eq:linear-L2-permutation-proof-tail-perfect}
\end{equation}

\smallskip
\noindent
\emph{Conclusion and order of limits.}

The small-\(q\) argument controls accuracy
up to \(\tau_\sigma(q_{\mathrm{sm}})\),
whereas the balanced reduction guarantees
perfect accuracy from
\(\tau_\sigma(\delta)\) onward.
Since
\(\delta<q_{\mathrm{sm}}\), these intervals
overlap. Thus, whenever the preceding
high-probability events hold, accuracy
is nondecreasing for all \(\tau\ge0\)
and equals one for all
\(\tau\ge\tau_{\mathrm{org}}\).

To justify the probability statements,
observe that the small-\(q\) and balanced-margin
events have probabilities tending to one
as \(p,H\to\infty\), independently of
\(\sigma\). For each fixed \(p,H\), the
balanced-reduction error vanishes almost
surely as \(\sigma\downarrow0\).
Combining these facts with the early-perfect
guarantee by a union bound, and taking
\(\sigma\downarrow0\) before
\(p,H\to\infty\), proves
\eqref{eq:linear-L2-permutation-global-accuracy-monotonicity}
and
\eqref{eq:linear-L2-permutation-permanent-perfect}.

Finally, for fixed \(p,H\), the interval
\([0,\tau_{\mathrm{org}}]\) has finite length
independent of \(\sigma\).
The microscopic linearization result
\eqref{eq:linear-L2-predictor-microscopic} of the Theorem~\ref{thm:linear-L2-microscopic-linearization}
implies that the predictor converges
uniformly to zero on this interval,
while
\eqref{eq:linear-L2-loss-microscopic}
gives uniform vanishing of its loss variation.
This proves
\eqref{eq:linear-L2-permutation-theorem-microscopic-predictor}
and
\eqref{eq:linear-L2-permutation-theorem-flat-loss}.
The initial-accuracy statement
\eqref{eq:linear-L2-permutation-theorem-initial-accuracy}
follows from
\eqref{eq:linear-L2-unary-initial-accuracy-limit}.
\end{proof}

The permanence result is particularly informative in the undercomplete
regime
\begin{equation}
0
<
\alpha
<
1.
\label{eq:linear-L2-permutation-undercomplete-alpha}
\end{equation}
Here
\begin{equation}
\operatorname{rank}(\mathbf T)
=
p
>
H,
\label{eq:linear-L2-permutation-undercomplete-rank}
\end{equation}
so the identity target cannot be represented exactly by an end-to-end
matrix
\(\mathbf V^{\top}\mathbf U\)
of rank at most \(H\). Nevertheless, the theorem shows that perfect
classification is reached microscopically and then maintained for all
later training times.

In the vanishing-initialization limit, the balanced trajectory converges
to the orthogonal projector onto the row space of \(\mathbf X\),
\begin{equation}
\mathbf P_X
=
\mathbf X^{\top}
\left(
\mathbf X\mathbf X^{\top}
\right)^{-1}
\mathbf X,
\label{eq:linear-L2-permutation-undercomplete-projector}
\end{equation}
which satisfies
\begin{equation}
\left\|
\mathbf I_p-\mathbf P_X
\right\|_{\mathrm F}^2
=
p-H.
\label{eq:linear-L2-permutation-undercomplete-projector-error}
\end{equation}
For the unary identity target,
\begin{equation}
\mathcal L(\mathbf M)
=
\frac{1}{2p}
\left\|
\mathbf M-\mathbf I_p
\right\|_{\mathrm F}^2.
\label{eq:linear-L2-permutation-undercomplete-loss}
\end{equation}
For every fixed \(p,H\) with \(H<p\), the global balanced reduction therefore
implies
\begin{equation}
\lim_{\sigma\downarrow0}
\limsup_{\tau\to\infty}
\left|
\mathcal L
\left(
\mathbf M^{(\sigma)}(\tau)
\right)
-
\frac{p-H}{2p}
\right|
=
0.
\label{eq:linear-L2-permutation-terminal-loss}
\end{equation}
Taking the large-system limit afterward with
\(H/p\to\alpha<1\) gives
\begin{equation}
\frac{p-H}{2p}
\longrightarrow
\frac{1-\alpha}{2}.
\label{eq:linear-L2-permutation-terminal-loss-large-system}
\end{equation}
Thus perfect classification does not require convergence to the
least-squares target itself: in the undercomplete regime, a nonvanishing
least-squares error remains even though the training accuracy is permanently
equal to one.

\subsection{Deeper linear networks: \texorpdfstring{\(L\ge3\)}{L >= 3}}
\label{app:linear-deep}

We now turn to depths \(L\ge3\). In contrast to the two-layer case, the
leading near-origin dynamics are already nonlinear, and escape from
vanishing initialization occurs on a diverging time scale. Nevertheless,
the same qualitative separation between target-driven organization and a
later predictive bottleneck persists. On the first near-origin scale, the
normalized dynamics develop a coherent component in the direction of the
full target \(\mathbf T\), so that the target classifier can emerge while
the end-to-end predictor remains microscopic and the loss is asymptotically
unchanged.

This distinction is important when relating our picture to the saddle-escape
analysis of \citet{jacot2021saddle}. From the viewpoint of the loss and the
parameter-scale escape from the rank-zero saddle, the width-one trajectory
identified there is the first nontrivial escape dynamics. For classification,
however, nontrivial behavior appears strictly earlier: the predictor can
already reorganize enough to recover the classifier of the full target
while remaining microscopic and producing a vanishing change in loss.

At a later scale, when the largest singular value of \(\mathbf T\) is simple,
\begin{equation}
s_1>s_2,
\label{eq:linear-deep-intro-simple-leading-mode}
\end{equation}
we use \citet[Theorem~2]{jacot2021saddle} to show that the first escape from
the origin is asymptotically governed by a width-one trajectory aligned with
the leading singular mode. In our accuracy-based picture, this escape is
therefore not the first stage of the dynamics, but rather the subsequent
degeneracy bottleneck.

Recovery of the same Gaussian-initialized trajectory is not known in the
generality needed here for \(L\ge3\). We therefore treat the first two
stages rigorously and return at the end of the subsection to a spectrally
aligned setting in which recruitment of the remaining modes and final
recovery can be followed exactly.

\paragraph{Near-origin dynamics.}

We retain the isotropic setting and the rescaled time
\begin{equation}
\tau
=
\gamma t,
\label{eq:linear-deep-rescaled-time}
\end{equation}
so that, up to an additive constant, the loss is
\begin{equation}
\mathcal L
=
\frac{\gamma}{2}
\left\|
\mathbf M-\mathbf T
\right\|_{\mathrm F}^2,
\qquad
\mathbf M
=
\mathbf W_L\cdots\mathbf W_1.
\label{eq:linear-deep-loss}
\end{equation}
For simplicity, all hidden layers have width \(H\):
\begin{equation}
\mathbf W_1
\in
\mathbb R^{H\times p},
\qquad
\mathbf W_\ell
\in
\mathbb R^{H\times H},
\quad
2\le\ell\le L-1,
\qquad
\mathbf W_L
\in
\mathbb R^{C\times H}.
\label{eq:linear-deep-layer-shapes}
\end{equation}
For \(b\ge a\), write
\begin{equation}
\mathbf W_{b:a}
=
\mathbf W_b
\mathbf W_{b-1}
\cdots
\mathbf W_a,
\label{eq:linear-deep-product-notation}
\end{equation}
with the natural identity convention for empty products. Gradient flow in
the rescaled time is
\begin{equation}
\dot{\mathbf W}_\ell
=
\mathbf W_{L:\ell+1}^{\top}
\left(
\mathbf T-\mathbf M
\right)
\mathbf W_{\ell-1:1}^{\top},
\qquad
\ell=1,\ldots,L.
\label{eq:linear-deep-factor-flow}
\end{equation}

We initialize
\begin{equation}
\mathbf W_\ell(0)
=
\sigma\mathbf G_\ell,
\qquad
\ell=1,\ldots,L,
\label{eq:linear-deep-Gaussian-init}
\end{equation}
where all entries of
\(\mathbf G_1,\ldots,\mathbf G_L\)
are independent standard Gaussian variables. For \(L\ge3\), the right-hand
side of
\eqref{eq:linear-deep-factor-flow}
is of order
\(\sigma^{L-1}\)
near the origin. Since the factors themselves have size \(O(\sigma)\), the
first \(O(1)\) evolution of their normalized values occurs on the
diverging physical-time scale
\begin{equation}
s
=
\sigma^{L-2}\tau.
\label{eq:linear-deep-slow-time}
\end{equation}
Define normalized factors by
\begin{equation}
\mathbf W_\ell(\tau)
=
\sigma
\mathbf Z_\ell^{(\sigma)}(s),
\qquad
s
=
\sigma^{L-2}\tau,
\label{eq:linear-deep-normalized-factors}
\end{equation}
and let
\begin{equation}
\boldsymbol{\Pi}^{(\sigma)}(s)
=
\mathbf Z_L^{(\sigma)}(s)
\cdots
\mathbf Z_1^{(\sigma)}(s).
\label{eq:linear-deep-normalized-predictor}
\end{equation}
Substitution into
\eqref{eq:linear-deep-factor-flow}
gives the exact normalized dynamics
\begin{equation}
\frac{d\mathbf Z_\ell^{(\sigma)}}{ds}
=
\left(
\mathbf Z_{L:\ell+1}^{(\sigma)}
\right)^{\top}
\left(
\mathbf T
-
\sigma^L\boldsymbol{\Pi}^{(\sigma)}
\right)
\left(
\mathbf Z_{\ell-1:1}^{(\sigma)}
\right)^{\top},
\qquad
\mathbf Z_\ell^{(\sigma)}(0)
=
\mathbf G_\ell.
\label{eq:linear-deep-normalized-flow}
\end{equation}
Hence, unlike for \(L=2\), the vanishing-initialization limit is already a
nonlinear dynamical system:
\begin{equation}
\frac{d\mathbf Z_\ell}{ds}
=
\mathbf Z_{L:\ell+1}^{\top}
\mathbf T
\mathbf Z_{\ell-1:1}^{\top},
\qquad
\mathbf Z_\ell(0)
=
\mathbf G_\ell.
\label{eq:linear-deep-homogeneous-flow}
\end{equation}
We refer to
\eqref{eq:linear-deep-homogeneous-flow}
as the homogeneous near-origin flow.
Unlike the exact gradient flow, this homogeneous
system may exhibit finite-time blow-up for
\(L\ge3\), since its vector field is
superlinear in the factors.
The following lemma therefore justifies the
near-origin approximation on compact intervals
strictly preceding the maximal existence time
of the homogeneous system.
This restriction will not affect our subsequent
organization result, which takes place on a
short initial interval where existence is
guaranteed.

\begin{lemma}[Near-origin reduction for \(L\ge3\)]
\label{lem:linear-deep-near-origin-reduction}
Fix \(p,C,H,L\), the target \(\mathbf T\), and a realization of
\(\mathbf G_1,\ldots,\mathbf G_L\). Let
\((\mathbf Z_1,\ldots,\mathbf Z_L)\)
be the maximal solution of
\eqref{eq:linear-deep-homogeneous-flow},
with maximal forward existence time
\(s_*\in(0,\infty]\).
Then, for every \(S<s_*\),
\begin{equation}
\max_{1\le\ell\le L}
\sup_{0\le s\le S}
\left\|
\mathbf Z_\ell^{(\sigma)}(s)
-
\mathbf Z_\ell(s)
\right\|_{\mathrm F}
\longrightarrow
0
\qquad
\text{as }
\sigma\downarrow0.
\label{eq:linear-deep-factor-convergence}
\end{equation}
Consequently,
\begin{equation}
\sup_{0\le s\le S}
\left\|
\frac{
\mathbf M^{(\sigma)}
\left(
s/\sigma^{L-2}
\right)
}{
\sigma^L
}
-
\mathbf Z_L(s)\cdots\mathbf Z_1(s)
\right\|_{\mathrm F}
\longrightarrow
0.
\label{eq:linear-deep-predictor-convergence}
\end{equation}
In particular,
\begin{equation}
\sup_{0\le s\le S}
\left\|
\mathbf M^{(\sigma)}
\left(
s/\sigma^{L-2}
\right)
\right\|_{\mathrm F}
=
O(\sigma^L),
\label{eq:linear-deep-microscopic-predictor}
\end{equation}
and
\begin{equation}
\sup_{0\le s\le S}
\left|
\mathcal L
\left(
\mathbf M^{(\sigma)}
\left(
s/\sigma^{L-2}
\right)
\right)
-
\mathcal L
\left(
\mathbf M^{(\sigma)}(0)
\right)
\right|
\longrightarrow
0.
\label{eq:linear-deep-flat-loss}
\end{equation}
\end{lemma}

\begin{proof}
The proof is based on continuous dependence of ODE solutions
on their vector fields.
The exact normalized dynamics
\eqref{eq:linear-deep-normalized-flow}
differ from the homogeneous system
\eqref{eq:linear-deep-homogeneous-flow}
by terms of order \(\sigma^L\).
We show that this perturbation produces a vanishing
difference between the two trajectories on every
compact interval on which the homogeneous solution exists.

The main point is that the vector fields are polynomial
and therefore only locally Lipschitz.
We first restrict attention to a bounded neighborhood
of the homogeneous trajectory and use Gronwall's
inequality to control the difference there.
A first-exit argument then guarantees that the
perturbed trajectory remains in this neighborhood
throughout the prescribed interval.

\smallskip
\noindent
\emph{Step 1: comparison of the two normalized flows.}

Collect the normalized factors into
\begin{equation}
\mathcal Z
=
(\mathbf Z_1,\ldots,\mathbf Z_L),
\qquad
\|\mathcal Z\|
=
\max_{1\le\ell\le L}
\|\mathbf Z_\ell\|_{\mathrm F}.
\label{eq:linear-deep-collection-norm}
\end{equation}
Denote the vector fields of the exact normalized
system and its homogeneous limit by
\(F_\sigma\) and \(F_0\), respectively:
\begin{equation}
\frac{d\mathcal Z^{(\sigma)}}{ds}
=
F_\sigma(\mathcal Z^{(\sigma)}),
\qquad
\frac{d\mathcal Z}{ds}
=
F_0(\mathcal Z).
\label{eq:linear-deep-vector-fields}
\end{equation}
Both systems have the same initial condition,
\begin{equation}
\mathcal Z^{(\sigma)}(0)
=
\mathcal Z(0)
=
(\mathbf G_1,\ldots,\mathbf G_L).
\end{equation}

Fix any \(S<s_*\).
By definition of the maximal existence time,
the homogeneous solution is well-defined and
continuous on \([0,S]\).
Consequently,
\begin{equation}
R_0
:=
\sup_{0\le s\le S}
\|\mathcal Z(s)\|
<
\infty.
\label{eq:linear-deep-limit-bound}
\end{equation}
Choose \(R>R_0+1\) and consider the ball
\begin{equation}
\|\mathcal Z\|\le R.
\label{eq:linear-deep-R-ball}
\end{equation}

Since \(F_0\) is a polynomial vector field,
it is Lipschitz on this bounded ball.
Thus there exists \(K_R<\infty\) such that
\begin{equation}
\|F_0(\mathcal Z)
-
F_0(\widetilde{\mathcal Z})\|
\le
K_R
\|\mathcal Z-\widetilde{\mathcal Z}\|
\label{eq:linear-deep-limit-Lipschitz}
\end{equation}
whenever both arguments belong to the ball.

Moreover, comparing
\eqref{eq:linear-deep-normalized-flow}
and
\eqref{eq:linear-deep-homogeneous-flow},
we see that their vector fields differ only
through the term involving
\(\sigma^L\boldsymbol{\Pi}^{(\sigma)}\).
Each component of this difference is
\(\sigma^L\) times a product containing
\(2L-1\) normalized factors.
Therefore, on the same bounded ball,
there exists \(C_R<\infty\), independent
of sufficiently small \(\sigma\), such that
\begin{equation}
\sup_{\|\mathcal Z\|\le R}
\|F_\sigma(\mathcal Z)-F_0(\mathcal Z)\|
\le
C_R\sigma^L.
\label{eq:linear-deep-vector-field-error}
\end{equation}

These two estimates provide the ingredients
for controlling the difference between
the trajectories, provided that both remain
inside the chosen ball.

\smallskip
\noindent
\emph{Step 2: Gronwall estimate and first-exit argument.}

Define the trajectory difference
\begin{equation}
E_\sigma(s)
=
\|\mathcal Z^{(\sigma)}(s)-\mathcal Z(s)\|.
\end{equation}
Let \(s_\sigma\) be the first time in
\([0,S]\) at which \(E_\sigma(s)=1\),
or set \(s_\sigma=S\) if no such time exists.

By construction,
\begin{equation}
E_\sigma(s)\le1
\qquad
\text{for }0\le s\le s_\sigma.
\end{equation}
Hence, throughout this interval,
\begin{equation}
\begin{aligned}
\|\mathcal Z^{(\sigma)}(s)\|
&\le
\|\mathcal Z(s)\|+E_\sigma(s)
\\
&\le
R_0+1
<
R.
\end{aligned}
\end{equation}
Thus both trajectories remain in the region
where
\eqref{eq:linear-deep-limit-Lipschitz}
and
\eqref{eq:linear-deep-vector-field-error}
apply.

Subtracting the integral equations for
the two trajectories, and using their
identical initial conditions, gives
\begin{equation}
\begin{aligned}
E_\sigma(s)
&\le
\int_0^s
\left\|
F_\sigma\left(
\mathcal Z^{(\sigma)}(u)
\right)
-
F_0\left(
\mathcal Z(u)
\right)
\right\|du
\\
&\le
\int_0^s
\left[
K_R E_\sigma(u)
+
C_R\sigma^L
\right]du,
\qquad
0\le s\le s_\sigma.
\end{aligned}
\label{eq:linear-deep-difference-inequality}
\end{equation}
Here we added and subtracted
\(F_0(\mathcal Z^{(\sigma)}(u))\),
then applied the Lipschitz and
vector-field perturbation bounds.

The integral form of Gronwall's inequality
therefore yields
\begin{equation}
\sup_{0\le s\le s_\sigma}
E_\sigma(s)
\le
C_R\sigma^L
\frac{e^{K_RS}-1}{K_R},
\label{eq:linear-deep-Gronwall}
\end{equation}
where the fraction is interpreted as \(S\)
when \(K_R=0\).

Since \(R\) and \(S\) are fixed,
the right-hand side tends to zero
as \(\sigma\downarrow0\).
In particular, it is strictly smaller
than one for all sufficiently small \(\sigma\).

But if the first-exit time occurred
at or before \(S\), we would have
\(E_\sigma(s_\sigma)=1\), contradicting
\eqref{eq:linear-deep-Gronwall}.
Therefore no first exit occurs on \([0,S]\).

It follows that, for sufficiently small
\(\sigma\), the exact normalized solution
exists throughout \([0,S]\) and
\begin{equation}
\sup_{0\le s\le S}
\|\mathcal Z^{(\sigma)}(s)-\mathcal Z(s)\|
\le
C_R\sigma^L
\frac{e^{K_RS}-1}{K_R}
\longrightarrow0.
\end{equation}
This proves
\eqref{eq:linear-deep-factor-convergence}.

Notice that no global existence assumption
on the homogeneous system was needed:
the comparison takes place entirely within
the compact interval \([0,S]\), where its
solution is known to remain bounded.

\smallskip
\noindent
\emph{Step 3: convergence of the predictor
and microscopicity of the loss.}

By the normalization
\eqref{eq:linear-deep-normalized-factors},
the exact end-to-end predictor satisfies
\begin{equation}
\mathbf M^{(\sigma)}
\left(
\frac{s}{\sigma^{L-2}}
\right)
=
\sigma^L
\mathbf Z_L^{(\sigma)}(s)
\cdots
\mathbf Z_1^{(\sigma)}(s).
\label{eq:linear-deep-M-normalized-product}
\end{equation}
The factor convergence established in Step 2
is uniform on \([0,S]\).
Since all factors are uniformly bounded
on this interval, continuity of matrix
multiplication gives
\begin{equation}
\sup_{0\le s\le S}
\left\|
\mathbf Z_L^{(\sigma)}(s)
\cdots
\mathbf Z_1^{(\sigma)}(s)
-
\mathbf Z_L(s)
\cdots
\mathbf Z_1(s)
\right\|_{\mathrm F}
\longrightarrow0.
\end{equation}
This proves
\eqref{eq:linear-deep-predictor-convergence}.

Furthermore, the normalized product remains
uniformly bounded on \([0,S]\).
Consequently,
\begin{equation}
\sup_{0\le s\le S}
\left\|
\mathbf M^{(\sigma)}
\left(
\frac{s}{\sigma^{L-2}}
\right)
\right\|_{\mathrm F}
=
O(\sigma^L),
\end{equation}
which establishes
\eqref{eq:linear-deep-microscopic-predictor}.
Thus the entire trajectory on the
corresponding original-time interval
\([0,S/\sigma^{L-2}]\) remains microscopic
in predictor space.

Finally, recall that
\begin{equation}
\mathcal L(\mathbf M)
=
\frac{\gamma}{2}
\|\mathbf M-\mathbf T\|_{\mathrm F}^2.
\end{equation}
For any two predictors
\(\mathbf M,\mathbf M_0\),
expanding the squared norms gives
\begin{equation}
\mathcal L(\mathbf M)-\mathcal L(\mathbf M_0)=
-\gamma
\langle
\mathbf T,\mathbf M-\mathbf M_0
\rangle_{\mathrm F}
+
\frac{\gamma}{2}
\left(
\|\mathbf M\|_{\mathrm F}^2
-
\|\mathbf M_0\|_{\mathrm F}^2
\right).
\label{eq:linear-deep-loss-difference-explicit}
\end{equation}
Taking
\begin{equation}
\mathbf M
=
\mathbf M^{(\sigma)}
\left(
\frac{s}{\sigma^{L-2}}
\right),
\qquad
\mathbf M_0
=
\mathbf M^{(\sigma)}(0),
\end{equation}
we know from
\eqref{eq:linear-deep-microscopic-predictor}
that both matrices have Frobenius norm
\(O(\sigma^L)\), uniformly over
\(s\in[0,S]\).

Because \(\gamma\) and \(\mathbf T\)
are fixed, Cauchy--Schwarz in
\eqref{eq:linear-deep-loss-difference-explicit}
therefore gives
\begin{equation}
\sup_{0\le s\le S}
\left|
\mathcal L\left(
\mathbf M^{(\sigma)}
\left(
\frac{s}{\sigma^{L-2}}
\right)
\right)
-
\mathcal L\left(
\mathbf M^{(\sigma)}(0)
\right)
\right|
=
O(\sigma^L)
\longrightarrow0.
\end{equation}
This proves
\eqref{eq:linear-deep-flat-loss}
and completes the proof.
\end{proof}

Thus vanishing initialization separates the near-origin evolution into two
scales. The original factors remain of size \(O(\sigma)\), and the
end-to-end predictor remains microscopic,
\(\mathbf M=O(\sigma^L)\), while the normalized factors undergo a
nontrivial nonlinear evolution on the slow time
\(s=\sigma^{L-2}\tau\). We next show that, at large width, the first
classifier selected by this homogeneous flow is governed by the full target
\(\mathbf T\).

\paragraph{Target-driven organization.}

We now show that the homogeneous near-origin dynamics can recover
the classifier induced by the full target \(\mathbf T\), before
the first rank-one escape from the origin. The mechanism is a
competition between the initial random predictor and a coherent
target-directed drift generated by gradient flow.

Although the width-normalized initial end-to-end product
is centered, its first derivative with respect to the
width-rescaled time has expectation \(L\mathbf T\).
At large width, both the initial normalized product and
the fluctuations of this derivative become small when
evaluated on individual training-set score comparisons.
Consequently, after a short evolution of the normalized
dynamics, the target-directed contribution dominates the
random fluctuations. The resulting classifier agrees with
that of the full target, even though the original end-to-end
predictor remains microscopic and the loss has changed
by a vanishing amount.

To analyze this early evolution, we introduce a second
rescaling that absorbs the natural width dependence of the
homogeneous system. Define
\begin{equation}
u
=
H^{(L-2)/2}s
=
H^{(L-2)/2}\sigma^{L-2}\tau,
\qquad
\overline{\mathbf Z}_\ell(u)
=
\frac{1}{\sqrt H}\mathbf Z_\ell(s).
\label{eq:linear-deep-width-time}
\end{equation}
In these variables, the homogeneous flow
\eqref{eq:linear-deep-homogeneous-flow}
becomes
\begin{equation}
\frac{d\overline{\mathbf Z}_\ell}{du}
=
\overline{\mathbf Z}_{L:\ell+1}^{\top}
\mathbf T
\overline{\mathbf Z}_{\ell-1:1}^{\top},
\qquad
\overline{\mathbf Z}_\ell(0)
=
\overline{\mathbf G}_\ell
:=
\frac{1}{\sqrt H}\mathbf G_\ell.
\label{eq:linear-deep-width-normalized-flow}
\end{equation}
We denote its end-to-end predictor by
\begin{equation}
\overline{\boldsymbol{\Pi}}(u)
=
\overline{\mathbf Z}_L(u)
\cdots
\overline{\mathbf Z}_1(u).
\label{eq:linear-deep-Q}
\end{equation}

For every fixed \(H\), the near-origin reduction
of Lemma~\ref{lem:linear-deep-near-origin-reduction}
implies
\begin{equation}
\frac{
\mathbf M^{(\sigma)}
\left(
\sigma^{-(L-2)}H^{-(L-2)/2}u
\right)
}{
\sigma^L H^{L/2}
}
\longrightarrow
\overline{\boldsymbol{\Pi}}(u)
\qquad
\text{as }\sigma\downarrow0,
\label{eq:linear-deep-M-Q-relation}
\end{equation}
uniformly on compact intervals where the homogeneous
flow exists. The normalization
\(\sigma^L H^{L/2}\) is positive and therefore
does not affect classification. In particular,
strict score margins of \(\overline{\boldsymbol{\Pi}}(u)\) transfer
to the exact gradient flow for sufficiently small
initialization.

For each training example \(\mathbf x_i\), let
\begin{equation}
\widehat c_i
=
\arg\max_{c\in[C]}
\left(
\mathbf T\mathbf x_i
\right)_c
\label{eq:linear-deep-target-class}
\end{equation}
be the class predicted by the full target.
We assume that these maximizers are unique and
define the minimum target decision gap and maximum
input norm by
\begin{equation}
\delta_{\mathbf T}
=
\min_i
\min_{c\neq\widehat c_i}
\left[
\left(
\mathbf T\mathbf x_i
\right)_{\widehat c_i}
-
\left(
\mathbf T\mathbf x_i
\right)_c
\right],
\qquad
R
=
\max_i\|\mathbf x_i\|_2.
\label{eq:linear-deep-target-gap}
\end{equation}

The following theorem establishes that, under
a nonvanishing target decision gap, the exact
gradient flow recovers all target predictions
while remaining microscopic throughout the
organization interval.

\begin{theorem}[Target-driven organization for \(L\ge3\)]
\label{thm:linear-deep-target-organization}

Fix the depth \(L\ge3\), and consider a sequence
of problems indexed by the hidden width
\(H\to\infty\). Suppose that
\begin{equation}
\frac{p}{H}=O(1),
\qquad
\frac{C}{H}=O(1),
\qquad
R=O(1),
\qquad
\|\mathbf T\|_{\mathrm{op}}=O(1),
\qquad
\delta_{\mathbf T}\ge\delta_0>0.
\label{eq:linear-deep-organization-regularity}
\end{equation}
Here \(\delta_0\) is independent of \(H\).

Define
\begin{equation}
\ell_H
=
\log(2NCH)
\label{eq:linear-deep-organization-log}
\end{equation}
and assume that
\begin{equation}
\frac{\ell_H^{\,L-1}}{\sqrt H}
\longrightarrow0.
\label{eq:linear-deep-organization-growth-condition}
\end{equation}
Choose a deterministic sequence \(b_H\to\infty\)
sufficiently slowly that
\begin{equation}
u_{\mathrm{org}}
:=
b_H
\frac{\ell_H^{\,L/2}}{\sqrt H}
\longrightarrow0.
\label{eq:linear-deep-organization-u}
\end{equation}

Then there exist events \(\mathcal E_H\),
depending only on the Gaussian initialization,
such that
\begin{equation}
\mathbb P(\mathcal E_H)
\longrightarrow1
\qquad
\text{as }H\to\infty,
\label{eq:linear-deep-organization-good-event-probability}
\end{equation}
and, for all sufficiently large \(H\), on
\(\mathcal E_H\), the homogeneous flow
\eqref{eq:linear-deep-width-normalized-flow}
exists throughout \([0,u_{\mathrm{org}}]\)
and satisfies
\begin{equation}
\arg\max_{c\in[C]}
\left(
\overline{\boldsymbol{\Pi}}(u_{\mathrm{org}})
\mathbf x_i
\right)_c
=
\widehat c_i
\qquad
\text{for every }i.
\label{eq:linear-deep-Q-target-classifier}
\end{equation}
Thus the homogeneous predictor at
\(u_{\mathrm{org}}\) induces exactly the
same classifier as the full target
\(\mathbf T\) on the training set.

Now couple the exact gradient flows for
different initialization scales by
\begin{equation}
\mathbf W_\ell^{(\sigma)}(0)
=
\sigma\mathbf G_\ell,
\qquad
\ell=1,\ldots,L,
\label{eq:linear-deep-organization-coupling}
\end{equation}
and define the corresponding organization
time in the rescaled gradient-flow variable
\(\tau\) by
\begin{equation}
\tau_{\mathrm{org}}^{(\sigma,H)}
=
\sigma^{-(L-2)}
H^{-(L-2)/2}
u_{\mathrm{org}}.
\label{eq:linear-deep-organization-tau}
\end{equation}

For every sufficiently large \(H\) and every
realization in \(\mathcal E_H\), the exact
gradient flow satisfies, for all sufficiently
small \(\sigma>0\),
\begin{equation}
\arg\max_{c\in[C]}
\left(
\mathbf M^{(\sigma)}
\left(
\tau_{\mathrm{org}}^{(\sigma,H)}
\right)
\mathbf x_i
\right)_c
=
\widehat c_i
\qquad
\text{for every }i.
\label{eq:linear-deep-organization-exact-classifier}
\end{equation}
Consequently,
\begin{equation}
\Acc\left(
\mathbf M^{(\sigma)}
\left(
\tau_{\mathrm{org}}^{(\sigma,H)}
\right)
\right)
=
A_{\mathbf T}
:=
\Acc(\mathbf T).
\label{eq:linear-deep-organization-accuracy}
\end{equation}

Moreover, the entire organization interval
is microscopic. For every fixed sufficiently
large \(H\) and every realization in
\(\mathcal E_H\),
\begin{equation}
\sup_{
0\le\tau\le\tau_{\mathrm{org}}^{(\sigma,H)}
}
\left\|
\mathbf M^{(\sigma)}(\tau)
\right\|_{\mathrm F}
\longrightarrow0
\qquad
\text{as }\sigma\downarrow0,
\label{eq:linear-deep-organization-microscopic}
\end{equation}
and
\begin{equation}
\sup_{
0\le\tau\le\tau_{\mathrm{org}}^{(\sigma,H)}
}
\left|
\mathcal L\left(
\mathbf M^{(\sigma)}(\tau)
\right)
-
\mathcal L\left(
\mathbf M^{(\sigma)}(0)
\right)
\right|
\longrightarrow0.
\label{eq:linear-deep-organization-flat-loss}
\end{equation}

Equivalently, for every fixed
\(\varepsilon>0\),
\begin{equation}
\begin{aligned}
\lim_{H\to\infty}
\liminf_{\sigma\downarrow0}
\mathbb P\Bigg(
&
\arg\max_{c\in[C]}
\left(
\mathbf M^{(\sigma)}
\left(
\tau_{\mathrm{org}}^{(\sigma,H)}
\right)
\mathbf x_i
\right)_c
=
\widehat c_i
\quad\text{for every }i,
\\
&
\sup_{
0\le\tau\le\tau_{\mathrm{org}}^{(\sigma,H)}
}
\left\|
\mathbf M^{(\sigma)}(\tau)
\right\|_{\mathrm F}
<
\varepsilon,
\\
&
\sup_{
0\le\tau\le\tau_{\mathrm{org}}^{(\sigma,H)}
}
\left|
\mathcal L\left(
\mathbf M^{(\sigma)}(\tau)
\right)
-
\mathcal L\left(
\mathbf M^{(\sigma)}(0)
\right)
\right|
<
\varepsilon
\Bigg)
=
1.
\end{aligned}
\label{eq:linear-deep-organization-sequential-limit}
\end{equation}

Hence, in the sequential limit
\(\sigma\downarrow0\) followed by
\(H\to\infty\), the gradient flow reaches
the classifier of the full target while
the end-to-end predictor remains microscopic
and the loss remains asymptotically unchanged
throughout the preceding organization stage.
\end{theorem}

\begin{proof}
The proof is based on a short-time expansion of the
homogeneous predictor \(\overline{\boldsymbol{\Pi}}(u)\).
The key observation is the decomposition
\begin{equation}
\overline{\boldsymbol{\Pi}}(u)
=
\underbrace{\overline{\boldsymbol{\Pi}}(0)}_{\text{random initialization}}
+
\underbrace{uL\mathbf T}_{\text{coherent target drift}}
+
\underbrace{
u\left(\overline{\boldsymbol{\Pi}}'(0)-L\mathbf T\right)
}_{\text{drift fluctuations}}
+
\underbrace{\boldsymbol{\mathcal R}(u)}_{\text{nonlinear remainder}},
\label{eq:linear-deep-proof-strategy}
\end{equation}
where \(\boldsymbol{\mathcal R}(u)\) is the
second-order Taylor remainder. We will show that,
on a high-probability event,
\begin{equation}
\|\boldsymbol{\mathcal R}(u)\|_{\mathrm{op}}
\le
C u^2
\end{equation}
on a sufficiently short interval, with a constant
\(C\) independent of the width \(H\).

The four terms in
\eqref{eq:linear-deep-proof-strategy}
play different roles.
The initial predictor is random and centered,
whereas the expected first-order drift is
\(L\mathbf T\).
We will show that, at large width, the initial
random scores are of order
\(\ell_H^{L/2}/\sqrt H\), while fluctuations
of the first-order drift vanish on every
training-set score comparison.

The organization time
\begin{equation}
u_{\mathrm{org}}
=
b_H\frac{\ell_H^{L/2}}{\sqrt H}
\end{equation}
is chosen to satisfy two requirements.
Since \(b_H\to\infty\), the coherent drift
dominates the initial random scores.
At the same time, \(u_{\mathrm{org}}\to0\),
so the nonlinear remainder is negligible
compared with the first-order contribution.

To make this argument precise, for every
training example \(i\) and competing class
\(c\neq\widehat c_i\), define
\begin{equation}
\mathbf d_{ic}
=
\mathbf e_{\widehat c_i}
-
\mathbf e_c.
\label{eq:linear-deep-score-vector}
\end{equation}
Then
\(\mathbf d_{ic}^{\top}\overline{\boldsymbol{\Pi}}(u)\mathbf x_i\)
is the difference between the score assigned
to the target-predicted class and that assigned
to class \(c\).
By definition of the target decision gap,
\begin{equation}
\|\mathbf d_{ic}\|_2
=
\sqrt2,
\qquad
\mathbf d_{ic}^{\top}\mathbf T\mathbf x_i
\ge
\delta_{\mathbf T}
\ge
\delta_0.
\label{eq:linear-deep-score-vector-properties}
\end{equation}

Our main goal is to establish
\begin{equation}
\mathbf d_{ic}^{\top}
\overline{\boldsymbol{\Pi}}(u_{\mathrm{org}})
\mathbf x_i
=
Lu_{\mathrm{org}}
\mathbf d_{ic}^{\top}
\mathbf T\mathbf x_i
+
o(u_{\mathrm{org}})
\label{eq:linear-deep-Q-first-order-score}
\end{equation}
uniformly over all training examples and
competing classes, on events whose probability
tends to one.
The positive target decision gap will then give
\begin{equation}
\mathbf d_{ic}^{\top}
\overline{\boldsymbol{\Pi}}(u_{\mathrm{org}})
\mathbf x_i
>
0
\label{eq:linear-deep-positive-score-gap}
\end{equation}
simultaneously for every \(i\) and
\(c\neq\widehat c_i\).
This proves that the homogeneous predictor
induces the classifier of the full target.

Finally, we will transfer this conclusion
to the exact gradient flow using
Lemma~\ref{lem:linear-deep-near-origin-reduction}
and prove that the entire organization interval
remains microscopic, with asymptotically
vanishing loss variation.

\smallskip
\noindent
\emph{Step 1: the expected first-order drift is
\(L\mathbf T\).}

Define
\begin{equation}
\overline{\boldsymbol{\Pi}}_0
:=
\overline{\boldsymbol{\Pi}}(0)
=
\overline{\mathbf G}_L
\cdots
\overline{\mathbf G}_1,
\qquad
\overline{\boldsymbol{\Pi}}_1
:=
\left.
\frac{d\overline{\boldsymbol{\Pi}}}{du}
\right|_{u=0}.
\label{eq:linear-deep-Q0-Q1}
\end{equation}

Differentiating the end-to-end product
\(\overline{\boldsymbol{\Pi}}(u)\) one factor at a time
and substituting the homogeneous flow
\eqref{eq:linear-deep-width-normalized-flow}
gives
\begin{equation}
\overline{\boldsymbol{\Pi}}_1
=
\sum_{\ell=1}^{L}
\left(
\overline{\mathbf G}_{L:\ell+1}
\overline{\mathbf G}_{L:\ell+1}^{\top}
\right)
\mathbf T
\left(
\overline{\mathbf G}_{\ell-1:1}^{\top}
\overline{\mathbf G}_{\ell-1:1}
\right),
\label{eq:linear-deep-Q1}
\end{equation}
with empty products interpreted as identity
matrices of the appropriate dimensions.

Each summand contains two Gram matrices,
one on either side of \(\mathbf T\).
Since these matrices involve disjoint groups
of Gaussian layers, they are independent.
Moreover, each normalized Gaussian layer
preserves squared norms in expectation.
Successive conditioning therefore yields
\begin{equation}
\mathbb E
\left[
\overline{\mathbf G}_{L:\ell+1}
\overline{\mathbf G}_{L:\ell+1}^{\top}
\right]
=
\mathbf I_C,
\qquad
\mathbb E
\left[
\overline{\mathbf G}_{\ell-1:1}^{\top}
\overline{\mathbf G}_{\ell-1:1}
\right]
=
\mathbf I_p.
\label{eq:linear-deep-product-Gram-mean}
\end{equation}

By independence, the expectation of each
summand in \eqref{eq:linear-deep-Q1}
is exactly \(\mathbf T\).
Since there are \(L\) summands,
\begin{equation}
\mathbb E\,\overline{\boldsymbol{\Pi}}_1
=
L\mathbf T.
\label{eq:linear-deep-Q1-mean}
\end{equation}

Thus gradient flow generates a coherent
first-order component in the direction
of the full target.
We next show that, when evaluated on
training-set score comparisons, this component
concentrates around its expectation and
eventually dominates the initial random predictor.

\smallskip
\noindent
\emph{Step 2: concentration of the initial
predictor and the first-order drift.}

The main technical ingredient is a concentration
estimate for scalar bilinear forms of Gaussian
product Gram matrices.
Importantly, we do not need these Gram matrices
to converge to the identity in operator norm.
We only need to control their action on
particular vectors appearing in the score
comparisons.

\smallskip
\noindent
\emph{A Gaussian-chain estimate.}

Fix an integer \(m\ge1\).
Let
\begin{equation}
\mathbf X_1
\in
\mathbb R^{H\times d},
\qquad
\mathbf X_j
\in
\mathbb R^{H\times H},
\quad
2\le j\le m,
\label{eq:linear-deep-chain-shapes}
\end{equation}
be independent matrices with entries
distributed as \(\mathcal N(0,1/H)\),
and define
\begin{equation}
\mathbf B_m
=
\mathbf X_m\cdots\mathbf X_1.
\label{eq:linear-deep-chain-product}
\end{equation}

We claim that for any deterministic
\(\mathbf u,\mathbf v\in\mathbb R^d\),
\begin{equation}
\mathbb E
\left|
\mathbf u^{\top}
\left(
\mathbf B_m^{\top}\mathbf B_m
-
\mathbf I_d
\right)
\mathbf v
\right|^2
\le
\frac{C_m}{H}
\|\mathbf u\|_2^2
\|\mathbf v\|_2^2,
\label{eq:linear-deep-one-sided-Gram-concentration}
\end{equation}
where \(C_m<\infty\) depends only on \(m\).

To prove this estimate, propagate the test
vectors through the Gaussian chain:
\begin{equation}
\mathbf z_0=\mathbf u,
\qquad
\mathbf w_0=\mathbf v,
\qquad
\mathbf z_j=\mathbf X_j\mathbf z_{j-1},
\qquad
\mathbf w_j=\mathbf X_j\mathbf w_{j-1}.
\label{eq:linear-deep-chain-vectors}
\end{equation}
Let
\begin{equation}
S_j
=
\mathbf z_j^{\top}\mathbf w_j.
\label{eq:linear-deep-chain-inner-product}
\end{equation}

Conditional on the vectors entering
layer \(j\), the Gaussian second-moment
identity gives
\begin{equation}
\mathbb E
\left[
S_j
\mid
\mathbf z_{j-1},\mathbf w_{j-1}
\right]
=
S_{j-1}.
\label{eq:linear-deep-chain-martingale-mean}
\end{equation}
Thus \((S_j)_{j=0}^m\) is a martingale:
each Gaussian layer preserves the inner
product in expectation.

The Gaussian fourth-moment identity gives
its conditional variance,
\begin{equation}
\operatorname{Var}
\left(
S_j
\mid
\mathbf z_{j-1},\mathbf w_{j-1}
\right)
=
\frac1H
\left(
\|\mathbf z_{j-1}\|_2^2
\|\mathbf w_{j-1}\|_2^2
+
S_{j-1}^2
\right).
\label{eq:linear-deep-chain-martingale-var}
\end{equation}

To control this variance, observe that
for every deterministic vector \(\mathbf z\),
\begin{equation}
\mathbb E
\left[
\|\mathbf X_j\mathbf z\|_2^4
\,\middle|\,
\mathbf z
\right]
=
\left(
1+\frac2H
\right)
\|\mathbf z\|_2^4.
\label{eq:linear-deep-chain-fourth-norm}
\end{equation}
Indeed, conditionally on \(\mathbf z\),
the squared norm
\(\|\mathbf X_j\mathbf z\|_2^2\)
has the distribution
\(\|\mathbf z\|_2^2\chi_H^2/H\).

Since the number of layers \(m\) is fixed,
iteration gives
\begin{equation}
\max_{0\le j\le m}
\mathbb E\|\mathbf z_j\|_2^4
\le
C_m\|\mathbf u\|_2^4,
\qquad
\max_{0\le j\le m}
\mathbb E\|\mathbf w_j\|_2^4
\le
C_m\|\mathbf v\|_2^4.
\label{eq:linear-deep-chain-fourth-norm-bound}
\end{equation}

Martingale increments are orthogonal in
\(L^2\).
Using the conditional variance formula,
Cauchy--Schwarz, and the fourth-moment
bounds above, we obtain
\begin{equation}
\begin{aligned}
\mathbb E|S_m-S_0|^2
&=
\sum_{j=1}^m
\mathbb E
\operatorname{Var}
\left(
S_j
\mid
\mathbf z_{j-1},\mathbf w_{j-1}
\right)
\\
&\le
\frac{C_m}{H}
\|\mathbf u\|_2^2
\|\mathbf v\|_2^2.
\end{aligned}
\label{eq:linear-deep-chain-martingale-bound}
\end{equation}
Finally,
\begin{equation}
S_m-S_0
=
\mathbf u^{\top}
\left(
\mathbf B_m^{\top}\mathbf B_m
-
\mathbf I_d
\right)
\mathbf v,
\label{eq:linear-deep-chain-martingale-identification}
\end{equation}
which proves
\eqref{eq:linear-deep-one-sided-Gram-concentration}.

Thus a product of finitely many normalized
Gaussian matrices preserves any fixed
bilinear form up to an \(L^2\)-error
of order \(H^{-1/2}\).

\smallskip
\noindent
\emph{Fluctuations of the first-order drift.}

For each layer index \(\ell\), define
\begin{equation}
\mathbf L_\ell
=
\overline{\mathbf G}_{L:\ell+1}
\overline{\mathbf G}_{L:\ell+1}^{\top},
\qquad
\mathbf R_\ell
=
\overline{\mathbf G}_{\ell-1:1}^{\top}
\overline{\mathbf G}_{\ell-1:1}.
\label{eq:linear-deep-left-right-Grams}
\end{equation}
Then
\begin{equation}
\overline{\boldsymbol{\Pi}}_1
=
\sum_{\ell=1}^L
\mathbf L_\ell\mathbf T\mathbf R_\ell.
\end{equation}

The Gaussian-chain estimate applies
directly to \(\mathbf R_\ell\), and to
\(\mathbf L_\ell\) after transposing
the corresponding Gaussian product.
Empty products contribute identity matrices
and have zero fluctuations.

Consequently, for deterministic vectors
of appropriate dimensions,
\begin{equation}
\mathbb E
\left|
\mathbf u^{\top}
(\mathbf L_\ell-\mathbf I_C)
\mathbf v
\right|^2
\le
\frac{C_L}{H}
\|\mathbf u\|_2^2
\|\mathbf v\|_2^2,
\label{eq:linear-deep-left-Gram-concentration}
\end{equation}
and
\begin{equation}
\mathbb E
\left|
\mathbf u^{\top}
(\mathbf R_\ell-\mathbf I_p)
\mathbf v
\right|^2
\le
\frac{C_L}{H}
\|\mathbf u\|_2^2
\|\mathbf v\|_2^2.
\label{eq:linear-deep-right-Gram-concentration}
\end{equation}

We also require a second-moment bound
for the vector \(\mathbf R_\ell\mathbf x\).
Applying
\eqref{eq:linear-deep-right-Gram-concentration}
coordinatewise gives
\begin{equation}
\begin{aligned}
\mathbb E
\left\|
(\mathbf R_\ell-\mathbf I_p)
\mathbf x
\right\|_2^2
&=
\sum_{j=1}^p
\mathbb E
\left|
\mathbf e_j^{\top}
(\mathbf R_\ell-\mathbf I_p)
\mathbf x
\right|^2
\\
&\le
C_L\frac{p}{H}
\|\mathbf x\|_2^2.
\end{aligned}
\label{eq:linear-deep-right-Gram-vector-bound}
\end{equation}
Since \(p/H=O(1)\), it follows that
\begin{equation}
\mathbb E
\|\mathbf R_\ell\mathbf x\|_2^2
\le
C_L\|\mathbf x\|_2^2,
\label{eq:linear-deep-right-Gram-vector-second-moment}
\end{equation}
where the constant may also depend on
the uniform upper bound for \(p/H\).

Now fix deterministic vectors
\(\mathbf a\in\mathbb R^C\)
and \(\mathbf x\in\mathbb R^p\).
For each summand of \(\overline{\boldsymbol{\Pi}}_1\),
we have the decomposition
\begin{equation}
\mathbf L_\ell\mathbf T\mathbf R_\ell
-
\mathbf T
=
(\mathbf L_\ell-\mathbf I_C)
\mathbf T\mathbf R_\ell
+
\mathbf T(\mathbf R_\ell-\mathbf I_p).
\label{eq:linear-deep-Q1-summand-decomposition}
\end{equation}

The left and right Gram matrices are
independent.
We may therefore condition on
\(\mathbf R_\ell\) when estimating
the first term.
Using
\eqref{eq:linear-deep-left-Gram-concentration}
and
\eqref{eq:linear-deep-right-Gram-vector-second-moment},
we obtain
\begin{equation}
\begin{aligned}
&
\mathbb E
\left|
\mathbf a^{\top}
(\mathbf L_\ell-\mathbf I_C)
\mathbf T\mathbf R_\ell
\mathbf x
\right|^2
\\
&\qquad\le
\frac{C_L}{H}
\|\mathbf a\|_2^2
\mathbb E
\|\mathbf T\mathbf R_\ell\mathbf x\|_2^2
\\
&\qquad\le
\frac{C_L}{H}
\|\mathbf a\|_2^2
\|\mathbf T\|_{\mathrm{op}}^2
\|\mathbf x\|_2^2.
\end{aligned}
\label{eq:linear-deep-Q1-left-fluctuation}
\end{equation}
Likewise,
\eqref{eq:linear-deep-right-Gram-concentration}
gives
\begin{equation}
\mathbb E
\left|
\mathbf a^{\top}
\mathbf T
(\mathbf R_\ell-\mathbf I_p)
\mathbf x
\right|^2
\le
\frac{C_L}{H}
\|\mathbf a\|_2^2
\|\mathbf T\|_{\mathrm{op}}^2
\|\mathbf x\|_2^2.
\label{eq:linear-deep-Q1-right-fluctuation}
\end{equation}

Since \(L\) is fixed, summing the
\(L\) contributions to
\(\overline{\boldsymbol{\Pi}}_1-L\mathbf T\)
and using the triangle inequality
in \(L^2\) yields
\begin{equation}
\left\|
\mathbf a^{\top}
(\overline{\boldsymbol{\Pi}}_1-L\mathbf T)
\mathbf x
\right\|_{L^2}
\le
\frac{C_L}{\sqrt H}
\|\mathbf a\|_2
\|\mathbf T\|_{\mathrm{op}}
\|\mathbf x\|_2.
\label{eq:linear-deep-Q1-L2}
\end{equation}

Thus, for any fixed pair of test vectors,
the first-order drift concentrates
around \(L\mathbf T\).

\smallskip
\noindent
\emph{Fluctuations of the initial predictor.}

The initial predictor \(\overline{\boldsymbol{\Pi}}_0\)
is a product of independent centered
Gaussian matrices.
Repeated conditioning through the layers
gives
\begin{equation}
\mathbb E
\left|
\mathbf a^{\top}\overline{\boldsymbol{\Pi}}_0\mathbf x
\right|^2
=
\frac1H
\|\mathbf a\|_2^2
\|\mathbf x\|_2^2.
\label{eq:linear-deep-Q0-L2}
\end{equation}

We now have \(L^2\)-estimates for both
the initial predictor and the fluctuations
of its first-order drift.
To control all training-set score comparisons
simultaneously, we upgrade these estimates
using Gaussian hypercontractivity.

\smallskip
\noindent
\emph{Uniform concentration over score comparisons.}

The scalar random variables
\(\mathbf a^{\top}\overline{\boldsymbol{\Pi}}_0\mathbf x\)
and
\(\mathbf a^{\top}(\overline{\boldsymbol{\Pi}}_1-L\mathbf T)\mathbf x\)
are polynomials in independent standard
Gaussian variables, of degrees at most
\(L\) and \(2L-2\), respectively.

Gaussian hypercontractivity gives,
for any Gaussian polynomial \(F\)
of degree at most \(d\),
\begin{equation}
\|F\|_{L^q}
\le
(q-1)^{d/2}
\|F\|_{L^2},
\qquad
q\ge2.
\end{equation}
In our setting, this yields
\begin{equation}
\left\|
\mathbf a^{\top}\overline{\boldsymbol{\Pi}}_0\mathbf x
\right\|_{L^q}
\le
(q-1)^{L/2}
\left\|
\mathbf a^{\top}\overline{\boldsymbol{\Pi}}_0\mathbf x
\right\|_{L^2},
\label{eq:linear-deep-Q0-hypercontractive}
\end{equation}
and
\begin{equation}
\left\|
\mathbf a^{\top}
(\overline{\boldsymbol{\Pi}}_1-L\mathbf T)\mathbf x
\right\|_{L^q}
\le
(q-1)^{L-1}
\left\|
\mathbf a^{\top}
(\overline{\boldsymbol{\Pi}}_1-L\mathbf T)\mathbf x
\right\|_{L^2}.
\label{eq:linear-deep-Q1-hypercontractive}
\end{equation}

There are at most \(NC\) training-set
score comparisons.
We choose the moment order
\begin{equation}
q_H
=
\max\{2,2\ell_H\},
\qquad
\ell_H=\log(2NCH).
\label{eq:linear-deep-hypercontractive-q}
\end{equation}
This logarithmic moment order allows us
to take a union bound over all comparisons
while retaining an \(H^{-1/2}\) concentration
rate, up to logarithmic factors.

Indeed, Markov's inequality gives
\begin{equation}
\mathbb P
\left(
|F|>
e\|F\|_{L^{q_H}}
\right)
\le
e^{-q_H}.
\end{equation}
Combining this with
\eqref{eq:linear-deep-Q0-L2},
\eqref{eq:linear-deep-Q1-L2},
and the hypercontractive estimates,
then taking a union bound, shows that,
with probability tending to one,
\begin{equation}
\max_i
\max_{c\neq\widehat c_i}
\left|
\mathbf d_{ic}^{\top}
\overline{\boldsymbol{\Pi}}_0\mathbf x_i
\right|
\le
C_L R
\frac{\ell_H^{L/2}}{\sqrt H},
\label{eq:linear-deep-Q0-uniform}
\end{equation}
and
\begin{equation}
\max_i
\max_{c\neq\widehat c_i}
\left|
\mathbf d_{ic}^{\top}
(\overline{\boldsymbol{\Pi}}_1-L\mathbf T)
\mathbf x_i
\right|
\le
C_L R\|\mathbf T\|_{\mathrm{op}}
\frac{\ell_H^{L-1}}{\sqrt H}.
\label{eq:linear-deep-Q1-uniform}
\end{equation}
The constants depend only on the fixed
depth and the uniform aspect-ratio bounds.

The joint failure probability is at most
\begin{equation}
2NCe^{-q_H}
\le
2NCe^{-2\ell_H}
=
\frac{1}{2NCH^2}
\longrightarrow0.
\end{equation}
By
\eqref{eq:linear-deep-organization-growth-condition},
the right-hand side of
\eqref{eq:linear-deep-Q1-uniform}
tends to zero.

We have therefore proved that the random
initial scores are uniformly of order
\(\ell_H^{L/2}/\sqrt H\), and that
the first-order drift converges to
\(L\mathbf T\) on all training-set
score comparisons.

\smallskip
\noindent
\emph{Step 3: uniform control of the
nonlinear remainder.}

We next justify the short-time Taylor
expansion in
\eqref{eq:linear-deep-proof-strategy}
with a remainder bound independent of \(H\).

Under the assumptions
\(p/H=O(1)\) and \(C/H=O(1)\),
standard Gaussian operator-norm bounds
imply that there exists a deterministic
constant \(K_G<\infty\) such that
\begin{equation}
\mathbb P
\left(
\max_{1\le\ell\le L}
\|\overline{\mathbf G}_\ell\|_{\mathrm{op}}
\le K_G
\right)
\longrightarrow1.
\label{eq:linear-deep-initial-op-bound}
\end{equation}
We work on this event.

Because \(L\) is fixed and
\(\|\mathbf T\|_{\mathrm{op}}=O(1)\),
the homogeneous vector field is
uniformly bounded whenever all
factor operator norms are bounded.
In particular, while
\begin{equation}
\max_\ell
\|\overline{\mathbf Z}_\ell(u)\|_{\mathrm{op}}
\le
2K_G,
\end{equation}
the factor equations imply
\begin{equation}
\left\|
\frac{d\overline{\mathbf Z}_\ell}{du}
\right\|_{\mathrm{op}}
\le
(2K_G)^{L-1}
\|\mathbf T\|_{\mathrm{op}}.
\label{eq:linear-deep-local-derivative-bound}
\end{equation}

The right-hand side is bounded
independently of \(H\).
A first-exit argument therefore gives
a constant \(u_0>0\), independent of \(H\),
such that the homogeneous flow exists
on \([0,u_0]\) and
\begin{equation}
\max_\ell
\sup_{0\le u\le u_0}
\|\overline{\mathbf Z}_\ell(u)\|_{\mathrm{op}}
\le
2K_G.
\label{eq:linear-deep-local-factor-bound}
\end{equation}
Integrating the derivative bound also gives
\begin{equation}
\|\overline{\mathbf Z}_\ell(u)
-
\overline{\mathbf G}_\ell\|_{\mathrm{op}}
\le
K_1u,
\qquad
0\le u\le u_0,
\label{eq:linear-deep-factor-small-u}
\end{equation}
for a constant \(K_1<\infty\)
independent of \(H\).

Differentiating \(\overline{\boldsymbol{\Pi}}(u)\)
expresses \(\overline{\boldsymbol{\Pi}}'(u)\) as a
polynomial in the normalized factors
and one copy of \(\mathbf T\).
Since the factor operator norms remain
uniformly bounded on \([0,u_0]\),
this expression is Lipschitz in
the factors, with a constant independent
of the width.

Using
\eqref{eq:linear-deep-factor-small-u},
we therefore obtain
\begin{equation}
\|\overline{\boldsymbol{\Pi}}'(u)-\overline{\boldsymbol{\Pi}}_1\|_{\mathrm{op}}
\le
K_2u,
\qquad
0\le u\le u_0,
\label{eq:linear-deep-Qprime-Lipschitz}
\end{equation}
for some \(K_2<\infty\)
independent of \(H\).

Integrating this estimate gives
\begin{equation}
\overline{\boldsymbol{\Pi}}(u)
=
\overline{\boldsymbol{\Pi}}_0
+
u\overline{\boldsymbol{\Pi}}_1
+
\boldsymbol{\mathcal R}(u),
\qquad
\|\boldsymbol{\mathcal R}(u)\|_{\mathrm{op}}
\le
\frac{K_2}{2}u^2.
\label{eq:linear-deep-Q-Taylor}
\end{equation}
This establishes the remainder bound
announced at the beginning of the proof.

Finally, since \(u_{\mathrm{org}}\to0\),
we have \(u_{\mathrm{org}}<u_0\)
for all sufficiently large \(H\).
Thus the homogeneous flow exists throughout
the organization interval, and the Taylor
expansion applies at \(u=u_{\mathrm{org}}\).

\smallskip
\noindent
\emph{Step 4: the target-directed drift
determines the classifier.}

Let \(\mathcal E_H\) be the intersection
of the operator-norm event
\eqref{eq:linear-deep-initial-op-bound}
and the two uniform concentration events
\eqref{eq:linear-deep-Q0-uniform}
and
\eqref{eq:linear-deep-Q1-uniform}.
The preceding estimates imply
\begin{equation}
\mathbb P(\mathcal E_H)
\longrightarrow1.
\label{eq:linear-deep-good-event}
\end{equation}

Fix a realization in \(\mathcal E_H\),
a training example \(i\), and a
competing class \(c\neq\widehat c_i\).

Using the Taylor expansion
\eqref{eq:linear-deep-Q-Taylor},
and adding and subtracting
\(Lu_{\mathrm{org}}\mathbf T\),
we obtain
\begin{equation}
\begin{aligned}
\mathbf d_{ic}^{\top}
\overline{\boldsymbol{\Pi}}(u_{\mathrm{org}})
\mathbf x_i
&=
Lu_{\mathrm{org}}
\mathbf d_{ic}^{\top}
\mathbf T\mathbf x_i
+
\mathbf d_{ic}^{\top}
\overline{\boldsymbol{\Pi}}_0\mathbf x_i
\\
&\qquad+
u_{\mathrm{org}}
\mathbf d_{ic}^{\top}
(\overline{\boldsymbol{\Pi}}_1-L\mathbf T)
\mathbf x_i
+
\mathbf d_{ic}^{\top}
\boldsymbol{\mathcal R}(u_{\mathrm{org}})
\mathbf x_i.
\end{aligned}
\label{eq:linear-deep-margin-decomposition}
\end{equation}

We now compare the three error terms
with the target-directed contribution.

First, the target term satisfies
\begin{equation}
Lu_{\mathrm{org}}
\mathbf d_{ic}^{\top}
\mathbf T\mathbf x_i
\ge
L\delta_0u_{\mathrm{org}}.
\label{eq:linear-deep-target-drift-margin}
\end{equation}

Second, the initial random score is
negligible compared with
\(u_{\mathrm{org}}\).
Indeed, by
\eqref{eq:linear-deep-Q0-uniform}
and the definition of \(u_{\mathrm{org}}\),
\begin{equation}
\frac{1}{u_{\mathrm{org}}}
\max_i
\max_{c\neq\widehat c_i}
\left|
\mathbf d_{ic}^{\top}
\overline{\boldsymbol{\Pi}}_0\mathbf x_i
\right|
\le
\frac{C_LR}{b_H}
\longrightarrow0.
\label{eq:linear-deep-Q0-negligible}
\end{equation}
This is precisely where
\(b_H\to\infty\) is used.

Third, the first-order drift
fluctuations vanish uniformly:
\begin{equation}
\max_i
\max_{c\neq\widehat c_i}
\left|
\mathbf d_{ic}^{\top}
(\overline{\boldsymbol{\Pi}}_1-L\mathbf T)
\mathbf x_i
\right|
\le
C_LR\|\mathbf T\|_{\mathrm{op}}
\frac{\ell_H^{L-1}}{\sqrt H}
\longrightarrow0.
\label{eq:linear-deep-Q1-negligible}
\end{equation}
Here we use
\eqref{eq:linear-deep-organization-growth-condition}.

Finally, the nonlinear remainder
is quadratic in \(u_{\mathrm{org}}\).
Using
\eqref{eq:linear-deep-Q-Taylor}
and the bounds on
\(\|\mathbf d_{ic}\|_2\)
and \(\|\mathbf x_i\|_2\),
we have
\begin{equation}
\begin{aligned}
\left|
\mathbf d_{ic}^{\top}
\boldsymbol{\mathcal R}(u_{\mathrm{org}})
\mathbf x_i
\right|
&\le
\|\mathbf d_{ic}\|_2
\|\mathbf x_i\|_2
\|\boldsymbol{\mathcal R}(u_{\mathrm{org}})\|_{\mathrm{op}}
\\
&\le
\frac{K_2R}{\sqrt2}
u_{\mathrm{org}}^2
=
o(u_{\mathrm{org}}).
\end{aligned}
\label{eq:linear-deep-remainder-negligible}
\end{equation}

Combining these estimates with
\eqref{eq:linear-deep-margin-decomposition}
establishes
\eqref{eq:linear-deep-Q-first-order-score},
uniformly over all training examples
and competing classes, on
\(\mathcal E_H\).

In particular,
\begin{equation}
\mathbf d_{ic}^{\top}
\overline{\boldsymbol{\Pi}}(u_{\mathrm{org}})
\mathbf x_i
\ge
L\delta_0u_{\mathrm{org}}
-
o(u_{\mathrm{org}})
>
0
\label{eq:linear-deep-margin-final}
\end{equation}
for all sufficiently large \(H\),
simultaneously for all \(i\) and
\(c\neq\widehat c_i\).

Every target-class score therefore
exceeds all competing scores.
This proves
\eqref{eq:linear-deep-Q-target-classifier}:
the homogeneous predictor at
\(u_{\mathrm{org}}\) induces exactly
the same classifier as \(\mathbf T\).

\smallskip
\noindent
\emph{Step 5: transfer to the exact gradient
flow and uniform microscopicity.}

The preceding steps establish the
classification result for the homogeneous
system.
We now transfer it to the exact gradient
flow with small but positive initialization
scale \(\sigma\).

Recall the corresponding homogeneous
slow time
\begin{equation}
s_{\mathrm{org}}
=
H^{-(L-2)/2}u_{\mathrm{org}}.
\label{eq:linear-deep-organization-s}
\end{equation}
On \(\mathcal E_H\), Step 3 guarantees
that the width-normalized homogeneous
flow exists on \([0,u_0]\).
Since \(u_{\mathrm{org}}<u_0\)
for all sufficiently large \(H\),
the original homogeneous flow
\eqref{eq:linear-deep-homogeneous-flow}
exists on a compact interval containing
\([0,s_{\mathrm{org}}]\).

For every fixed sufficiently large \(H\)
and every realization in \(\mathcal E_H\),
we may therefore apply
Lemma~\ref{lem:linear-deep-near-origin-reduction}.
Converting between the \(s\)- and
\(u\)-rescalings gives
\begin{equation}
\sup_{0\le u\le u_{\mathrm{org}}}
\Bigg\|
\frac{
\mathbf M^{(\sigma)}
\left(
\sigma^{-(L-2)}
H^{-(L-2)/2}u
\right)
}{
\sigma^L H^{L/2}
}
-
\overline{\boldsymbol{\Pi}}(u)
\Bigg\|_{\mathrm F}
\longrightarrow0
\qquad
\text{as }\sigma\downarrow0.
\label{eq:linear-deep-organization-transfer}
\end{equation}

In particular, at \(u=u_{\mathrm{org}}\),
the normalized exact predictor converges
to \(\overline{\boldsymbol{\Pi}}(u_{\mathrm{org}})\).
By
\eqref{eq:linear-deep-margin-final},
all score comparisons of the limiting
predictor are strictly positive.
Since the training set and the number
of classes are finite at fixed \(H\),
these strict inequalities persist for
all sufficiently small \(\sigma\).

Multiplication by the positive scalar
\(\sigma^L H^{L/2}\) does not affect
predicted classes.
Consequently,
\begin{equation}
\arg\max_{c\in[C]}
\left(
\mathbf M^{(\sigma)}
\left(
\tau_{\mathrm{org}}^{(\sigma,H)}
\right)
\mathbf x_i
\right)_c
=
\widehat c_i
\qquad
\text{for every }i.
\end{equation}
This proves
\eqref{eq:linear-deep-organization-exact-classifier}
and
\eqref{eq:linear-deep-organization-accuracy}.

It remains to establish microscopicity
and vanishing loss variation uniformly
over the entire organization interval.

By Step 3, the homogeneous factors
are bounded on \([0,u_{\mathrm{org}}]\).
Their product \(\overline{\boldsymbol{\Pi}}(u)\)
is therefore bounded on this interval.
For fixed \(H\), the uniform convergence
\eqref{eq:linear-deep-organization-transfer}
then implies
\begin{equation}
\sup_{
0\le\tau\le
\tau_{\mathrm{org}}^{(\sigma,H)}
}
\|\mathbf M^{(\sigma)}(\tau)\|_{\mathrm F}
=
O(\sigma^L)
\longrightarrow0
\qquad
\text{as }\sigma\downarrow0.
\label{eq:linear-deep-organization-microscopic-proof}
\end{equation}
The implicit constant may depend on
the fixed problem and initialization
realization, but not on \(\sigma\).

This proves that the entire organization
interval remains microscopic, rather
than only its endpoint.

Finally, recall that
\begin{equation}
\mathcal L(\mathbf M)
=
\frac{\gamma}{2}
\|\mathbf M-\mathbf T\|_{\mathrm F}^2.
\end{equation}
Writing
\(\mathbf M_0=\mathbf M^{(\sigma)}(0)\),
the difference in loss satisfies
\begin{equation}
\mathcal L(\mathbf M)
-
\mathcal L(\mathbf M_0)=
-\gamma
\left\langle
\mathbf T,
\mathbf M-\mathbf M_0
\right\rangle_{\mathrm F}
+
\frac{\gamma}{2}
\left(
\|\mathbf M\|_{\mathrm F}^2
-
\|\mathbf M_0\|_{\mathrm F}^2
\right).
\label{eq:linear-deep-organization-loss-difference}
\end{equation}

By
\eqref{eq:linear-deep-organization-microscopic-proof},
both
\(\mathbf M^{(\sigma)}(\tau)\)
and \(\mathbf M_0\) have Frobenius norm
\(O(\sigma^L)\), uniformly for
\(0\le\tau\le\tau_{\mathrm{org}}^{(\sigma,H)}\).
Since \(\gamma\) and \(\mathbf T\)
are fixed in the inner limit,
Cauchy--Schwarz yields
\begin{equation}
\sup_{
0\le\tau\le
\tau_{\mathrm{org}}^{(\sigma,H)}
}
\Big|
\mathcal L
\left(
\mathbf M^{(\sigma)}(\tau)
\right)
-
\mathcal L
\left(
\mathbf M^{(\sigma)}(0)
\right)
\Big|
=
O(\sigma^L)
\longrightarrow0.
\end{equation}
This establishes
\eqref{eq:linear-deep-organization-flat-loss}.

We have thus shown that, on
\(\mathcal E_H\), the exact gradient
flow recovers the target classifier
at \(\tau_{\mathrm{org}}^{(\sigma,H)}\)
for all sufficiently small \(\sigma\),
while both microscopicity conclusions
hold uniformly over the preceding
organization interval.

Since
\(\mathbb P(\mathcal E_H)\to1\),
taking first \(\sigma\downarrow0\)
at fixed \(H\) and then \(H\to\infty\)
establishes
\eqref{eq:linear-deep-organization-sequential-limit}
and completes the proof.
\end{proof}

The theorem identifies a stage of the dynamics that is invisible at the
level of macroscopic loss escape. Starting from the centered random
end-to-end product, the homogeneous gradient flow generates a coherent
target-directed component, and large width suppresses the competing
initialization fluctuations sufficiently for this component to determine
all training predictions. Consequently, the classifier induced by the full
target can emerge strictly before the rank-one escape from the origin.

\paragraph{Rank-one escape and the predictive bottleneck.}

The target-driven organization established above occurs
while the end-to-end predictor is still microscopic.
We now consider the subsequent escape from the rank-zero
saddle, using the results of \citet{jacot2021saddle}.
Their analysis shows that, under a simple leading
singular value, gradient flow initialized at vanishing
scale first escapes along a balanced rank-one direction
determined by the leading singular component of the
target.

For our purposes, we do not need to characterize the
full escape trajectory. It suffices to identify the
end-to-end predictor when the parameter norm first
reaches a small prescribed radius. This formulation
will also allow us to control the entire trajectory
preceding the predictive bottleneck.

After the time rescaling used throughout this section,
the end-to-end objective is
\begin{equation}
\mathcal C(\mathbf M)
=
\frac12
\|\mathbf M-\mathbf T\|_{\mathrm F}^2,
\qquad
\nabla\mathcal C(\mathbf 0)
=
-\mathbf T.
\label{eq:linear-deep-end-to-end-objective}
\end{equation}
Write the singular value decomposition of the target as
\begin{equation}
\mathbf T
=
\sum_{k=1}^r
s_k\mathbf p_k\mathbf q_k^\top,
\qquad
s_1>s_2,
\label{eq:linear-deep-simple-leading-mode}
\end{equation}
where \(s_2:=0\) if \(r=1\).
Define the leading singular direction
\begin{equation}
\mathbf D_1
=
\mathbf p_1\mathbf q_1^\top
\label{eq:linear-deep-leading-direction}
\end{equation}
and the rank-one truncated target
\begin{equation}
\mathbf T_1
=
s_1\mathbf D_1.
\label{eq:linear-deep-leading-truncation}
\end{equation}
The condition \(s_1>s_2\) ensures that the optimal
first escape direction is unique up to hidden-layer
orthogonal transformations.

\begin{proposition}[First rank-one escape]
\label{prop:linear-deep-rank-one-escape}
Fix \(L\ge3\), the dimensions \(p,C,H\), and a target
\(\mathbf T\) satisfying
\eqref{eq:linear-deep-simple-leading-mode}.
Couple the Gaussian initializations for all
\(\sigma>0\) by
\begin{equation}
\mathbf W_\ell^{(\sigma)}(0)
=
\sigma\mathbf G_\ell,
\qquad
\ell=1,\ldots,L,
\label{eq:linear-deep-escape-coupling}
\end{equation}
where the matrices \(\mathbf G_\ell\) have independent
standard Gaussian entries.

There exists an event \(\mathcal E_{\mathrm{esc}}\),
depending only on the Gaussian initialization, such that
\begin{equation}
\mathbb P(\mathcal E_{\mathrm{esc}})
=
1
\quad\text{if }L=3,
\qquad
\mathbb P(\mathcal E_{\mathrm{esc}})
\ge\frac12
\quad\text{if }L>3.
\label{eq:linear-deep-escape-probability}
\end{equation}

Define the Euclidean norm of the network parameters by
\begin{equation}
\|\boldsymbol\theta^{(\sigma)}(\tau)\|_2^2
=
\sum_{\ell=1}^L
\|\mathbf W_\ell^{(\sigma)}(\tau)\|_{\mathrm F}^2.
\label{eq:linear-deep-parameter-norm}
\end{equation}
There exists \(\rho_0>0\) such that, for every fixed
\(\rho\in(0,\rho_0)\), the first hitting time
\begin{equation}
\tau_\rho^{(\sigma)}
=
\inf\left\{
\tau\ge0:
\|\boldsymbol\theta^{(\sigma)}(\tau)\|_2
=
\rho
\right\}
\label{eq:linear-deep-first-hit-time}
\end{equation}
is finite for all sufficiently small \(\sigma\)
on \(\mathcal E_{\mathrm{esc}}\), and
\begin{equation}
\mathbf M^{(\sigma)}
\left(\tau_\rho^{(\sigma)}\right)
\longrightarrow
\left(\frac{\rho}{\sqrt L}\right)^L
\mathbf D_1
\qquad
\text{as }\sigma\downarrow0.
\label{eq:linear-deep-first-hit-predictor}
\end{equation}

In particular, the predictor at the first escape
from a sufficiently small parameter-space
neighborhood converges to a positive multiple
of the rank-one truncated target \(\mathbf T_1\).
\end{proposition}

\begin{proof}
We apply \citet[Theorem~2]{jacot2021saddle}
to the objective
\eqref{eq:linear-deep-end-to-end-objective}.
Its gradient at the origin is \(-\mathbf T\),
whose largest singular value is simple
by assumption.

Let
\(\boldsymbol\theta^{(\sigma)}(\tau)\)
denote the collection of all network weights.
The cited theorem states that, on an event
\(\mathcal E_{\mathrm{esc}}\) with probability
\eqref{eq:linear-deep-escape-probability},
there exist escape times
\(\tau_{\mathrm{esc}}^{(\sigma)}\to\infty\),
a width-one gradient-flow trajectory
\(\boldsymbol\theta^*(t)\), and hidden-layer
orthogonal transformations such that
\begin{equation}
\boldsymbol\theta^{(\sigma)}
\left(
\tau_{\mathrm{esc}}^{(\sigma)}+t
\right)
\longrightarrow
\boldsymbol\theta^*(t)
\qquad
\text{as }\sigma\downarrow0,
\end{equation}
for every fixed \(t\in\mathbb R\),
after embedding the width-one trajectory
into the original width-\(H\) network and
applying the corresponding orthogonal
transformations.

Here \(t\) is time measured relative to
the escape time, rather than the original
training time \(\tau\).
The limiting trajectory approaches the
rank-zero saddle only in backward time:
\begin{equation}
\lim_{t\to-\infty}
\boldsymbol\theta^*(t)
=
\mathbf0.
\end{equation}
It is nonzero at every finite time.

We now identify this limiting trajectory
in our setting. The fact that the escape
trajectory has width one does not, by itself,
imply alignment with the leading singular
vectors of \(\mathbf T\). We establish this
alignment using an invariant submanifold
of the exact gradient flow and the uniqueness
of the optimal escape path.

Let \(\mathbf e_1\in\mathbb R^H\)
be the first standard basis vector.
Consider the spectrally aligned
rank-one configurations
\begin{equation}
\begin{aligned}
\mathbf W_1(t)
&=
\alpha_1(t)\mathbf e_1\mathbf q_1^\top,
\\
\mathbf W_\ell(t)
&=
\alpha_\ell(t)\mathbf e_1\mathbf e_1^\top,
\qquad 2\le\ell\le L-1,
\\
\mathbf W_L(t)
&=
\alpha_L(t)\mathbf p_1\mathbf e_1^\top.
\end{aligned}
\label{eq:linear-deep-width-one-factors}
\end{equation}
Their end-to-end predictor is
\begin{equation}
\mathbf M(t)
=
m(t)\mathbf D_1,
\qquad
m(t)
=
\prod_{\ell=1}^L\alpha_\ell(t).
\end{equation}

We first verify that this family is
invariant under the exact gradient flow.
For the objective
\(\mathcal C(\mathbf M)
=\frac12\|\mathbf M-\mathbf T\|_{\mathrm F}^2\),
the factor dynamics satisfy
\begin{equation}
\dot{\mathbf W}_\ell
=
\left(
\mathbf W_L\cdots\mathbf W_{\ell+1}
\right)^\top
(\mathbf T-\mathbf M)
\left(
\mathbf W_{\ell-1}\cdots\mathbf W_1
\right)^\top,
\end{equation}
with empty products interpreted as
identity matrices. Since
\begin{equation}
\mathbf T\mathbf q_1=s_1\mathbf p_1,
\qquad
\mathbf T^\top\mathbf p_1=s_1\mathbf q_1,
\end{equation}
substitution of
\eqref{eq:linear-deep-width-one-factors}
shows that the vector field preserves
the displayed input, output, and hidden
directions. The dynamics reduce exactly to
\begin{equation}
\dot\alpha_\ell
=
\left(
s_1-\prod_{j=1}^L\alpha_j
\right)
\prod_{\substack{j=1\\j\neq\ell}}^L
\alpha_j,
\qquad
\ell=1,\ldots,L.
\label{eq:linear-deep-width-one-scalar-flow}
\end{equation}
Thus spectral alignment is preserved
at every time, not merely to leading
order near the origin.

Moreover, deep-linear gradient flow
preserves the adjacent-layer imbalance
quantities
\begin{equation}
\mathbf W_{\ell+1}^{\top}\mathbf W_{\ell+1}
-
\mathbf W_\ell\mathbf W_\ell^{\top}.
\label{eq:linear-deep-balance-invariant-general}
\end{equation}
Within the aligned family, these invariants
reduce to the differences
\(\alpha_{\ell+1}^2-\alpha_\ell^2\).
An escape trajectory approaching the origin
as \(t\to-\infty\) must therefore be
balanced. For the optimal escape direction,
the scalar factors can be chosen positive
by hidden-layer orthogonal transformations,
so the balanced trajectory satisfies
\begin{equation}
\alpha_1(t)
=
\cdots
=
\alpha_L(t)
=
a(t)>0.
\label{eq:linear-deep-width-one-balanced}
\end{equation}
Equation~\eqref{eq:linear-deep-width-one-scalar-flow}
then becomes
\begin{equation}
\dot a
=
(s_1-a^L)a^{L-1}.
\label{eq:linear-deep-width-one-amplitude-flow}
\end{equation}
This equation admits a positive
trajectory with
\(0<a(t)<s_1^{1/L}\) and
\(a(t)\to0\) as \(t\to-\infty\).
Its backward-time asymptotics are
\begin{equation}
a(t)
\sim
\left(
(L-2)s_1(-t)
\right)^{-1/(L-2)},
\qquad
t\to-\infty.
\label{eq:linear-deep-width-one-backward-asymptotics}
\end{equation}
Hence this aligned trajectory escapes
the origin at the optimal rate, along
the balanced leading-singular direction.

To identify it with the limiting trajectory
provided by \citet[Theorem~2]{jacot2021saddle},
we use the uniqueness of optimal escape
paths established in the proof of that
theorem, through their Theorem~5.
When \(s_1>s_2\), the optimal escape
direction of the homogeneous leading-order
dynamics is unique up to hidden-layer
orthogonal transformations. The uniqueness
correspondence for fast escape paths then
identifies the limiting trajectory of the
full objective with the spectrally aligned
trajectory constructed above, up to these
transformations and a translation of time.
In particular, alignment holds along the
entire limiting escape trajectory.

Consequently, writing
\(\boldsymbol\theta^*(t)\)
for the limiting path, its end-to-end
predictor and parameter norm satisfy
\begin{equation}
\mathbf M^*(t)
=
a(t)^L\mathbf D_1,
\qquad
\|\boldsymbol\theta^*(t)\|_2^2
=
La(t)^2.
\label{eq:linear-deep-width-one-parameter-norm}
\end{equation}

It remains to relate this limiting path
to our first hitting time.
In the proof of
\citet[Theorem~2]{jacot2021saddle},
the escape time is constructed as the
first time the parameter trajectory
reaches a sufficiently small fixed radius;
see Proposition~21 of that work.
In particular, their construction allows
the shifted time origin to be chosen at
the first hitting time
\(\tau_\rho^{(\sigma)}\), for every
sufficiently small fixed \(\rho>0\).

With this choice,
\begin{equation}
\left\|
\boldsymbol\theta^*(0)
\right\|_2
=
\rho.
\end{equation}
Indeed, the exact trajectory has norm
\(\rho\) at the first hitting time, and
the parameter convergence above preserves
this equality in the limit.

By balance,
\begin{equation}
a(0)
=
\frac{\rho}{\sqrt L}.
\label{eq:linear-deep-width-one-radius-amplitude}
\end{equation}
Hence the corresponding limiting
end-to-end predictor is
\begin{equation}
\mathbf M^*(0)
=
\left(
\frac{\rho}{\sqrt L}
\right)^L
\mathbf D_1.
\label{eq:linear-deep-width-one-radius-predictor-amplitude}
\end{equation}

Finally, hidden-layer orthogonal
transformations preserve the
end-to-end predictor.
Taking the product of the convergent
factors at shifted time \(t=0\)
therefore gives
\begin{equation}
\mathbf M^{(\sigma)}
\left(
\tau_\rho^{(\sigma)}
\right)
\longrightarrow
\left(
\frac{\rho}{\sqrt L}
\right)^L
\mathbf D_1,
\qquad
\sigma\downarrow0.
\end{equation}
This proves
\eqref{eq:linear-deep-first-hit-predictor}
and completes the proof.
\end{proof}

The rank-one escape has an immediate consequence for
classification. Recall the leading singular direction
\(\mathbf D_1\) and the rank-one truncated target
\(\mathbf T_1=s_1\mathbf D_1\), defined in
\eqref{eq:linear-deep-leading-direction}
and \eqref{eq:linear-deep-leading-truncation}.
Since \(s_1>0\), these matrices induce the same
classifier. We denote its accuracy by
\begin{equation}
A_1
=
\Acc(\mathbf D_1)
=
\Acc(\mathbf T_1).
\label{eq:linear-deep-A1}
\end{equation}

To ensure that convergence of predictors implies
agreement of their classifiers, we assume that
\(\mathbf D_1\) has a unique top-scoring class
on every training example. Specifically, define
\begin{equation}
\widehat c_i^{(1)}
=
\arg\max_{c\in[C]}
(\mathbf D_1\mathbf x_i)_c
\label{eq:linear-deep-D1-class}
\end{equation}
and the minimum decision gap
\begin{equation}
\delta_{1,H}
=
\min_i
\min_{c\neq\widehat c_i^{(1)}}
\left[
(\mathbf D_1\mathbf x_i)_{\widehat c_i^{(1)}}
-
(\mathbf D_1\mathbf x_i)_c
\right]
>
0.
\label{eq:linear-deep-D1-gap}
\end{equation}
Unlike the target decision gap in the organization
theorem, \(\delta_{1,H}\) need not be bounded away
from zero uniformly in \(H\). This is because
the initialization scale \(\sigma\) tends to zero
at fixed \(H\) before the large-system limit.

The following theorem shows that the first
rank-one escape produces a predictive bottleneck
whose accuracy is \(A_1\), while the entire
preceding trajectory remains microscopic and
its loss stays asymptotically unchanged.

\begin{theorem}[Predictive bottleneck for \(L\ge3\)]
\label{thm:linear-deep-predictive-bottleneck}

Fix \(L\ge3\) and consider a sequence of
finite problems indexed by \(H\), satisfying
the simple-leading-singular-value condition
\eqref{eq:linear-deep-simple-leading-mode}.
Assume that, for every member of the sequence,
the leading direction \(\mathbf D_1\) has
a unique top-scoring class on every
training example.

For the Gaussian initializations coupled as in
\eqref{eq:linear-deep-escape-coupling},
there exist deterministic radii
\(\rho_H>0\), with \(\rho_H\to0\), and
corresponding realization-dependent bottleneck
times \(\tau_{\mathrm{bot}}^{(\sigma,H)}\)
such that, for every fixed \(\varepsilon>0\),
\begin{equation}
\begin{aligned}
\liminf_{H\to\infty}
\liminf_{\sigma\downarrow0}
\mathbb P\Bigg(
&
\Acc\left(
\mathbf M^{(\sigma)}
\left(
\tau_{\mathrm{bot}}^{(\sigma,H)}
\right)
\right)
=
A_1,
\\
&
\sup_{
0\le\tau\le\tau_{\mathrm{bot}}^{(\sigma,H)}
}
\|\mathbf M^{(\sigma)}(\tau)\|_{\mathrm F}
<
\varepsilon,
\\
&
\sup_{
0\le\tau\le\tau_{\mathrm{bot}}^{(\sigma,H)}
}
\left|
\mathcal L\left(
\mathbf M^{(\sigma)}(\tau)
\right)
-
\mathcal L(\mathbf0)
\right|
<
\varepsilon
\Bigg)
\ge
\pi_L,
\end{aligned}
\label{eq:linear-deep-bottleneck-joint}
\end{equation}
where
\begin{equation}
\pi_L
=
\begin{cases}
1, & L=3,\\
\frac12, & L>3.
\end{cases}
\label{eq:linear-deep-bottleneck-probability}
\end{equation}

Thus, with asymptotic probability at least
\(\pi_L\), gradient flow reaches the classifier
of the leading singular mode while the
end-to-end predictor and the change in loss
remain asymptotically vanishing throughout
the preceding trajectory.
\end{theorem}

\begin{proof}
The proof combines the first-hit convergence
of Proposition~\ref{prop:linear-deep-rank-one-escape}
with a deterministic bound on the predictor
before the first hitting time. Loss monotonicity
will then allow us to control the entire
preceding loss trajectory using only its
initial and final values.

\smallskip
\noindent
\emph{Step 1: choosing a microscopic escape radius.}

Fix \(H\), and write
\(\gamma=\gamma_H>0\) and \(s_1=s_{1,H}>0\).
Let \(\rho_{0,H}>0\) be the radius supplied by
Proposition~\ref{prop:linear-deep-rank-one-escape}.

We choose a deterministic amplitude
\begin{equation}
a_H
=
\min\left\{
\frac{s_1}{2},
\frac1H,
\frac{1}{H\gamma(1+s_1)},
\left(
\frac{\rho_{0,H}}{2\sqrt L}
\right)^L
\right\}.
\label{eq:linear-deep-bottleneck-amplitude}
\end{equation}
This choice ensures
\begin{equation}
0<a_H<s_1,
\qquad
a_H\longrightarrow0,
\label{eq:linear-deep-bottleneck-amplitude-properties}
\end{equation}
as well as
\begin{equation}
\gamma a_Hs_1\le\frac1H,
\qquad
\gamma a_H^2\le\frac1H.
\label{eq:linear-deep-bottleneck-loss-scales}
\end{equation}
The first property will make the predictor
microscopic; the last two bounds will control
the loss even if \(\gamma\) or \(s_1\)
depend on \(H\).

Define the associated parameter-space radius
\begin{equation}
\rho_H
=
\sqrt L\,a_H^{1/L}.
\label{eq:linear-deep-bottleneck-radius}
\end{equation}
By construction,
\begin{equation}
0<\rho_H<\rho_{0,H},
\qquad
\rho_H\longrightarrow0.
\label{eq:linear-deep-bottleneck-radius-local}
\end{equation}

Let \(\mathcal E_{\mathrm{esc}}\) be the
escape event from
Proposition~\ref{prop:linear-deep-rank-one-escape}.
On this event, for sufficiently small
\(\sigma\), define
\begin{equation}
\tau_{\mathrm{bot}}^{(\sigma,H)}
=
\inf\left\{
\tau\ge0:
\|\boldsymbol\theta^{(\sigma)}(\tau)\|_2
=
\rho_H
\right\}.
\label{eq:linear-deep-bottleneck-time}
\end{equation}
For realizations where this hitting time
is not defined, we set
\(\tau_{\mathrm{bot}}^{(\sigma,H)}=0\).

The first-escape proposition gives,
on \(\mathcal E_{\mathrm{esc}}\),
\begin{equation}
\mathbf M^{(\sigma)}
\left(
\tau_{\mathrm{bot}}^{(\sigma,H)}
\right)
\longrightarrow
a_H\mathbf D_1
\qquad
\text{as }\sigma\downarrow0.
\label{eq:linear-deep-bottleneck-limit}
\end{equation}

\smallskip
\noindent
\emph{Step 2: classification at the bottleneck.}

Since \(a_H>0\), the limiting predictor
\(a_H\mathbf D_1\) induces exactly the
same classifier as \(\mathbf T_1\).
More precisely, its score margins satisfy
\begin{equation}
\begin{aligned}
&
(a_H\mathbf D_1\mathbf x_i)_{\widehat c_i^{(1)}}
-
(a_H\mathbf D_1\mathbf x_i)_c
\\
&\qquad\ge
a_H\delta_{1,H}>0,
\qquad c\neq\widehat c_i^{(1)}.
\end{aligned}
\label{eq:linear-deep-bottleneck-limiting-margin}
\end{equation}

At fixed \(H\), there are finitely many
training examples and score comparisons.
Thus convergence in
\eqref{eq:linear-deep-bottleneck-limit}
and the strictly positive limiting
margins imply that, on
\(\mathcal E_{\mathrm{esc}}\), for all
sufficiently small \(\sigma\),
\begin{equation}
\Acc\left(
\mathbf M^{(\sigma)}
\left(
\tau_{\mathrm{bot}}^{(\sigma,H)}
\right)
\right)
=
A_1.
\label{eq:linear-deep-bottleneck-accuracy-proof}
\end{equation}
No uniform lower bound on
\(\delta_{1,H}\) is needed.

\smallskip
\noindent
\emph{Step 3: microscopicity before the bottleneck.}

For each fixed \(H\), the coupled
initialization satisfies
\begin{equation}
\|\boldsymbol\theta^{(\sigma)}(0)\|_2
=
\sigma
\left(
\sum_{\ell=1}^L
\|\mathbf G_\ell\|_{\mathrm F}^2
\right)^{1/2}
\longrightarrow0.
\label{eq:linear-deep-bottleneck-initial-parameter-norm}
\end{equation}
Hence, for all sufficiently small
\(\sigma\), the initial parameters lie
inside the sphere of radius \(\rho_H\).

By definition of the first hitting time,
on \(\mathcal E_{\mathrm{esc}}\),
\begin{equation}
\|\boldsymbol\theta^{(\sigma)}(\tau)\|_2
\le
\rho_H,
\qquad
0\le\tau\le
\tau_{\mathrm{bot}}^{(\sigma,H)}.
\label{eq:linear-deep-bottleneck-parameter-control}
\end{equation}

Submultiplicativity of the Frobenius norm
and the arithmetic--geometric mean
inequality therefore give
\begin{equation}
\begin{aligned}
\|\mathbf M^{(\sigma)}(\tau)\|_{\mathrm F}
&\le
\prod_{\ell=1}^L
\|\mathbf W_\ell^{(\sigma)}(\tau)\|_{\mathrm F}
\\
&\le
\left(
\frac1L
\sum_{\ell=1}^L
\|\mathbf W_\ell^{(\sigma)}(\tau)\|_{\mathrm F}^2
\right)^{L/2}
\\
&\le
\left(
\frac{\rho_H}{\sqrt L}
\right)^L
=
a_H.
\end{aligned}
\label{eq:linear-deep-bottleneck-uniform-microscopic}
\end{equation}
Consequently,
\begin{equation}
\sup_{
0\le\tau\le
\tau_{\mathrm{bot}}^{(\sigma,H)}
}
\|\mathbf M^{(\sigma)}(\tau)\|_{\mathrm F}
\le
a_H
\longrightarrow0
\qquad
\text{as }H\to\infty.
\label{eq:linear-deep-bottleneck-uniform-microscopic-limit}
\end{equation}

Importantly, this bound holds for the
entire trajectory preceding the first
hitting time, without requiring any
further information about the intermediate
gradient-flow dynamics.

\smallskip
\noindent
\emph{Step 4: uniform flatness of the loss.}

By definition of the leading singular
direction,
\begin{equation}
\|\mathbf D_1\|_{\mathrm F}=1,
\qquad
\langle\mathbf T,\mathbf D_1\rangle_{\mathrm F}
=
s_1.
\label{eq:linear-deep-bottleneck-D1-identities}
\end{equation}
Therefore the limiting bottleneck predictor
has loss
\begin{equation}
\begin{aligned}
\mathcal L(a_H\mathbf D_1)
-
\mathcal L(\mathbf0)
&=
\frac{\gamma}{2}
\left(
\|\mathbf T-a_H\mathbf D_1\|_{\mathrm F}^2
-
\|\mathbf T\|_{\mathrm F}^2
\right)
\\
&=
-\gamma a_Hs_1
+
\frac{\gamma}{2}a_H^2.
\end{aligned}
\label{eq:linear-deep-bottleneck-loss-difference}
\end{equation}
Using
\eqref{eq:linear-deep-bottleneck-loss-scales},
we obtain
\begin{equation}
\left|
\mathcal L(a_H\mathbf D_1)
-
\mathcal L(\mathbf0)
\right|
\le
\frac{3}{2H}
\longrightarrow0.
\label{eq:linear-deep-bottleneck-loss-bound}
\end{equation}

For fixed \(H\), the predictor convergence
\eqref{eq:linear-deep-bottleneck-limit}
and continuity of the loss imply
\begin{equation}
\mathcal L\left(
\mathbf M^{(\sigma)}
\left(
\tau_{\mathrm{bot}}^{(\sigma,H)}
\right)
\right)
\longrightarrow
\mathcal L(a_H\mathbf D_1)
\label{eq:linear-deep-bottleneck-loss-inner-limit}
\end{equation}
as \(\sigma\downarrow0\), on
\(\mathcal E_{\mathrm{esc}}\).
Similarly, since
\(\mathbf M^{(\sigma)}(0)
=\sigma^L\mathbf G_L\cdots\mathbf G_1
\to\mathbf0\),
\begin{equation}
\mathcal L\left(
\mathbf M^{(\sigma)}(0)
\right)
\longrightarrow
\mathcal L(\mathbf0).
\label{eq:linear-deep-bottleneck-initial-loss}
\end{equation}

Finally, gradient flow makes the loss
nonincreasing. Hence, throughout the
interval preceding the bottleneck,
\begin{equation}
\mathcal L\left(
\mathbf M^{(\sigma)}
\left(
\tau_{\mathrm{bot}}^{(\sigma,H)}
\right)
\right)
\le
\mathcal L\left(
\mathbf M^{(\sigma)}(\tau)
\right)
\le
\mathcal L\left(
\mathbf M^{(\sigma)}(0)
\right).
\label{eq:linear-deep-bottleneck-loss-sandwich}
\end{equation}
It follows that
\begin{equation}
\begin{aligned}
&
\sup_{
0\le\tau\le
\tau_{\mathrm{bot}}^{(\sigma,H)}
}
\left|
\mathcal L\left(
\mathbf M^{(\sigma)}(\tau)
\right)
-
\mathcal L(\mathbf0)
\right|
\\
&\qquad\le
\max\Bigg\{
\left|
\mathcal L\left(
\mathbf M^{(\sigma)}(0)
\right)
-
\mathcal L(\mathbf0)
\right|,
\left|
\mathcal L\left(
\mathbf M^{(\sigma)}
\left(
\tau_{\mathrm{bot}}^{(\sigma,H)}
\right)
\right)
-
\mathcal L(\mathbf0)
\right|
\Bigg\}.
\end{aligned}
\label{eq:linear-deep-bottleneck-uniform-flat-loss}
\end{equation}

The right-hand side is asymptotically
bounded by \(3/(2H)\) after taking
\(\sigma\downarrow0\) at fixed \(H\),
on \(\mathcal E_{\mathrm{esc}}\).
It therefore vanishes in the subsequent
limit \(H\to\infty\).

\smallskip

We have established all three conclusions
of \eqref{eq:linear-deep-bottleneck-joint}
on the escape event, for sufficiently small
\(\sigma\) and sufficiently large \(H\).
By Proposition~\ref{prop:linear-deep-rank-one-escape},
\begin{equation}
\mathbb P(\mathcal E_{\mathrm{esc}})
\ge
\pi_L.
\label{eq:linear-deep-bottleneck-escape-probability-final}
\end{equation}
Taking first \(\sigma\downarrow0\)
and then \(H\to\infty\) proves the
claimed probability bound.
\end{proof}

The preceding results can now be combined to
describe the first two stages of the accuracy
dynamics. Theorem~\ref{thm:linear-deep-target-organization}
establishes an early stage in which the classifier
agrees with the full target, whereas
Theorem~\ref{thm:linear-deep-predictive-bottleneck}
identifies a later stage governed by its leading
rank-one singular component.

We show that these stages occur in the stated
temporal order and that the predictor and loss
remain microscopic throughout. The only
depth-dependent qualification is the probability
guarantee inherited from the saddle-escape result
of \citet{jacot2021saddle}.

\begin{linkedtheorem}
{Target-driven organization and predictive bottleneck for \(L\ge3\)}
{thm:main-linear-deep-accuracy-well}
\label{thm:linear-deep-accuracy-well}

Fix \(L\ge3\), and consider the coupled Gaussian
initialization
\eqref{eq:linear-deep-escape-coupling}.
Suppose that the assumptions of
Theorem~\ref{thm:linear-deep-target-organization}
and
Theorem~\ref{thm:linear-deep-predictive-bottleneck}
hold. In particular, the target has a simple
largest singular value \(s_1>s_2\), and its
leading singular direction \(\mathbf D_1\)
has a unique top-scoring class on every
training example.

Recall the accuracies
\begin{equation}
A_{\mathbf T}
=
\Acc(\mathbf T),
\qquad
A_1
=
\Acc(\mathbf T_1)
=
\Acc(\mathbf D_1),
\label{eq:linear-deep-linked-accuracies}
\end{equation}
where \(\mathbf T_1=s_1\mathbf D_1\).

Let \(\tau_{\mathrm{org}}^{(\sigma,H)}\)
be the deterministic organization time from
Theorem~\ref{thm:linear-deep-target-organization}.
Then there exist initialization-dependent
bottleneck times
\(\tau_{\mathrm{bot}}^{(\sigma,H)}\)
such that, for every fixed \(\varepsilon>0\),
\begin{equation}
\begin{aligned}
\liminf_{H\to\infty}
\liminf_{\sigma\downarrow0}
\mathbb P\Bigg(
&
0<
\tau_{\mathrm{org}}^{(\sigma,H)}
<
\tau_{\mathrm{bot}}^{(\sigma,H)},
\\
&
\Acc\left(
\mathbf M^{(\sigma)}
\left(
\tau_{\mathrm{org}}^{(\sigma,H)}
\right)
\right)
=
A_{\mathbf T},
\\
&
\Acc\left(
\mathbf M^{(\sigma)}
\left(
\tau_{\mathrm{bot}}^{(\sigma,H)}
\right)
\right)
=
A_1,
\\
&
\sup_{
0\le\tau\le\tau_{\mathrm{bot}}^{(\sigma,H)}
}
\|\mathbf M^{(\sigma)}(\tau)\|_{\mathrm F}
<
\varepsilon,
\\
&
\sup_{
0\le\tau\le\tau_{\mathrm{bot}}^{(\sigma,H)}
}
\left|
\mathcal L\left(
\mathbf M^{(\sigma)}(\tau)
\right)
-
\mathcal L(\mathbf0)
\right|
<
\varepsilon
\Bigg)
\ge
\pi_L,
\end{aligned}
\label{eq:linear-deep-linked-joint}
\end{equation}
where
\begin{equation}
\pi_L
=
\begin{cases}
1, & L=3,\\[2mm]
\frac12, & L>3.
\end{cases}
\label{eq:linear-deep-linked-probability}
\end{equation}

In particular, if
\begin{equation}
A_{\mathbf T}>A_1,
\label{eq:linear-deep-linked-strict-gap}
\end{equation}
then the trajectory reaches the classifier
of the full target before entering a
bottleneck with strictly smaller accuracy.
Both stages occur while the predictor
remains microscopic and the loss stays
asymptotically equal to that of the zero
predictor.

This conclusion holds with probability
tending to one for \(L=3\), and with
asymptotic probability at least \(1/2\)
for \(L>3\).
\end{linkedtheorem}

\begin{proof}
The two preceding theorems establish the
respective accuracy statements and microscopicity
bounds. It remains to show that organization
occurs before the bottleneck, and that the
two stages occur simultaneously with the
claimed probability.

We first establish the time ordering.
Let \(\mathcal E_H^{\mathrm{org}}\)
be the high-probability event from
Theorem~\ref{thm:linear-deep-target-organization},
and let \(\mathcal E_{\mathrm{esc},H}\)
be the escape event from
Proposition~\ref{prop:linear-deep-rank-one-escape}.
By the preceding results,
\begin{equation}
\mathbb P(\mathcal E_H^{\mathrm{org}})
\longrightarrow1,
\qquad
\mathbb P(\mathcal E_{\mathrm{esc},H})
\ge
\pi_L.
\label{eq:linear-deep-linked-event-probabilities}
\end{equation}

Fix a sufficiently large \(H\) and a
realization in \(\mathcal E_H^{\mathrm{org}}\).
Recall that the homogeneous slow time
corresponding to organization is
\begin{equation}
s_{\mathrm{org}}
=
H^{-(L-2)/2}u_{\mathrm{org}}.
\end{equation}
The proof of
Theorem~\ref{thm:linear-deep-target-organization}
shows that the homogeneous flow exists
on an interval containing
\([0,s_{\mathrm{org}}]\).

We may therefore apply
Lemma~\ref{lem:linear-deep-near-origin-reduction}.
In particular, the normalized factors
\(\mathbf Z_\ell^{(\sigma)}(s)\)
remain uniformly bounded on
\([0,s_{\mathrm{org}}]\)
as \(\sigma\downarrow0\).

Recalling the normalization
\(\mathbf W_\ell^{(\sigma)}(\tau)
=\sigma\mathbf Z_\ell^{(\sigma)}(s)\),
with \(s=\sigma^{L-2}\tau\), we obtain
\begin{equation}
\begin{aligned}
&
\sup_{
0\le\tau\le\tau_{\mathrm{org}}^{(\sigma,H)}
}
\|\boldsymbol\theta^{(\sigma)}(\tau)\|_2
\\
&\quad=
\sigma
\sup_{0\le s\le s_{\mathrm{org}}}
\left(
\sum_{\ell=1}^L
\|\mathbf Z_\ell^{(\sigma)}(s)\|_{\mathrm F}^2
\right)^{1/2}
\\
&\quad=
O(\sigma)
\longrightarrow0.
\end{aligned}
\label{eq:linear-deep-linked-org-parameter-bound}
\end{equation}
Here \(H\) is fixed, and the implicit
constant may depend on the initialization
realization.

Now recall that the bottleneck time is
defined as the first time the parameter
norm reaches the deterministic radius
\(\rho_H>0\) from
Theorem~\ref{thm:linear-deep-predictive-bottleneck}:
\begin{equation}
\tau_{\mathrm{bot}}^{(\sigma,H)}
=
\inf\left\{
\tau\ge0:
\|\boldsymbol\theta^{(\sigma)}(\tau)\|_2
=
\rho_H
\right\}
\end{equation}
on the escape event.

At fixed \(H\), the radius \(\rho_H\)
is positive and independent of \(\sigma\).
Thus
\eqref{eq:linear-deep-linked-org-parameter-bound}
implies that, for all sufficiently small
\(\sigma\), the parameter trajectory
remains strictly inside this sphere
throughout the organization interval.

Consequently, on
\(\mathcal E_H^{\mathrm{org}}
\cap\mathcal E_{\mathrm{esc},H}\),
the first hitting time must satisfy
\begin{equation}
0<
\tau_{\mathrm{org}}^{(\sigma,H)}
<
\tau_{\mathrm{bot}}^{(\sigma,H)}
\label{eq:linear-deep-linked-time-order}
\end{equation}
for all sufficiently small \(\sigma\).

We next collect the remaining conclusions.
On \(\mathcal E_H^{\mathrm{org}}\),
Theorem~\ref{thm:linear-deep-target-organization}
guarantees, for sufficiently small \(\sigma\),
\begin{equation}
\Acc\left(
\mathbf M^{(\sigma)}
\left(
\tau_{\mathrm{org}}^{(\sigma,H)}
\right)
\right)
=
A_{\mathbf T}.
\label{eq:linear-deep-linked-org-accuracy}
\end{equation}
Similarly, on \(\mathcal E_{\mathrm{esc},H}\),
Theorem~\ref{thm:linear-deep-predictive-bottleneck}
gives
\begin{equation}
\Acc\left(
\mathbf M^{(\sigma)}
\left(
\tau_{\mathrm{bot}}^{(\sigma,H)}
\right)
\right)
=
A_1.
\label{eq:linear-deep-linked-bot-accuracy}
\end{equation}

The same bottleneck theorem supplies the
uniform bound (see \eqref{eq:linear-deep-bottleneck-uniform-microscopic-limit})
\begin{equation}
\sup_{
0\le\tau\le\tau_{\mathrm{bot}}^{(\sigma,H)}
}
\|\mathbf M^{(\sigma)}(\tau)\|_{\mathrm F}
\le
a_H,
\qquad
a_H\longrightarrow0
\quad\text{as }H\to\infty,
\label{eq:linear-deep-linked-uniform-microscopic}
\end{equation}
for all sufficiently small \(\sigma\)
on the escape event.

Furthermore, its proof establishes (see \eqref{eq:linear-deep-bottleneck-loss-bound})
\begin{equation}
\limsup_{\sigma\downarrow0}
\sup_{
0\le\tau\le\tau_{\mathrm{bot}}^{(\sigma,H)}
}
\Big|
\mathcal L\left(
\mathbf M^{(\sigma)}(\tau)
\right)
-
\mathcal L(\mathbf0)
\Big|
\le
\frac{3}{2H}.
\label{eq:linear-deep-linked-uniform-flat-loss}
\end{equation}
Thus, for every fixed \(\varepsilon>0\),
both uniform microscopicity and flatness
hold for all sufficiently large \(H\)
and then all sufficiently small \(\sigma\),
on the escape event.

It remains to combine the probabilities.
Define the simultaneous good event
\begin{equation}
\mathcal E_H^{\mathrm{joint}}
=
\mathcal E_H^{\mathrm{org}}
\cap
\mathcal E_{\mathrm{esc},H}.
\label{eq:linear-deep-linked-joint-good-event}
\end{equation}
No independence between these events
is required. By the union bound,
\begin{equation}
\begin{aligned}
\mathbb P(\mathcal E_H^{\mathrm{joint}})
&\ge
\mathbb P(\mathcal E_{\mathrm{esc},H})
-
\mathbb P\left(
(\mathcal E_H^{\mathrm{org}})^c
\right)
\\
&\ge
\pi_L-o(1).
\end{aligned}
\label{eq:linear-deep-linked-joint-good-probability}
\end{equation}

For every fixed sufficiently large \(H\)
and every realization in
\(\mathcal E_H^{\mathrm{joint}}\),
all the required statements hold
simultaneously for sufficiently small
\(\sigma\), with the threshold allowed
to depend on the realization.

Therefore, taking first the limit
\(\sigma\downarrow0\) at fixed \(H\),
and then \(H\to\infty\), yields
\eqref{eq:linear-deep-linked-joint}.
\end{proof}

\paragraph{Recovery: open generically, explicit under spectral alignment.}
\label{app:recovery-L3}

Theorem~\ref{thm:linear-deep-accuracy-well} establishes the first
two stages of the accuracy dynamics for Gaussian initialization:
early target-driven organization followed by the rank-one
predictive bottleneck. It does not establish recovery of the
same Gaussian-initialized trajectory. For \(L\ge3\), global
convergence of unregularized deep matrix factorization from
the multivariate vanishing-Gaussian initialization considered
here is not established in the generality required for such
a conclusion.

For gradient flow, \citet{bah2022learning} prove convergence
to a critical point and show that, for almost every
initialization, the limiting end-to-end matrix is a global
minimizer of the loss restricted to the manifold of matrices
of some rank \(k\). For \(L=2\), the limiting rank can be
shown to be maximal, yielding global convergence. For
\(L\ge3\), however, their argument does not determine the
limiting rank and therefore does not establish convergence
to the unrestricted global minimizer. An analogous limitation
arises in the analysis of gradient descent by
\citet{nguegnang2024convergence}.

Global convergence has been established under stronger or
more restrictive assumptions. For gradient descent,
\citet{arora2019convergence} obtain linear convergence under
approximate balance and a positive deficiency margin,
requiring the initial loss to lie below that of any
rank-deficient solution. For gradient flow, the deep-linear
convergence result used by \citet{elkabetz2021continuous}
concerns scalar regression with balanced initialization.
These results do not establish global convergence for the
general multivariate, unregularized vanishing-Gaussian
gradient flow studied here.

The missing recovery mechanism is closely related to the
saddle-to-saddle picture proposed by \citet{jacot2021saddle}.
After the rigorously established first escape analyzed above,
they conjecture that the vanishing-initialization trajectory
successively visits critical points of increasing rank,
\begin{equation}
0
\longrightarrow
1
\longrightarrow
2
\longrightarrow
\cdots,
\label{eq:linear-deep-rank-sequence}
\end{equation}
eventually reaching a global minimizer. In our setting,
subsequent escapes would correspond to the recruitment
of singular modes absent from the rank-one bottleneck
and could provide the missing recovery mechanism for the
Gaussian trajectory. This increasing-rank scenario,
however, is not established in the required generality.

The mechanism can nevertheless be followed exactly in
the classical spectrally aligned setting of
\citet{saxe2014exact}. We use a balanced
specialization of their invariant manifold, adapted to
our notation and isotropic-input setting. Consider a target with singular
value decomposition
\begin{equation}
\mathbf T
=
\sum_{k=1}^r
s_k\mathbf p_k\mathbf q_k^\top,
\qquad
s_k>0,
\label{eq:linear-deep-aligned-target}
\end{equation}
and define its left and right singular-vector bases by
\begin{equation}
\mathbf P
=
\left(
\mathbf p_1,\ldots,\mathbf p_r
\right),
\qquad
\mathbf Q
=
\left(
\mathbf q_1,\ldots,\mathbf q_r
\right).
\label{eq:linear-deep-aligned-singular-bases}
\end{equation}
For \(H\ge r\), choose
\(\mathbf B\in\mathbb R^{H\times r}\)
with orthonormal columns,
\begin{equation}
\mathbf B^\top\mathbf B
=
\mathbf I_r.
\label{eq:linear-deep-aligned-B}
\end{equation}
Initialize the network on the balanced, spectrally aligned
manifold by setting
\begin{equation}
\mathbf W_1(0)
=
\mathbf B\mathbf A(0)\mathbf Q^\top,
\label{eq:linear-deep-aligned-W1}
\end{equation}
\begin{equation}
\mathbf W_\ell(0)
=
\mathbf B\mathbf A(0)\mathbf B^\top,
\qquad
2\le\ell\le L-1,
\label{eq:linear-deep-aligned-hidden}
\end{equation}
and
\begin{equation}
\mathbf W_L(0)
=
\mathbf P\mathbf A(0)\mathbf B^\top,
\label{eq:linear-deep-aligned-WL}
\end{equation}
where
\begin{equation}
\mathbf A(0)
=
\operatorname{diag}
\left(
a_1(0),\ldots,a_r(0)
\right),
\qquad
a_k(0)>0.
\label{eq:linear-deep-aligned-A0}
\end{equation}

This manifold is invariant under gradient flow
\citep{saxe2014exact}. The singular directions remain
fixed, and the end-to-end predictor takes the form
\begin{equation}
\mathbf M(\tau)
=
\sum_{k=1}^r
m_k(\tau)\mathbf p_k\mathbf q_k^\top,
\qquad
m_k(\tau)
=
a_k(\tau)^L.
\label{eq:linear-deep-aligned-predictor}
\end{equation}
Substituting the aligned factorization into the gradient-flow
equations shows that the modes evolve independently:
\begin{equation}
\dot m_k
=
L(s_k-m_k)m_k^{2-2/L}.
\label{eq:linear-deep-aligned-mode-dynamics}
\end{equation}

Equation~\eqref{eq:linear-deep-aligned-mode-dynamics}
makes the competition between singular modes explicit.
Their relative growth rates satisfy
\begin{equation}
\frac{d}{d\tau}\log m_k
=
L(s_k-m_k)m_k^{(L-2)/L}.
\label{eq:linear-deep-aligned-relative-growth}
\end{equation}
For \(L\ge3\), modes that are already larger can enjoy
a multiplicative growth advantage near the origin,
while larger target singular values provide an
additional advantage. Consequently, the leading mode
may separate strongly from the remaining modes,
producing an approximate rank-one predictive bottleneck.

Such dominance, when it occurs, cannot prevent eventual
recovery. As a mode approaches its target value, the
residual \(s_k-m_k\) vanishes and its growth slows.
Subleading modes that remain below their target values
continue to grow. More generally, the scalar dynamics
guarantee that every strictly positive mode converges
to its target value, regardless of whether it starts
above or below that value. Thus the full target is
eventually reconstructed.

\begin{corollary}[Permanent classification recovery in the aligned regime]
\label{cor:linear-deep-aligned-recovery}

Consider the aligned balanced initialization
\eqref{eq:linear-deep-aligned-W1}--%
\eqref{eq:linear-deep-aligned-WL},
with \(a_k(0)>0\) for every \(k\).
Suppose that both \(\mathbf T\) and its leading
singular component
\begin{equation}
\mathbf T_1
=
s_1\mathbf p_1\mathbf q_1^\top
\label{eq:linear-deep-aligned-T1}
\end{equation}
have unique top-scoring classes on every training
example. Define
\begin{equation}
A_{\mathbf T}
=
\Acc(\mathbf T),
\qquad
A_1
=
\Acc(\mathbf T_1),
\label{eq:linear-deep-aligned-accuracies}
\end{equation}
let \(\delta_1>0\) be the minimum decision gap
of \(\mathbf T_1\), and set
\begin{equation}
R
=
\max_i\|\mathbf x_i\|_2.
\label{eq:linear-deep-aligned-R}
\end{equation}

Suppose that, at some time \(\tau_{\mathrm{bot}}\),
the leading mode dominates the remaining modes
in the sense that
\begin{equation}
\max_{k\ge2}
m_k(\tau_{\mathrm{bot}})
<
\frac{\delta_1}
{2\sqrt2\,R\,s_1}
m_1(\tau_{\mathrm{bot}}),
\label{eq:linear-deep-aligned-bottleneck-condition}
\end{equation}
where the maximum is interpreted as \(0\) when \(r=1\).
Then
\begin{equation}
\Acc\left(
\mathbf M(\tau_{\mathrm{bot}})
\right)
=
A_1.
\label{eq:linear-deep-aligned-bottleneck-accuracy}
\end{equation}

Moreover,
\begin{equation}
\mathbf M(\tau)
\longrightarrow
\mathbf T
\qquad
\text{as }\tau\to\infty,
\label{eq:linear-deep-aligned-recovery}
\end{equation}
and consequently there exists a finite time
\(\tau_{\mathrm{rec}}>\tau_{\mathrm{bot}}\)
such that
\begin{equation}
\Acc\left(
\mathbf M(\tau)
\right)
=
A_{\mathbf T}
\qquad
\text{for every }\tau\ge\tau_{\mathrm{rec}}.
\label{eq:linear-deep-aligned-permanent-recovery}
\end{equation}

Thus, whenever
\begin{equation}
A_{\mathbf T}>A_1,
\label{eq:linear-deep-aligned-accuracy-gap}
\end{equation}
an aligned trajectory that enters the
leading-mode predictive bottleneck subsequently
recovers permanently to the target accuracy.
\end{corollary}

\begin{proof}
We first show that the dominance condition makes
the classifier agree with that of the leading
singular component. We then use the independent
mode dynamics to establish convergence to the
full target and permanent recovery of its classifier.

\smallskip
\noindent
\emph{Classification at the bottleneck.}

By invariance of the aligned manifold,
\begin{equation}
\mathbf M(\tau)
=
\sum_{k=1}^r
m_k(\tau)\mathbf p_k\mathbf q_k^\top.
\label{eq:linear-deep-aligned-proof-M}
\end{equation}
At the proposed bottleneck time, decompose
the predictor into its leading mode and
the remaining components:
\begin{equation}
\mathbf M(\tau_{\mathrm{bot}})
=
\frac{m_1(\tau_{\mathrm{bot}})}{s_1}
\mathbf T_1
+
\mathbf E,
\label{eq:linear-deep-aligned-bottleneck-decomposition}
\end{equation}
where
\begin{equation}
\mathbf E
=
\sum_{k=2}^r
m_k(\tau_{\mathrm{bot}})
\mathbf p_k\mathbf q_k^\top.
\label{eq:linear-deep-aligned-bottleneck-error}
\end{equation}
Because the singular directions are orthonormal,
\begin{equation}
\|\mathbf E\|_{\mathrm{op}}
=
\max_{k\ge2}
m_k(\tau_{\mathrm{bot}}),
\label{eq:linear-deep-aligned-error-norm}
\end{equation}
with the right-hand side equal to zero
when \(r=1\).

For each example, let
\begin{equation}
\widehat c_i^{(1)}
=
\arg\max_{c\in[C]}
(\mathbf T_1\mathbf x_i)_c.
\label{eq:linear-deep-aligned-T1-class}
\end{equation}
For any competing class
\(c\neq\widehat c_i^{(1)}\),
we obtain
\begin{equation}
\begin{aligned}
\left(
\mathbf e_{\widehat c_i^{(1)}}-\mathbf e_c
\right)^\top
\mathbf M(\tau_{\mathrm{bot}})
\mathbf x_i
&=
\frac{m_1(\tau_{\mathrm{bot}})}{s_1}
\left(
\mathbf e_{\widehat c_i^{(1)}}-\mathbf e_c
\right)^\top
\mathbf T_1\mathbf x_i
+
\left(
\mathbf e_{\widehat c_i^{(1)}}-\mathbf e_c
\right)^\top
\mathbf E\mathbf x_i
\\
&\ge
\frac{m_1(\tau_{\mathrm{bot}})}{s_1}
\delta_1
-
\sqrt2\,R\|\mathbf E\|_{\mathrm{op}}
\\
&>
\frac{m_1(\tau_{\mathrm{bot}})}{2s_1}
\delta_1
>
0.
\end{aligned}
\label{eq:linear-deep-aligned-bottleneck-margin}
\end{equation}
The last inequality follows from
\eqref{eq:linear-deep-aligned-bottleneck-condition}.
Thus every score comparison agrees with that
of \(\mathbf T_1\), proving
\eqref{eq:linear-deep-aligned-bottleneck-accuracy}.

\smallskip
\noindent
\emph{Convergence and permanent recovery.}

Each mode satisfies
\eqref{eq:linear-deep-aligned-mode-dynamics}.
Because \(m_k(0)>0\), its amplitude remains
positive, and
\begin{equation}
\operatorname{sign}(\dot m_k)
=
\operatorname{sign}(s_k-m_k).
\label{eq:linear-deep-aligned-mode-sign}
\end{equation}
Therefore each mode moves monotonically
toward \(s_k\), increasing if initially
below its target value and decreasing
if initially above it.

The scalar ODE is locally Lipschitz on
\((0,\infty)\), and \(s_k\) is an equilibrium.
Consequently, a trajectory cannot cross
\(s_k\). It remains between its initial
value and \(s_k\), and hence has a finite
positive limit.

If this limit were different from \(s_k\),
the right-hand side of
\eqref{eq:linear-deep-aligned-mode-dynamics}
would remain bounded away from zero,
contradicting convergence. Therefore,
\begin{equation}
m_k(\tau)
\longrightarrow
s_k
\qquad
\text{for every }k.
\label{eq:linear-deep-aligned-mode-limit}
\end{equation}

Since the singular directions are fixed
and orthonormal, we have the exact identity
\begin{equation}
\|\mathbf M(\tau)-\mathbf T\|_{\mathrm{op}}
=
\max_{1\le k\le r}
|m_k(\tau)-s_k|
\longrightarrow0.
\label{eq:linear-deep-aligned-M-to-T}
\end{equation}
This proves
\eqref{eq:linear-deep-aligned-recovery}.

It remains to translate predictor convergence
into permanent classification recovery.
Since \(\mathbf T\) has a unique top-scoring
class on each example, define
\begin{equation}
\widehat c_i^{\,T}
=
\arg\max_{c\in[C]}
(\mathbf T\mathbf x_i)_c
\label{eq:linear-deep-aligned-target-class}
\end{equation}
and its minimum decision gap
\begin{equation}
\delta_{\mathbf T}
=
\min_i
\min_{c\neq\widehat c_i^{\,T}}
\left[
(\mathbf T\mathbf x_i)_{\widehat c_i^{\,T}}
-
(\mathbf T\mathbf x_i)_c
\right]
>
0.
\label{eq:linear-deep-aligned-target-gap}
\end{equation}
Because the training set is finite,
this minimum is strictly positive.

For every example and competing class,
the perturbation of its target score
margin is bounded by
\begin{equation}
\begin{aligned}
&
\left|
\left(
\mathbf e_{\widehat c_i^{\,T}}-\mathbf e_c
\right)^\top
(\mathbf M(\tau)-\mathbf T)
\mathbf x_i
\right|
\\
&\qquad\le
\sqrt2\,R
\|\mathbf M(\tau)-\mathbf T\|_{\mathrm{op}}.
\end{aligned}
\end{equation}
Hence all target score margins remain
positive whenever
\begin{equation}
\|\mathbf M(\tau)-\mathbf T\|_{\mathrm{op}}
<
\frac{\delta_{\mathbf T}}
{2\sqrt2\,R}.
\label{eq:linear-deep-aligned-recovery-margin-condition}
\end{equation}

Since
\(\|\mathbf M(\tau)-\mathbf T\|_{\mathrm{op}}\to0\),
this inequality holds for every sufficiently
large \(\tau\). We may therefore choose a
finite \(\tau_{\mathrm{rec}}>\tau_{\mathrm{bot}}\)
such that the classifier agrees with that of
\(\mathbf T\) for all
\(\tau\ge\tau_{\mathrm{rec}}\), proving
\eqref{eq:linear-deep-aligned-permanent-recovery}.
\end{proof}

The aligned result should not be interpreted as
completing Theorem~\ref{thm:linear-deep-accuracy-well}
for the same Gaussian-initialized trajectory.
Rather, it isolates the recruitment and recovery
mechanism in a setting where the singular modes
can be followed exactly. The aligned dynamics
guarantee eventual reconstruction of the full
target, while the appearance of an intermediate
rank-one predictive bottleneck depends on the
relative mode amplitudes.

This picture is consistent with the saddle-to-saddle
scenario of \citet{jacot2021saddle}: depth can amplify
differences between initially weak modes and create
a low-rank predictive bottleneck; as the leading mode
approaches saturation, modes that remain below their
target strengths continue evolving, eventually
restoring the full target and its classifier.

\section{Phenomenology and Experimental Details}
\label{app:experiments}

This appendix provides the experimental details and additional controls
underlying the phenomenological observations of
Section~\ref{sec:phenomenology}.  We organize the experiments from the most
controlled setting to progressively more conventional learning problems.
Appendix~\ref{app:unary-experiments} studies unary-operation learning, where
the target structure and the factorization can be varied directly.
Appendix~\ref{app:mnist-experiments} then moves to MNIST and tests the
dependence of the well on initialization scale, width, loss, biases,
factorization depth, and optimization method.  Finally,
Appendix~\ref{app:cifar10-cnn} considers a convolutional network on CIFAR-10
and compares both training and test dynamics.

Figures~\ref{fig:main-cifar-mnist} and~\ref{fig:main-unary} show
representative examples in the main text.  The figures below provide the
corresponding broader sweeps and controls.  Our aim here is therefore not to
repeat the phenomenological discussion of Section~\ref{sec:phenomenology},
but to document the experimental setups and make clear which comparisons
support the empirical conclusions used in the paper.

\subsection{Controlled unary-operation experiments}
\label{app:unary-experiments}

The unary setting provides the cleanest controlled experiments because the
target geometry and the factorization can be modified independently.  We
consider three settings.  First, a shallow factorization trained on a
permutation target gives a flat-spectrum control for which the rigorous theory
predicts no well.  Second, we keep the same permutation target but use a
three-factor parameterization, testing whether a well can occur without target
spectral heterogeneity.  Third, we return to the shallow factorization but
replace the permutation by an iid random unary target, restoring the
heterogeneous target spectrum analyzed in
Appendix~\ref{app:linear-L2-random-unary}.  For each setting we vary both
initialization scale and hidden width.

\begin{figure}[t]
    \centering

    \includegraphics[width=0.32\linewidth]
    {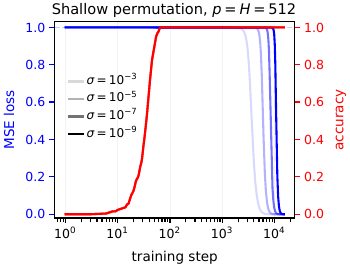}
    \hfill
    \includegraphics[width=0.32\linewidth]
    {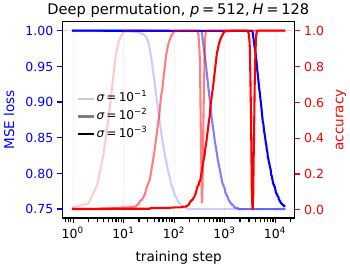}
    \hfill
    \includegraphics[width=0.32\linewidth]
    {unary_imgs/random_unary_shallow_sigma_sweep.pdf}

    \vspace{0.35em}

    \includegraphics[width=0.32\linewidth]
    {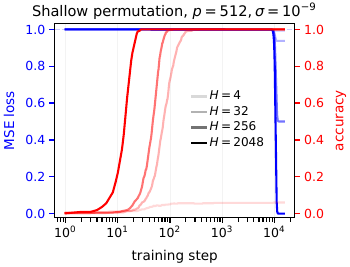}
    \hfill
    \includegraphics[width=0.32\linewidth]
    {unary_imgs/permutation_deep_H_sweep.pdf}
    \hfill
    \includegraphics[width=0.32\linewidth]
    {unary_imgs/random_unary_shallow_H_sweep.pdf}

    \caption{
    \textbf{Controlled unary-operation experiments.}
    Columns correspond to the shallow permutation model (left), the
    three-factor permutation model (middle), and the shallow model on an iid
    random unary target (right).  The top row varies the initialization scale
    \(\sigma\); the bottom row varies the hidden width \(H\).  Darker curves
    denote smaller \(\sigma\) in the top row and larger \(H\) in the bottom
    row.  Blue curves show full-population MSE and red curves show training
    accuracy.  The shallow permutation model reaches high accuracy while the loss remains near its initial value, without developing a well. Changing to the three-factor
    parameterization produces a pronounced well for the same flat-spectrum
    target, while a heterogeneous random unary target produces a well already
    in the shallow model, consistently with the heterogeneous-spectrum
    mechanism analyzed theoretically.
    }
    \label{fig:unary-controlled-sweeps}
\end{figure}

\paragraph{Setup, optimization, and measurement.}
Unless stated otherwise, our unary experiments use \(p=512\) inputs and \(p=512\) output classes and
train on the complete operation table.  There is therefore no train--test
split and no minibatch sampling.  We optimize
\begin{equation}
\mathcal L_{\mathbf T}(\mathbf M)
=
\frac{1}{p}\|\mathbf M-\mathbf T\|_{\mathrm F}^{2}.
\label{eq:unary-exp-loss}
\end{equation}
Full-batch gradient descent is implemented using
\texttt{torch.optim.SGD} with zero momentum and zero weight decay.

The implementation stores examples as columns: for input \(i\),
\begin{equation}
\mathbf T_{ci}
=
\mathbf 1\{f(i)=c\}.
\label{eq:unary-exp-target-convention}
\end{equation}
This is the same target convention as in Section~\ref{sec:linear}.  The prediction for input \(i\) is obtained by
taking the largest entry of column \(i\).

Every unary run contains \(15{,}000\) parameter updates, with loss and
top-one accuracy recorded throughout optimization.

The shallow permutation and shallow random-unary experiments use learning
rate
\begin{equation}
\eta_{\rm shallow}=0.5,
\label{eq:unary-exp-shallow-lr}
\end{equation}
whereas the three-factor permutation experiments use
\begin{equation}
\eta_{\rm deep}=50.
\label{eq:unary-exp-deep-lr}
\end{equation}
The learning rates were chosen separately for the two parameterizations.
Accordingly, the shallow-versus-deep permutation comparison should be read as
showing that the flat-spectrum target can exhibit a well under a deeper
factorized optimization dynamics, rather than as a matched learning-rate
depth ablation.

\paragraph{Initialization.}
We use fan-in-normalized Gaussian initialization multiplied by a global scale
\(\sigma\).  For the shallow model
\begin{equation}
\mathbf M=\mathbf U^{\top} \mathbf V,
\qquad
\mathbf U,\mathbf V\in\mathbb R^{H\times p},
\label{eq:unary-exp-shallow-model}
\end{equation}
the entries are initialized as
\begin{equation}
\mathbf V_{hi}(0)
\sim
\mathcal N\!\left(0,\frac{\sigma^2}{p}\right),
\qquad
\mathbf U_{hc}(0)
\sim
\mathcal N\!\left(0,\frac{\sigma^2}{H}\right).
\label{eq:unary-exp-shallow-init}
\end{equation}
For the deeper model
\begin{equation}
\mathbf M= \mathbf U^{\top} \mathbf W \mathbf V,
\qquad
\mathbf W\in\mathbb R^{H\times H},
\label{eq:unary-exp-deep-model}
\end{equation}
we additionally use
\begin{equation}
\mathbf W_{hk}(0)
\sim
\mathcal N\!\left(0,\frac{\sigma^2}{H}\right),
\label{eq:unary-exp-deep-init}
\end{equation}
while \(\mathbf U\) and \(\mathbf V\) retain the scalings in
\eqref{eq:unary-exp-shallow-init}.  There is no additional factor of \(1/H\)
in the forward map.

This normalization is useful for the width sweeps because the leading scale
of the initial predictor is independent of \(H\):
\begin{equation}
\mathbb E[\mathbf M_{ci}(0)^2]
=
\frac{\sigma^4}{p}
\qquad
\text{for }\mathbf M=\mathbf U^{\top}\mathbf V,
\label{eq:unary-exp-shallow-score-variance}
\end{equation}
and
\begin{equation}
\mathbb E[\mathbf M_{ci}(0)^2]
=
\frac{\sigma^6}{p}
\qquad
\text{for }\mathbf M=\mathbf U^{\top}\mathbf W\mathbf V.
\label{eq:unary-exp-deep-score-variance}
\end{equation}
Thus varying \(H\) changes the latent dimension without simultaneously
introducing an explicit growth of the initial logit scale.

This initialization differs from the unnormalized Gaussian initialization
used in the rigorous unary theory.  The experiments should therefore be
viewed as finite-width qualitative tests of the same phenomenology rather
than direct finite-\(p\) evaluations of the asymptotic formulas.

\paragraph{Sweep parameters and random seeds.}
The same permutation target is used in all permutation experiments.  It is
generated once using PyTorch seed \(0\).  For the shallow permutation width
sweep we fix
\(\sigma=10^{-9}\) and use
\(H\in\{4,32,256,2048\}\), with initialization seed \(1\).
For the corresponding initialization sweep we fix \(H=512\), use
\(\sigma\in\{10^{-3},10^{-5},10^{-7},10^{-9}\}\), and use initialization
seed \(100\).  The same seed is reset for every value of \(\sigma\), so the
same underlying Gaussian tensors are rescaled rather than independently
resampled.

For the three-factor permutation model, the width sweep fixes
\(\sigma=10^{-3}\), uses \(H\in\{16,64,512\}\), and uses initialization
seed \(200\).  The initialization sweep fixes \(H=128\), uses
\(\sigma\in\{10^{-1},10^{-2},10^{-3}\}\), and uses initialization seed
\(300\), again rescaling the same underlying Gaussian realization across
the sweep.  All of these runs contain \(15{,}000\) updates; for visual
clarity, the plotted three-factor width panel displays the first
\(8{,}000\) steps.

For the random-unary experiments, the target is sampled independently as
\begin{equation}
f(i)\overset{\mathrm{iid}}{\sim}\operatorname{Unif}([p])
\label{eq:unary-exp-random-target}
\end{equation}
using target seed \(1000\).  The particular realization used in the figures
has exactly \(328\) occupied output classes, and hence
\(\operatorname{rank}(T)=328\).  The same target is used in both random-unary
sweeps.  The initialization seed is \(2000\) throughout.  The initialization
sweep fixes \(H=512\) and uses
\(\sigma\in\{10^{-1},10^{-4},10^{-7},10^{-10}\}\); the width sweep fixes
\(\sigma=10^{-10}\) and uses \(H\in\{8,64,512\}\).

\paragraph{Shallow permutation: microscopic classification without a well.}
The left column of Figure~\ref{fig:unary-controlled-sweeps} gives the
flat-spectrum control.  Decreasing \(\sigma\) strongly postpones the
macroscopic decrease of the MSE, while the early classification transition
changes much less.  For sufficiently small initialization there is therefore
a long interval in which accuracy is already perfect while the loss remains
close to its zero-predictor value.  Nevertheless, none of the displayed
shallow permutation trajectories develops an accuracy well: once perfect
classification is reached, it is retained.  This is the finite-width analogue
of the behavior proved in
Theorem~\ref{thm:main-permutation-unary}.

The width sweep also illustrates the distinction between matrix fitting and
classification.  The permutation target has rank \(512\), whereas the
predictor \({\mathbf U}^{\top}\mathbf V\) has rank at most \(H\).  Hence the
\(H=4,32,\) and \(256\) runs cannot fit the target matrix exactly, but can
nevertheless attain perfect top-one accuracy (except for the $H=4$ run).  Increasing width makes the
early classifier form more rapidly and eventually removes the rank
obstruction to exact matrix fitting.

\paragraph{Three-factor permutation: a well without target spectral heterogeneity.}
The middle column keeps exactly the same permutation target but changes the
parameterization to \(\mathbf M={\mathbf U}^{\top}\mathbf W \mathbf V\).  The behavior is qualitatively
different.  At the larger initialization, accuracy evolves comparatively
monotonically, whereas sufficiently small initialization reveals a pronounced
high--low--high trajectory.  The width sweep shows the complementary
dependence on latent dimension: wider models develop a stronger early
classifier and a more pronounced subsequent nonmonotonicity.

Because the permutation target has a flat positive spectrum, the
heterogeneous-spectrum interference mechanism proved in
Appendix~\ref{app:linear-L2-random-unary} cannot operate here in its original form.
The experiment therefore demonstrates that target spectral heterogeneity is
not necessary for an accuracy well once the parameterization is changed.
This is qualitatively consistent with the broader bottleneck picture of
Section~\ref{sec:heuristic}.

\paragraph{Random unary target: heterogeneous spectra in the shallow model.}
The right column returns to \(\mathbf M=\mathbf U^{\top}\mathbf V\) but replaces the permutation
by the random target \eqref{eq:unary-exp-random-target}.  Sufficiently small
initialization produces a clear early accuracy hill followed by a substantial
well.  In the realizable \(H=512\) runs, accuracy later recovers to one.
The stagewise form of the trajectory is qualitatively consistent with the
occupancy-dependent spectral dynamics analyzed in
Appendix~\ref{app:linear-L2-random-unary}.

The width sweep also exposes behavior outside the realizable regime.  Since
the realized target has rank \(328\), the \(H=8\) and \(H=64\) runs cannot
fit it exactly.  These narrow networks reach very high accuracy during the
early stage but settle at substantially lower accuracy after much longer MSE
optimization.  Thus reduction of the optimized Frobenius error can destroy
class-ordering information that supported a better intermediate classifier.
The \(H=512\) model is realizable and eventually recovers perfect accuracy.

\paragraph{Evolution of the score matrix.}
To visualize the mechanism underlying the random-unary well more directly,
Figure~\ref{fig:main-mechanism} (left) repeats the shallow experiment in a
small system with \(p=10\), \(H=50\), and \(\sigma=10^{-9}\), using full-batch
gradient descent with learning rate \(0.1\).  The output labels are permuted
in decreasing order of target occupancy, and the inputs are grouped by their
target output for visualization.  At selected points along the trajectory we
display the end-to-end score matrix
\begin{equation}
\mathbf M(t)=\mathbf U(t)^{\top}\mathbf V(t),
\end{equation}
normalized separately at each time by
\begin{equation}
\widetilde{\mathbf M}(t)
=
\frac{\mathbf M(t)}
{\max_{c,i}|\mathbf M_{ci}(t)|}.
\end{equation}
Squares indicate the target output and dots the current prediction for each
input.  The snapshots include three states with the same accuracy \(A=0.6\),
taken respectively before the early peak, during the subsequent decline, and
during recovery.  They show that equal classification accuracy can correspond
to qualitatively different score-matrix structures at different stages of the
well.  The last four snapshots use a symmetric logarithmic color scale to make
weaker score components visible.

Taken together, the unary experiments provide two complementary departures
from the shallow permutation control.  A well appears after changing the
factorized dynamics while keeping the target spectrum flat, and it also
appears after changing to a heterogeneous target while keeping the shallow
factorization.  Across these settings, decreasing initialization separates
the microscopic accuracy dynamics from macroscopic loss decay, while
increasing width generally strengthens the early classifier.

\subsection{Accuracy wells in fully connected networks on MNIST}
\label{app:mnist-experiments}

The unary experiments allow unusually direct control of the target geometry
and factorization. We next ask which aspects of the accuracy-well
phenomenology persist in a standard multiclass classification problem. We use
MNIST to examine the effects of initialization scale, width, loss,
architectural biases, factorization depth, optimization method, and random
seed. We additionally use saved checkpoints from one fixed trajectory to probe
how the participation of individual hidden neurons changes through the
accuracy hill, well, and recovery.

\paragraph{Dataset and common training protocol.}
We use the standard MNIST training set of \(60{,}000\) grayscale
\(28\times28\) images from ten classes. Pixel values are divided by \(255\)
and each image is flattened to
\begin{equation}
x\in\mathbb R^{784}.
\label{eq:mnist-input}
\end{equation}
We apply no additional centering, normalization, or data augmentation. The
MNIST test set is not used for optimization or model selection. It is used
only for the post-hoc checkpoint diagnostics described below.

Unless stated otherwise, models are trained for \(10{,}000\) minibatch
updates with batch size
\begin{equation}
B=1024,
\end{equation}
using plain SGD with learning rate
\begin{equation}
\eta=0.005,
\end{equation}
zero momentum, zero weight decay, and no learning-rate schedule.
Minibatches are obtained from random permutations of the full training set.
Consecutive blocks of \(1024\) examples are used; when fewer than \(1024\)
examples remain, the remainder is discarded and a new random permutation is
generated.

Loss and top-one accuracy are evaluated on the current minibatch immediately
before the corresponding parameter update. Unless explicitly identified as
test accuracy, the curves therefore report minibatch training quantities.

The random streams are controlled across experimental conditions. For the
main sweeps, the global PyTorch and NumPy seed is \(1\), the default
weight-initialization seed is \(1234\), and the minibatch-permutation seed is
\(5678\). The minibatch generator is reset for every run, so different
experimental conditions see the same minibatch sequence. Within an
initialization-scale sweep of a fixed architecture, the initialization
generator is also reset for every value of \(\sigma\), so that the same
underlying Gaussian realization is merely rescaled. The experiments were run
in single precision on an NVIDIA Tesla P100-PCIE-16GB GPU.

\paragraph{ReLU model, initialization, and losses.}
Our nonlinear model is a one-hidden-layer ReLU network
\begin{equation}
f_\theta(x)
=
W_2\,\operatorname{ReLU}(W_1x+b_1)+b_2,
\qquad
W_1\in\mathbb R^{H\times784},
\qquad
W_2\in\mathbb R^{10\times H}.
\label{eq:mnist-relu-model}
\end{equation}
Depending on the experiment, either bias, both biases, or neither bias is
included. Whenever present, a bias is trainable and initialized at zero.

The weights use fan-in-normalized Gaussian initialization,
\begin{equation}
(W_1)_{hi}(0)
\sim
\mathcal N\!\left(0,\frac{\sigma^2}{784}\right),
\qquad
(W_2)_{ch}(0)
\sim
\mathcal N\!\left(0,\frac{\sigma^2}{H}\right).
\label{eq:mnist-relu-init}
\end{equation}

For the squared-loss experiments, the target for an example of class \(y\)
is the one-hot vector \(e_y\), and we optimize
\begin{equation}
\mathcal L_{\mathrm{MSE}}
=
\frac{1}{B}
\sum_{i=1}^{B}
\left\|
f_\theta(x_i)-e_{y_i}
\right\|_2^2.
\label{eq:mnist-mse}
\end{equation}
With this normalization, the zero predictor has loss one.

For the cross-entropy experiments we use
\begin{equation}
\mathcal L_{\mathrm{CE}}
=
-\frac{1}{B}
\sum_{i=1}^{B}
\log
\frac{
\exp f_{\theta,y_i}(x_i)
}{
\sum_{c=1}^{10}\exp f_{\theta,c}(x_i)
}.
\label{eq:mnist-ce}
\end{equation}
The corresponding zero-logit value is \(\log 10\).

\paragraph{Architectural sensitivity: hidden and output biases.}
We first use squared loss to test how the location of a trainable bias affects
the phenomenon. We compare four architectures:
\begin{center}
\begin{tabular}{@{}lcc@{}}
\toprule
Architecture & hidden bias \(b_1\) & output bias \(b_2\) \\
\midrule
No biases & absent & absent \\
Hidden bias only & present & absent \\
Output bias only & absent & present \\
Hidden and output biases & present & present \\
\bottomrule
\end{tabular}
\end{center}
Apart from the specified biases, the architectures use the same weight
initialization, loss, minibatch sequence, and optimization protocol.

\begin{figure}[t]
    \centering

    \includegraphics[width=0.48\linewidth]
    {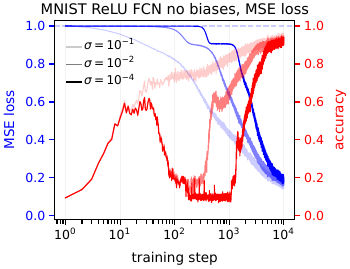}
    \hfill
    \includegraphics[width=0.48\linewidth]
    {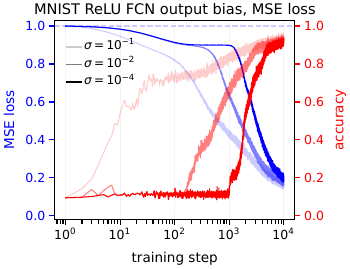}

    \vspace{0.45em}

    \includegraphics[width=0.48\linewidth]
    {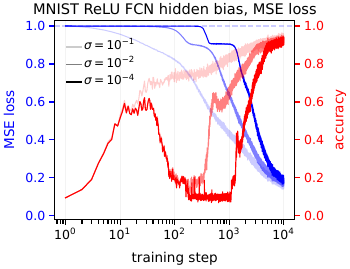}
    \hfill
    \includegraphics[width=0.48\linewidth]
    {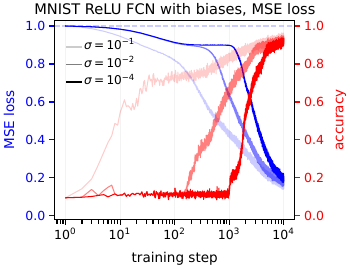}

    \caption{
    \textbf{Initialization-scale sweeps for the MNIST ReLU bias ablation
    under squared loss.}
    Columns distinguish the absence (left) and presence (right) of an output
    bias, while rows distinguish the absence (top) and presence (bottom) of a
    hidden bias. All panels use \(H=8192\) and
    \(\sigma\in\{10^{-1},10^{-2},10^{-4}\}\), with darker curves denoting
    smaller initialization. Blue curves show minibatch MSE and red curves
    show minibatch top-one accuracy. Without an output bias, sufficiently
    small initialization produces a pronounced accuracy well regardless of
    whether a hidden bias is present. Introducing an output bias suppresses
    the pronounced well in both corresponding architectures.
    }
    \label{fig:mnist-mse-bias-sigma}
\end{figure}

\begin{figure}[t]
    \centering

    \includegraphics[width=0.48\linewidth]
    {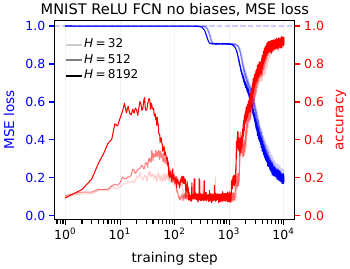}
    \hfill
    \includegraphics[width=0.48\linewidth]
    {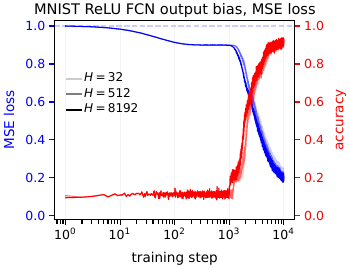}

    \vspace{0.45em}

    \includegraphics[width=0.48\linewidth]
    {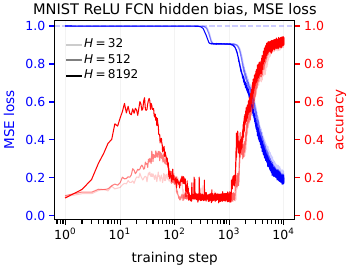}
    \hfill
    \includegraphics[width=0.48\linewidth]
    {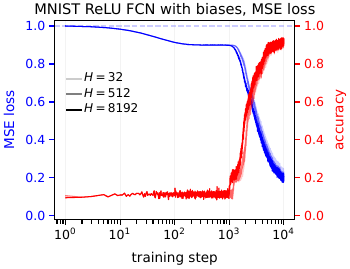}

    \caption{
    \textbf{Width sweeps for the MNIST ReLU bias ablation under squared
    loss.}
    The panel arrangement is the same as in
    Figure~\ref{fig:mnist-mse-bias-sigma}. All runs use
    \(\sigma=10^{-4}\) and \(H\in\{32,512,8192\}\), with darker curves
    denoting larger width. The no-bias and hidden-bias-only networks exhibit
    pronounced nonmonotone accuracy trajectories, whereas architectures
    containing an output bias do not develop a comparably pronounced well.
    }
    \label{fig:mnist-mse-bias-width}
\end{figure}

\begin{figure}[t]
    \centering

    \includegraphics[width=0.45\linewidth]
    {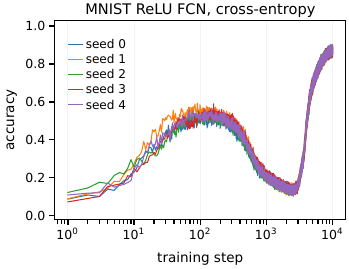}
    \hfill
    \includegraphics[width=0.45\linewidth]
    {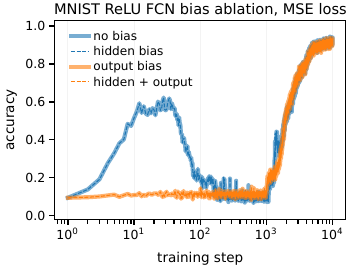}

    \caption{
    \textbf{Robustness and bias ablation for the MNIST accuracy
    well.}
    \emph{Left}: seed robustness for the bias-free ReLU network trained with
    cross-entropy at \(H=8192\), \(\sigma=10^{-4}\), showing raw minibatch
    accuracy trajectories for five independent initializations and minibatch
    streams. All five runs exhibit the same hill--well--recovery pattern.
    \emph{Right}: four-way bias ablation under squared loss at
    \(H=8192\), \(\sigma=10^{-4}\). The pronounced well is preserved with no
    bias or a hidden bias alone, but is suppressed when an output bias is
    present.
    }
    \label{fig:mnist-seeds-bias}
\end{figure}

For each architecture we perform an initialization-scale sweep with
\begin{equation}
H=8192,
\qquad
\sigma\in\{10^{-1},10^{-2},10^{-4}\},
\label{eq:mnist-mse-sigma-values}
\end{equation}
and a width sweep with
\begin{equation}
\sigma=10^{-4},
\qquad
H\in\{32,512,8192\}.
\label{eq:mnist-mse-width-values}
\end{equation}
The full sweeps are shown in
Figures~\ref{fig:mnist-mse-bias-sigma} and
\ref{fig:mnist-mse-bias-width}. A direct four-way comparison at
\(H=8192\) and \(\sigma=10^{-4}\) is shown in the right panel of
Figure~\ref{fig:mnist-seeds-bias}.

The distinction between hidden and output biases is robust across the two
sweeps. Adding a hidden bias leaves the characteristic early hill and
subsequent drop largely intact. A directly trainable class-wise output bias,
by contrast, suppresses the early hill and removes the pronounced well.
Adding a hidden bias together with the output bias does not restore the
effect. Thus the relevant architectural observation is not that biases in
general remove the well; rather, the location at which a bias enters the
predictor matters.

The initialization sweeps also reproduce the strong dependence on
initialization scale seen in the controlled unary experiments. For the
architectures that exhibit a well, decreasing \(\sigma\) makes substantial
accuracy evolution visible while the MSE remains close to its zero-predictor
baseline. The width sweeps show that the strength and timing of the early
classifier also depend on capacity.

\paragraph{Changing the objective: cross-entropy.}
We next keep the bias-free ReLU architecture and replace squared loss by
cross-entropy. The training protocol remains \(10{,}000\) SGD updates with
\(\eta=0.005\) and batch size \(1024\). The initialization sweep fixes
\begin{equation}
H=8192,
\qquad
\sigma\in\{10^{-1},10^{-2},10^{-4}\},
\end{equation}
while the width sweep fixes
\begin{equation}
\sigma=10^{-4},
\qquad
H\in\{32,512,8192\}.
\end{equation}
The two sweeps are shown in the middle column of
Figure~\ref{fig:main-cifar-mnist}.

The same qualitative accuracy well persists under cross-entropy. At smaller
initialization, accuracy rises substantially before the cross-entropy has
moved far from its zero-logit value, then falls before the later
high-accuracy state is reached. Increasing width changes the timing and
prominence of the early classifier while preserving the nonmonotone
trajectory. Thus the phenomenon is not tied to the one-hot squared objective
used in the unary experiments.

\paragraph{Robustness to random seed.}
To verify that the observed cross-entropy trajectory is not a peculiarity of
one initialization or one minibatch ordering, we repeat the bias-free ReLU
experiment for five independent random seeds. We fix
\begin{equation}
H=8192,
\qquad
\sigma=10^{-4},
\qquad
\eta=0.005,
\qquad
B=1024,
\end{equation}
and train for \(10{,}000\) SGD updates with cross-entropy. The five runs use
seeds \(0,1,2,3,4\); for each run both the Gaussian initialization and the
minibatch-permutation stream are changed. Thus this experiment probes
robustness to the two principal stochastic ingredients of the training
procedure rather than varying only the initial weights.

The resulting accuracy curves are shown in the left panel of
Figure~\ref{fig:mnist-seeds-bias}. All five runs display the same
qualitative sequence: an early accuracy hill, a substantial subsequent drop,
and a late sharp recovery. The location and height of the early maximum vary
moderately across seeds, but the existence and approximate timing of the well
are highly reproducible.

\paragraph{Parameterization versus small initialization: linear controls.}
The next experiments isolate the role of factorization from nonlinear
activation functions. We consider depth-\(L\) linear networks
\begin{equation}
f_\theta(x)
=
W_LW_{L-1}\cdots W_1x
\label{eq:mnist-deep-linear}
\end{equation}
without biases. The architectures are
\begin{equation}
784\to10
\quad (L=1),
\qquad
784\to H\to10
\quad (L=2),
\qquad
784\to H\to H\to10
\quad (L=3),
\end{equation}
with \(H=512\) for \(L\geq2\). Every layer is initialized independently
with
\begin{equation}
(W_k)_{ij}(0)
\sim
\mathcal N\!\left(
0,\frac{\sigma^2}{d_{k-1}}
\right),
\label{eq:mnist-linear-init}
\end{equation}
where \(d_{k-1}\) is the input dimension of layer \(k\). Independent
layer-wise Gaussian draws are generated using successive seeds
\(1234+k\).

The depth comparison uses
\begin{equation}
L\in\{1,2,3\},
\qquad
\eta=0.005,
\qquad
B=1024,
\qquad
10{,}000\ \text{updates},
\end{equation}
with depth-dependent initialization scales
\begin{equation}
\sigma_{L=1}=10^{-9},
\qquad
\sigma_{L=2}=10^{-3},
\qquad
\sigma_{L=3}=10^{-1}.
\label{eq:mnist-linear-depth-sigmas}
\end{equation}
The scales are chosen so that the dynamics of each parameterization is
clearly resolved within the common optimization horizon. Since the
initialization scale changes with depth, this comparison is intended as a
qualitative parameterization comparison rather than as a depth-scaling law at
fixed \(\sigma\).

Importantly, all three architectures represent the same class of end-to-end
functions: each defines a linear map from \(\mathbb R^{784}\) to
\(\mathbb R^{10}\). Because \(H=512>10\), the factorized models do not
suffer an end-to-end rank restriction relative to the direct \(L=1\)
classifier. The principal difference is therefore the parameterization of
the linear map and the gradient dynamics that this parameterization induces.

As a complementary control, we fix the direct \(L=1\) model and vary only its
initialization scale:
\begin{equation}
W_{ci}(0)
\sim
\mathcal N\!\left(0,\frac{\sigma^2}{784}\right),
\qquad
\sigma\in\{10^{-1},10^{-5},10^{-9}\}.
\label{eq:mnist-linear-L1-sigmas}
\end{equation}
These runs again use learning rate \(0.005\), batch size \(1024\), and
\(10{,}000\) updates.

\begin{figure}[t]
    \centering

    \includegraphics[width=0.48\linewidth]
    {mnist_imgs/mnist_mse_linear_depth.pdf}
    \hfill
    \includegraphics[width=0.48\linewidth]
    {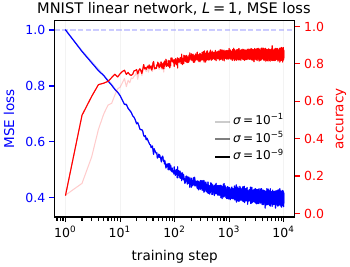}

    \caption{
    \textbf{Linear-network controls on MNIST under squared loss.}
    \emph{Left}: comparison of the direct \(L=1\) model with two- and
    three-factor parameterizations \(L=2,3\), using
    \(\sigma_{L=1}=10^{-9}\), \(\sigma_{L=2}=10^{-3}\), and
    \(\sigma_{L=3}=10^{-1}\). Although all three parameterizations represent
    the same class of end-to-end linear maps, the factorized models develop
    pronounced accuracy wells while the direct model has much simpler
    accuracy dynamics.
    \emph{Right}: initialization-scale sweep for the direct classifier with
    \(\sigma\in\{10^{-1},10^{-5},10^{-9}\}\). Even when the direct model is
    initialized extremely close to zero, its accuracy evolves comparatively
    monotonically and does not develop the pronounced well observed in the
    factorized models. Blue curves show minibatch MSE and red curves show
    minibatch top-one accuracy.
    }
    \label{fig:mnist-linear-controls}
\end{figure}

The left panel of Figure~\ref{fig:mnist-linear-controls} shows that changing
the parameterization can qualitatively alter the accuracy dynamics. Both
factorized models develop an early hill, subsequent deterioration, and later
recovery, despite having exactly the same end-to-end function class as the
direct linear classifier. The effect therefore does not require a nonlinear
activation or additional representational power.

The right panel provides the complementary control. We keep the direct
\(L=1\) parameterization fixed and reduce the initialization scale from
\(10^{-1}\) to \(10^{-9}\). Despite this eight-order-of-magnitude change, the
direct classifier does not develop the pronounced hill--well--recovery
trajectory seen in the factorized models. In particular, the two smallest
initializations produce very similar dynamics. Thus small initialization
alone is not sufficient to generate an accuracy well in this setting; the
factorized parameterization qualitatively changes the optimization dynamics.

\paragraph{Optimizer robustness.}
We next test the sensitivity of the phenomenon to the optimization method.
We use the bias-free ReLU network with cross-entropy,
\begin{equation}
H=1024,
\qquad
\sigma=10^{-4},
\qquad
B=1024,
\end{equation}
and train for \(10{,}000\) updates. The optimizer configurations are
\begin{center}
\begin{tabular}{@{}lll@{}}
\toprule
Optimizer & Learning rate & Additional parameters \\
\midrule
SGD & \(10^{-2}\) & momentum \(=0\) \\
SGD with momentum & \(10^{-2}\) & momentum \(=0.9\) \\
Adam & \(10^{-5}\) &
\(\beta_1=0.9,\ \beta_2=0.999,\ \epsilon=10^{-8}\) \\
\bottomrule
\end{tabular}
\end{center}
All three runs use zero weight decay and matched initialization and minibatch
streams. The momentum run uses standard PyTorch SGD momentum, while the Adam
run uses the standard PyTorch values of \(\beta_1,\beta_2\), and
\(\epsilon\). The resulting accuracy trajectories are shown in the
top-right panel of Figure~\ref{fig:main-cifar-mnist}.

A pronounced well appears for both SGD variants, and nonmonotone accuracy is
also visible under Adam, although the drop is substantially shallower for the
Adam hyperparameters used here. Because the three optimizers use different
learning rates, this experiment is best viewed as evidence that the
phenomenon is not specific to plain SGD rather than as a quantitative
optimizer comparison.

\paragraph{Neuron participation and output diversity through the well.}
The preceding experiments establish the phenomenology but do not by
themselves probe the internal reorganization associated with the proposed
target-driven organization--degeneracy bottleneck--recruitment picture.
We therefore analyze two complementary diagnostics along one fixed bias-free
ReLU trajectory trained with squared loss,
\begin{equation}
H=8192,
\qquad
\sigma=10^{-4},
\qquad
\eta=0.005,
\qquad
B=1024.
\label{eq:mnist-participation-run}
\end{equation}
The model is trained for \(10{,}000\) updates, and checkpoints are saved at
20 approximately logarithmically spaced times,
\begin{equation}
\begin{split}
t\in\{&
0,1,2,3,5,8,13,22,36,60,100,167,278,464,774,\\
&1292,2154,3594,5995,10000
\}.
\end{split}
\label{eq:mnist-checkpoint-times}
\end{equation}
For this diagnostic run we use initialization seed \(0\) and a separately
seeded minibatch stream.  All quantities below are evaluated post hoc on the
full MNIST test set of \(10{,}000\) examples and do not affect training.

We first quantify how broadly the class-dependent score contribution is
distributed across hidden neurons.  Write the bias-free network as
\begin{equation}
f_\theta(x)
=
\sum_{h=1}^{H}
v_h\,\phi(w_h^\top x),
\qquad
\phi(z)=\max\{z,0\},
\label{eq:mnist-neuron-decomposition}
\end{equation}
where \(v_h\in\mathbb R^{C}\), with \(C=10\), is the output-weight vector
associated with hidden neuron \(h\).  For each neuron, define its
class-conditional mean activation
\begin{equation}
a_h(c)
=
\frac{1}{N_c}
\sum_{i:\,y_i=c}
\phi(w_h^\top x_i),
\qquad
c=1,\ldots,C,
\label{eq:mnist-class-mean-activation}
\end{equation}
where \(N_c\) is the number of test examples of class \(c\), and let
\(a_h\in\mathbb R^C\) denote the vector with entries \(a_h(c)\).

A component of \(v_h\) proportional to the all-ones vector shifts all output
logits equally and therefore cannot affect their ordering.  We remove this
classification-irrelevant direction by defining
\begin{equation}
\widetilde v_h
=
v_h
-
\frac{\mathbf 1^\top v_h}{C}\mathbf 1.
\label{eq:mnist-centered-output-weight}
\end{equation}
The corresponding contribution of neuron \(h\) to the centered
class-conditional score matrix is
\begin{equation}
\widetilde B_h
=
a_h\widetilde v_h^\top
\in\mathbb R^{C\times C}.
\label{eq:mnist-neuron-score-contribution}
\end{equation}
We associate to neuron \(h\) the contribution energy
\begin{equation}
e_h
=
\|\widetilde B_h\|_{\mathrm F}^2
=
\|a_h\|_2^2
\|\widetilde v_h\|_2^2.
\label{eq:mnist-neuron-energy}
\end{equation}
This measures the magnitude of the neuron's contribution to class-dependent
score differences; it does not assume that the contribution is individually
aligned with the correct classifier.

We quantify the spread of this contribution over the hidden layer using the
participation ratio
\begin{equation}
N_{\mathrm{eff}}
=
\frac{
\left(\sum_{h=1}^{H}e_h\right)^2
}{
\sum_{h=1}^{H}e_h^2
},
\label{eq:mnist-neff}
\end{equation}
and report the normalized \emph{effective participation fraction}
\begin{equation}
\rho_{\mathrm{eff}}
=
\frac{N_{\mathrm{eff}}}{H}.
\label{eq:mnist-participation-fraction}
\end{equation}
Whenever the total contribution energy is nonzero,
\(1/H\leq\rho_{\mathrm{eff}}\leq1\).  A larger value means that
class-dependent score contribution is distributed over a larger effective
fraction of the hidden layer, whereas a smaller value indicates concentration
onto fewer neurons.  Importantly, \(\rho_{\mathrm{eff}}\) measures
concentration of contribution magnitudes rather than their correctness or
mutual coherence, and is therefore interpreted jointly with classification
accuracy.

As a complementary function-space diagnostic, we measure how broadly the
network uses the available output classes.  At checkpoint \(t\), let
\begin{equation}
\widehat y_i(t)
=
\operatorname*{arg\,max}_{c}
f_{\theta(t),c}(x_i)
\label{eq:mnist-predicted-label}
\end{equation}
be the predicted class of test example \(i\), and define the marginal
predicted-class frequencies
\begin{equation}
p_c(t)
=
\frac{1}{N_{\mathrm{test}}}
\sum_{i=1}^{N_{\mathrm{test}}}
\mathbf 1\!\left\{
\widehat y_i(t)=c
\right\},
\qquad
c=1,\ldots,C.
\label{eq:mnist-output-frequencies}
\end{equation}
We define the effective number of used output classes by the participation
ratio
\begin{equation}
C_{\mathrm{eff}}(t)
=
\frac{1}{
\sum_{c=1}^{C}p_c(t)^2
},
\label{eq:mnist-effective-output-number}
\end{equation}
and normalize it by the total number of classes,
\begin{equation}
\rho_{\mathrm{out}}(t)
=
\frac{C_{\mathrm{eff}}(t)}{C}
=
\frac{1}{
C\sum_{c=1}^{C}p_c(t)^2
}.
\label{eq:mnist-output-diversity}
\end{equation}
We refer to \(\rho_{\mathrm{out}}\) as the \emph{output diversity}.  It lies
between \(1/C\) and \(1\).  Uniform use of all \(C\) output classes gives
\(\rho_{\mathrm{out}}=1\), while complete collapse onto a single predicted
class gives \(\rho_{\mathrm{out}}=1/C\).  More generally, approximately
uniform use of only \(k\) classes gives
\(\rho_{\mathrm{out}}\approx k/C\).  Unlike accuracy, output diversity does
not measure whether predictions are correct: a classifier may distribute its
predictions evenly across classes while assigning the wrong label to every
example.  It therefore supplies information complementary to accuracy.

The right panel of Figure~\ref{fig:main-mechanism} shows full-test accuracy,
the hidden-neuron participation fraction \(\rho_{\mathrm{eff}}\), and the
output diversity \(\rho_{\mathrm{out}}\) at the saved checkpoints.  The three
quantities reveal a structured sequence consistent with the heuristic picture
of Section~\ref{sec:heuristic}.  During the initial target-driven
organization, accuracy rises while output predictions remain distributed
across several classes and the class-dependent score contribution is spread
over a relatively large effective fraction of the hidden layer.  As the
trajectory enters the accuracy well, both diagnostics collapse:
\(\rho_{\mathrm{eff}}\) falls sharply, showing that the centered
class-dependent score contribution becomes concentrated on a much smaller
effective subset of hidden neurons, while \(\rho_{\mathrm{out}}\) approaches
its minimum value \(1/C\), indicating that predictions themselves become
concentrated onto very few output classes.  Thus the accuracy minimum
coincides with a bottleneck both in hidden-unit participation and in the
diversity of the resulting classifier outputs.

The recovery stage reverses both forms of degeneracy.  Output diversity rises
rapidly toward one as the classifier begins to use the full set of output
classes again.  Hidden-neuron participation also rebounds, indicating renewed
involvement of additional components of the hidden layer, although it does
not return to its initial value.  Accuracy rises concurrently to its final
high level.  This simultaneous restoration of output diversity and partial
increase in neuronal participation is consistent with the recruitment stage
of the heuristic picture.

The two diagnostics probe different aspects of the same transition.
\(\rho_{\mathrm{eff}}\) is a parameterization-dependent measure of how
class-dependent score contribution is distributed across hidden neurons,
whereas \(\rho_{\mathrm{out}}\) is a purely functional statistic describing
the marginal diversity of the classifier's predictions.  Their simultaneous collapse near the bottom of the well therefore gives more specific evidence for a transient degeneracy bottleneck than accuracy alone. 

Taken together, the MNIST experiments show that the accuracy well persists
across two losses, a wide range of widths, multiple random seeds, and several
optimization methods, while remaining sensitive to parameterization.
A class-wise output bias suppresses the pronounced well, whereas a hidden
bias does not; factorizing a linear classifier can create a strong well
without changing its end-to-end function class, while simply shrinking the
initialization of the direct classifier does not.

The checkpoint diagnostics provide a complementary mechanistic observation.
As the early classifier deteriorates, class-dependent score contribution
becomes sharply concentrated onto a smaller effective subset of hidden
neurons, while the marginal distribution of predicted labels simultaneously
collapses onto very few output classes.  Both forms of degeneracy are reversed
during recovery: output diversity rapidly approaches its maximum and hidden
participation partially rebounds as accuracy rises.  This gives quantitative
support for the heuristic
picture developed in Section~\ref{sec:heuristic}, while leaving open the
precise nonlinear dynamics responsible for the bottleneck.

\subsection{Accuracy wells in a convolutional network on CIFAR-10}
\label{app:cifar10-cnn}

The controlled experiments above establish the accuracy-well phenomenon in
analytically tractable matrix-factorization models and in fully connected
networks on MNIST.  We next ask whether the same qualitative dynamics remain
visible in a more conventional nonlinear architecture trained on natural
images.  To this end, we train a simple two-hidden-layer convolutional neural
network on CIFAR-10 \citep{krizhevsky2009learning}.  Our objective is not to
optimize benchmark performance, but rather to test for the phenomenon in a
minimal genuinely convolutional setting.  We therefore omit pooling,
normalization layers, residual connections, dropout, and data augmentation,
and study separately the dependence on initialization scale and model
capacity.  We record both training and held-out test dynamics.

\begin{figure}[t]
    \centering

    \includegraphics[width=0.48\linewidth]
    {cifar10_imgs/sigma_sweep_train.pdf}
    \hfill
    \includegraphics[width=0.48\linewidth]
    {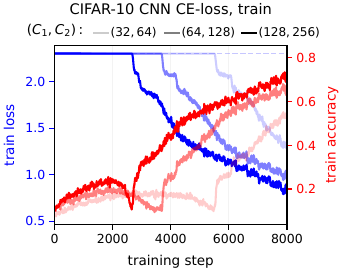}

    \vspace{0.5em}

    \includegraphics[width=0.48\linewidth]
    {cifar10_imgs/sigma_sweep_test.pdf}
    \hfill
    \includegraphics[width=0.48\linewidth]
    {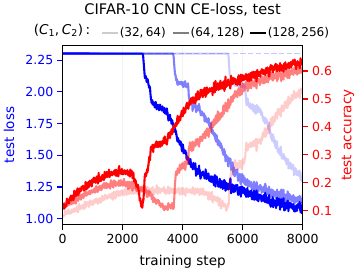}

    \caption{
    \textbf{Accuracy-well dynamics in a two-hidden-layer CNN on CIFAR-10.}
    Columns correspond to the initialization-scale sweep (left) and the
    model-capacity sweep (right), while rows show training dynamics (top) and
    held-out test dynamics (bottom).
    In the left column, the channel widths are fixed to
    \((C_1,C_2)=(128,256)\), and darker curves correspond to smaller
    initialization scale \(\sigma\).
    In the right column, \(\sigma=10^{-4}\) is fixed, and darker curves
    correspond to larger channel widths.
    Blue curves show cross-entropy loss and red curves show top-one accuracy.
    The horizontal dashed blue line denotes the zero-logit cross-entropy
    \(\log 10\).
    Each displayed trajectory is a 25-step moving average of the corresponding
    per-minibatch measurements.
    Small initialization exposes a pronounced early accuracy hill followed by
    an accuracy well, with substantial accuracy dynamics occurring while the
    cross-entropy remains close to its zero-logit value.
    In the capacity sweep, larger models form a stronger early classifier and
    undergo an earlier and more pronounced hill--well--recovery trajectory.
    The corresponding held-out test curves closely reproduce the qualitative
    training dynamics.
    }
    \label{fig:cifar10-cnn-sweeps}
\end{figure}

\paragraph{Architecture and data preprocessing.}
We use the standard CIFAR-10 train--test split, consisting of \(50{,}000\)
training and \(10{,}000\) test RGB images of spatial resolution
\(32\times32\) from ten classes.
The dataset is loaded through the Hugging Face
\texttt{uoft-cs/cifar10} interface.
Pixel intensities are first divided by \(255\) and subsequently normalized
channelwise using
\begin{equation}
\mu
=
(0.4914,\,0.4822,\,0.4465),
\qquad
s
=
(0.2470,\,0.2435,\,0.2616).
\label{eq:cifar-normalization}
\end{equation}
No data augmentation is applied.

The network consists of two convolutional hidden layers followed by a
fully-connected classifier:
\begin{equation}
h_1(x)
=
\operatorname{ReLU}
\!\left(
\operatorname{Conv}^{3\rightarrow C_1}_{3\times3}(x)
\right),
\label{eq:cifar-cnn-layer1}
\end{equation}
\begin{equation}
h_2(x)
=
\operatorname{ReLU}
\!\left(
\operatorname{Conv}^{C_1\rightarrow C_2}_{3\times3}(h_1(x))
\right),
\label{eq:cifar-cnn-layer2}
\end{equation}
and
\begin{equation}
f_\theta(x)
=
W_{\rm out}\operatorname{vec}(h_2(x)),
\qquad
W_{\rm out}
\in
\mathbb R^{10\times(32\cdot32\cdot C_2)}.
\label{eq:cifar-cnn-output}
\end{equation}
Both convolutions have kernel size \(3\times3\), stride one, and padding one,
so the spatial resolution remains \(32\times32\) through both convolutional
layers.  All convolutional and fully-connected layers are bias-free.  The
model contains no pooling, batch normalization, dropout, or residual
connections.

\paragraph{Optimization and initialization.}
We minimize the mean cross-entropy loss using plain minibatch SGD with fixed
learning rate
\begin{equation}
\eta=0.025,
\label{eq:cifar-learning-rate}
\end{equation}
batch size \(128\), zero momentum, and zero weight decay.  No learning-rate
schedule is used.

Every scalar entry of each convolutional and fully-connected weight tensor is
initialized independently according to
\begin{equation}
\theta_j(0)
\sim
\mathcal N(0,\sigma^2).
\label{eq:cifar-entrywise-init}
\end{equation}
Thus, in these experiments, \(\sigma\) denotes a common \emph{entrywise}
standard deviation and is not divided by the fan-in of the corresponding
layer.  This choice deliberately differs from the parameterizations used in
the preceding unary and MNIST experiments and therefore provides an
additional robustness test: the CIFAR-10 experiments ask whether the
accuracy-well phenomenology persists under a substantially different
initialization convention rather than being tied to one particular
normalization.

\paragraph{Measurement protocol and reproducibility.}
At every optimization step, the implementation draws one training minibatch
and one test minibatch, each containing \(128\) examples.  Separate shuffled
streams are used for the two splits.  Training loss and accuracy and test loss
and accuracy are all evaluated at the same parameter state, immediately
before the SGD update.  Only the training cross-entropy is differentiated and
used to update the parameters; the test data therefore play no role in
optimization.

Within each epoch, a fresh random permutation of the corresponding split is
generated and divided into complete minibatches of size \(128\); an incomplete
remainder is omitted before a new permutation is generated.  Loss and
accuracy are stored at every optimization step.  The curves in
Figure~\ref{fig:cifar10-cnn-sweeps} show 25-step moving averages of these
per-minibatch measurements.

We match the pseudorandom streams across experimental conditions.  The global
seed is \(1\), the dedicated initialization generator uses seed \(123\), and
the training and test permutation generators use seeds \(456\) and \(789\),
respectively.  The initialization generator is reset at the beginning of each
run.  Consequently, within the fixed-architecture initialization-scale sweep,
the same underlying Gaussian realization is multiplied by the different
values of \(\sigma\).  The training and test data streams are likewise reset
for every run, so corresponding conditions encounter the same sequences of
minibatch indices.  Each curve displayed in
Figure~\ref{fig:cifar10-cnn-sweeps} corresponds to one run with these fixed
seeds.

\paragraph{Initialization-scale sweep.}
We first fix the architecture to
\begin{equation}
(C_1,C_2)
=
(128,256)
\label{eq:cifar-fixed-width}
\end{equation}
and train for \(5{,}000\) SGD updates using
\begin{equation}
\sigma
\in
\left\{
4\times10^{-3},
\;
2\times10^{-4},
\;
10^{-4}
\right\}.
\label{eq:cifar-sigma-sweep}
\end{equation}
The left column of Figure~\ref{fig:cifar10-cnn-sweeps} shows that decreasing
the initialization scale progressively separates the early accuracy dynamics
from the macroscopic decay of the cross-entropy.

For the largest initialization, \(\sigma=4\times10^{-3}\), the
cross-entropy begins decreasing almost immediately and the accuracy increases
without a pronounced accuracy well.  At the two smaller initialization
scales, the behavior is qualitatively different.  The loss remains close to
the zero-logit value
\begin{equation}
\mathcal L_0
=
\log 10,
\label{eq:cifar-zero-logit-loss}
\end{equation}
for an extended interval, even though accuracy has already risen appreciably
above chance.

The separation is clearest for \(\sigma=10^{-4}\).  While the loss is still
essentially on its initial plateau, the accuracy first rises from chance and
forms a clear early hill.  It then drops sharply back toward chance before
recovering again as the cross-entropy enters its macroscopic descent.  Thus,
the nonmonotonicity is already present in a regime in which the continuous
training objective changes very little: useful class-ordering information is
formed and subsequently degraded before conventional loss-based learning
becomes macroscopically visible.

Reducing \(\sigma\) also delays the characteristic transitions.  The accuracy
well occurs earlier for \(\sigma=2\times10^{-4}\) and later for
\(\sigma=10^{-4}\), consistently with the general role of small
initialization in opening a longer microscopic dynamical regime.

The held-out trajectories in the bottom-left panel show the same qualitative
behavior at closely corresponding training times.  In particular, for small
initialization, test accuracy also develops an early hill, subsequently
collapses, and later recovers.  Hence the transient early predictor is not
merely fitting the minibatches used for the SGD updates: it already captures
class-dependent structure that generalizes to unseen CIFAR-10 examples.

\paragraph{Model-capacity sweep.}
We next fix
\begin{equation}
\sigma
=
10^{-4}
\label{eq:cifar-capacity-fixed-sigma}
\end{equation}
and vary the convolutional channel widths according to
\begin{equation}
(C_1,C_2)
\in
\left\{
(32,64),
\;
(64,128),
\;
(128,256)
\right\}.
\label{eq:cifar-capacity-sweep}
\end{equation}
Each condition is trained for \(10{,}000\) SGD updates.  To focus on the
transient regime of interest, the right column of
Figure~\ref{fig:cifar10-cnn-sweeps} displays the first \(8{,}000\) updates.

The capacity sweep shows a strong systematic dependence of the transient
dynamics on model size.  The largest network,
\((C_1,C_2)=(128,256)\), forms the strongest early classifier and undergoes
the earliest pronounced accuracy well.  The intermediate network develops a
lower early hill and makes the corresponding transition later, while the
smallest network forms the weakest early classifier and remains in the
microscopic regime for considerably longer.  Nonmonotone structure is
nevertheless visible throughout the family.  The same ordering and
approximately the same transition times appear in the held-out test
trajectories.

The capacity sweep is not intended as a controlled infinite-width scaling
limit.  Because the same entrywise initialization variance
\(\sigma^2\) is used while \(C_1\) and \(C_2\) vary, changing the channel
widths also changes aggregate activation and initialization statistics.
Accordingly, the experiment does not isolate width while holding all other
initialization statistics fixed.  Its empirical conclusion is instead that,
within this family of increasingly large CNNs under a fixed entrywise
initialization convention, greater model capacity is associated with a
stronger early classifier, an earlier transition, and a more pronounced
hill--well--recovery trajectory.

This parametrization difference is also useful from the broader
phenomenological perspective.  The preceding unary and MNIST experiments and
the present CNN experiment use different initialization conventions, yet
closely related accuracy-well dynamics are observed across these settings.
The CIFAR-10 results therefore provide evidence that the phenomenon persists
across substantially different parameterizations; we do not, of course,
claim invariance under arbitrary choices of parametrization.

\begin{figure}[t]
    \centering
    \includegraphics[width=0.48\linewidth]
    {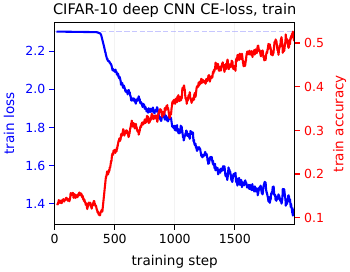}
    \hfill
    \includegraphics[width=0.48\linewidth]
    {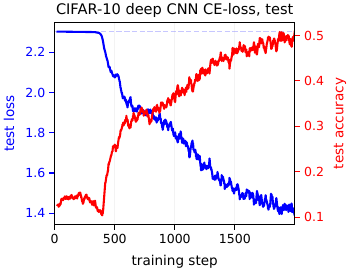}
    \caption{
    \textbf{Dynamics of a deeper convolutional network on CIFAR-10.}
    Training trajectories are shown on the left and held-out test trajectories
    on the right.  The network has five convolutional hidden layers with
    \((C_1,\ldots,C_5)=(256,256,256,256,256)\), followed by a fully connected
    ten-class output layer.  All weights use entrywise Gaussian initialization
    with \(\sigma=8\times10^{-3}\); optimization uses minibatch SGD with
    learning rate \(0.025\) and batch size \(128\).  Blue curves show
    cross-entropy loss and red curves show top-one accuracy; the horizontal
    dashed blue line denotes the zero-logit cross-entropy \(\log 10\).
    Curves are 25-step moving averages of the per-minibatch measurements.
    }
    \label{fig:cifar10-deep-cnn}
\end{figure}

\paragraph{Additional deep convolutional experiment.}
As an additional architectural stress test, we also consider a substantially
deeper convolutional network with five hidden convolutional layers followed
by a fully connected output layer.  We use
\begin{equation}
(C_1,C_2,C_3,C_4,C_5)
=
(256,256,256,256,256),
\label{eq:cifar-deep-cnn-widths}
\end{equation}
with a ReLU nonlinearity after every convolution.  As in the preceding CNN
experiments, all convolutions have \(3\times3\) kernels, stride one, padding
one, and no biases, pooling, normalization layers, dropout, or residual
connections.  We retain the same CIFAR-10 preprocessing, minibatch size
\(128\), SGD learning rate \(\eta=0.025\), and train--test measurement
protocol described above.  All weight entries are initialized independently
from
\begin{equation}
\mathcal N(0,\sigma^2),
\qquad
\sigma=8\times10^{-3},
\label{eq:cifar-deep-cnn-init}
\end{equation}
and the network is trained for \(2{,}000\) SGD updates.  Figure
\ref{fig:cifar10-deep-cnn} shows the resulting training and held-out test
trajectories.  Despite the considerable increase in depth relative to the
two-convolution architecture considered above, the dynamics remain strongly
structured at small initialization: accuracy begins to improve while the
network is still close to the zero-logit regime, and the corresponding
training and test trajectories evolve on closely related timescales.  This
additional experiment therefore shows that the qualitative separation between
early predictive dynamics and macroscopic loss evolution is not restricted
to the shallow convolutional architecture used in the parameter sweeps above.

\paragraph{Summary.}
The CIFAR-10 experiments extend the observed accuracy-well phenomenology to a
nonlinear convolutional architecture trained on natural images.  At
sufficiently small initialization, the CNN exhibits the characteristic
sequence
\begin{equation}
\text{chance}
\;\longrightarrow\;
\text{early accuracy hill}
\;\longrightarrow\;
\text{accuracy well}
\;\longrightarrow\;
\text{recovery},
\label{eq:cifar-accuracy-well-sequence}
\end{equation}
with substantial hill--well dynamics occurring while the cross-entropy
remains close to its zero-logit value.  Smaller initialization separates the
microscopic accuracy dynamics more clearly from macroscopic loss decay, while
the capacity sweep shows that larger models in the tested family form a
stronger early classifier and traverse the subsequent transition earlier.
The close correspondence between training and held-out trajectories further
shows that the transient classifier already generalizes beyond the
optimization minibatches.

We do not infer from these experiments that the internal mechanism producing
the CNN accuracy well is identical to the spectral mechanism proved for the
unary targets.  Rather, the results show that the same qualitative
nonmonotone accuracy dynamics survive substantial changes in architecture,
dataset, loss function, stochastic optimization, and initialization
parametrization, supporting the broader phenomenological picture developed
in this work.

\end{document}